\documentclass[smallextended]{svjour3}       
\smartqed  
\usepackage{graphicx}
\usepackage[utf8]{inputenc} 
\usepackage[T1]{fontenc}    
\usepackage{hyperref}       
\usepackage{booktabs}       
\usepackage{amsfonts}       
\usepackage{nicefrac}       
\usepackage{microtype}      
\usepackage{color}
\usepackage{amsmath,cancel,subfigure,pifont,tikz}
\usetikzlibrary{arrows,shapes,backgrounds,through,shadows}
\usepackage[percent]{overpic}
\usepackage{wrapfig}

\DeclareMathOperator{\argmax}{argmax}
\DeclareMathOperator{\sigmoid}{sigm}
\newcommand{\figref}[1]{Fig.~\ref{#1}}
\renewcommand{\eqref}[1]{Eq.~(\ref{#1})}

\newcommand{\calN}{{\cal N}}

\newcommand{\Sigmah}{\Sigma_H}

\newcommand{\SigmaH}{\Sigma_H}
\newcommand{\lstms}{{\sigma}}

\tikzstyle{disc}=[rectangle,
draw=blue!50,
thick,inner sep=0pt,
line width=1pt,
minimum size=6mm]  
\tikzstyle{rnn}=[diamond,draw=black,thick,inner sep=0pt,line width=1pt,minimum size=8mm]  
\tikzstyle{randomv}=[circle,draw=black,thick,inner sep=0pt,minimum size=7mm,>=stealth]  
\tikzstyle{obs}=[fill=blue!20,thick]  
\tikzstyle{ocont}=[circle,draw=blue!50,thick,inner sep=0pt,minimum size=8mm,>=stealth]  
\tikzstyle{dgraph}=[->, line width=1pt]
\tikzstyle{ugraph}=[line width=1pt]

\begin{document}

\title{{\LARGE Unsupervised Separation of Dynamics from Pixels}}

\titlerunning{Unsupervised Separation of Dynamics from Pixels}        


\author{Silvia Chiappa\thanks{Silvia Chiappa and Ulrich Paquet contributed equally.} \lastand
       Ulrich Paquet\\[8pt]
              DeepMind, London, UK\\
              {\small \{csilvia,upaq\}@google.com}
}

\institute{
}%

\date{}
\authorrunning{Silvia Chiappa and Ulrich Paquet}

\maketitle

\begin{abstract}
\vskip-1.5cm
We present an approach to learn the dynamics of multiple objects from image sequences in an unsupervised way. We introduce a probabilistic model that first generate noisy positions for each object through a separate linear state-space model, and then renders the positions of all objects in the same image through a highly non-linear process. Such a linear representation of the dynamics enables us to propose an inference method that uses exact and efficient inference tools and that can be deployed to query the model in different ways without retraining.
\end{abstract}

\section{Introduction}
\label{intro}
The dynamics of moving objects can be described on a very low-dimensional manifold of positions and velocities, e.g.~by
using a state-space model representation of Newtonian laws. A wealth of literature in probabilistic tracking \cite{barshalom93estimation,blackman99design} allows us to ask questions such as where objects are going to be in the future or where they have been in the past from noisy observations of their current positions, if the latent dynamics are known or can be learned.

In this paper, we consider the more challenging task of learning the presence of multiple objects and their dynamics in an unsupervised way from image observations. The problem of learning dynamics from pixels has been studied for the simpler case of one object \cite{fraccaro17disentangled,pearce2018comparing}.
The multiple-object scenario is much harder as an object-to-identity assignment problem is introduced. 
The unsupervised learning method should be able to reliably disentangle objects from a sequence of images---this requires a recurrent attention mechanism which should learn to track each object separately over time.

Deep recurrent neural networks have demonstrated impressive results on pixel-prediction and related tasks
\cite{babaeizadeh18stochastic,chiappa17recurrent,denton17unsupervised,Finn2016,Oh2015,srivastava15unsupervised,Sun2016}.
Whilst powerful and easy to design, in such methods the hidden states do not generally correspond to interpretable dynamics. Enforcing a desired hidden-state representation is challenging---attempts to get positions from pixels have so far succeeded only in the supervised scenario \cite{watters17visual}.
These methods also have shortcomings when asked to infer intermediate images from observed preceding and following images in a sequence.

Probabilistic extensions, such as modern stochastic variational approaches to hidden Markov models \cite{Fraccaro2016,Gao2016,Krishnan2017}, can achieve interpretable hidden-state representations and perform rich probabilistic reasoning. 
The capabilities of these methods can be further enhanced with the use of traditional probabilistic inference routines \cite{fraccaro17disentangled,johnson16composing,lin18variational}. 
This paper contributes
an approach in this direction. In Sec. \ref{sec:generative}, we introduce a model for generating images containing multiple objects in which positions are explicitly represented using auxiliary variables. In Sec. \ref{sec:IL}, we
leverage this representation to introduce a method for performing inference and learning that makes use of  exact and efficient inference techniques. Finally, in Sec. \ref{sec:results} we show how our approach performs on inferring latent positions from image sequences, and on image generation and interpolation, using an artificial dataset representing moving cannonballs.

\section{A model for rendering and inferring multiple objects dynamics}
\label{sec:model}

\begin{wrapfigure}[9]{r}{0.35\textwidth}
\vskip-0.75cm
\includegraphics[width=0.17\textwidth]{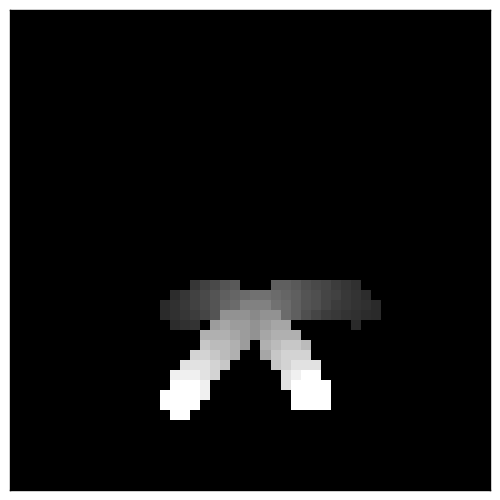} 
\includegraphics[width=0.17\textwidth]{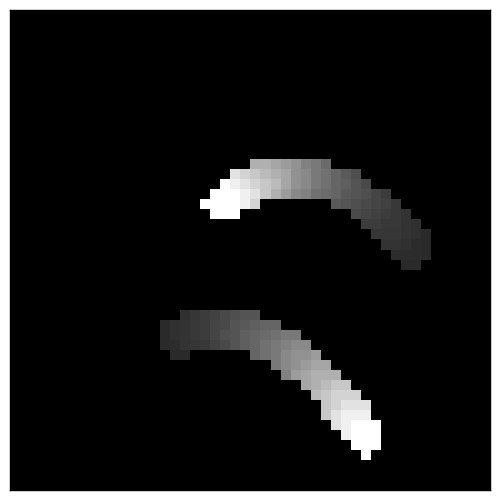}
\caption{Two sequences of images overlaid in time, each containing two cannonballs moving in opposite directions.}
\label{fig:examples}
\end{wrapfigure}

We wish to learn the dynamics of $N$ objects from sequences of images $v_{1:T} \equiv v_1, \ldots, v_T$ in an unsupervised way.
We restrict ourselves to the case in which the objects move independently in the two-dimensional plane, and $N$ is known; and assume that each image is formed by white pixels representing objects positions and a black background. 

Two examples of sequences, each containing two cannonballs moving in opposite directions, are given in \figref{fig:examples}---the images are overlaid in time such that lighter shades correspond to more recent images.

The observed dynamics in the pixel space are high-dimensional and non-linear.
However, the intrinsic dynamics can be described on the low-dimensional manifold of 
positions and velocities by simple dynamical systems.
In the sections that follow, we show that
the explicit representation of such latent dynamics enables us to infer positions using exact and efficient techniques. We also show that our approach enables us to answer different questions using the learned dynamics
without the need to retrain or modify the model.

\subsection{Generative model}\label{sec:generative}
We assume that the generative process underlying the observed images consists of two main parts: 
A part that describes the objects dynamics in the low-dimensional manifold of positions and velocities through a linear state-space model, and a part that renders the latent positions of the $N$ objects into the images through a highly non-linear process. 

\subsubsection{Rendering latent object positions into images}
\label{sec:rendering-objects}

We use a set of auxiliary variables $a^n_{1:T}$ to explicitly represent the latent positions of object $n$ in the two-dimensional plane.
Given the positions of all objects at time $t$, $a^{1:N}_t\equiv a^1_t,\ldots, a^N_t$,
an image $v_t$ is generated by
recurrently rendering each $a^n_t$ on $v_t$.
We start with $x^0 = \tanh(\theta_{x^0})$,
which represents a latent state vector from which an empty image canvas
is generated, with $\theta_{x^0}$ denoting an unknown parameter vector.
For $n=1,\ldots,N$, we iterate
\begin{align}
\alpha^n & = \mathrm{sigm}(W^{\alpha} a^n_t + b^{\alpha})   & & \textrm{(latent attention mask)} \nonumber\\
\hat{x}^n & = \tanh(W^x a^n_t + b^x) & & \textrm{(object $n$'s state contribution)} \nonumber\\
x^n & = (1 - \alpha^n) \circ x^{n-1} + \alpha^n \circ \hat{x}^n & & \textrm{(state update)} \,,
\label{eq:rendering}
\end{align}
and finally generate the image as
\begin{equation} \label{eq:v}
v_t \sim p_{\theta}(v_t | a_t^{1:N}) = 
\mathrm{Bernoulli}(\mathrm{sigm}(W^v x^N + b^v)) \, .
\end{equation}
The symbol $\circ$ indicates element-wise vector multiplication, and $\mathrm{sigm}(a) \equiv 1 / (1 + \mathrm{e}^{-a} )$.
The $W$'s and $b$'s indicate weight matrices and biases, and are all included in the unknown parameters $\theta$ of the generative model.

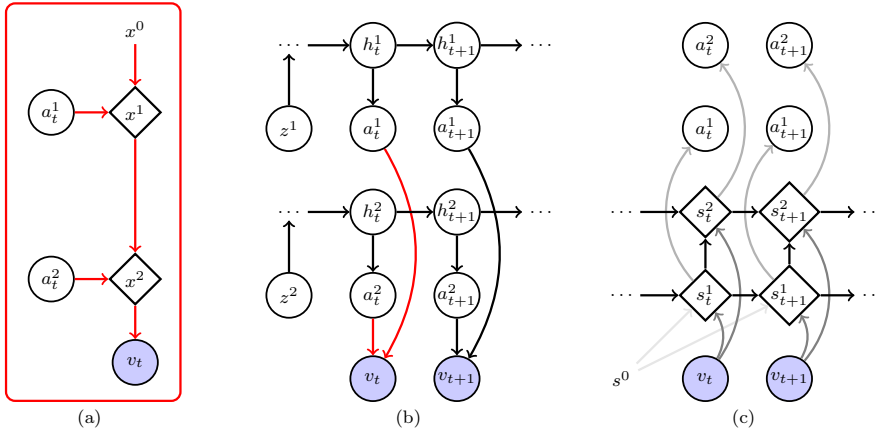
\begin{figure}[t]
\centering
\scalebox{0.85}{
\subfigure[]{
\begin{tikzpicture}[->, line width=1pt, node distance=1.3cm]
\draw[rounded corners, draw=red] (-0.7, 1.7) rectangle (2.0, -4.5) {};
\node[randomv] (gena1t) {$a_t^1$};
\node[rnn, right of = gena1t] (x1) {$x^1$};
\node[draw=none, fill=none, above of = x1] (x0) {$x^0$};
\node[draw=none, fill=none, below of = x1] (blank2) {};
\node[rnn, below of = blank2] (x2) {$x^2$};
\node[randomv, left of = x2] (gena2t) {$a_t^2$};
\node[randomv, obs, below of = x2] (genvt) {$v_t$};
\draw[line width=1pt, draw=red](x0)--(x1);
\draw[line width=1pt, draw=red](x1)--(x2);
\draw[line width=1pt, draw=red](x2)--(genvt);
\draw[line width=1pt, draw=red](gena1t)--(x1);
\draw[line width=1pt, draw=red](gena2t)--(x2);
\end{tikzpicture}
\label{fig:model:generative}
}
\hskip1.0cm
\subfigure[]{
\begin{tikzpicture}[->, line width=1pt, node distance=1.3cm]
\node[draw=none, fill=none] (htm1) {$\ldots$};
\node[randomv, below of = htm1] (z1) {$z^1$};
\node[randomv, right of = htm1] (ht1) {$h_t^1$};
\node[randomv, right of = ht1] (htp1) {$h_{t+1}^1$};
\draw[line width=1pt](z1)--(htm1);
\node[draw=none, fill=none, right of = htp1] (htpp1) {$\ldots$};
\node[randomv, below of = ht1] (at1) {$a_t^1$};
\node[randomv, below of = htp1] (atp1) {$a_{t+1}^1$};
\node[randomv, below of = at1] (ht2) {$h_t^2$};
\node[draw=none, fill=none, left of = ht2] (htm2) {$\ldots$};
\node[randomv, right of = ht2] (htp2) {$h_{t+1}^2$};
\node[draw=none, fill=none, right of = htp2] (htpp2) {$\ldots$};
\node[randomv, below of = ht2] (at2) {$a_t^2$};
\node[randomv, below of = htp2] (atp2) {$a_{t+1}^2$};
\node[randomv, left of = at2] (z2) {$z^2$};
\draw[line width=1pt](z2)--(htm2);
\node[randomv, obs, below of = at2] (vt) {$v_t$};
\node[randomv, obs, below of = atp2] (vtp) {$v_{t+1}$};
\draw[line width=1pt, draw=red](at2)--(vt);
\draw[line width=1pt](atp2)--(vtp);
\draw[line width=1pt, draw=red](at1) to [bend right=-30] (vt);
\draw[line width=1pt](atp1) to [bend right=-30] (vtp);
\draw[line width=1pt](htm1)--(ht1);
\draw[line width=1pt](ht1)--(htp1);
\draw[line width=1pt](htp1)--(htpp1);
\draw[line width=1pt](ht1)--(at1);
\draw[line width=1pt](htp1)--(atp1);
\draw[line width=1pt](htm2)--(ht2);
\draw[line width=1pt](ht2)--(htp2);
\draw[line width=1pt](htp2)--(htpp2);
\draw[line width=1pt](ht2)--(at2);
\draw[line width=1pt](htp2)--(atp2);
\end{tikzpicture}
\label{fig:model}}
\hskip0.4cm
\subfigure[]{
\begin{tikzpicture}[->, line width=1pt, node distance=1.3cm]
\node[randomv, obs] (vt) {$v_t$};
\node[draw=none, fill=none, left of = vt] (st0) {$s^0$};
\node[rnn, above of = vt] (st1) {$s_t^1$};
\node[rnn, above of = st1] (st2) {$s_t^2$};
\node[draw=none, fill=none, left of = st1] (stm1) {$\ldots$};
\node[draw=none, fill=none, left of = st2] (stm2) {$\ldots$};
\node[randomv, above of = st2] (at1) {$a_t^1$};
\node[randomv, above of = at1] (at2) {$a_t^2$};
\node[randomv, obs, right of = vt] (vtp) {$v_{t+1}$};
\node[rnn, above of = vtp] (stp1) {$s_{t+1}^1$};
\node[rnn, above of = stp1] (stp2) {$s_{t+1}^2$};
\node[randomv, above of = stp2] (atp1) {$a_{t+1}^1$};
\node[randomv, above of = atp1] (atp2) {$a_{t+1}^2$};
\node[draw=none, fill=none, right of = stp1] (stpp1) {$\ldots$};
\node[draw=none, fill=none, right of = stp2] (stpp2) {$\ldots$};
\draw[line width=1pt, draw=black!10](st0)--(st1);
\draw[line width=1pt, draw=black!10](st0)--(stp1);
\draw[line width=1pt, draw=black!30](st1) to [bend left=40] (at1);
\draw[line width=1pt, draw=black!30](st2) to [bend right=40] (at2);
\draw[line width=1pt, draw=black!30](stp1) to [bend left=45] (atp1);
\draw[line width=1pt, draw=black!30](stp2) to [bend right=40] (atp2);
\draw[line width=1pt, draw=black!50](vt) to [bend right=30] (st1);
\draw[line width=1pt, draw=black!50](vt) to [bend right=35] (st2);
\draw[line width=1pt, draw=black!50](vtp) to [bend right=30] (stp1);
\draw[line width=1pt, draw=black!50](vtp) to [bend right=40] (stp2);
\draw[line width=1pt](stm1)--(st1);
\draw[line width=1pt](stm2)--(st2);
\draw[line width=1pt](st1)--(stp1);
\draw[line width=1pt](st2)--(stp2);
\draw[line width=1pt](stp1)--(stpp1);
\draw[line width=1pt](stp2)--(stpp2);
\draw[line width=1pt](st1)--(st2);
\draw[line width=1pt](stp1)--(stp2);
\end{tikzpicture}
\label{fig:model:inference}}}
\caption{(a)
The generative model for an image $v_t$, $p_{\theta}(v_t | a_{t}^{1:N})$, for $N=2$, see \eqref{eq:v}. Random variables are indicated with circles; observed variables are shaded. Diamond nodes indicate recurrent neural network hidden states. 
(b)
The full generative model, where the initial $h_1^n$ depends on $z^n$, see \eqref{eq:mixture}.
Conditioned on $z^n$, the backbone for each object $n$ is modelled by a LGSSM, see \eqref{eq:lds}.
The two red arrows correspond to the conditional density in \figref{fig:model:generative}.
(c)
The inference network for $q_{\phi}(a^{1:N}_{1:T} | v_{1:T})$, given
as a recurrent set of equations in Eqs. (\ref{eq:inference:state-update}) and (\ref{eq:inference-of-a}).}
\label{fig:all-models}
\end{figure}

This rendering process is illustrated in \figref{fig:model:generative}.
The final state $x^N$, which is transformed in \eqref{eq:v} so that
$v_t$ can be sampled, should contain information from all $N$ objects. To achieve that, 
the state is iteratively \emph{updated} through \eqref{eq:rendering} to incorporate the \emph{contribution} from object $n$, $\hat{x}^n$. 
This is obtained through an \emph{attention mask} vector $\alpha^n$ with elements in the interval $[0, 1]$, which specifies what information from object $n$ should be included and
what information from $x^{n-1}$ should be retained.

\subsubsection{Latent dynamics}
\label{sec:latent}
We model the latent positions of each object, $a^n_{1:T}$, using a hidden Markov model with linear Gaussian hidden-state and output, also known as linear Gaussian state-space model (LGSSM) \cite{barber11inferenceA,chiappa06phd}, i.e. 
\begin{align}
h^n_t & = Ah^n_{t-1} + u_t + \eta_t^h,  \hskip0.3cm \eta^h_t \sim {\cal N} (\eta^h_t ; 0, \Sigmah)\,,\nonumber\\
a^n_t & = B h^n_t + \eta^a_t,  \hskip0.3cm 
\eta^a_t \sim {\cal N}(\eta^a_t ; 0, \Sigma_A) \,,
\label{eq:lgssm}
\end{align}
where ${\cal N}(x ; \mu, \Sigma)$ denotes the density of a Gaussian random variable $x$ with mean $\mu$ and covariance $\Sigma$. We use the constraints $A=[I,\ \ \delta I;\ \ 0, \ \ I] \in \mathbb{R}^{4 \times 4}$ (where $[\cdot, \cdot; \cdot, \cdot]$ indicates  horizontal and vertical matrix concatenation, $I \in \mathbb{R}^{2 \times 2}$ denotes the identity matrix, and $\delta$ denotes the sampling period) and $B = [I, \ \ 0] \in \mathbb{R}^{2 \times 4}$ to obtain a description of Newtonian laws, such that the vector $h_t$ represents positions and velocities, and $u_t$ the force (which is assumed to be a constant $u$ over time). We include $\delta$, $u$, $\Sigmah$ and $\Sigma_A$ in the generative model parameters $\theta$.

Objects may start moving
from disjoint sets of initial positions and velocities; for instance
from the left, the right but not the center of $v_1$.
To allow the model to \emph{not} put probability mass on initial positions and velocities that might never be helpful in explaining $a_{1:T}^n$,
the initial positions and velocities $h^n_1$ are drawn from a $K$ component Gaussian mixture, i.e.
\begin{align} 
h^n_1 \sim p_{\theta}(h^n_1)
 = \sum_{k=1}^K p_{\theta}(z^n=k) \, p_{\theta}(h^n_1 | z^n=k) 
 = \sum_{k=1}^K \pi_k \, \calN (h^n_1 ; \mu_k, \Sigma_k) \,. 
\label{eq:mixture}
\end{align}
We additionally include $\pi_{1:K}$, $\mu_{1:K}$ and $\Sigma_{1:K}$ in the generative model parameters $\theta$. 
The joint density of all random variables factorizes as
\begin{align} 
p_{\theta}(a^{1:N}_{1:T}, h^{1:N}_{1:T}, z^{1:N})=
\prod_{n=1}^N  \bigg\{
\prod_{t=1}^T p_{\theta}(a^n_t | h^n_t) \bigg\}p_{\theta}(h^n_1 | z^n) \, p_{\theta}(z^n)
\prod_{t=2}^T p_{\theta}(h^n_t | h^n_{t-1}) \,,
\label{eq:lds}
\end{align}
where $p_{\theta}(a^n_t | h^n_t) = \calN (a^n_t ; Bh^n_t, \Sigma_A)$ and $p_{\theta}(h^n_t | h^n_{t-1}) = {\cal N} (h^n_t ; Ah^n_{t-1} + u, \SigmaH)$. 
The full generative model combines \eqref{eq:v} with \eqref{eq:lds} to yield
\begin{equation*}
p_{\theta}(v_{1:T}, a^{1:N}_{1:T}, h^{1:N}_{1:T}, z^{1:N})
= \bigg\{\prod_{t=1}^T p_{\theta}(v_t | a^{1:N}_{t}) \bigg\} p_{\theta}(a^{1:N}_{1:T}, h^{1:N}_{1:T} , z^{1:N})\,.
\end{equation*}
The backbone of the model is illustrated in \figref{fig:model}.

The advantage of this formulation for the latent dynamics is that quantities such as the smoothed distribution $p_{\theta}(h^n_t | a^n_{1:T}, z^n)$, the likelihood $p_{\theta}(a^n_{1:T} | z^n)$, or the most likely mixture component
$\argmax_k p_{\theta}(z^n=k | a^n_{1:T})$, can be computed exactly in ${\cal O}(T)$ operations, using 
message passing algorithms such as the Kalman filtering and Rauch-Tung-Striebel smoothing \cite{barber11inferenceA,chiappa06phd}. 

\subsection{Inference and Learning} \label{sec:IL}
The non-linearity of the rendering process makes the computation of $p_{\theta}(v_{1:T})$, 
$p_{\theta}(a^{1:N}_{1:T}, h^{1:N}_{1:T}, z^{1:N} | v_{1:T})$, and 
of quantities like $p_{\theta}(a^n_t | v_{1:T})$ needed for estimating the positions of object $n$,
intractable. We address this problem using a recent approach to variational methods known as variational auto-encoding (VAE) \cite{kingma14autoencoding,rezende14stochastic}.

The basic principle of variational methods is to introduce a tractable approximating
distribution\footnote{Whilst in practice we need to consider all observed sequences in the KL, to simplify the notation we focus the exposition on one sequence only.} $q_{\phi}(a^{1:N}_{1:T}, h^{1:N}_{1:T}, z^{1:N} | v_{1:T})$ to the intractable distribution
$p_{\theta}(a^{1:N}_{1:T}, h^{1:N}_{1:T}, z^{1:N} | v_{1:T})$
via the Kullback-Leibler divergence 
\begin{align*}
\textrm{KL}\Big(q_{\phi}(a^{1:N}_{1:T}, h^{1:N}_{1:T}, z^{1:N} | v_{1:T})\, \Big\| \, p_{\theta}(a^{1:N}_{1:T}, h^{1:N}_{1:T}, z^{1:N} | v_{1:T})\Big)\,.
\end{align*}
Given that 
\begin{align*}
\textrm{KL}(q(a,h,z|v)\, \big\| \, p(a,h,z|v))&=\biggl<\log \frac{q(a,h,z|v)}{p(a,h,z|v)}\biggr>_{q(a,h,z|v)}\\
&\hskip-1.7cm=\bigl<\log q(a,h,z|v)-\log p(a,h,z,v)\bigr>_{q(a,h,z|v)} +\log p(v)\geq 0\,,
\end{align*}
where we omitted super and subscript indices and used the notation $\bigl<\cdot\bigr>_{q(\cdot)}$ to indicate averaging wrt $q(\cdot)$, we obtain a lower bound ${\cal F}_{\theta,\phi}$ on $\log p_{\theta}(v)$, i.e. $\log p_{\theta}(v)\geq {\cal F}_{\theta,\phi}$ with
\begin{align*}
{\cal F}_{\theta,\phi} =-\bigl<\log q(a,h,z|v)\bigr>_{q(a,h,z|v)}+\bigl<\log p(a,h,z,v)\bigr>_{q(a,h,z|v)}\,.
\end{align*}
If ${\cal F}_{\theta,\phi}$ were tractable and we were able to perform marginalization on $q_{\phi}(a,h,z|v)$, we could find the optimal $q_{\phi}(a,h,z|v)$ (parameters $\phi$) and $\theta$ by maximizing the bound; see \cite{chiappa08bayesian,chiappa14explicit} for traditional approaches to variational methods in the temporal setting. However, this is not the case for our generative model choice, and thus we instead use the more recent VAE approach to variational methods, where a Monte-Carlo approximation of the intractable ${\cal F}_{\theta,\phi}$ is deployed.

The VAE approach consists in rewriting the bound in the form ${\cal F}_{\theta,\phi}=\bigl<f_{\theta,\phi}(\epsilon_{1:T})\bigr>_{q_{\epsilon}(\epsilon_{1:T})}$ for a parameter free distribution $q_{\epsilon}(\epsilon_{1:T})$, such that the gradient of ${\cal F_{\theta,\phi}}$ with respect to $\phi$ is given by $\nabla_{\phi}{\cal F_{\theta,\phi}}=\bigl<\nabla_{\phi} f_{\theta,\phi}(\epsilon_{1:T})\bigr>_{q_{\epsilon}(\epsilon_{1:T})}$---this is often called reparemetrization trick. We can then approximate the gradient with the Monte-Carlo estimate
\begin{equation} \label{eq:reparam-monte-carlo}
\bigl<\nabla_{\phi} f_{\theta,\phi}(\epsilon_{1:T})\bigr>_{q_{\epsilon}(\epsilon_{1:T})} \approx  
\hskip0.0cm \frac{1}{M}\sum_{m=1}^M \nabla_{\phi} f_{\theta,\phi}(\epsilon^m_{1:T}), \hskip0.4cm \epsilon^m_{1:T}\sim q_{\epsilon}(\epsilon^m_{1:T})\,.
\end{equation}
In our case, the formulation of the latent dynamics described above enables us to avoid employing a full approximation of $p_{\theta}(a^{1:N}_{1:T}, h^{1:N}_{1:T}, z^{1:N} | v_{1:T})$, and instead to decompose this distribution as a product of the exact tractable distribution $p_{\theta}(h^{1:N}_{1:T}, z^{1:N} | a^{1:N}_{1:T})$ and an approximation $q_{\phi}(a^{1:N}_{1:T} | v_{1:T})$ of $p_{\theta}(a^{1:N}_{1:T} | v_{1:T})$, i.e. 
\begin{align*}
p_{\theta}(a^{1:N}_{1:T}, h^{1:N}_{1:T}, z^{1:N} | v_{1:T})
& = \underbrace{p_{\theta}(h^{1:N}_{1:T}, z^{1:N} | a^{1:N}_{1:T}, \cancel{v_{1:T}}) }_{= \prod_{n=1}^N p_{\theta}(h^n_{1:T}, z^{n} | a^n_{1:T}) }
p_{\theta}(a^{1:N}_{1:T} | v_{1:T}) \\
&\approx \bigg\{\prod_{n=1}^N p_{\theta}(h^n_{1:T}, z^{n} | a^n_{1:T})
\bigg\} q_{\phi}(a^{1:N}_{1:T} | v_{1:T}) \,.
\end{align*}
Thanks to this representation, the bound can be expressed as 
\begin{align*}
{\cal F}_{\theta,\phi} & = \bigl<- \log [q(a|v)\underbrace{q(h,z|a,\cancel{v})}_{p(h,z|a)}] +\log [p(v|a,\cancel{h,z})p(a,h,z)]\bigr>_{q(a| v) q(h,z|a,\cancel{ v})}\\
& = -\biggl<\log \frac{q(a| v)}{p( v|a)}\biggr>_{q(a| v)}+\Bigl<\underbrace{-\log p(h,z|a)+\log p(a,h,z)}_{\log p(a)}\Bigr>_{q(a|
v)p(h,z|a)}\\
& = -\biggl<\log \frac{q(a| v)}{p( v|a)}\biggr>_{q(a| v)}+\bigl<\log p(a)\bigr>_{q(a| v)}\,.
\end{align*}
This gives 
\begin{equation} \label{eq:elbo}
{\cal F}_{\theta,\phi} =
\bigl<
\log p_{\theta}(v_{1:T} | a^{1:N}_{1:T})\bigr>_{q_{\phi}(a^{1:N}_{1:T} | v_{1:T})}
- \textrm{KL} \Big( q_{\phi}(a^{1:N}_{1:T} | v_{1:T})
\, \Big\| \, p_{\theta}(a^{1:N}_{1:T}) \Big)\,.
\end{equation}
We model $q_{\phi}(a^{1:N}_{1:T} | v_{1:T})$
using a recurrent neural network with states $s^{1:N}_{1:T}$ as in \figref{fig:model:inference}.
More specifically we assume $a^n_t \sim \calN(a^n_t \, ; \, \mu_\phi(s^n_t), \sigma^2_{\phi}(s^n_t))$ and use the reparametrization $a^n_t=\mu_{\phi}(s^n_t)+\sigma_{\phi}(s^n_t)\circ\epsilon^n_t$, under the assumption  $q_{\epsilon}(\epsilon^{1:N}_{1:T})=\prod_{n,t} q_{\epsilon}(\epsilon^n_t)$ with $\epsilon^n_t \sim {\cal N}(\epsilon^n_t;0,I)$. We describe this inference network in more details in the next section.

\subsubsection{Inference network}
\label{sec:IN}
As shown in \figref{fig:model:inference}, the inference network iterates a latent state vector $s_t^n$ over $t=1,\ldots,T$ and $n=1,\ldots,N$.
Starting with $s^0 = s_t^0 = 0$ at time-step $t$, we recurrently iterate
\begin{align}
\beta^n_t & = \sigmoid(W^\beta [s_{t-1}^n, s_t^{n-1}, v_t] + b^\beta) & & \textrm{(latent attention mask)} \nonumber \\
\hat{s}^n_t & = 
\tanh(W^s [s_{t-1}^n, s_t^{n-1}, v_t] + b^s) & & 
\textrm{(object $n$'s state contribution)} \nonumber \\
s^n_t & = (1 - \beta^n_t) \circ s^{n-1}_t + \beta^n_t \circ  \hat{s}^n_t & &
\textrm{(state update)}\,, \label{eq:inference:state-update}
\end{align}
to compute a vector $s_t^{1:N}$ for each of the objects
at time step $t$.
Similar to the generative model in Sec.~\ref{sec:rendering-objects},
there is an \emph{attention mask} $\beta_t^n$
that specifies how much of each component of $s_t^{n-1}$
we should keep. The mask is a function of
the visible image $v_t$, as well as the recurrently computed values
for both the previous object at this time-step, and this object and the previous time-step.
There is also a \emph{contribution} $\hat{s}^n_t$
coming from image $v_t$ for object $n$.
The combination of $s^{n-1}_t$ and $\hat{s}^n_t$ is used to 
\emph{update} $s^{n}_t$. Samples from $q_{\phi}(a^{1:N}_{1:T} | v_{1:T})$ are generated as
\begin{equation} \label{eq:inference-of-a}
a^n_t = \mu_\phi(s^n_t) + \sigma_{\phi}(s^n_t) \circ \epsilon_t^n\,,
\end{equation}
where $\epsilon_t^n \sim \calN(\epsilon_t^n ; 0, I)$.
In this computation, external Gaussian noise is inserted in a computation graph, and transformed---this ensures that the Monte-Carlo estimate in \eqref{eq:reparam-monte-carlo} is fully differentiable.

The $W$'s and $b$'s denote weight matrices and biases, and are included in the inference network parameters $\phi$.
The recurrent process at time-step $t=1$ depends on $s_0^n = \tanh(\phi_{s_0^n})$,
and an initial state for each object $n$ is learned through parameters $\phi_{s_0^n}$\footnote{In practice,
as the state $s_0^n$ encodes which way we can interrogate $v_1$ to infer $a_1^n$, we have obtained better results by learning separate $\phi_{s_0^n}$ that depend on the number of objects $N$ in the image.}.

Note that $v_t$ appears in Eqs. (\ref{eq:inference:state-update}) and (\ref{eq:inference-of-a}) as input to
every step $n$ at time-step $t$. This is important: To infer the
position $a_t^n$, we need to consider the latent representation $s_{t-1}^n$ of object $n$ (actually 1 to $n$) in the previous image, as well as
$s_{t}^{n-1}$, which contains a rolled-up representation of
objects 1 to $n-1$ in this image. Both these representations need to act on $v_t$ to infer $a_t^n$.

\subsubsection{Learning}
\label{sec:learning}
In the Kullback-Leibler divergence term
in \eqref{eq:elbo}, $\bigl<\log q_{\phi}(a^{1:N}_{1:T} | | v_{1:T})\bigr>_{q}$ can be expressed in analytic form.
Both the first term in \eqref{eq:elbo} and $\bigl<\log p_{\theta}(a^{1:N}_{1:T})\bigr>_{q}$ in the KL divergence term
can be stochastically estimated using a sample
$a^{1:N}_{1:T} \sim q_{\phi}(a^{1:N}_{1:T} | v_{1:T})$. For such a sample,
\begin{equation} \label{eq:kalman-filter}
\log p_{\theta}(a^{1:N}_{1:T})
= \sum_{n=1}^N \log \Bigg( \sum_{k=1}^K p_{\theta} (z^n = k) 
p_{\theta} (a^{n}_{1:T} | z^n = k)
\Bigg) .
\end{equation}
To compute \eqref{eq:kalman-filter}, we take $a^{n}_{1:T}$ as
``observations'', and for each mixture component $z^n = k$ we run a Kalman filter to obtain the log-likelihood $\log p_{\theta} (a^{n}_{1:T} | z^n = k)$.
The objective function is the sum of the bounds in \eqref{eq:elbo}
over all sequences in the training dataset. The negative of this objective is minimized via a stochastic gradient descent algorithm, using mini-batches from the dataset.

\subsection{Limitations}  
One of the main limitations of our approach is that, as 
$q_{\phi}(a^{1:N}_{1:t-1} | v_{1:t-1})$ does not explicitly incorporate the LGSSM dynamics, 
the objective function can have many sub-optimal local maxima. 
As common in the VAE literature, we address such a decoupling between the generative model and variational
distribution by annealing the KL term in the bound ${\cal F}_{\theta,\phi}$ to ensure
that the dynamics are correctly accounted for during training. 
More advanced methods in the literature consist in incorporating the unknown latent dynamics into
$q_{\phi}(a^{1:N}_{1:t-1} | v_{1:t-1})$ by essentially letting
the dynamics be a ``regularizer'' to the parameters $\phi$.
``Structured inference networks''
provide a framework for achieving this \cite{lin18variational}.
In \cite{pearce2018comparing} we show that this approach leads to better inference and more stable results than annealing for the case of learning the dynamics of one object from pixels---it not obvious how this can be extended to the multiple-object scenario.

A more general limitation of our approach is its applicability to images that contain arbitrary backgrounds and objects, and nonlinear interactions between objects.

\section{Results}
\label{sec:results}
In this section, we evaluate our approach on inferring latent positions, and on image generation and interpolation using artificially generated images describing the movement of cannonballs, see the examples shown in \figref{fig:examples}.

\subsection{Dataset}
We generated image sequences of length $T=30$ describing the movement of up to three cannonballs.
Noisy positions $a_{1:T}^n$ were generated with an LGSSM formulation of Newtonian laws, as described in Sec. \ref{sec:latent}, with
sampling period $\delta=0.015$; force $u  = -g (0 \ \ 0.5\delta^2 \ \ 0 \ \ \delta)$, where $g=9.81$ is the gravitational constant; $\Sigma_H=0$; and $\Sigma_A=0.001I$.

\begin{figure}[t]
\begin{center}
\includegraphics[height=2.5cm,width=2.5cm]{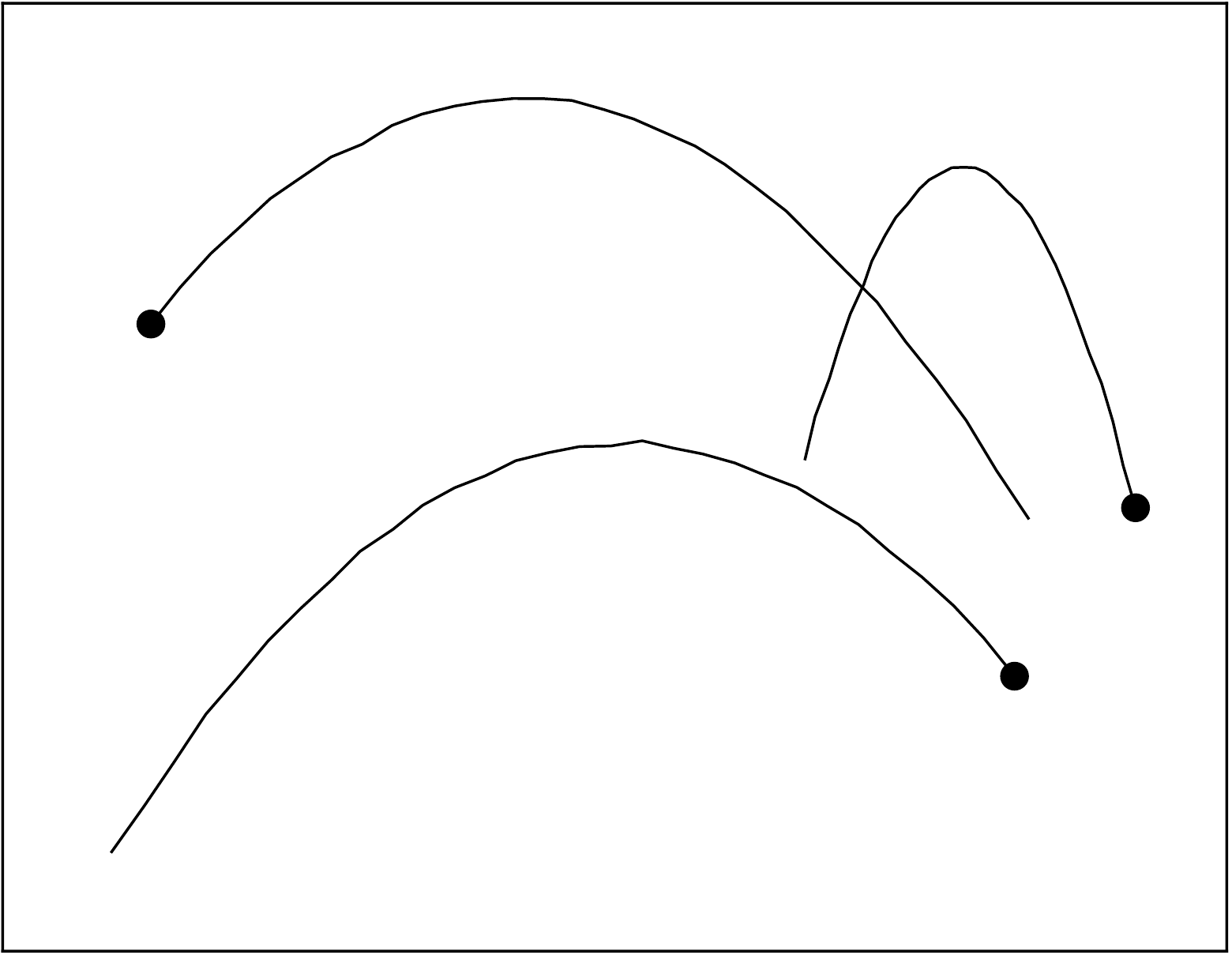}
\hskip0.05cm
\includegraphics[height=2.5cm,width=2.5cm]{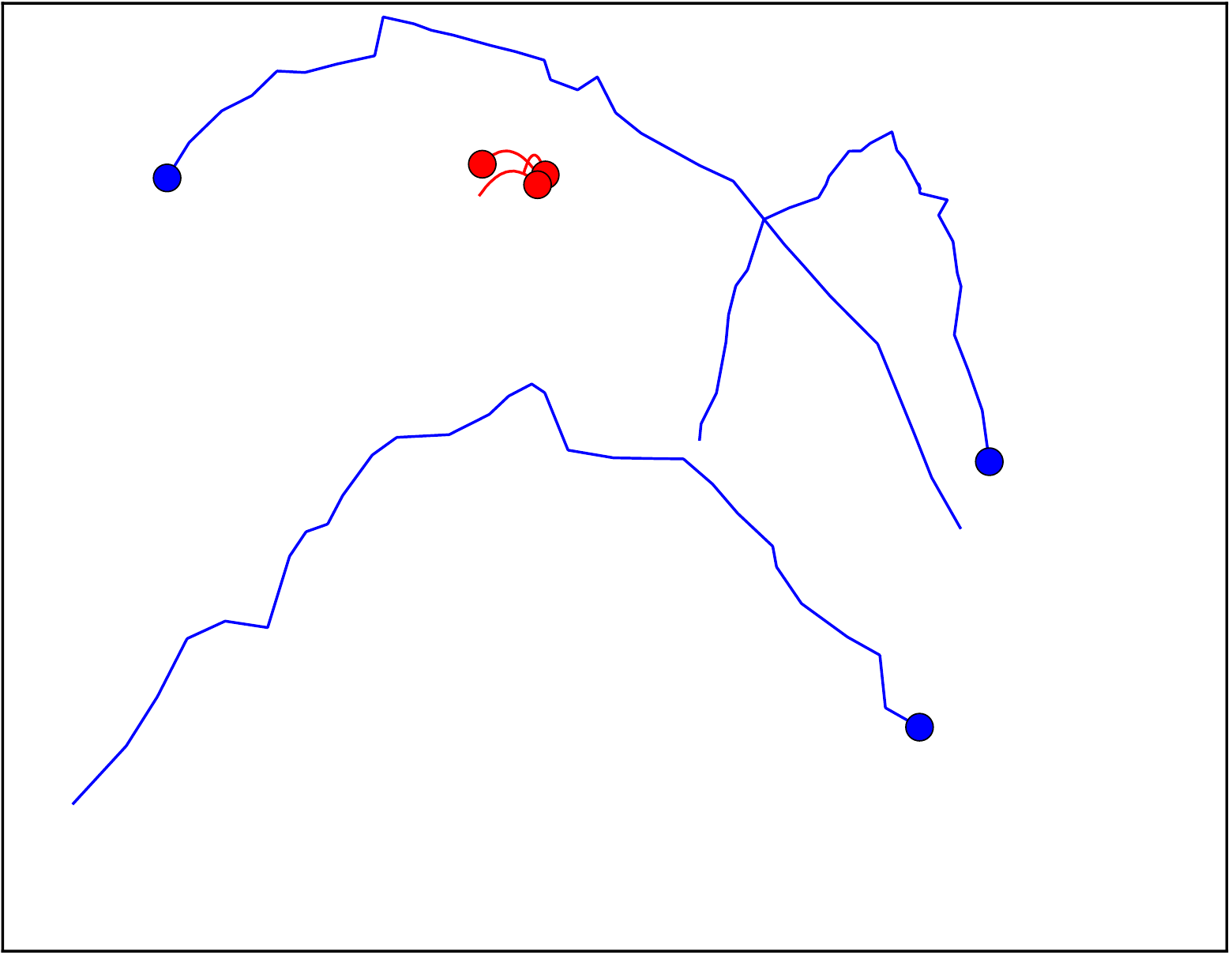}
\hskip0.2cm
\includegraphics[height=2.5cm,width=2.5cm]{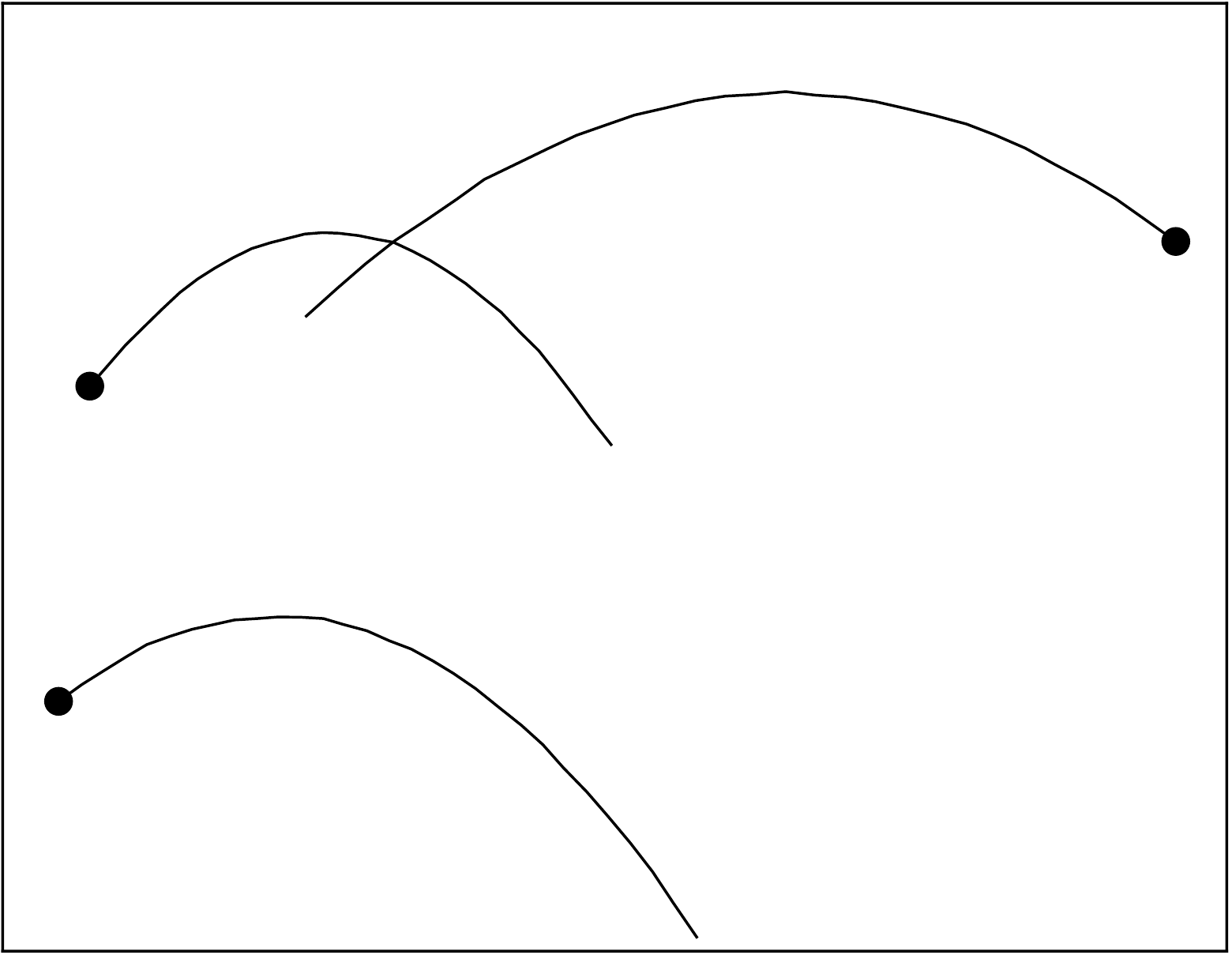}
\hskip0.05cm
\includegraphics[height=2.5cm,width=2.5cm]{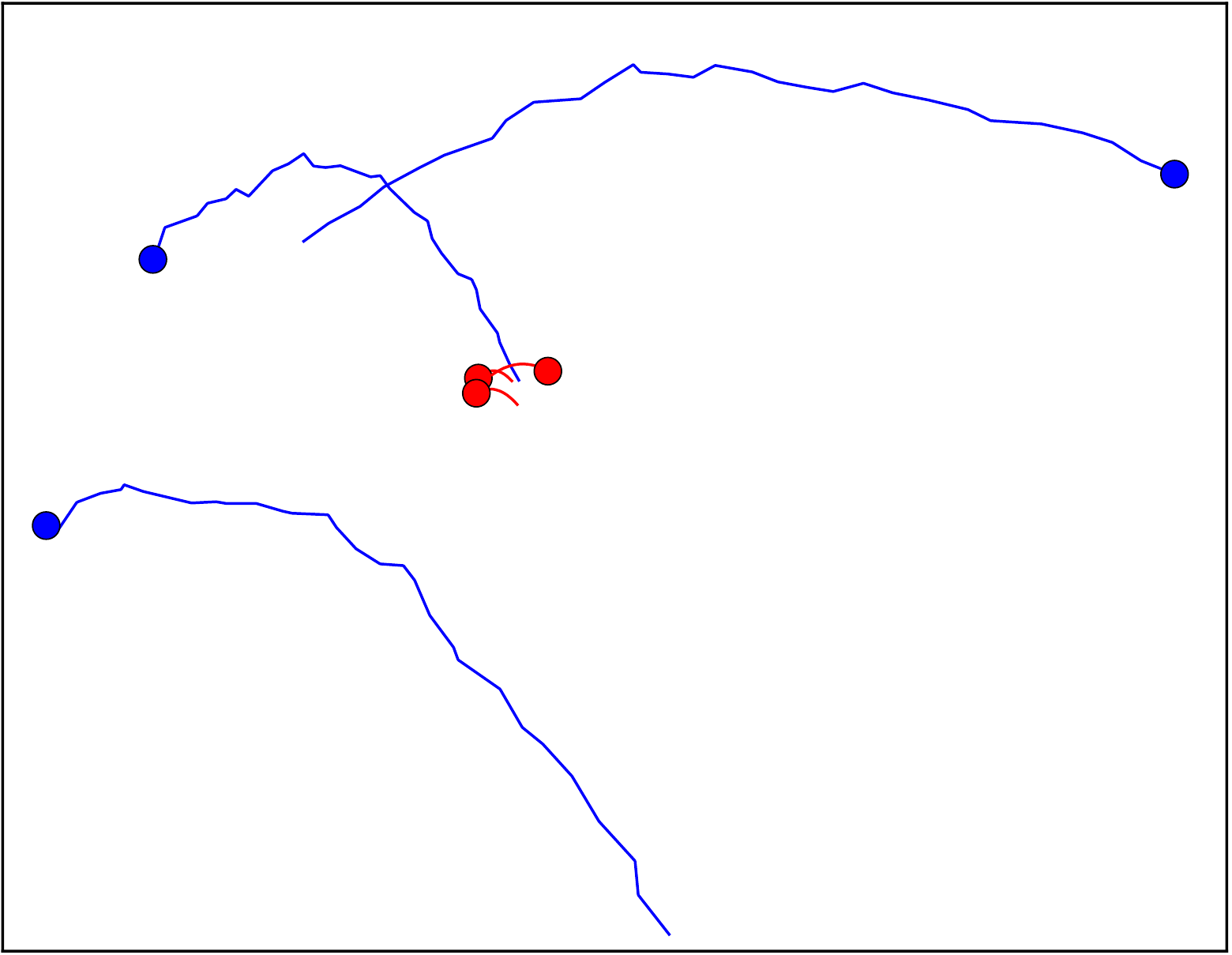}\\[8pt]
\includegraphics[height=2.5cm,width=2.5cm]{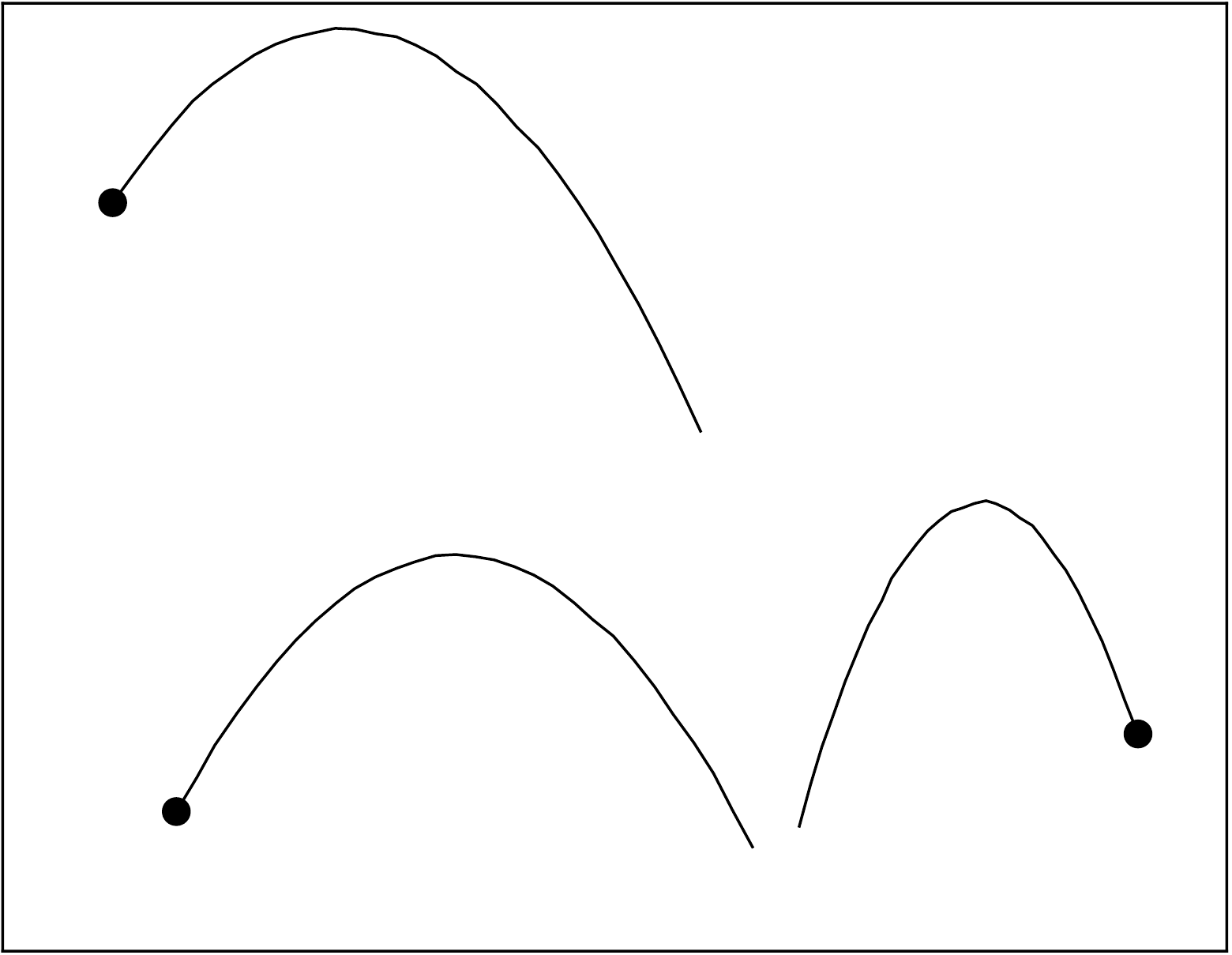}
\hskip0.05cm
\includegraphics[height=2.5cm,width=2.5cm]{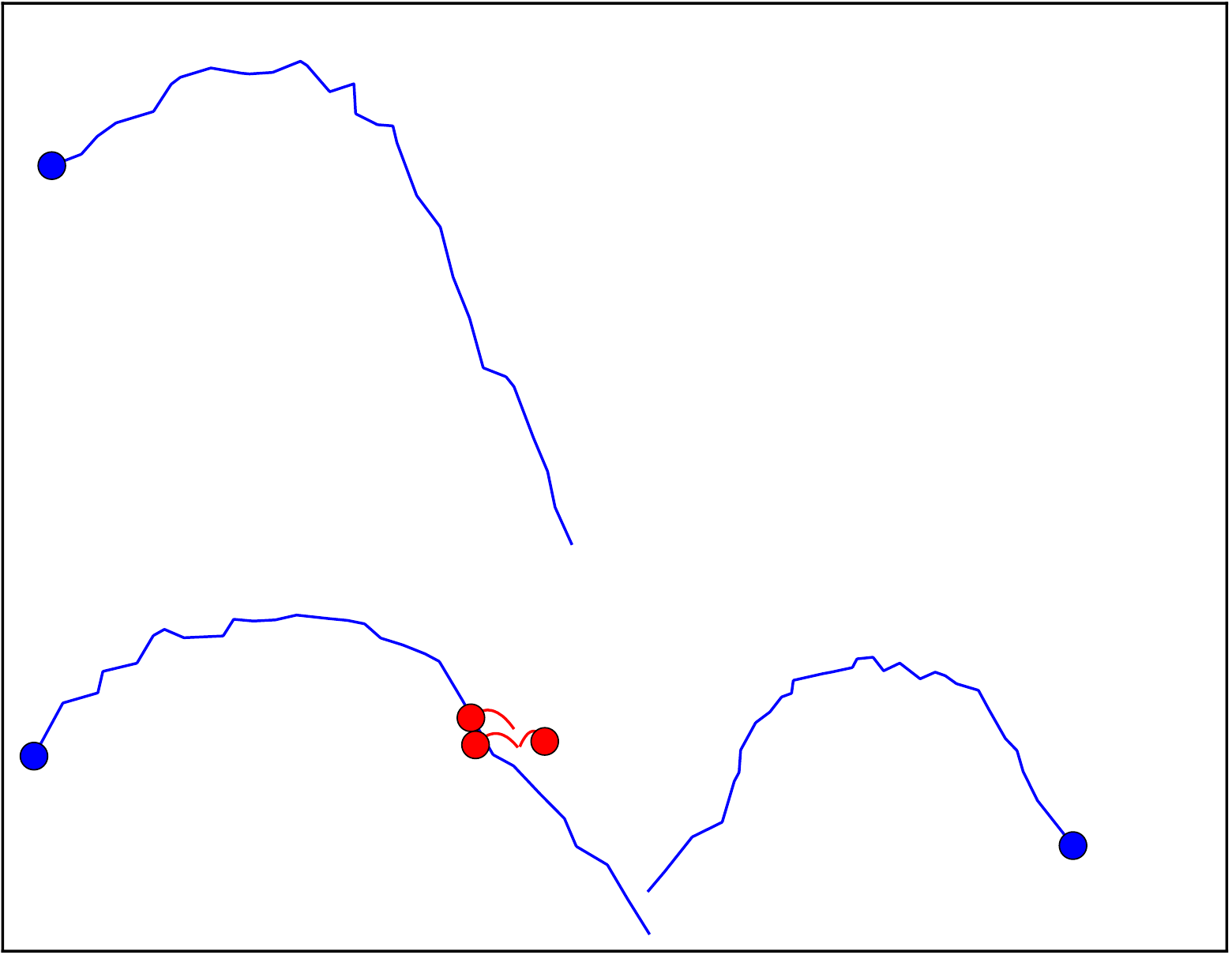}
\hskip0.2cm
\includegraphics[height=2.5cm,width=2.5cm]{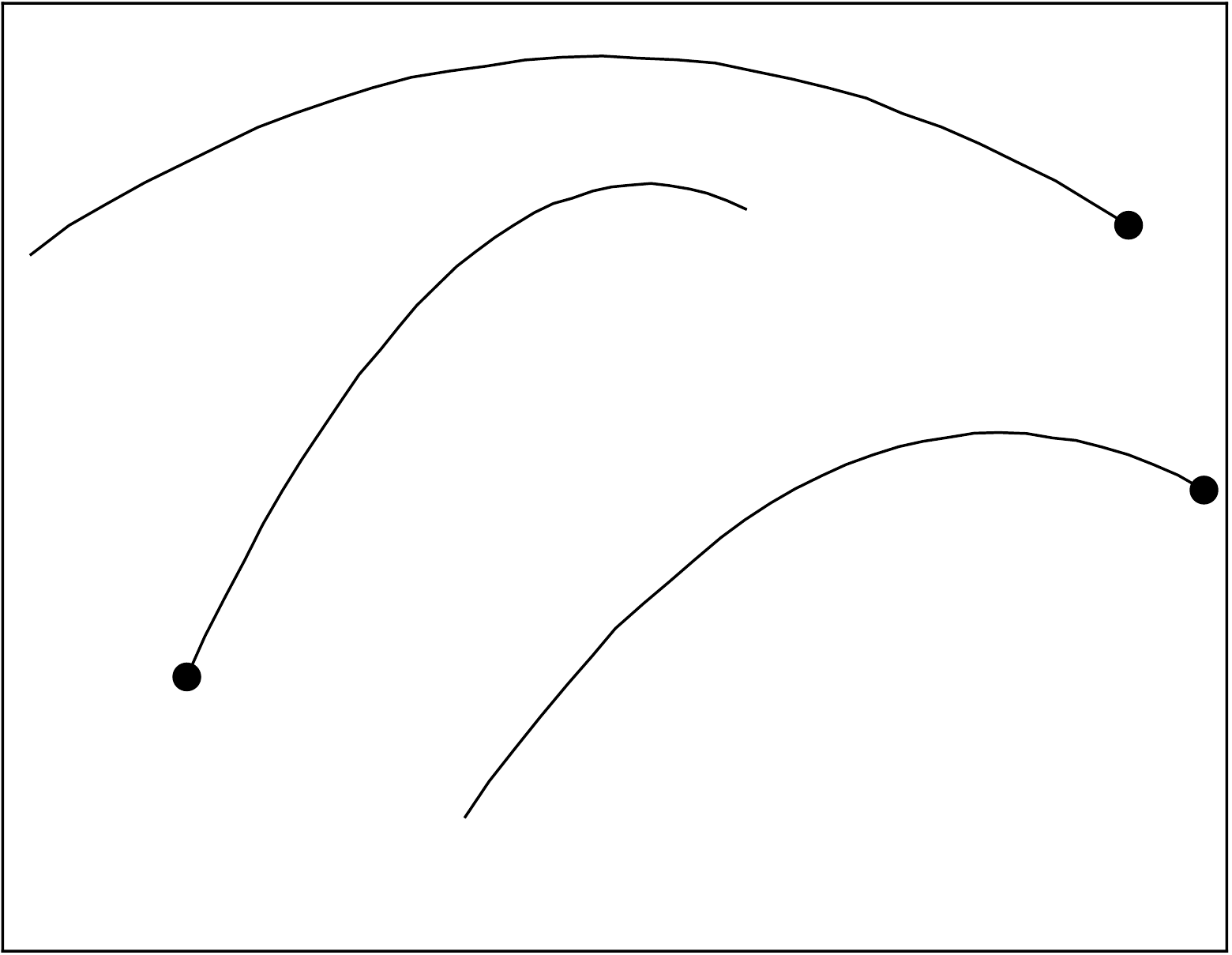}
\hskip0.05cm
\includegraphics[height=2.5cm,width=2.5cm]{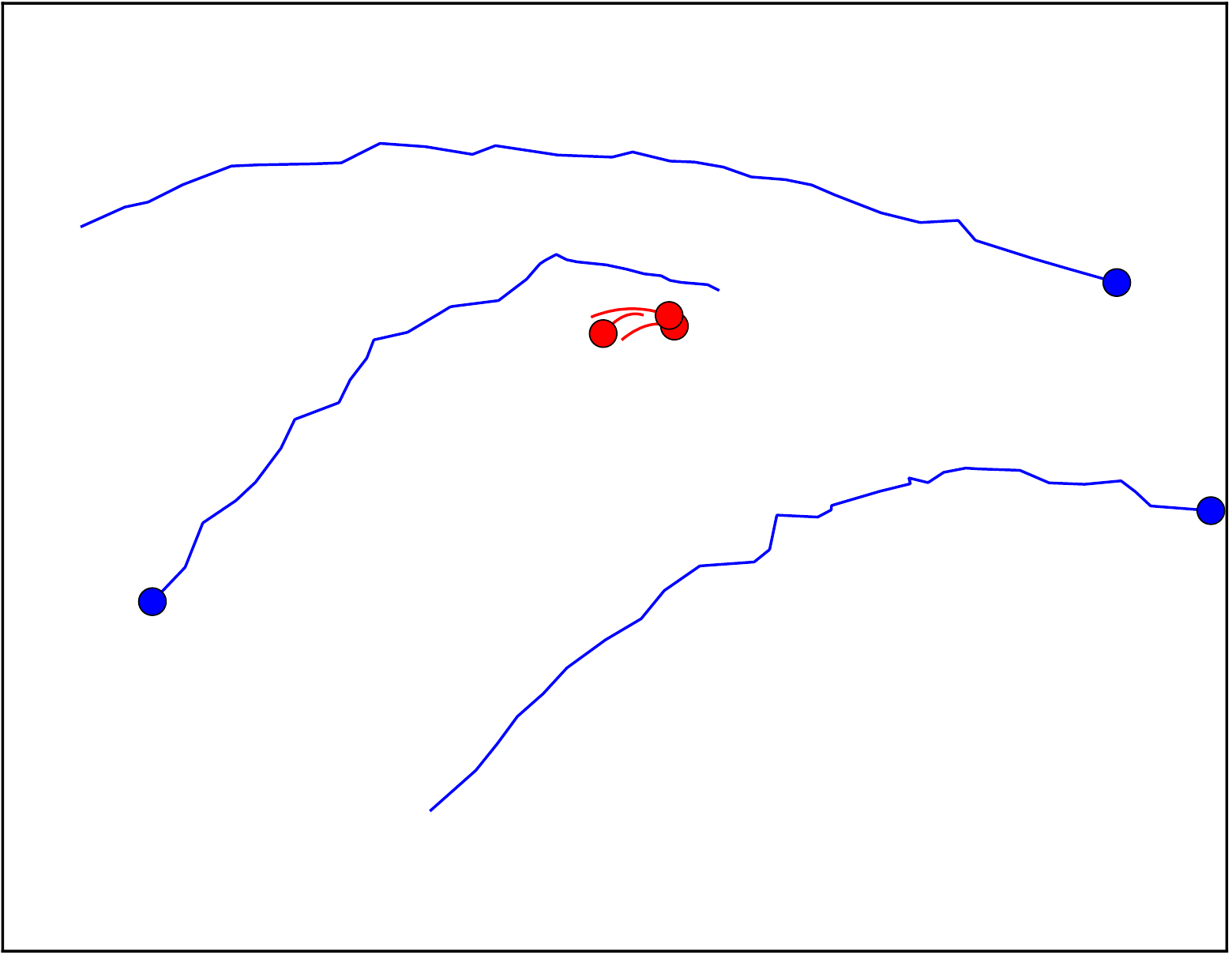}
\end{center}
\caption{Ground-truth (black; left) and inferred (blue; right) trajectories from image sequences containing three balls.
Initial positions are indicated with a circle.
The ground-truth trajectories are also showed in red in each of the right figures,
illustrating that the model can learn any arbitrary scaling of the dynamics, as long as it adequately explains the observed images.}
\label{fig:InferPos}
\end{figure}

Each ball was shot with random shooting angle $\gamma$ in the interval $[40^\circ, 60^\circ]$, from either the left side of the $x$-axis in the interval $[-0.5, -0.1]$ or the right side of the $x$-axis in the interval $[-0.5, -0.1] + 0.9\max_x$, where $\max_x$ indicates the maximum possible displacement at $T=30$ when starting in $[-0.5, -0.1]$.
The initial position on the $y$-axis was sampled in the interval $[-0.5, 0.5]$. The initial velocity $i_v$ was sampled in the interval $[2, 3]$. The resulting velocity on the $x$-axis, $i_v \cos(\gamma \pi / 180.0)$, was flipped in sign if the ball was shot from the right side of the image. Some examples of trajectories are shown in black in \figref{fig:InferPos},
with circles indicating initial positions.

To render the positions into white patches of radius $R=2$ in the image,
the generated positions $a_{1:T}^{1:N}$ were re-scaled to
lie in the interval $[R, H - 1 - R]\times [R, W - 1 - R]$,
where $H=48, W=48$ indicate the height and width of the image.
This re-scaling ensured that each ball was always fully contained in the image. 
We also experimented with similar datasets with $T=50$ and $H=W=32$, obtaining similar results.
With $H=W=32$ the problem is easier in terms of dimensionality, but the latent positions are less identifiable, as close positions in the latent space might induce the same position in the image.

\subsection{Initialization and training}
The dataset consists of sequences with $N \in \{1,2,3\}$ balls. Importantly, as the networks in \figref{fig:all-models} can dynamically unroll, the model was trained on 
all such sequences jointly.
We used $N$ to inform the networks of
how many steps to unroll for each image sequence.

As the initial cannonballs are roughly separated into two main clusters,
we assumed two mixture components, i.e. $K=2$. 
Although a higher $K$ would induce a more refined grouping of initial positions and velocities, our experiments indicate that $K=2$ was sufficient to obtain accurate results. 

\begin{wrapfigure}[11]{r}{0.36\textwidth}
\vskip-0.5cm
\hskip-0.2cm
\centering
\includegraphics[width=0.36\textwidth]{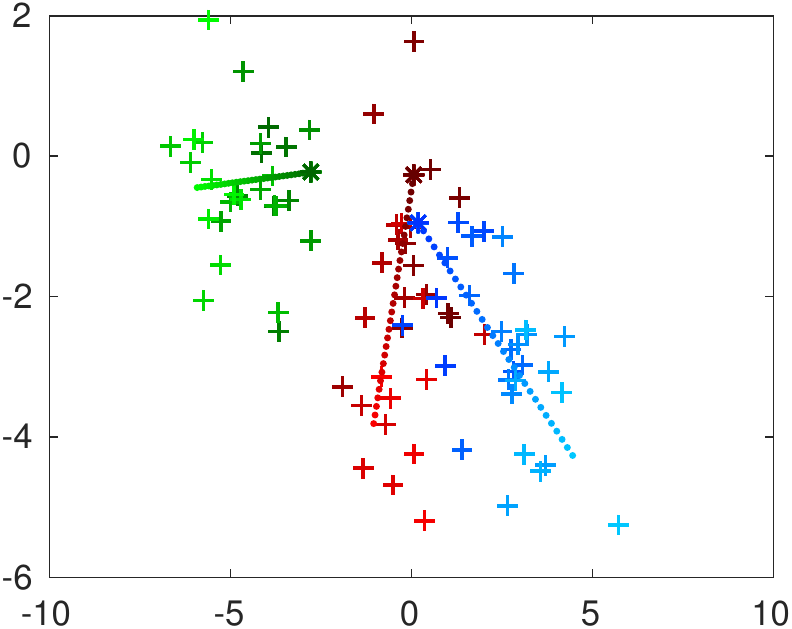}
\caption{Example of initialization.}
\label{fig:Initialization}
\end{wrapfigure}

We experimented with different types of initialization for the LGSSM.
Good results were obtained 
as long as smoothness in the dynamics was enforced so that we could guide the inference network toward smooth trajectories (as explained below). One example of such an initialization is to set $A$ and $B$ as above with $\delta=0.1$, $u=0$, $\Sigma_H=0.001I$, and $\Sigma_A=I$; sample the part of $\mu_k$ corresponding to positions from a standard Gaussian distribution and set the part corresponding to the velocities to zero; and set $\Sigma_k=I$: This ensures symmetry breaking without imposing any meaningful prior on the clusters.

\begin{figure}[t]
\subfigure[]{
\centering
\hskip-0.1cm
\includegraphics[height=2.25cm,width=2.25cm]{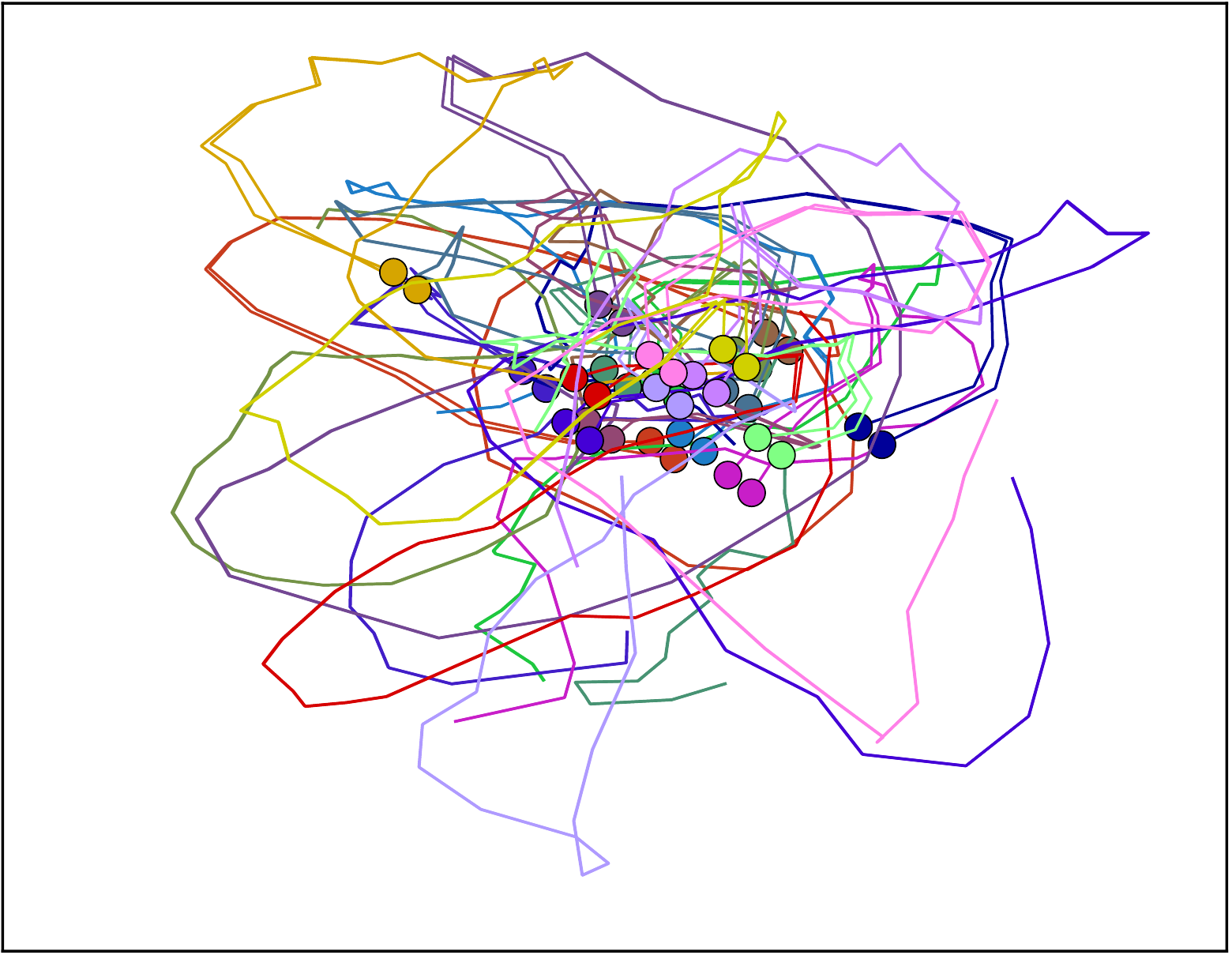}
\includegraphics[height=2.25cm,width=2.25cm]{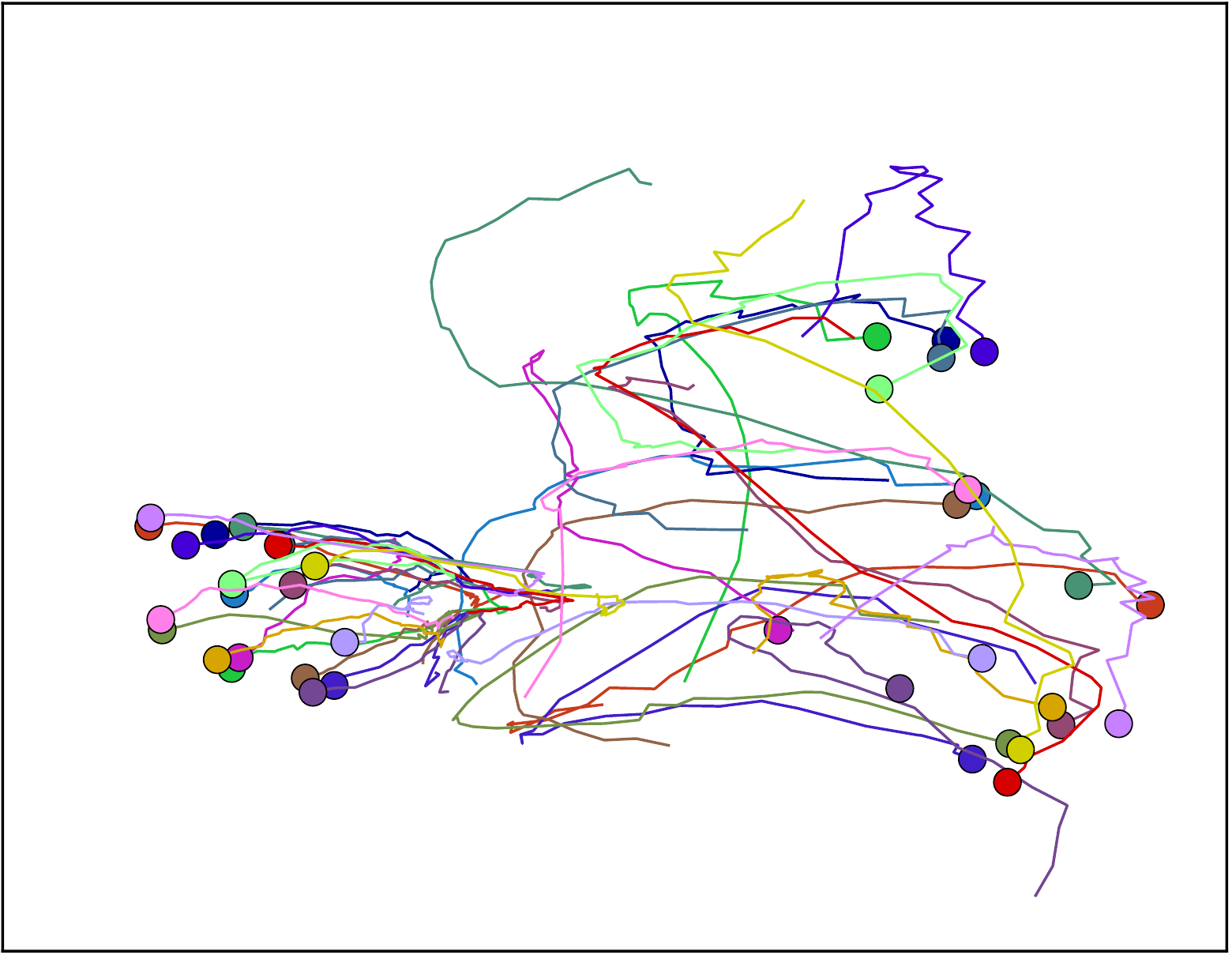}
\includegraphics[height=2.25cm,width=2.25cm]{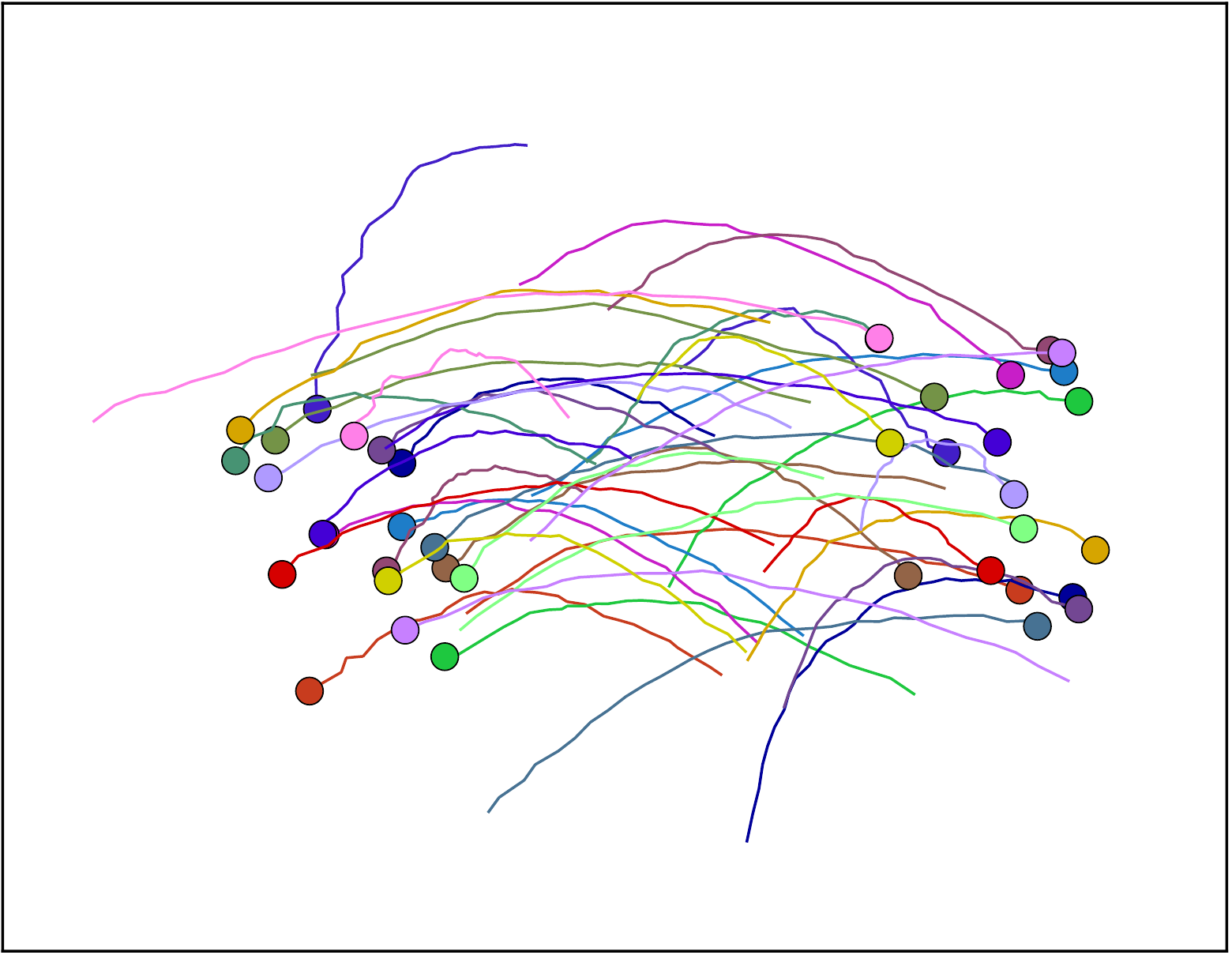}}
\hskip0.1cm
\subfigure[]{
\includegraphics[height=2.25cm,width=2.25cm]{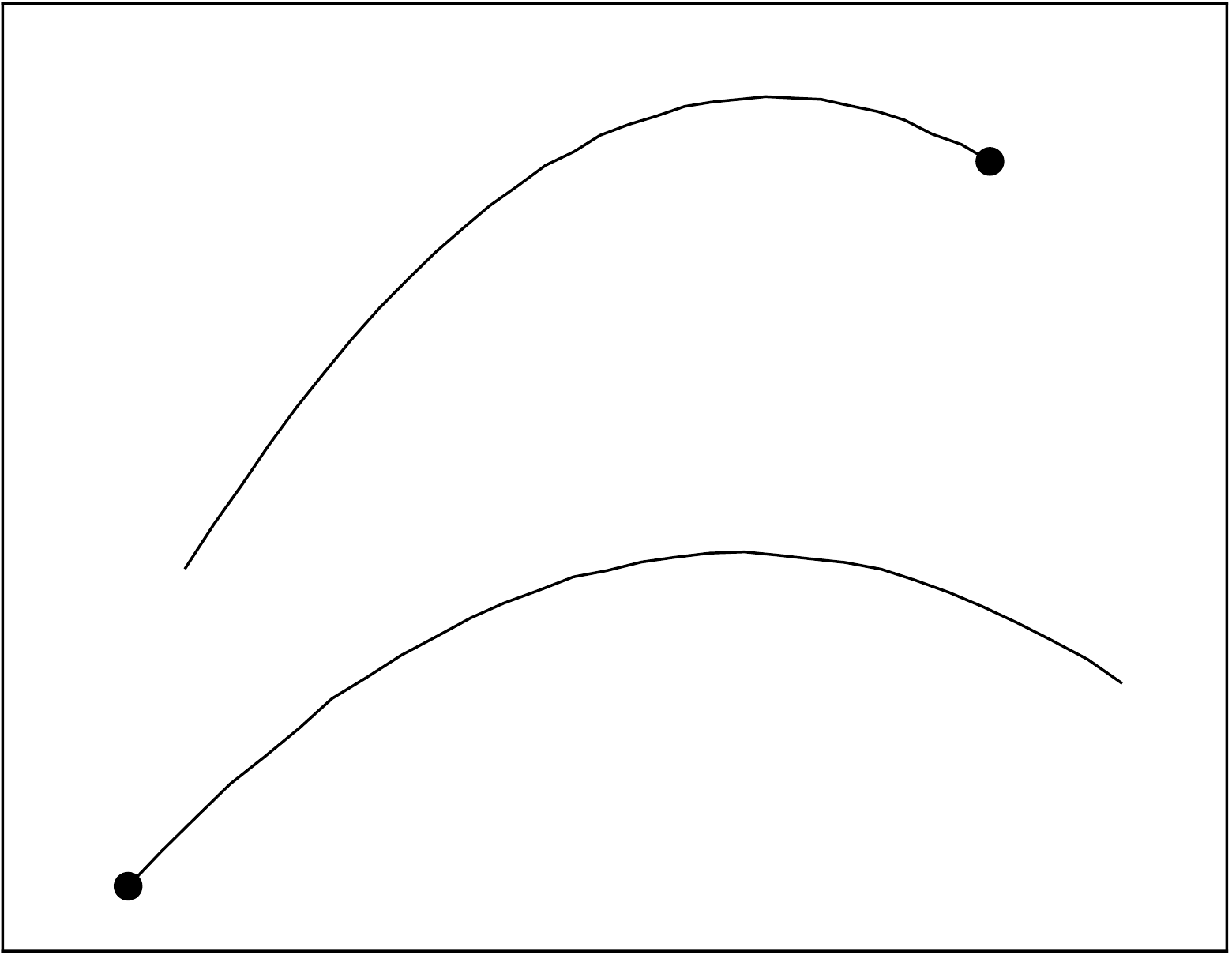}
\hskip0.01cm
\includegraphics[height=2.25cm,width=2.25cm]{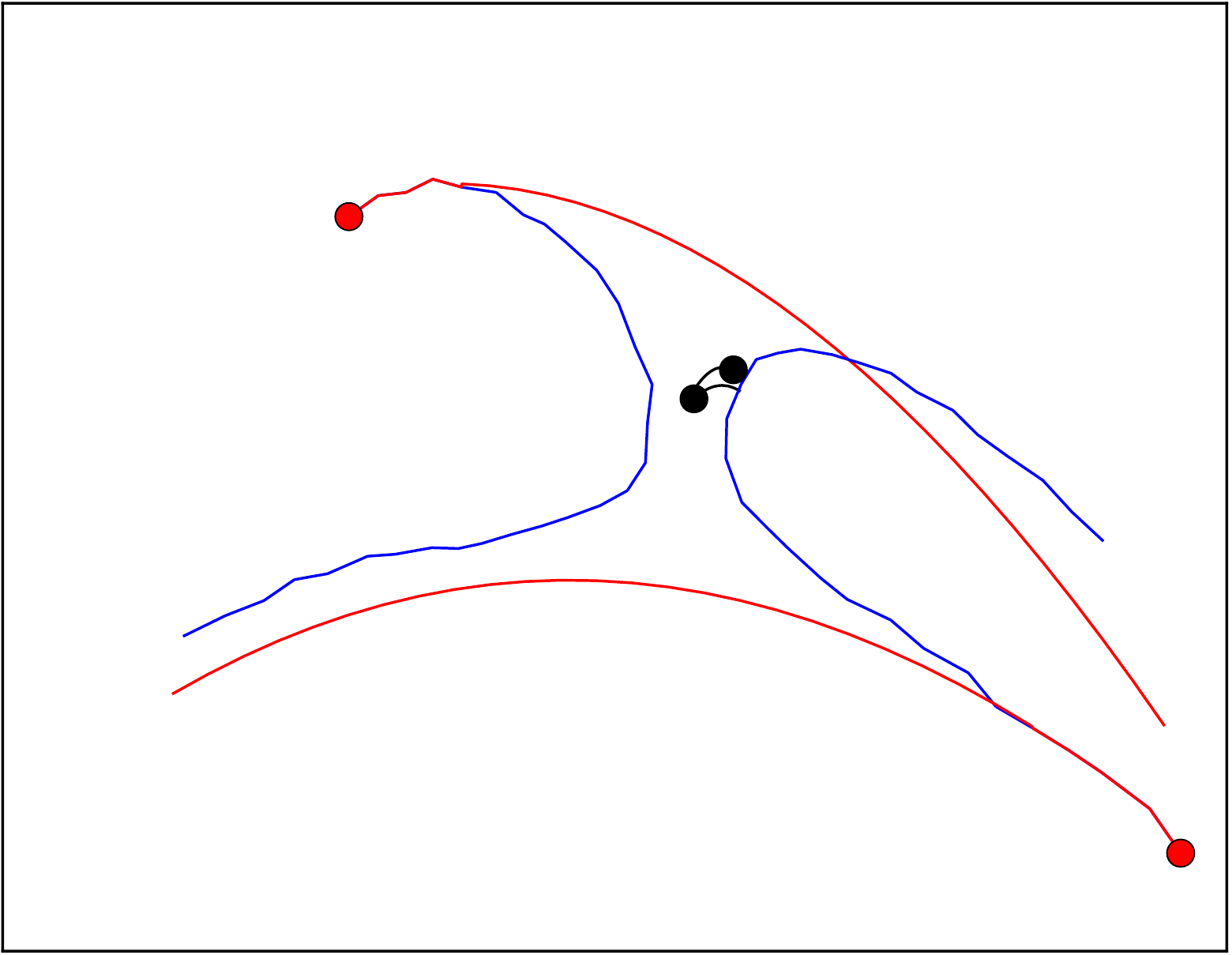}}
\caption{(a) Initial, middle, and later stages of training, showing the inference network means $\mu_{\phi}(s^n_1),\ldots,\mu_{\phi}(s^n_{30})$, $n=1,2$, for a batch of image sequences. (b) Ground-truth (black; left), generated (red; right), and inferred (blue; right) trajectories for a case in which we fail to learn an accurate inference network despite having learned an accurate generative model.}
\label{fig:training}
\end{figure}

The absence of force gives rise to positions that form straights lines, as the dots in \figref{fig:Initialization}, which represent the first two dimensions of $h^n_{1:T}$ for $n=1,\ldots,3$. The high emission noise gives rise to highly non-smooth $a^n_{1:T}$ (crosses). This initialization induces very different dynamics from the ground-truth and does not assume any clustering, but 
encodes prior information that objects move smoothly in time.

For the rendering and inference networks in Secs. \ref{sec:rendering-objects} and \ref{sec:IN},
the weight matrices entries were randomly initialized from $\mathcal{N}( \, \cdot \, ; 0, 1 / \sqrt{d})$,
where $d$ is the number of matrix elements. All biases were initialized to zero. 
The dimension of the latent state $s^n_t$ was set to 1024.
We used the Adam optimizer with learning rate 0.001, mini-batch size 20, and default values $\beta_1=0.9$, $\beta_2=0.999$, $\epsilon = 10^{-8}$. Training was stopped after $2 \times 10^5$ iterations.

To guide the inference network toward smooth trajectories we kept the LGSSM parameters
fixed to their initial values for the first $10^4$ iterations,
and only optimized for all other parameters.
This initialization loosely gives temporal coherence between the inference network and the renderer, for different images.
The model was then trained jointly, first by changing the objective function by multiplying the KL term in \eqref{eq:elbo}
with a weight $\beta$, starting at $\beta = 100$,
and annealing $\beta$ down to one.
This avoids the LGSSM parameters from too quickly modelling the output of a (still very sub-optimal) 
$q_{\phi}(a_{1:T}^{1:N} | v_{1:T})$, and stagnating at a local maximum.
This process initializes the model, after which end-to-end training proceeds.

\subsection{Inferring latent positions from image sequences}

In \figref{fig:InferPos} we show estimates of positions from the inference network.
The ground-truth trajectories from which each image sequence is generated are shown in black.
The inference network means $\mu_{\phi}(s^n_1), \ldots, \mu_{\phi}(s^n_T)$ are plotted in blue.
Notice that each plot is scaled differently to aid qualitative evaluation. For that reason, we 
also selected a run with minimum rotation, but the latent positions can only be retrieved up to re-scaling and rotation.
The larger scale of the inferred trajectories is highlighted by re-plotting the ground-truth trajectories (red lines) together with the inferred positions. 
These figures demonstrate that our model can accurately infer latent positions from the sequences of images in an unsupervised way. 

In \figref{fig:training}(a), we show the inference network means for a batch of image sequences at three stages of 
training (for a case in which $N=2$), illustrating how the model learns over time.
To highlight the challenge in learning to disentangle the objects dynamics,
and the importance of initially strongly regularizing it
to be close to the LGSSM during training,
\figref{fig:training}(b) shows an example in which we fail to learn an accurate inference network despite having learned accurate LGSSM dynamics.
The ground-truth trajectories (black; left) and the trajectories generated from the learned LGSSM dynamics (red; right) are very similar. On the other hand, the inference network means (blue; right) swap the ball when reaching the middle part of the image:
Rather than learning a successful attention mechanism, the inference network has learned to attend to the left part of the image for one ball and to the right part of the image for the other ball.
\begin{figure}[t]
\begin{center}
\hskip-0.05cm
\includegraphics[height=1.9cm,width=3.8cm]{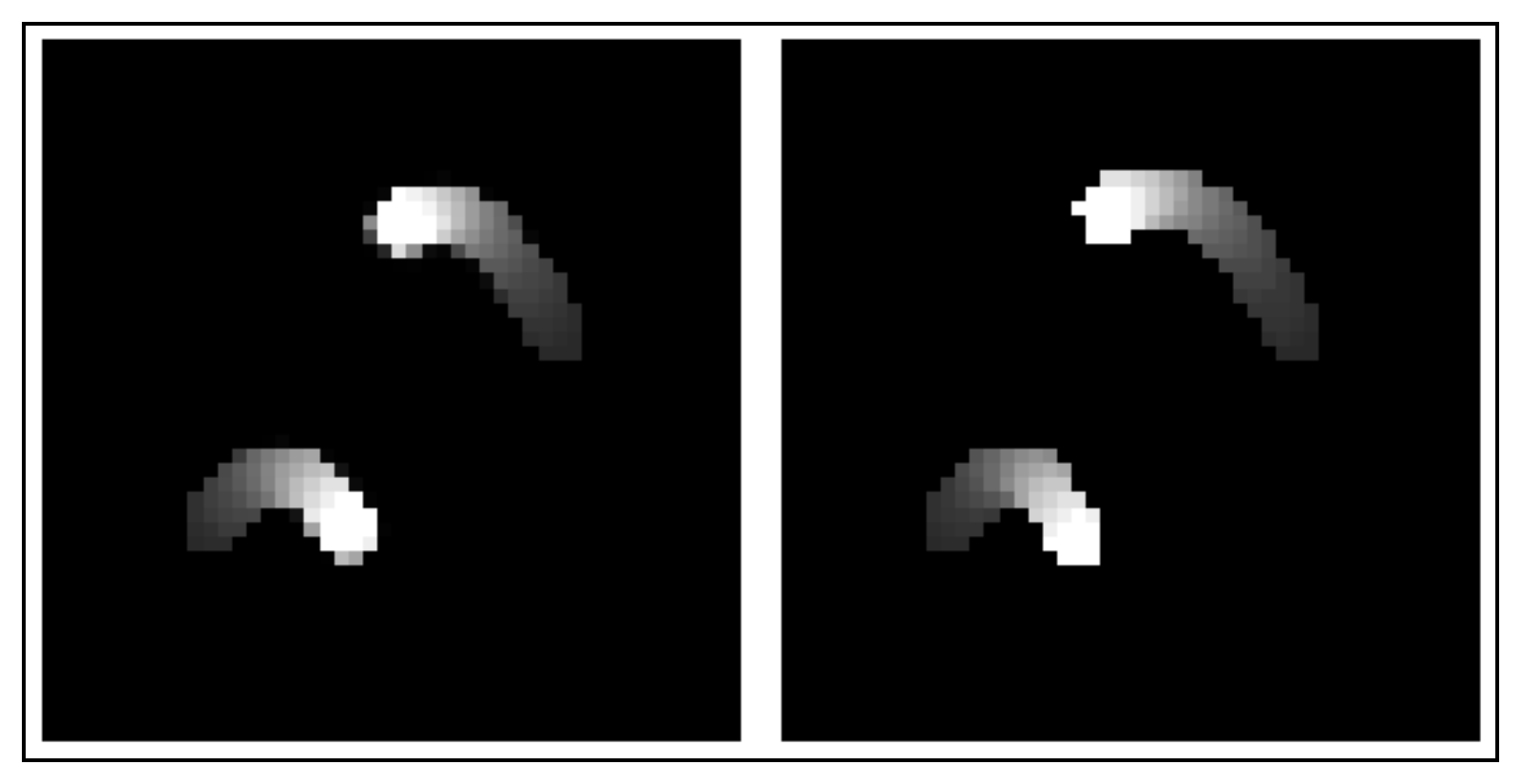}
\includegraphics[height=1.9cm,width=3.8cm]{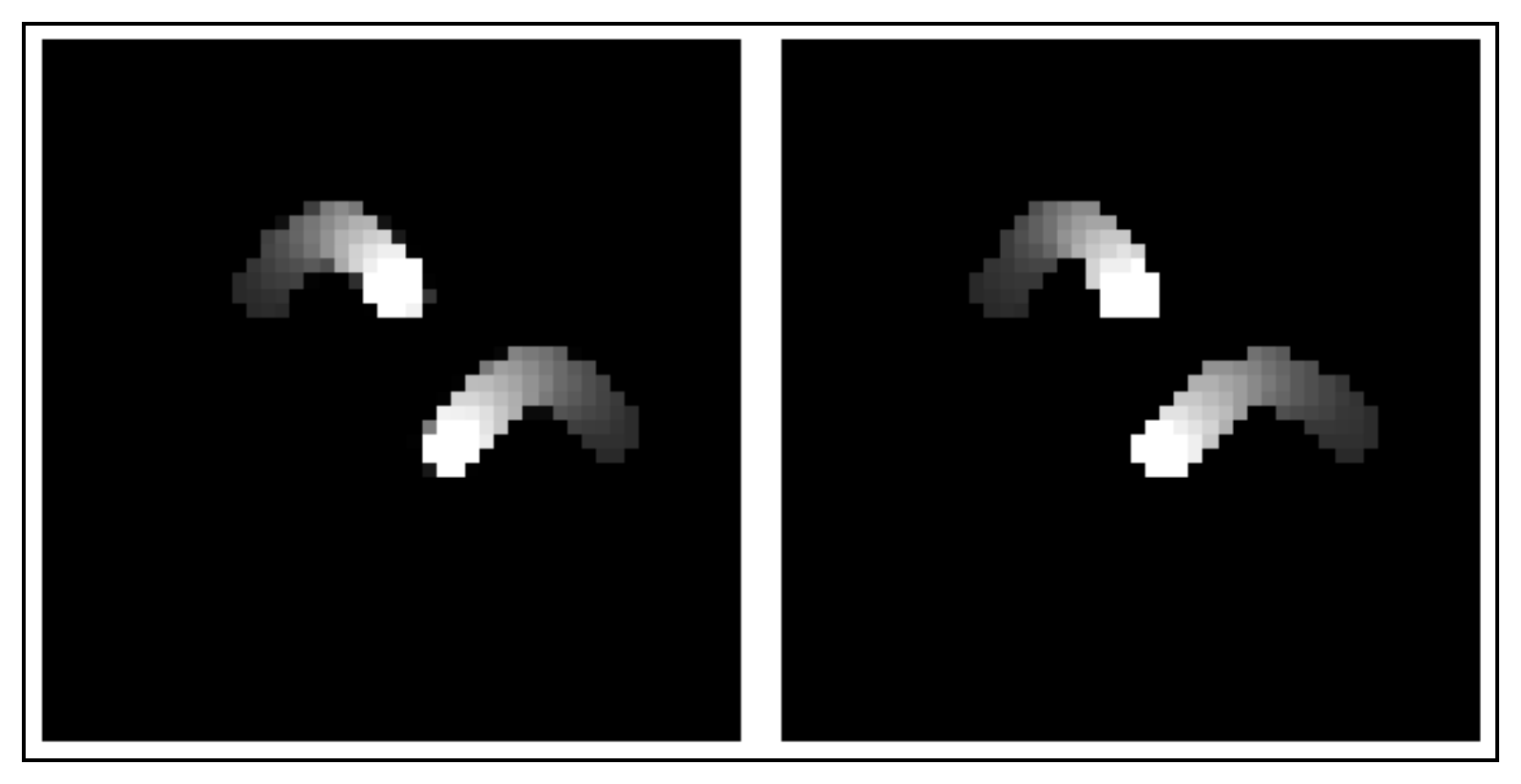}
\includegraphics[height=1.9cm,width=3.8cm]{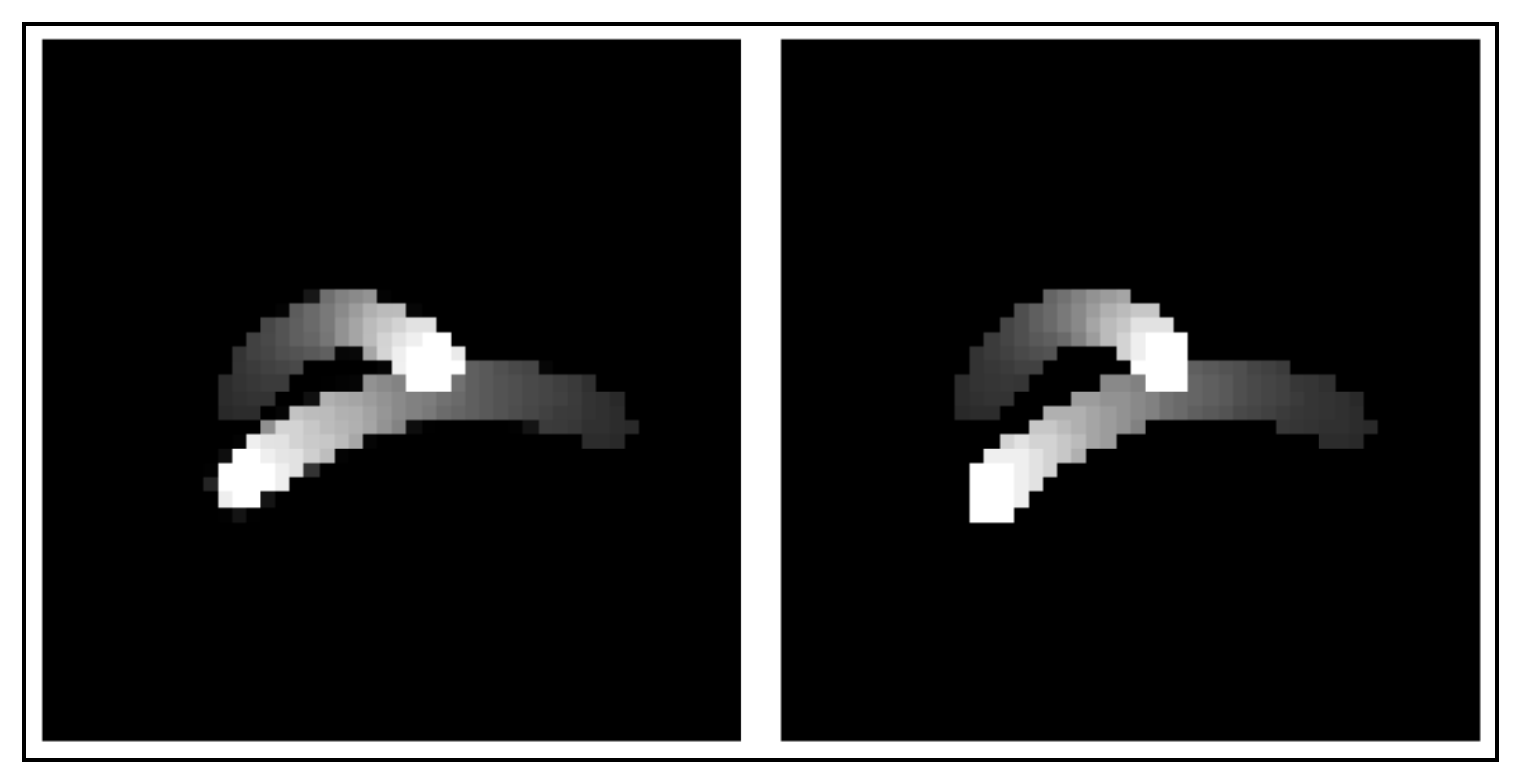}
\end{center}
\begin{center}
\hskip-0.05cm
\includegraphics[height=1.9cm,width=3.8cm]{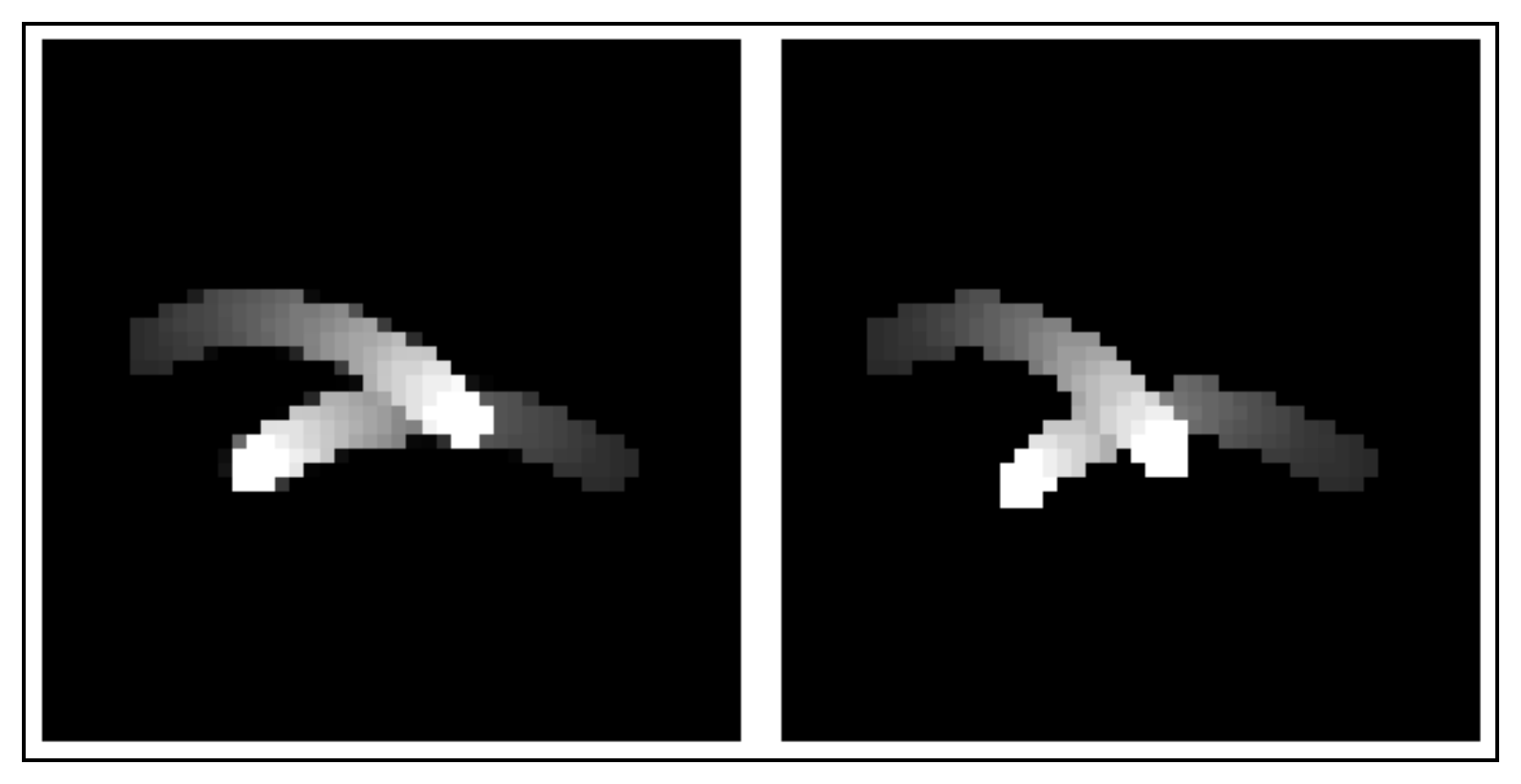}
\includegraphics[height=1.9cm,width=3.8cm]{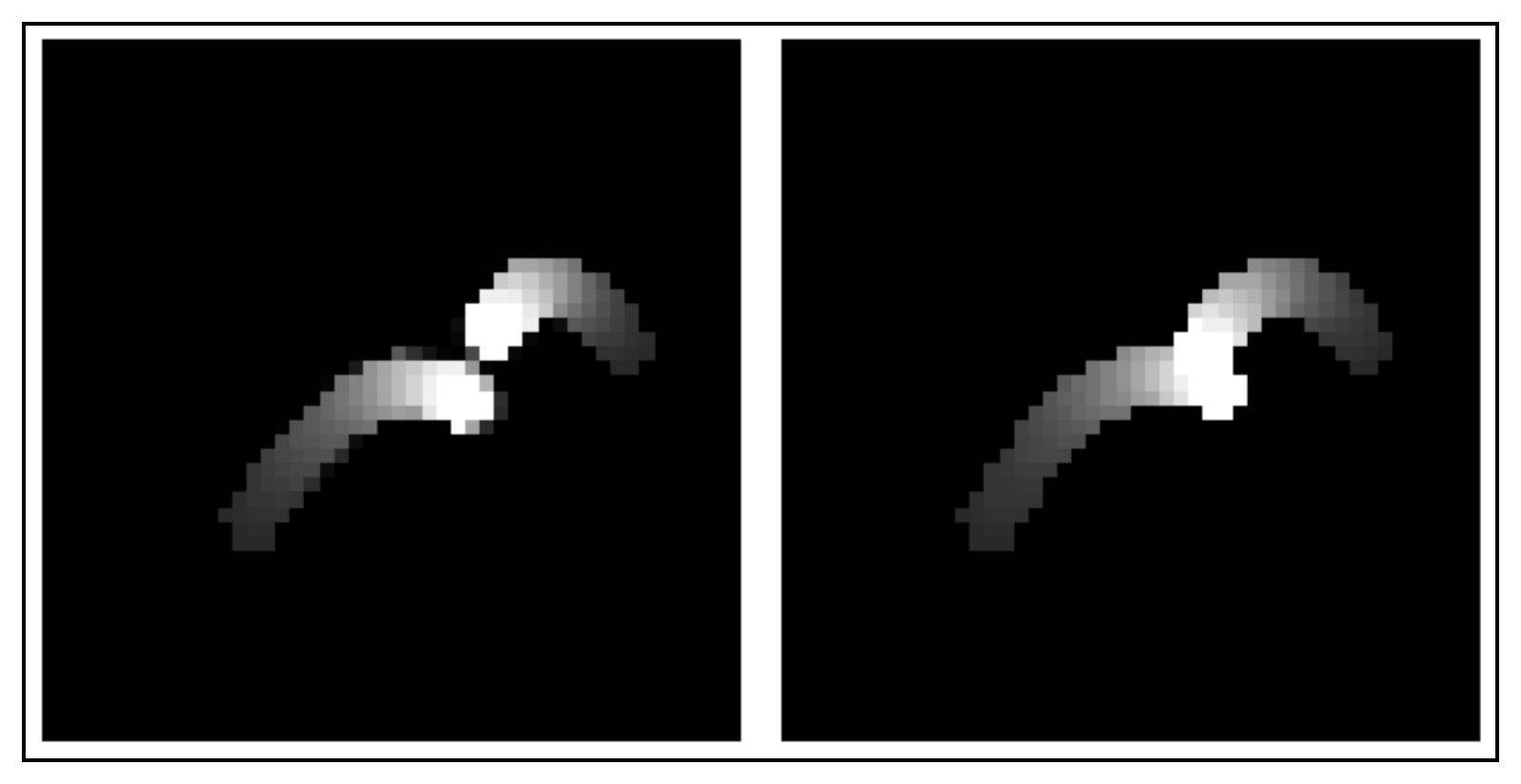}
\includegraphics[height=1.9cm,width=3.8cm]{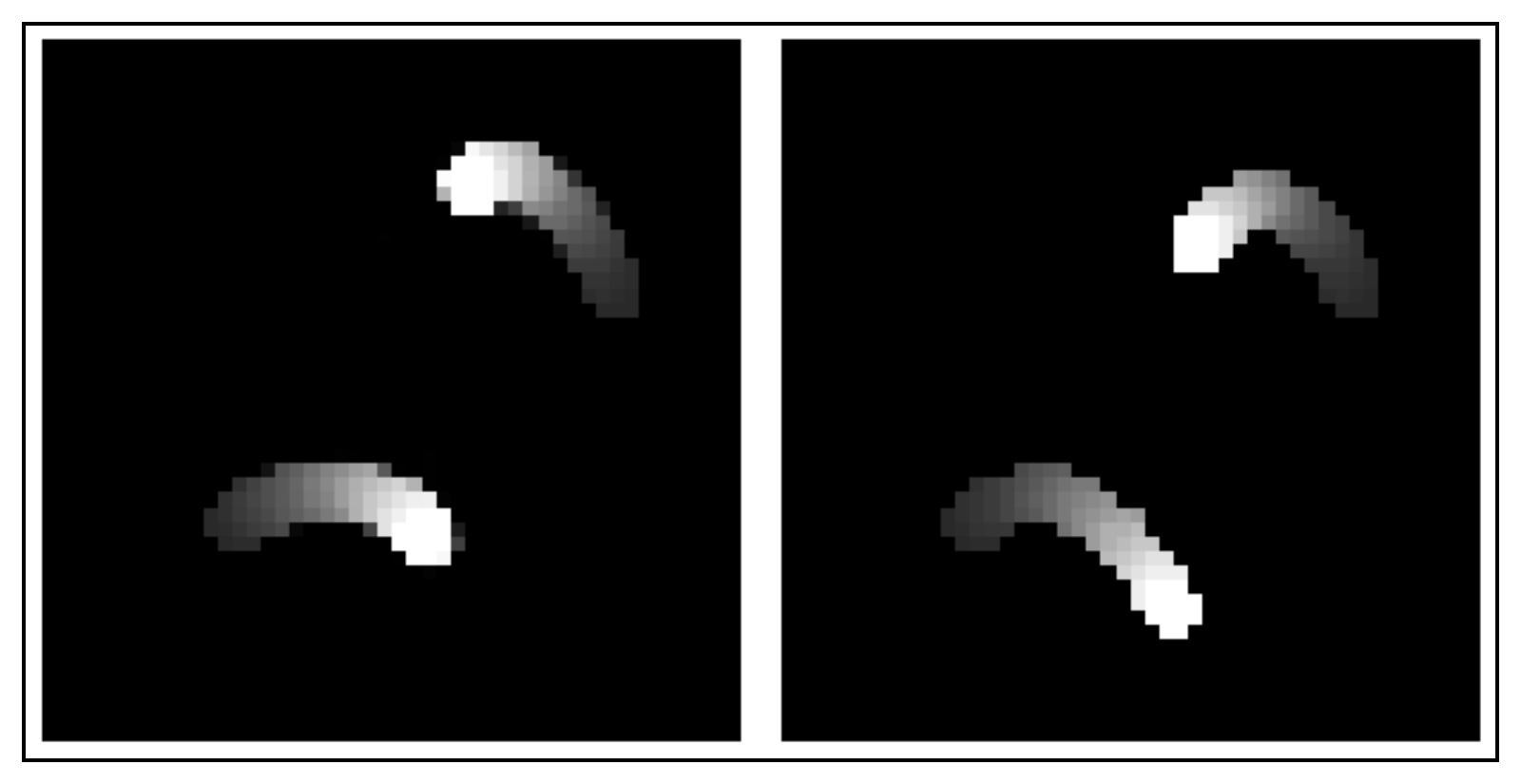}
\end{center}
\caption{Six examples of generated (left) versus ground-truth (right) images overlaid in time. Top row: Our model. Bottom row: ED-LSTM model.}
\label{fig:GenObsFramesPaper}
\end{figure}

\subsection{Multi-step ahead generation of images}
To benchmark against a standard deep-learning model for multi-step ahead generation of images,
we compared our model to an encoder-decoder long-short term memory model (ED-LSTM) \cite{hochreiter97long} on the task of generating the 25 images following five observed images $v_{1:5}$.

\begin{wrapfigure}[10]{r}{0.45\textwidth}
\vskip-0.9cm
\hskip-0.25cm
\centering
\scalebox{0.84}{
\begin{tikzpicture}[->, line width=1pt, node distance=1.3cm]
\node at (3.2,0) {$\cdots$};
\node at (8.9,0) {$\cdots$};
\node[rnn] (stau) at (4,0) {$\lstms_{\tau}$};
\node[rnn] (staup) at (6,0) {$\lstms_{\tau+1}$};
\node[rnn] (staupp) at (8,0) {$\lstms_{\tau+2}$};
\node[randomv] (hvtau) at (4,-2) {$\hat v_{\tau}$};
\node[randomv] (hvtaup) at (6,-2) {$\hat v_{\tau+1}$};
\node[randomv] (hvtaupp) at (8,-2) {$\hat v_{\tau+2}$};
\node[randomv,obs] (vtau) at (4,-3) {$v_{\tau}$};
\node[randomv,obs] (vtaup) at (6,-3) {$v_{\tau+1}$};
\node[randomv,obs] (vtaupp) at (8,-3) {$v_{\tau+2}$};
\draw[line width=1pt](stau)--(staup);
\draw[line width=1pt](staup)--(staupp);
\draw[line width=1pt](vtau)to[bend left=-15]node[sloped,above]{encod.}(staup);
\draw[line width=1pt,dashed](hvtaup)to[bend left=-15](staupp);
\draw[line width=1pt,dashed](vtaup)to[bend left=-15](staupp);
\draw[line width=1pt](stau)--(hvtau);
\draw[line width=1pt](staup)--node[sloped,above]{decod.}++(hvtaup);
\draw[line width=1pt](staupp)--(hvtaupp);
\end{tikzpicture}
}
\caption{Encoder-decoder long-short term memory structure.}
\label{fig: ED-LSTM}
\end{wrapfigure}
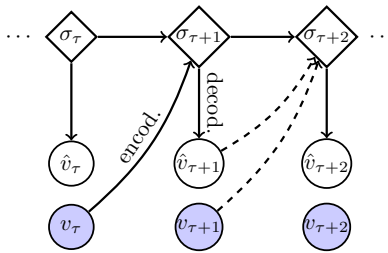

The structure of the ED-LSTM is represented in \figref{fig: ED-LSTM}: At each time-step $t$, the hidden state $\lstms_t$ generates an image $\hat v_t$ through a decoding transformation. 
For the first $\tau=5$ time-steps, $\lstms_t$
receives an encoded version of the previous ground-truth image $v_{t-1}$,
as well as $\lstms_{t-1}$, as input.
From time-step $\tau+1$ onward,
$\lstms_t$ receives an encoded version of the previous ground-truth image $v_{t-1}$ during training,
\emph{or} an encoded version of the previous generated image $\hat v_{t-1}$ when the model is used in a multi-step ahead generation mode.
We experimented with both convolutional and fully connected encoding and decoding transformations. The best results were obtained with one or two fully connected layers and dimension 2048 for $\lstms_t$.
(All layers except the last were followed by a ReLU activation. 
The decoder last layer was followed by a sigmoid activation.) 
We used the same weights and biases initialization and optimizer settings as for our model.

\begin{figure}[t]
\begin{center}
\includegraphics[height=0.2\textwidth,width=0.2\textwidth]{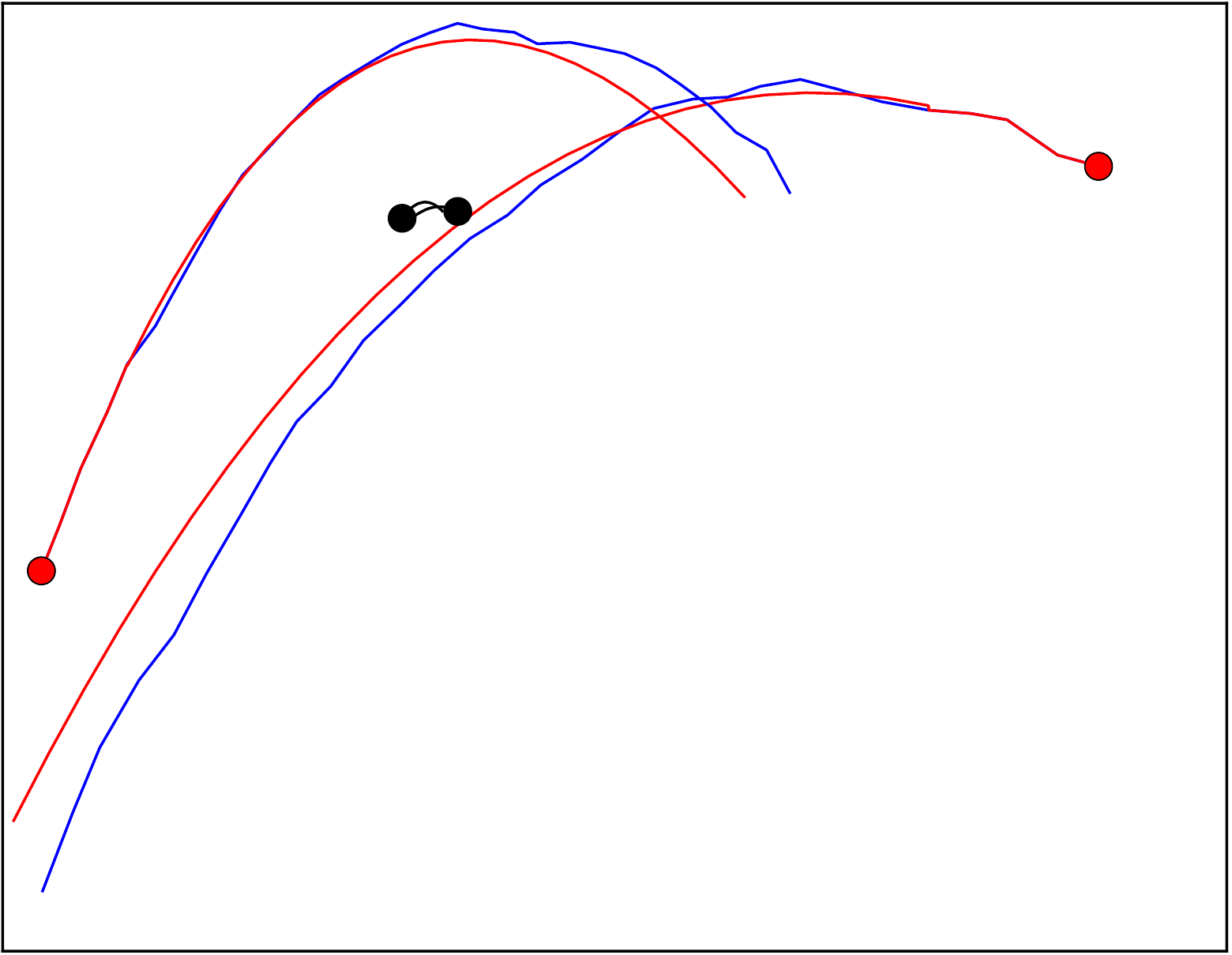}
\hskip-0.1cm
\includegraphics[height=0.2\textwidth,width=0.2\textwidth]{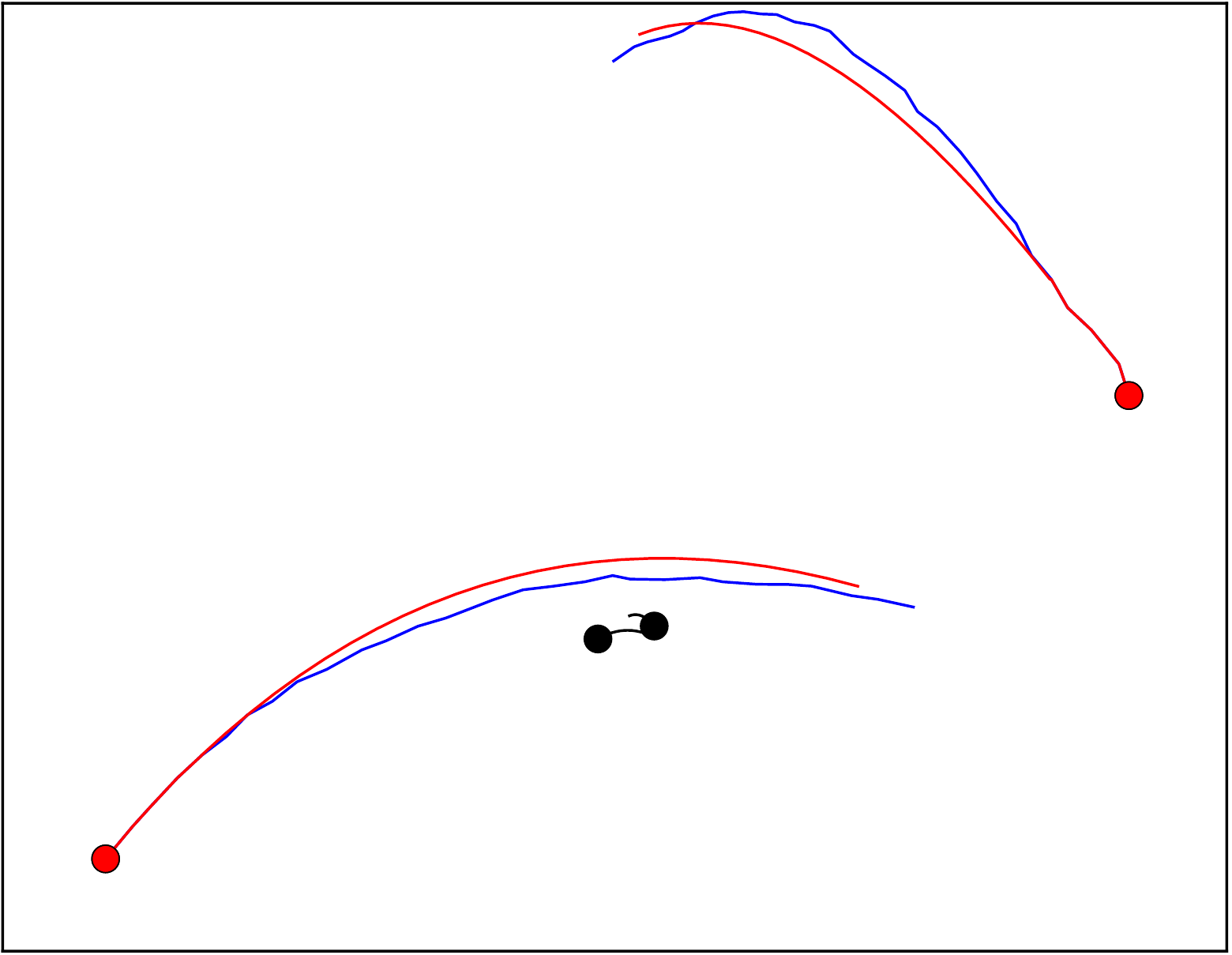}
\hskip-0.1cm
\includegraphics[height=0.2\textwidth,width=0.2\textwidth]{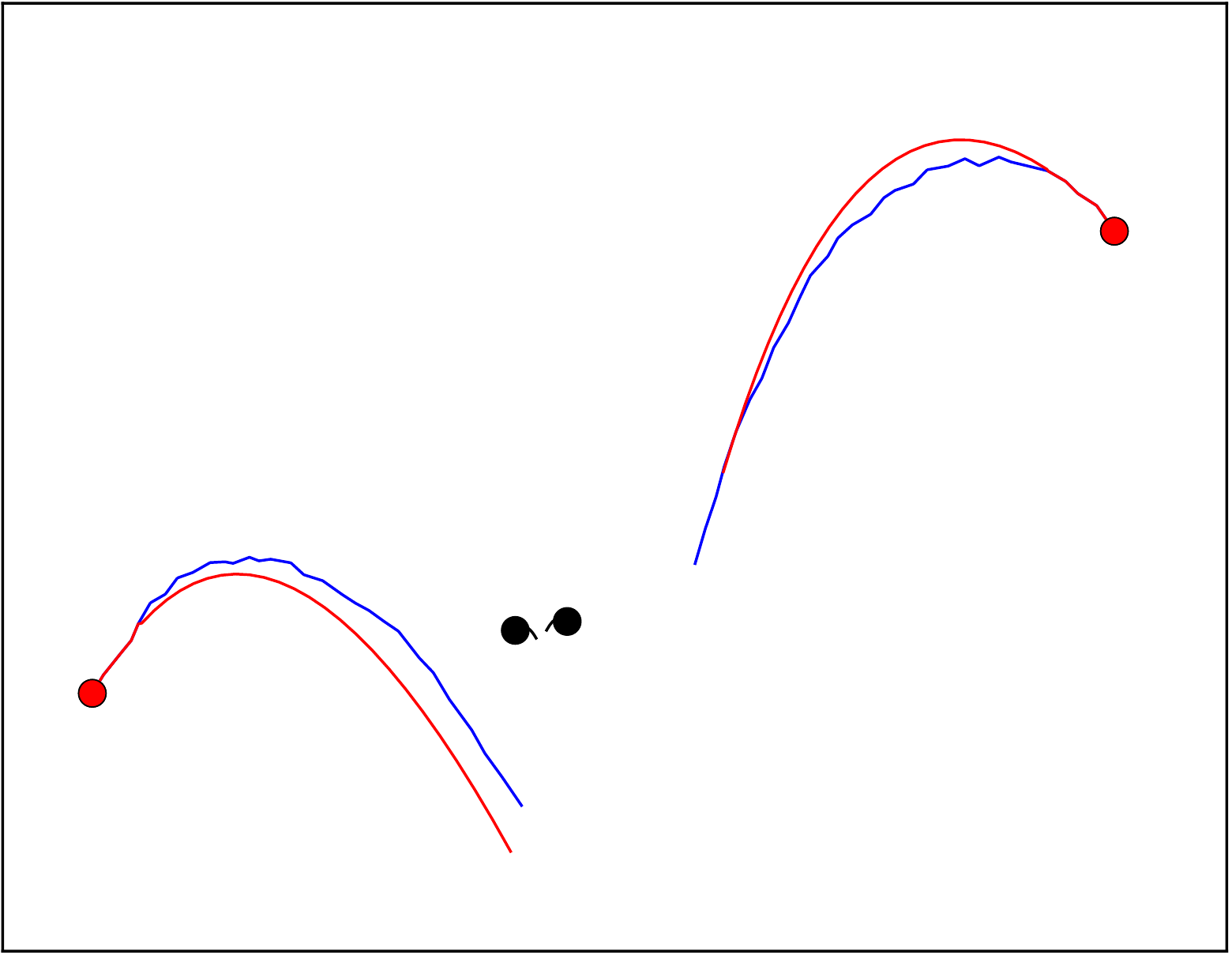}
\hskip-0.1cm
\includegraphics[height=0.2\textwidth,width=0.2\textwidth]{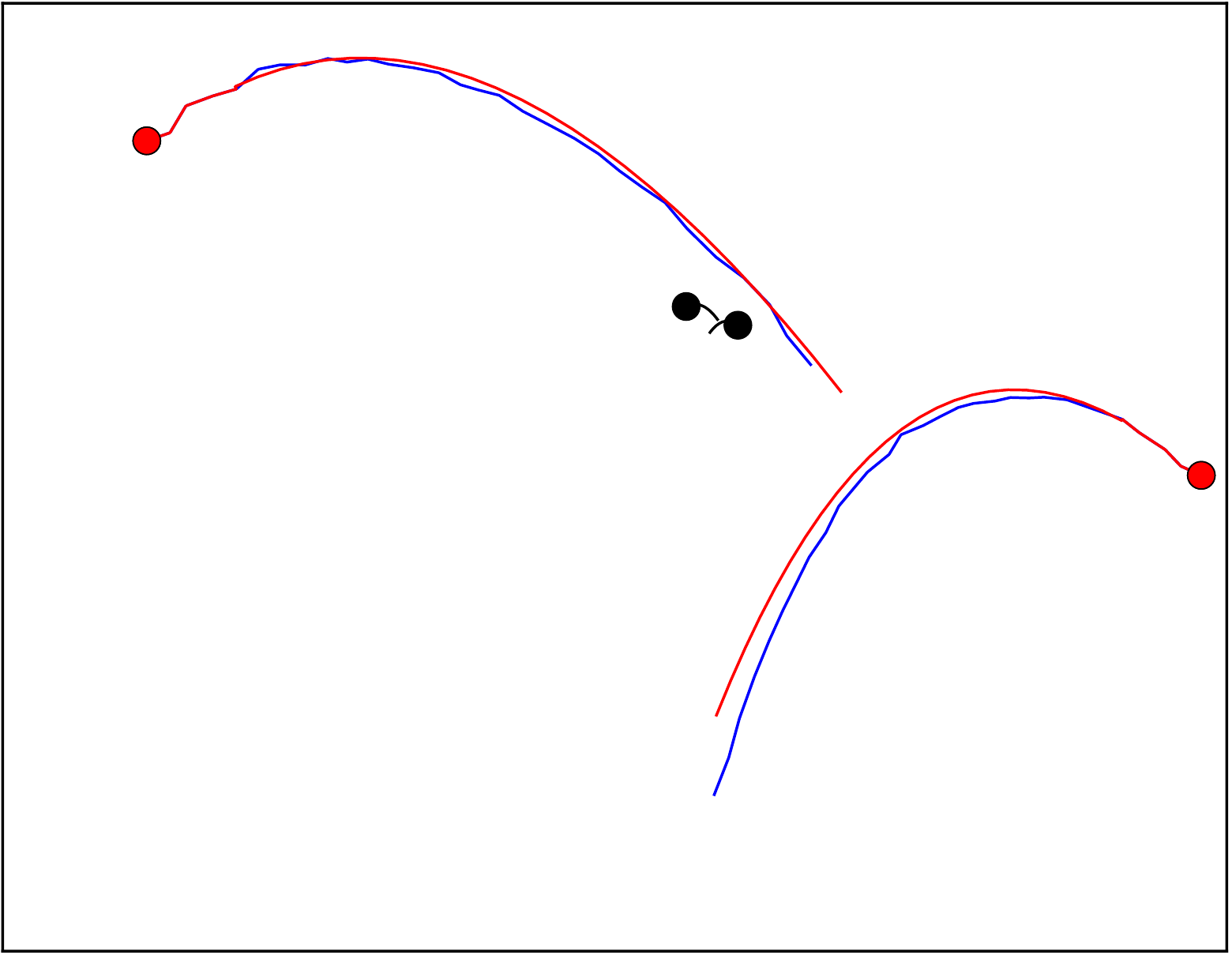}
\end{center}
\caption{
We infer $a_{1:5}^{n}$ for each ball from $v_{1:5}$, and use the filtered means of $p(h_{1:5}^{n} | a_{1:5}^{n}, z^n)$ for the most likely $z^n$ to forward-generate the rest of the red trajectories $a_{6:30}^n$. In blue we show the trajectories that would be obtained by the inference network 
if we observed the entire $v_{1:30}$.
The ground-truth trajectories underlying $v_{1:30}$ are shown in (tiny) black lines.
}
\label{fig:InferGenPos}
\end{figure}

To generate images with our model, we first inferred $a^n_{1:5}$ as the means $\mu_{\phi}(s^n_1),\ldots,\mu_{\phi}(s^n_5)$ of the inference network, for each object $n$.
Using the inferred $a^n_{1:5}$,
we computed the most likely mixture component as $k^* =\argmax_k [p(z^n = k | a^n_{1:5}) \propto p(a^n_{1:5} | z^n = k)\pi_k$] by running a Kalman filter.
The filtered means of $p(h^n_5 | a^n_{1:5}, z^n=k^*)$ were then used
as initial conditions to generate $a^n_{6:30}$ with the learned LGSSM dynamics (\eqref{eq:lgssm}). 
Finally images $v_{6:30}$ were generated through rendering of $a^{1:N}_{6:30}$ (Eqs. (\ref{eq:rendering}) and (\ref{eq:v})).

In \figref{fig:GenObsFramesPaper}, we show generated versus ground-truth images overlaid in time,
for four sequences (more examples are given in the Appendix). 
For generation, our averaged test loss over the 25 time-steps was 0.691, using
only eight as latent state dimension for the dynamics ($h_t^{1:2}$). In contrast,
the ED-LSTM's test loss was 0.693, with 2048 as latent state dimension for the dynamics.
Some examples of trajectories generated by our model are shown in \figref{fig:InferGenPos}.

\begin{figure}[t]
\centering
\begin{minipage}[t]{0.47\textwidth}
\centering
\begin{overpic}[height=0.420\textwidth,width=0.420\textwidth]{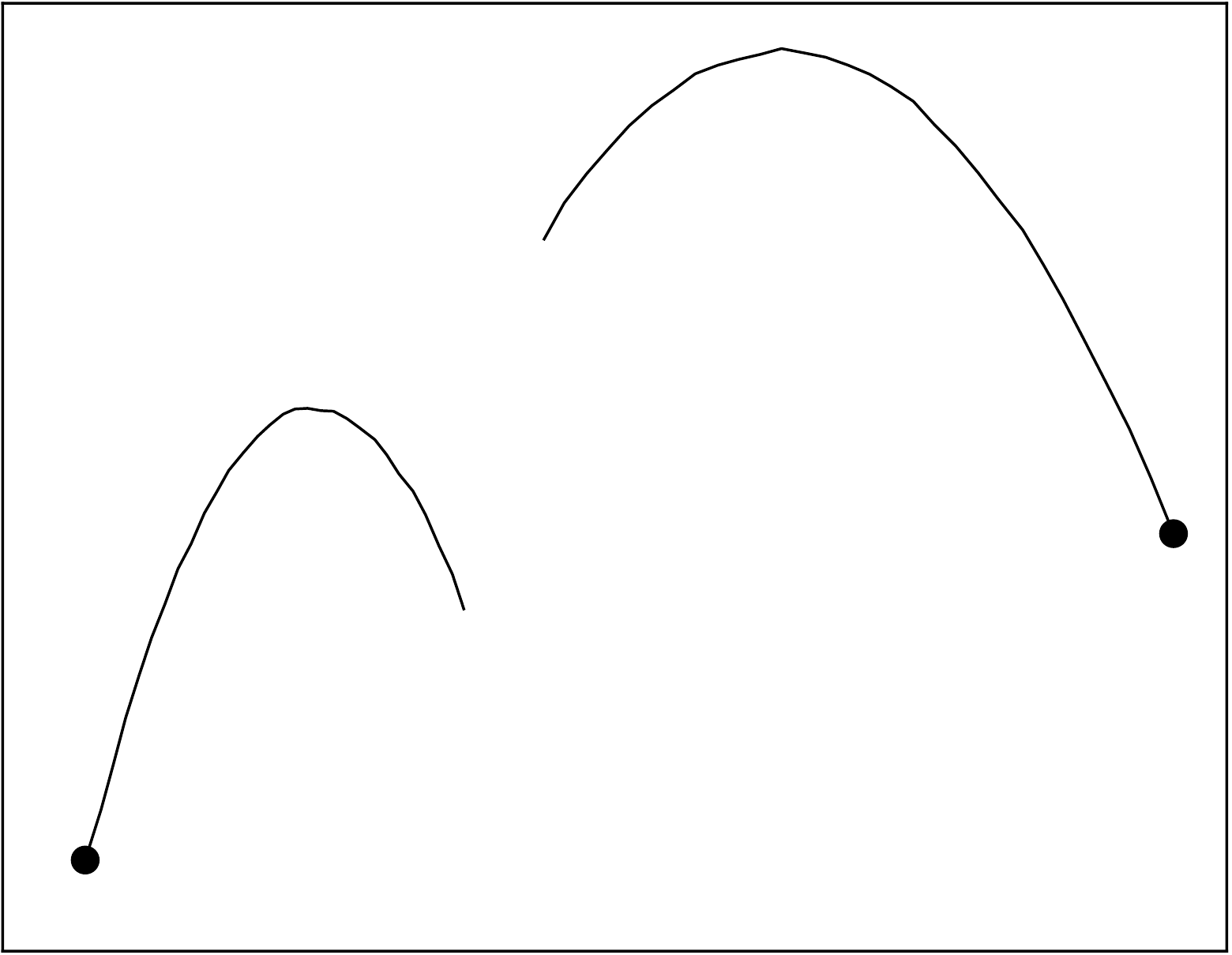}
 \put (10,63) {$a_{1:30}^{1}$}
 \put (62,60) {$a_{1:30}^{2}$}
\end{overpic}
\includegraphics[height=0.420\textwidth,width=0.420\textwidth]{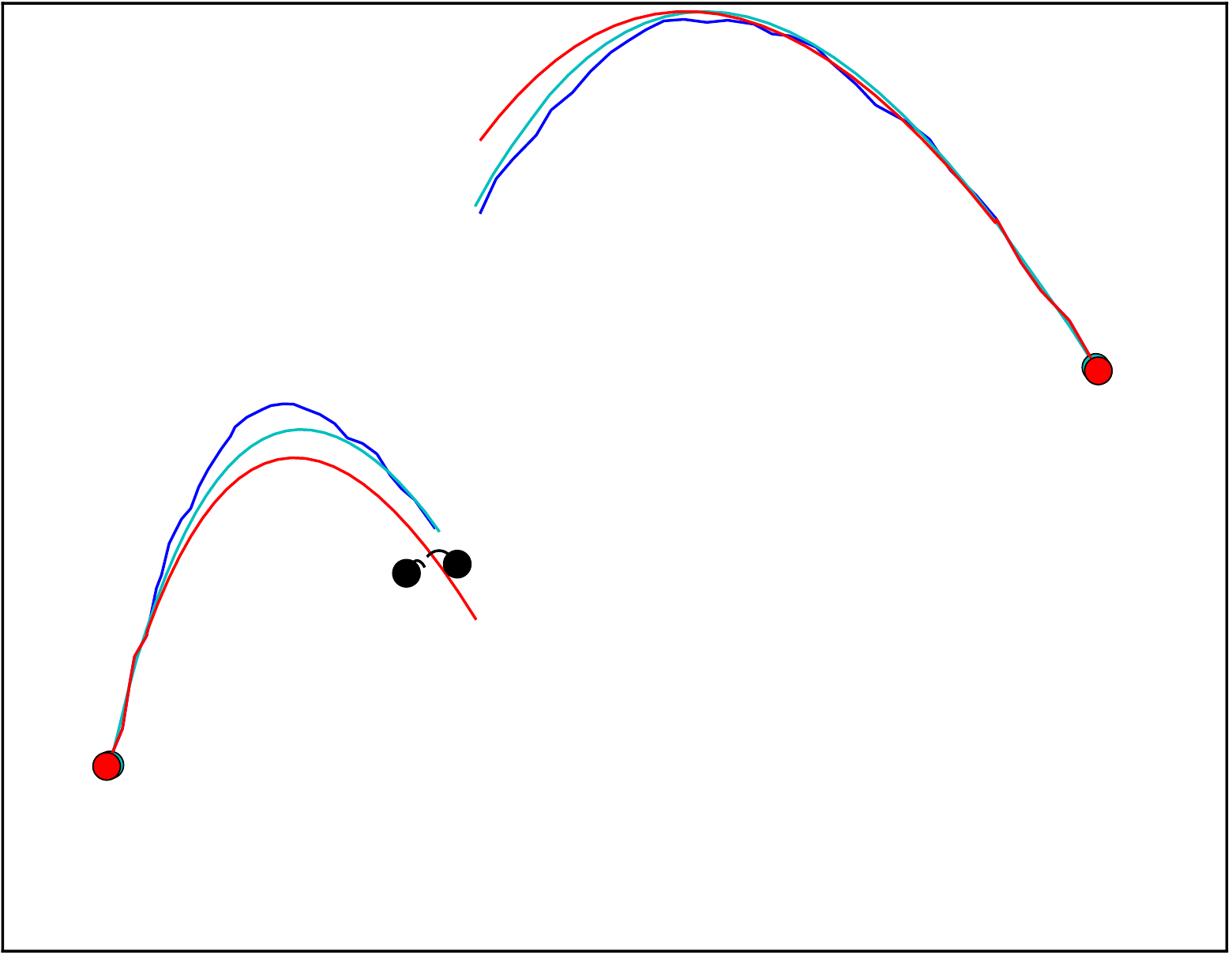} \\
\vspace{4pt}
\includegraphics[width=0.450\textwidth]{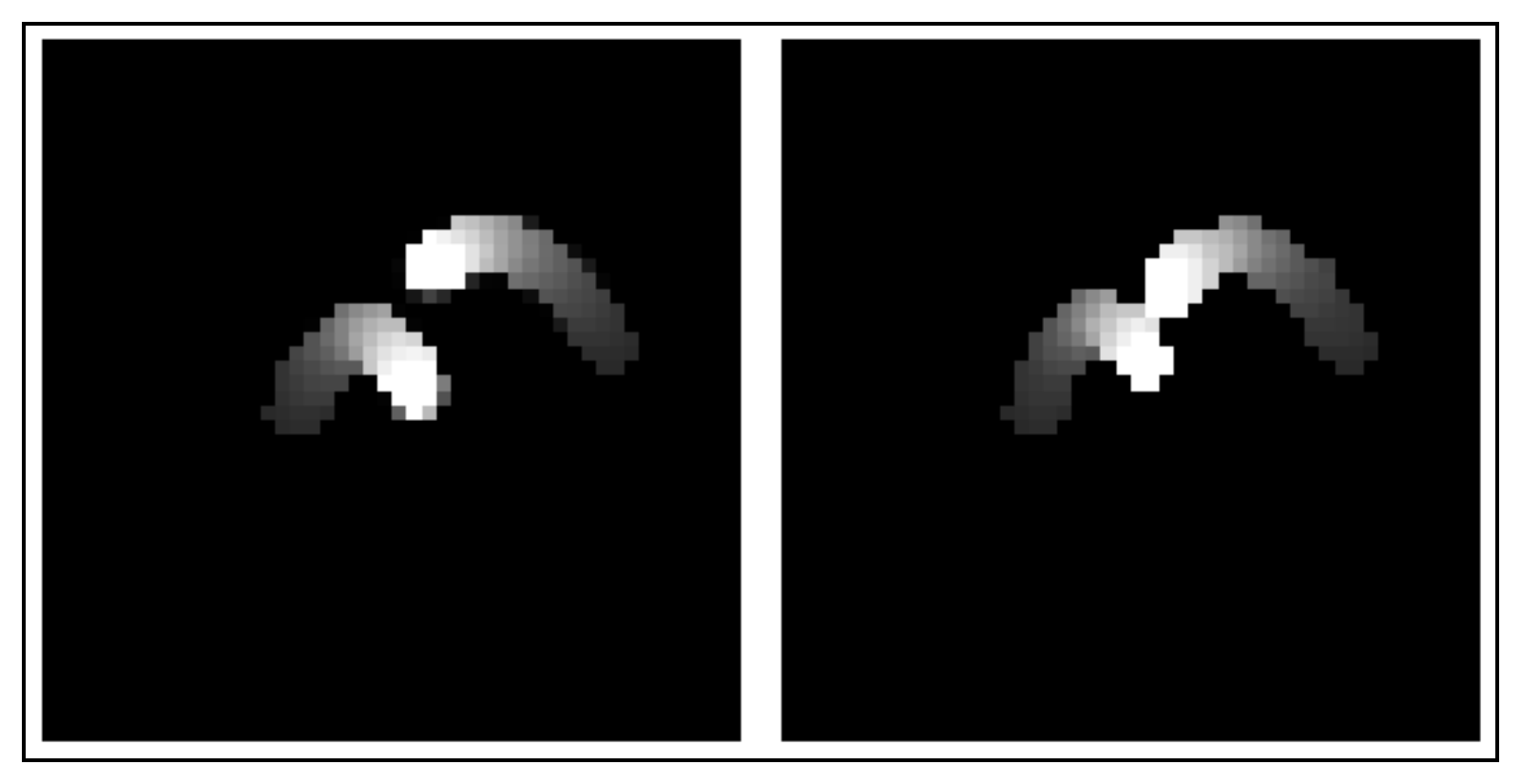}
\includegraphics[width=0.450\textwidth]{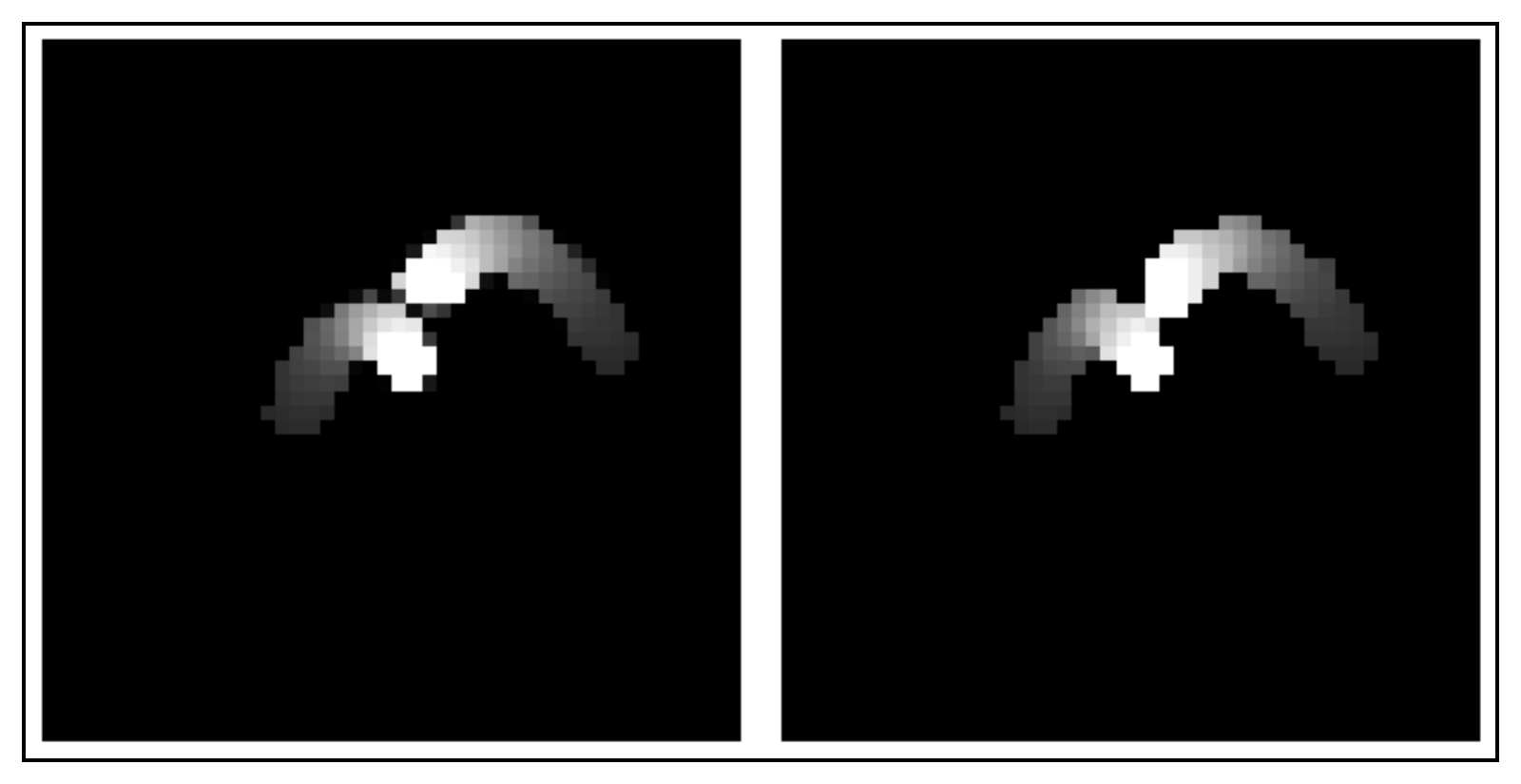} \\
$\phantom{.}$
Generated \emph{vs} $v_{1:30}$
| 
Interpolated \emph{vs} $v_{1:30}$
\end{minipage}
\begin{minipage}[t]{0.47\textwidth}
\centering
\includegraphics[height=0.420\textwidth,width=0.420\textwidth]{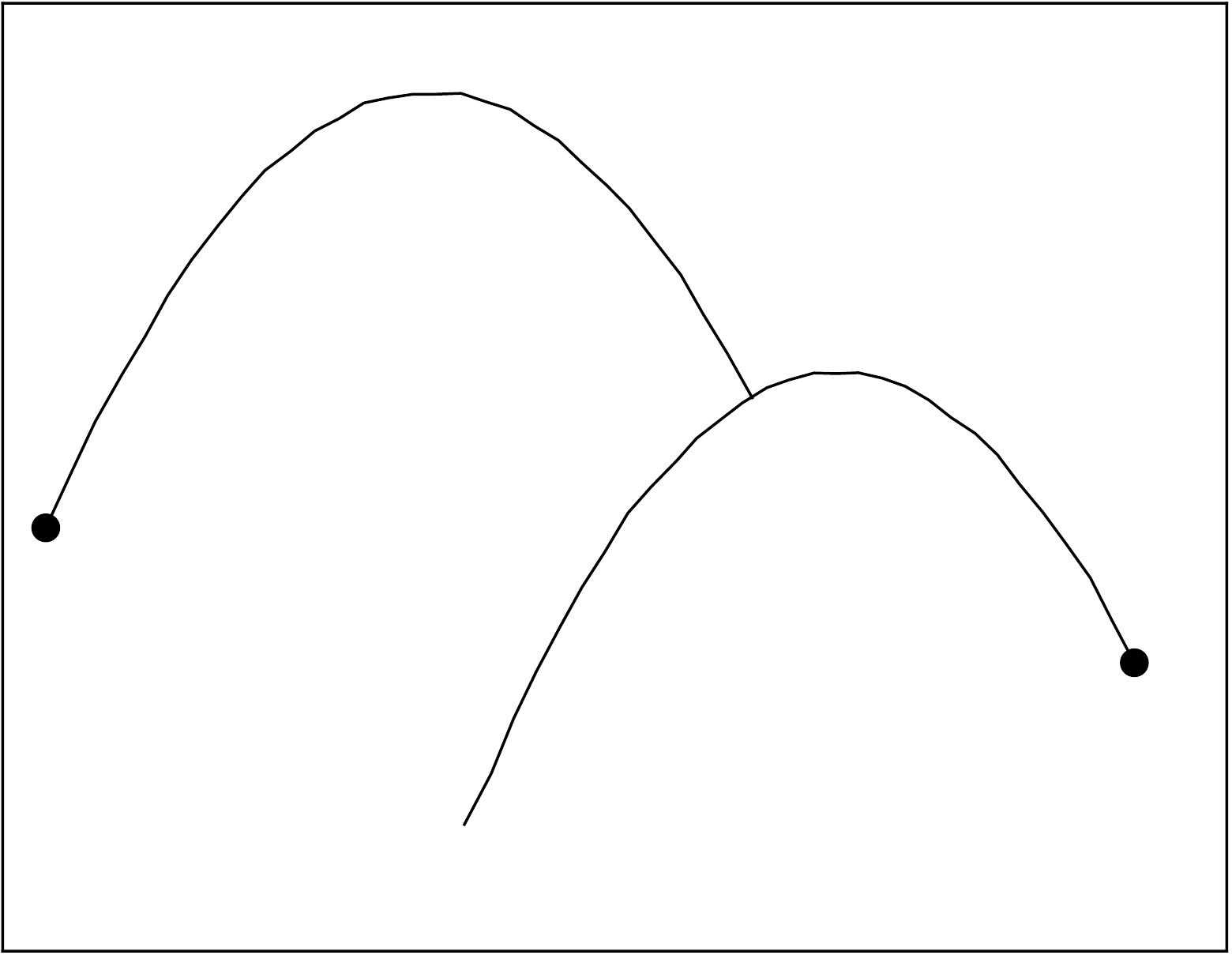}
\includegraphics[height=0.420\textwidth,width=0.420\textwidth]{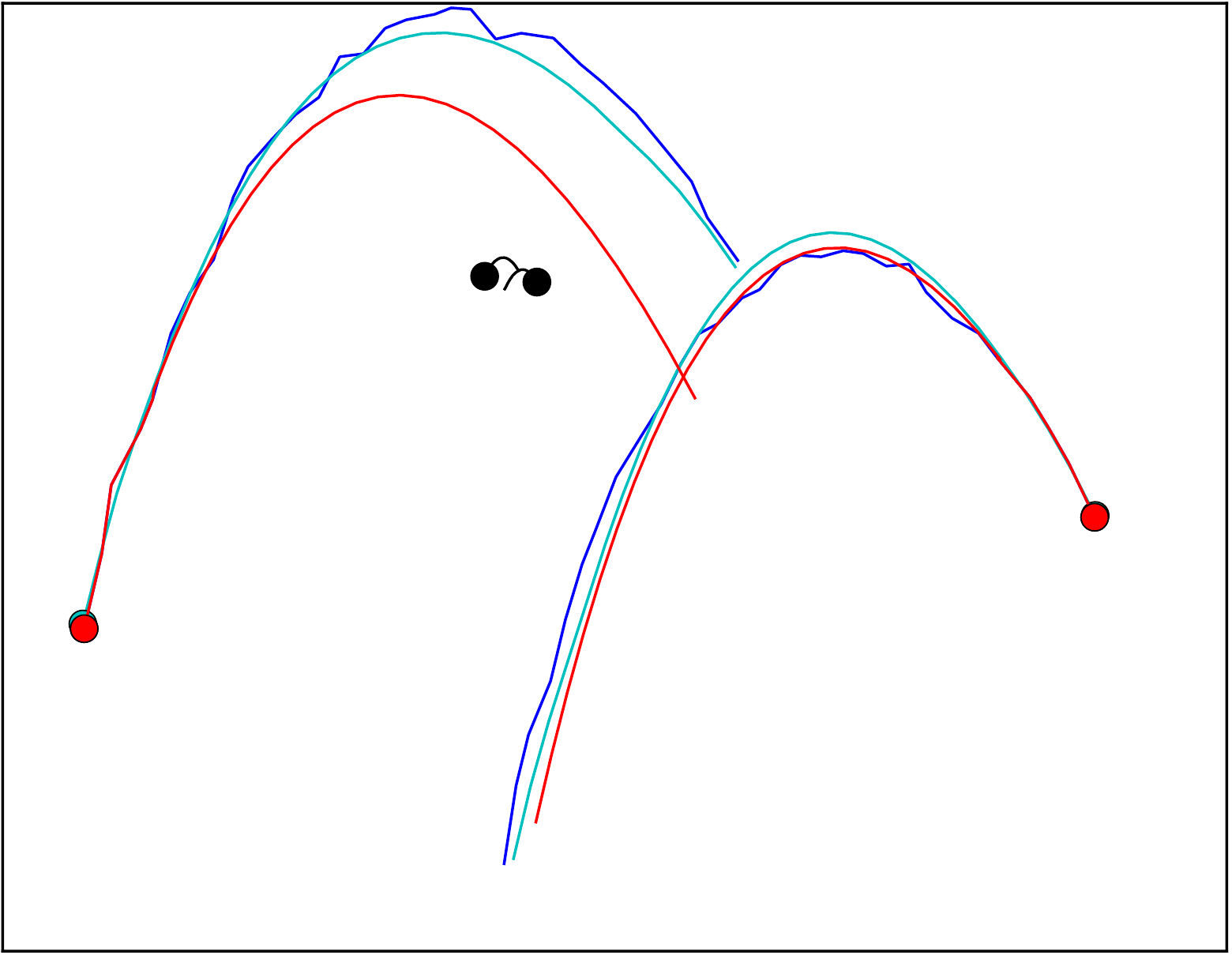} \\
\vspace{4pt}
\includegraphics[width=0.450\textwidth]{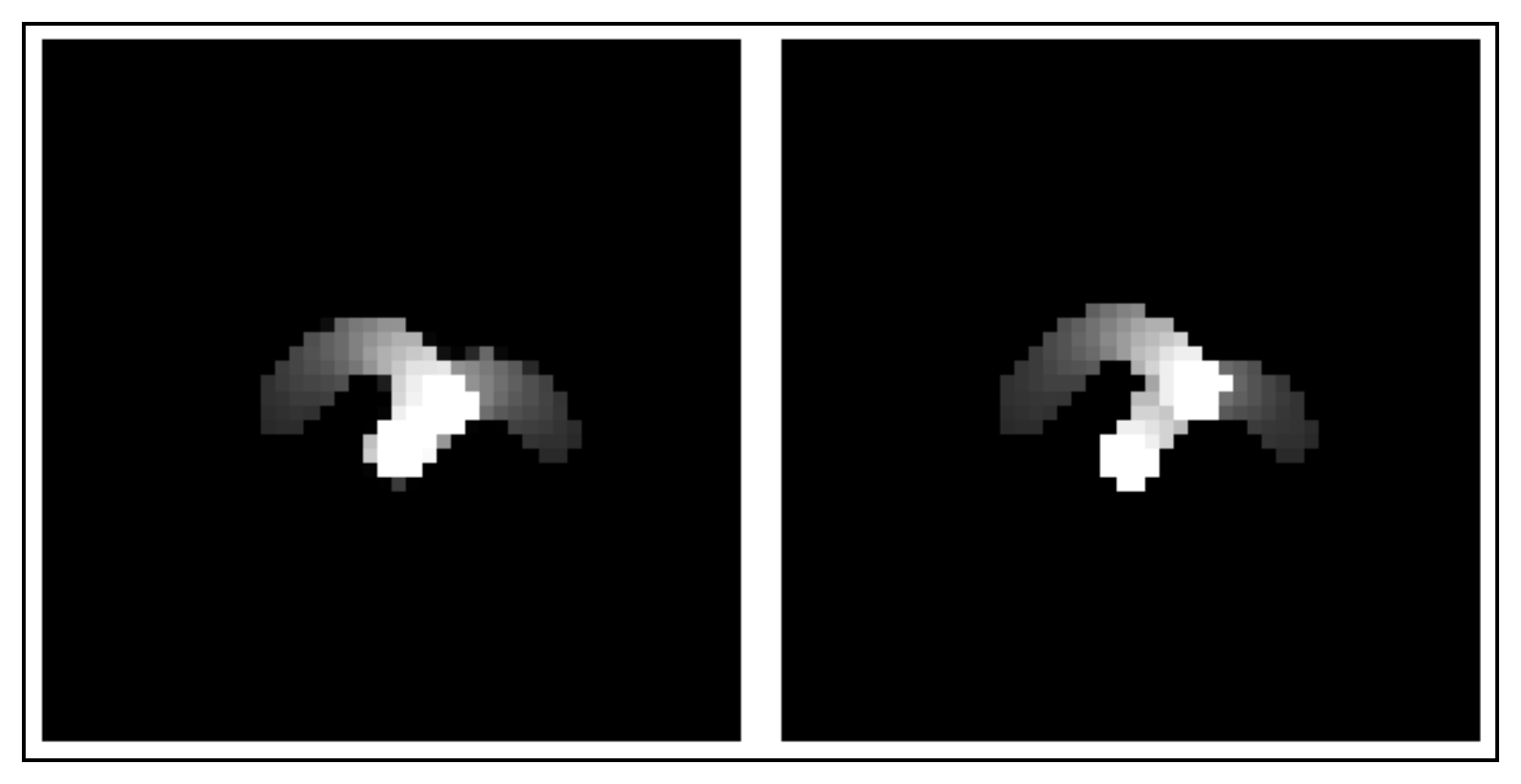}
\includegraphics[width=0.450\textwidth]{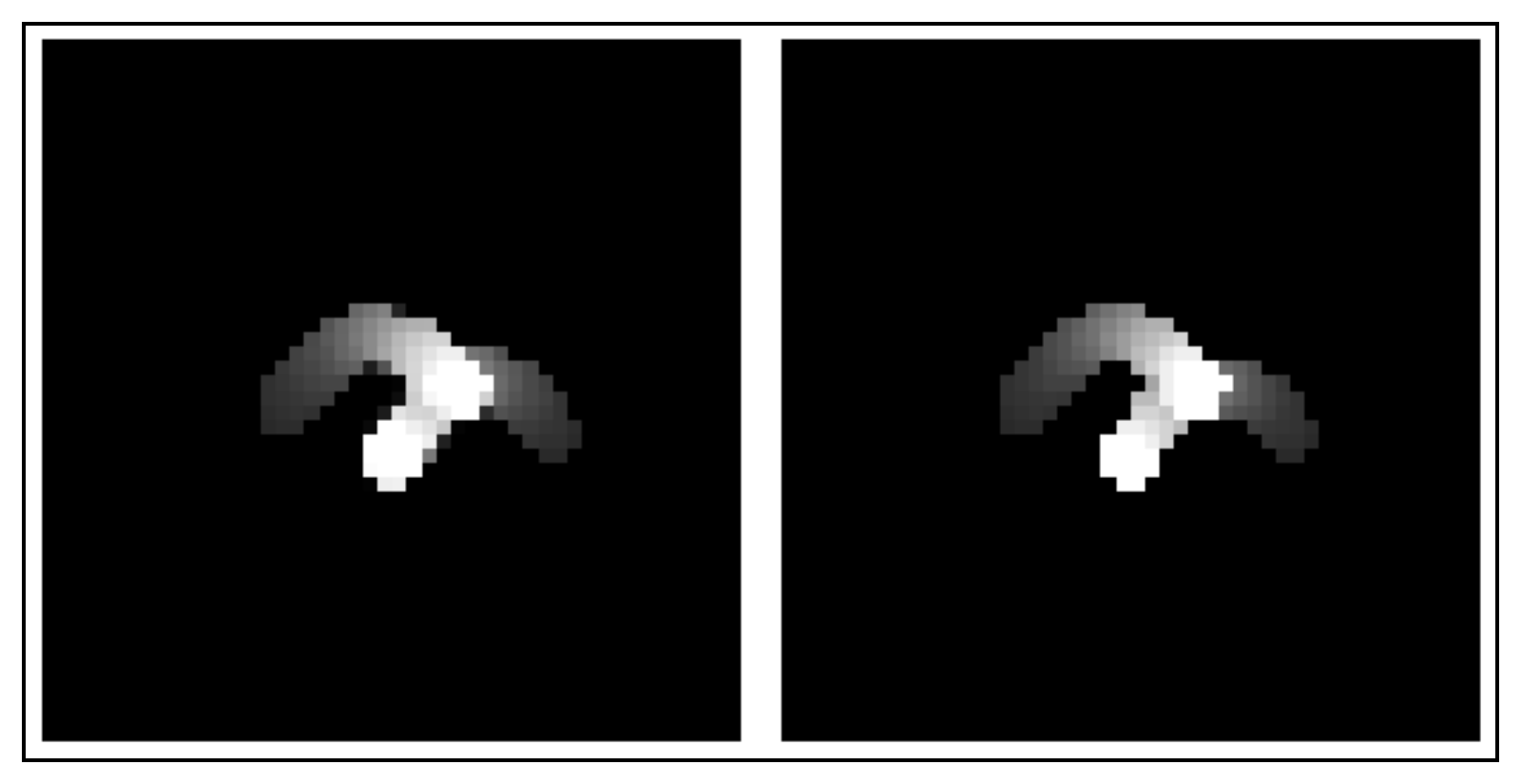}
\end{minipage} \\
\vskip0.4cm
\begin{minipage}{0.47\textwidth}
\centering
\includegraphics[height=0.420\textwidth,width=0.420\textwidth]{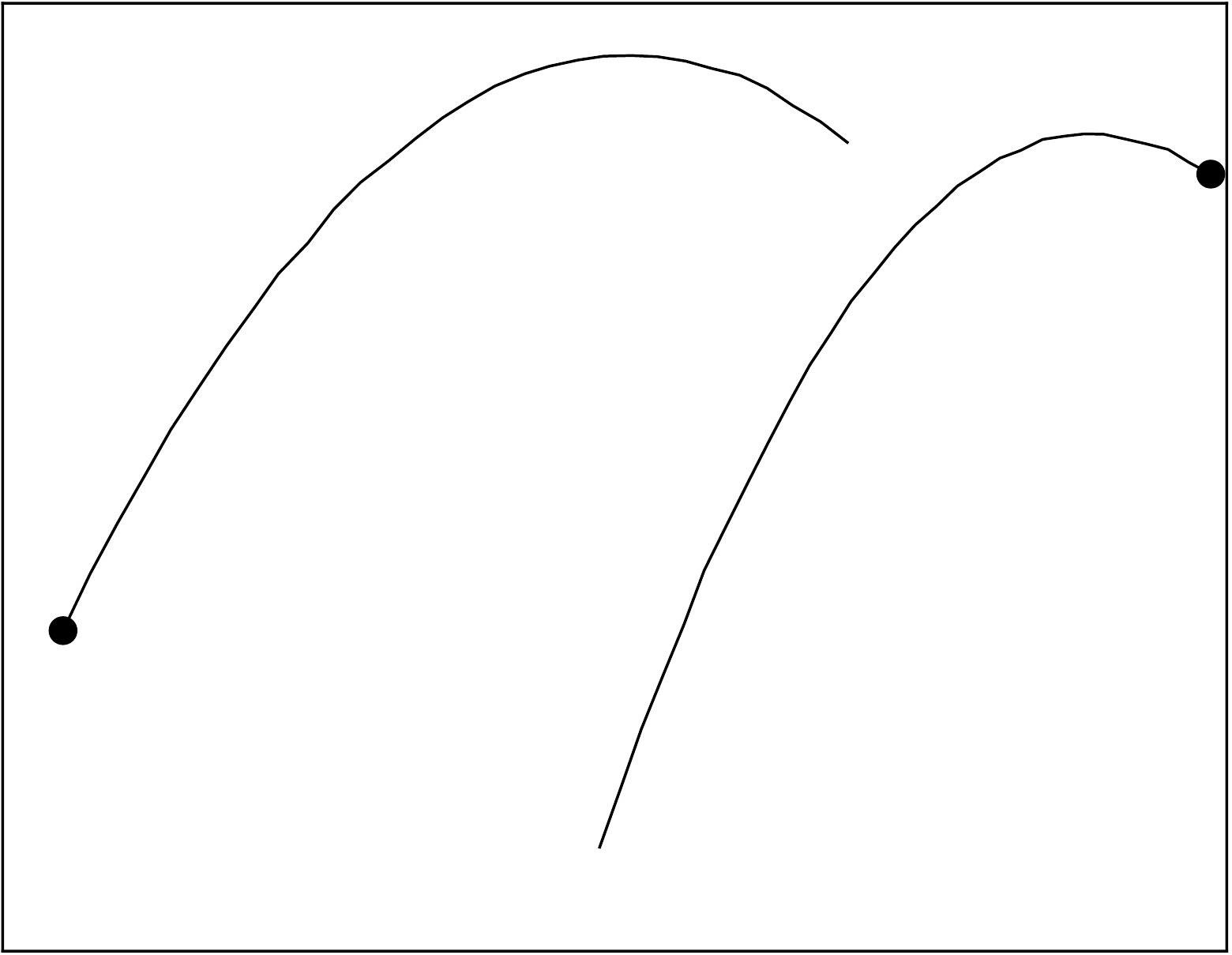}
\includegraphics[height=0.420\textwidth,width=0.420\textwidth]{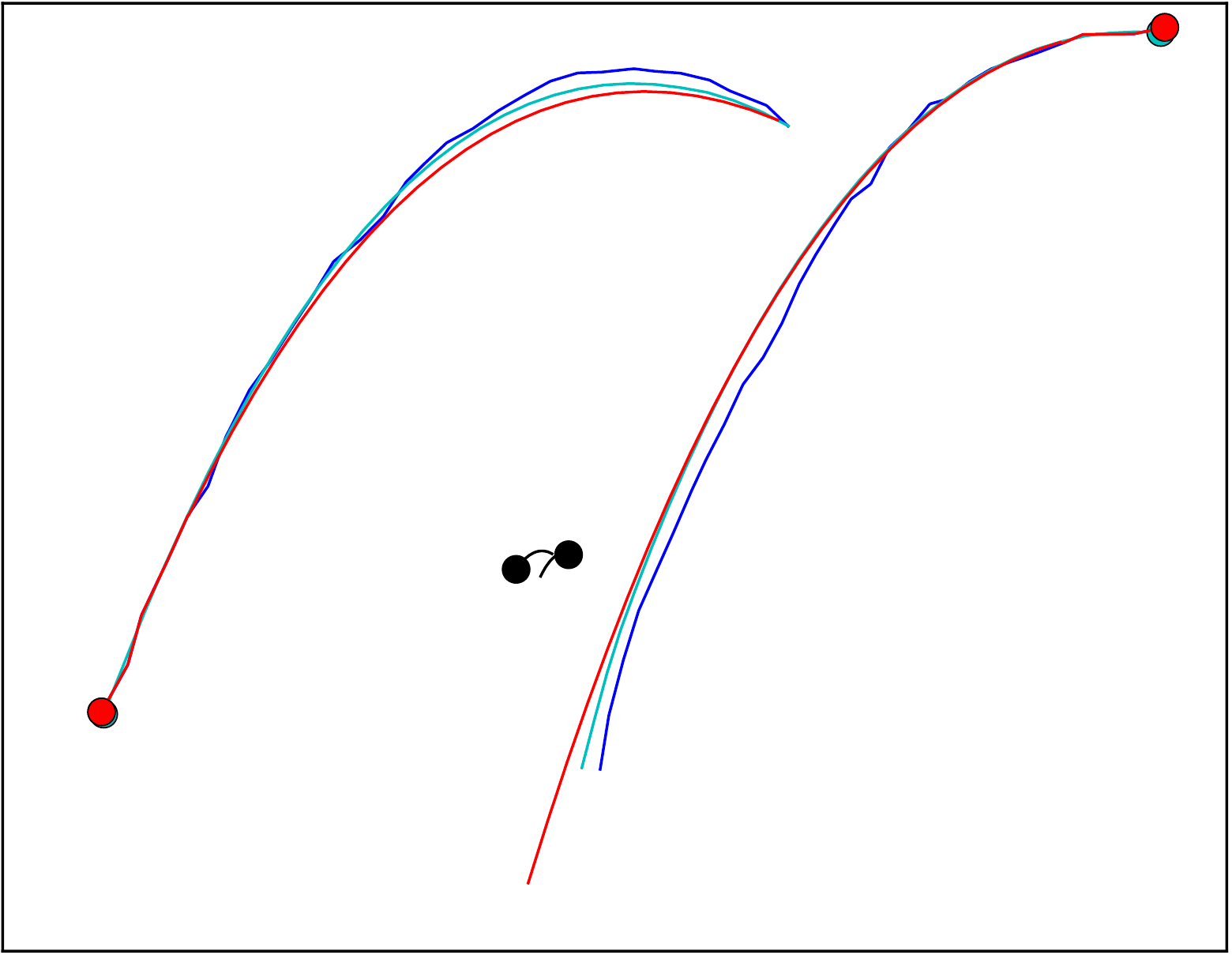} \\
\vspace{4pt}
\includegraphics[width=0.450\textwidth]{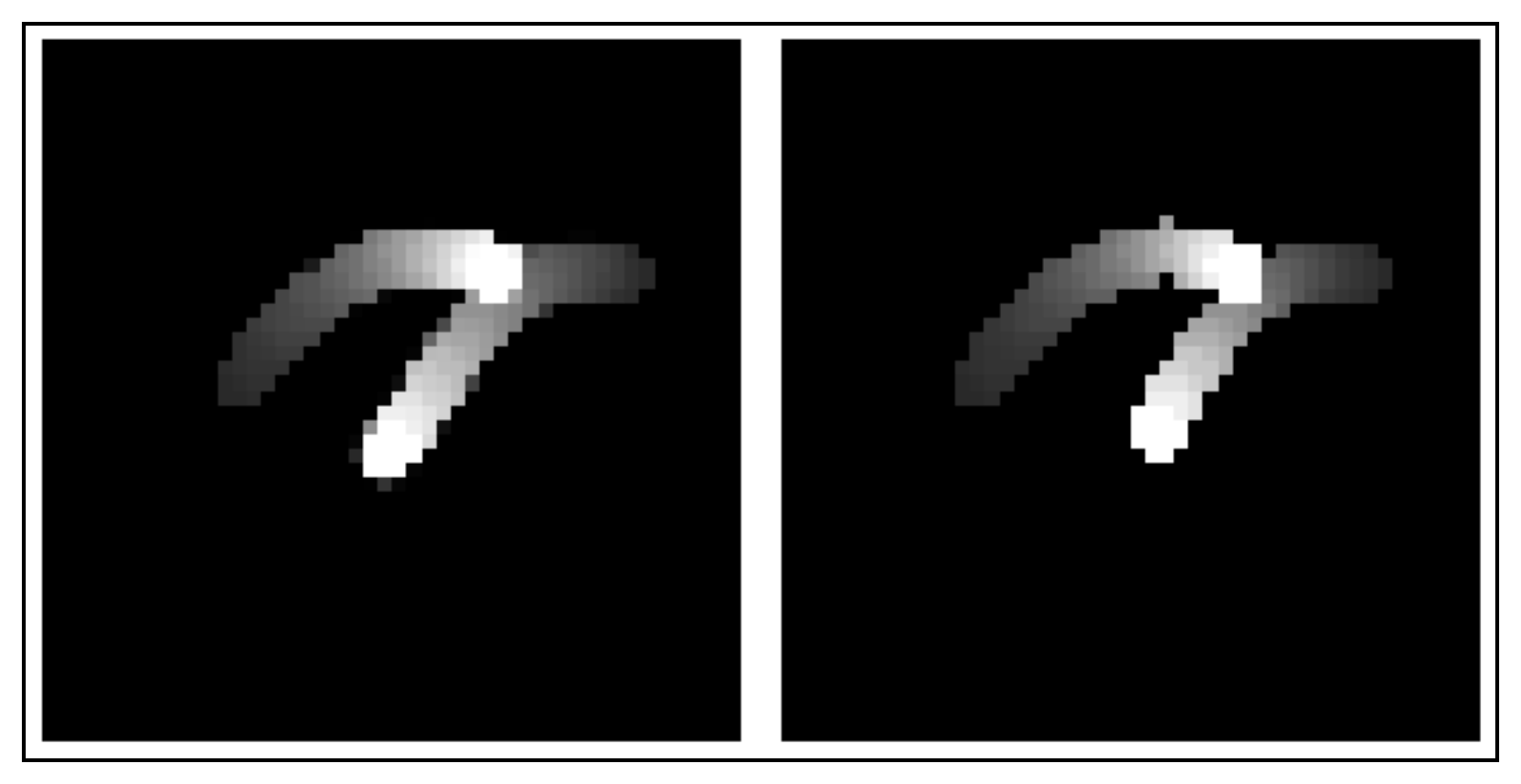}
\includegraphics[width=0.450\textwidth]{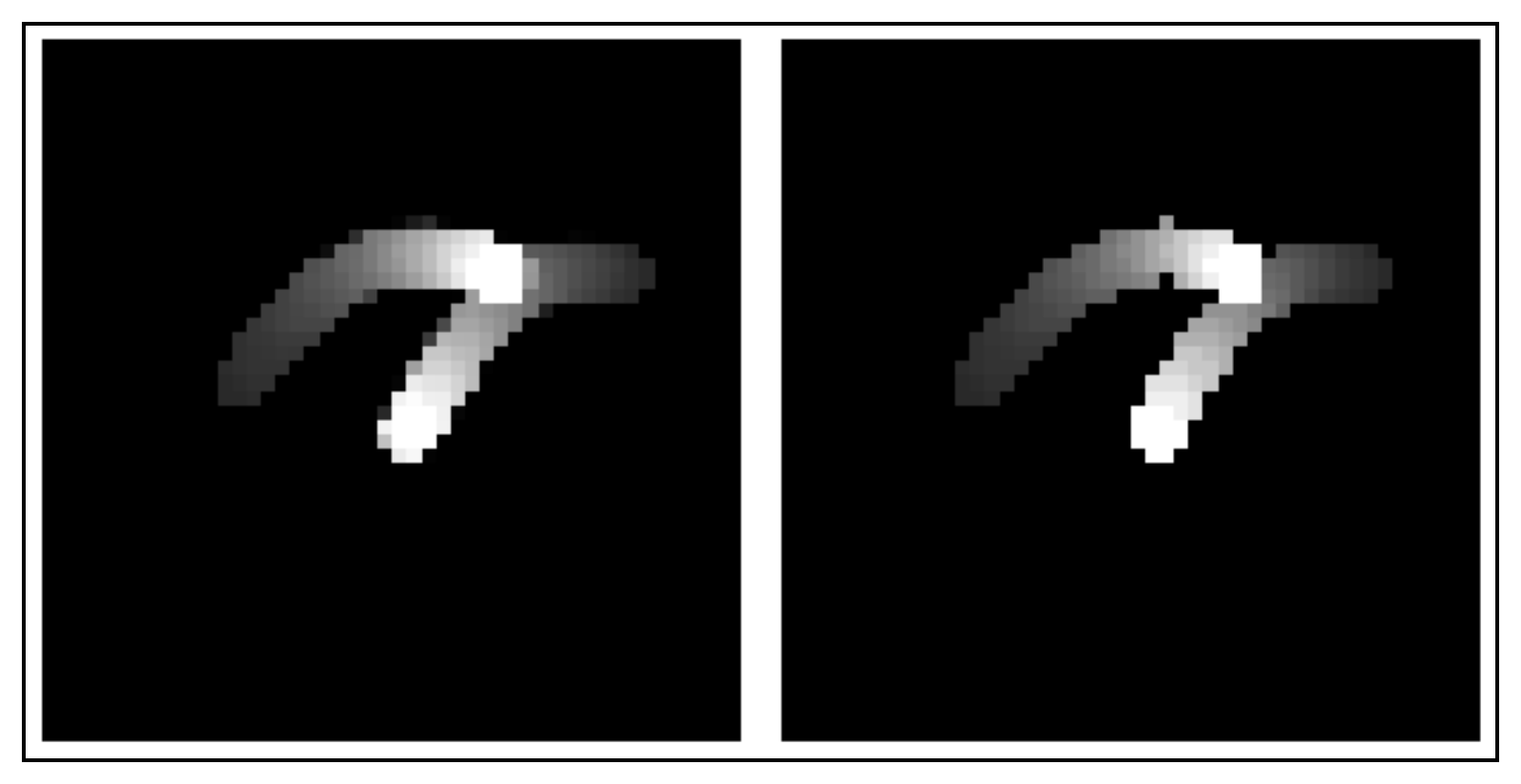}
\end{minipage}
\begin{minipage}{0.47\textwidth}
\centering
\includegraphics[height=0.420\textwidth,width=0.420\textwidth]{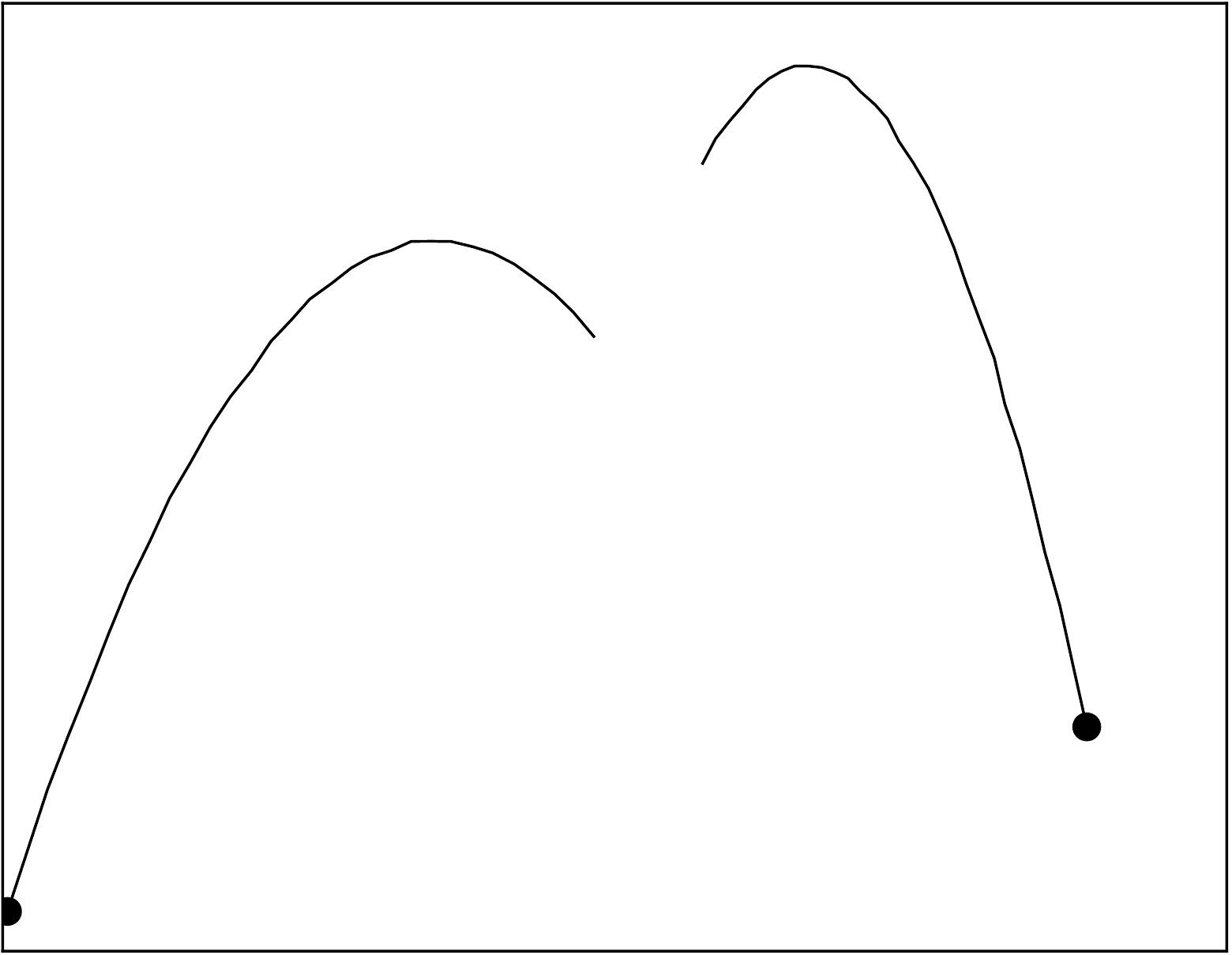}
\includegraphics[height=0.420\textwidth,width=0.420\textwidth]{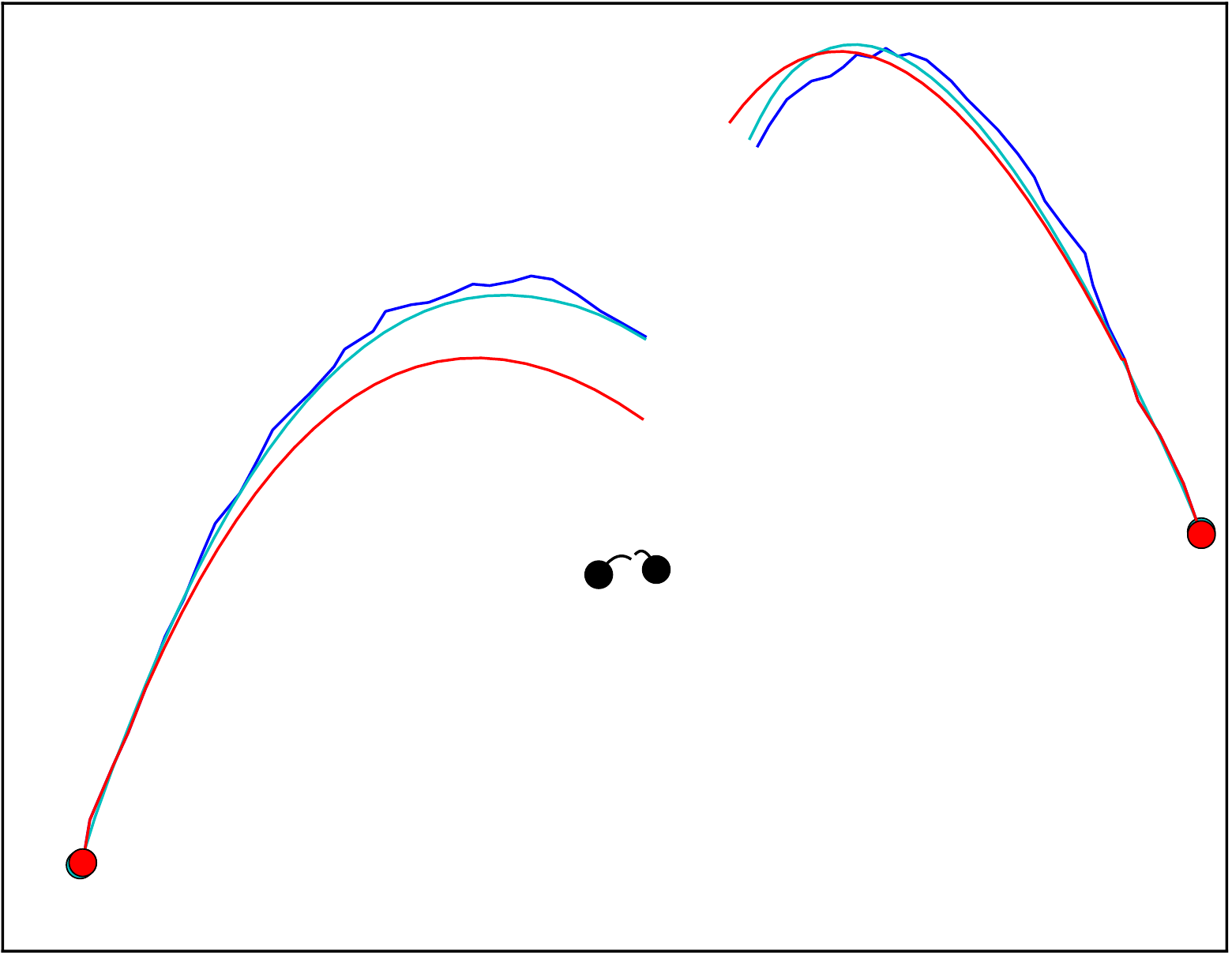} \\
\vspace{4pt}
\includegraphics[width=0.450\textwidth]{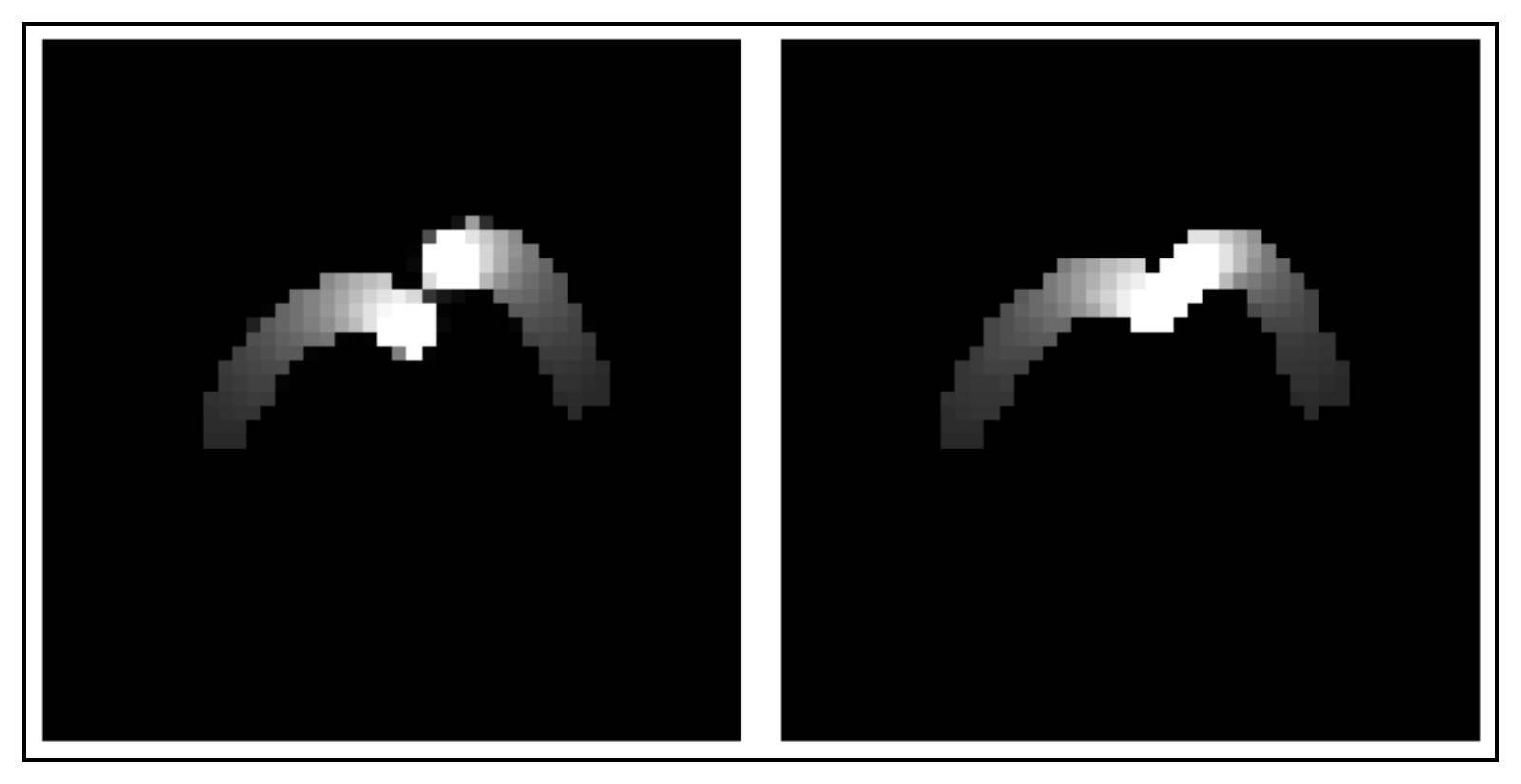}
\includegraphics[width=0.450\textwidth]{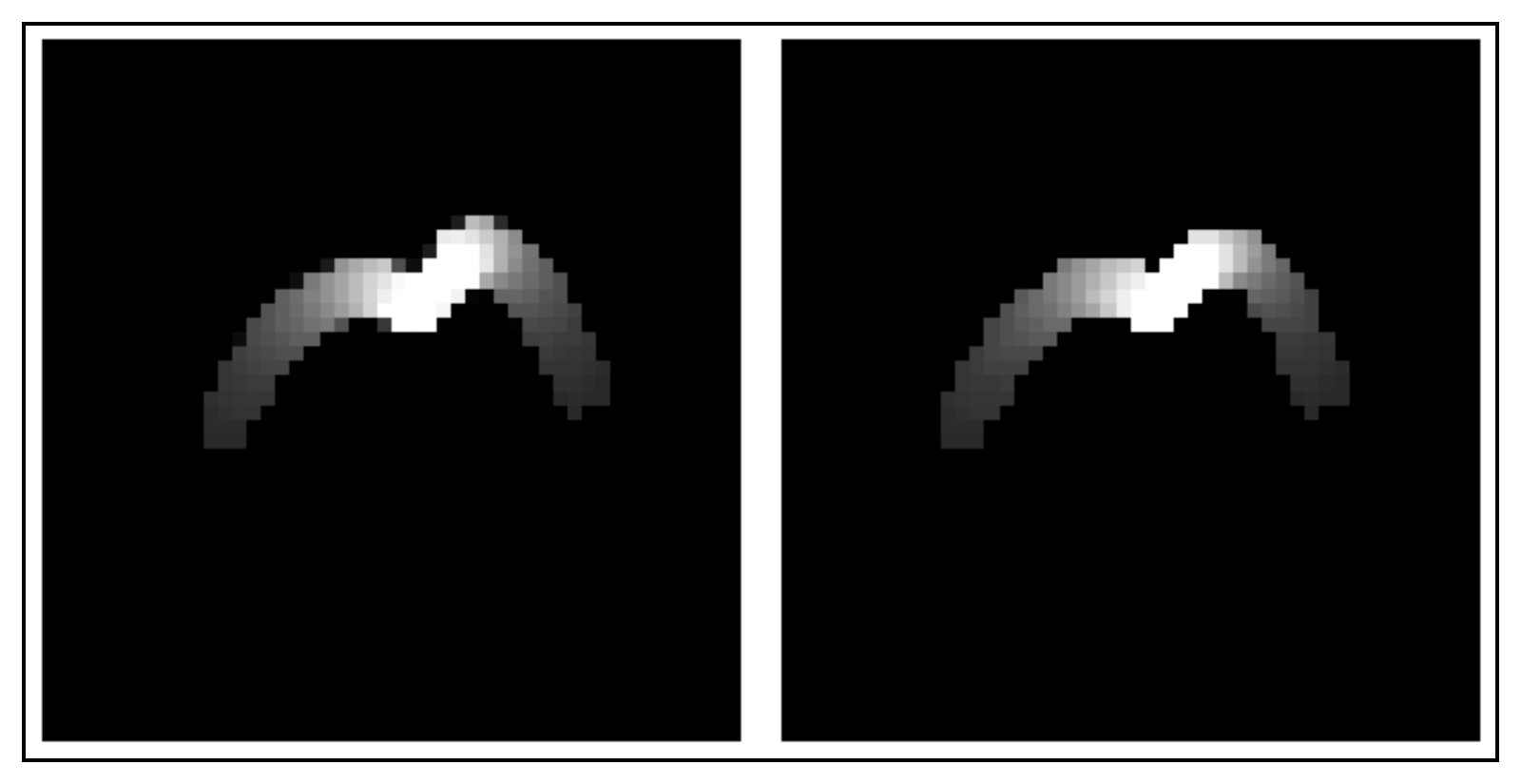}
\end{minipage} \\
\caption{Four examples of generation and interpolation. Top: Ground-truth (black), inferred (blue), generated (red), and interpolated (cyan) trajectories. Bottom: Generated (left) versus ground-truth (right) images, and interpolated (left) versus ground-truth (right) images overlaid in time.}
\label{fig:Inter}
\end{figure}
\subsection{Inference using past and future observations}
Our model can interpolate missing images (and positions) from past and future images, a task that cannot be solved by the ED-LSTM without model adjustment and retraining. 

We evaluated how our model performs in inferring the latent positions and images in the intermediate time-steps $t=6,\ldots,25$, based on observing the first and last five images $v_{1:5}$ and $v_{26:30}$. The model should be able to use information from the future to produce more accurate latent positions than the ones that would be estimated by forward generation as in \figref{fig:InferGenPos}.

To solve the task, we first used $v_{1:5}$ to obtain $\mu_{\phi}(s^n_{1}), \ldots, \mu_{\phi}(s^n_{5})$ and the most likely mixture component $k^*$; images at time-steps $t=6,\ldots,25$ were then generated by the mechanism explained above. 
The inference network was run with the generated images to obtain a warmed-in state $s_{25}$, which was then used as initial state for another run of the inference network with 
observed images $v_{26:30}$ to infer $\mu_{\phi}(s^n_{26}), \ldots, \mu_{\phi}(s^n_{30})$.
We finally used $\mu_{\phi}(s^n_{1}),\ldots,\mu_{\phi}(s^n_{5})$ and $\mu_{\phi}(s^n_{26}), \ldots, \mu_{\phi}(s^n_{30})$ as observations in a Rauch-Tung-Striebel smoother
to interpolate the missing trajectories $a_{6:25}^n$, assuming that the observations at the intermediate time-steps were missing (integrated out from the model). 

Some examples of obtained results are shown in \figref{fig:Inter} (more examples are given in the Appendix).
We show the ground-truth trajectories in black, $\mu_{\phi}(s^n_{1}),\ldots,\mu_{\phi}(s^n_{30})$ in blue, the generated trajectories in red, and the interpolated trajectories in cyan.
The interpolation \emph{corrects} the generated trajectories
by bringing them closer to $\mu_{\phi}(s^n_{1}), \ldots, \mu_{\phi}(s^n_{5})$ (obtained by observing the images at all time-steps) whilst maintaining the smoothness of dynamics constraints. 
To show the results of this correction mechanism in the pixel space, below the trajectories, from left to right, we show the generated versus ground-truth images and the interpolated versus ground-truth images, overlaid in time. 

\section{Conclusions}
This paper describes an unsupervised approach to disentangle the dynamics of objects from pixels. 
We showed that it is possible for an
inference network that is recurrent over both time and object number 
to sequentially parse each image to determine latent object positions.
The model is regularized with a mixture of linear Gaussian state-space models,
which encourages temporal coherence between latent positions extracted from images.
Whilst the considered images of cannonballs are much simpler than
those that would be encountered in most real-world applications, 
we nevertheless successfully demonstrated the usefulness of recovering interpretable latent structure in an unsupervised way and, more generally, of building structured generative models for high-dimensional visual stimuli.

\bibliography{main}

\begin{thebibliography}{10}

\bibitem{babaeizadeh18stochastic}
M.~Babaeizadeh, C.~Finn, D.~Erhan, R.~Campbell, and S.~Levine.
\newblock Stochastic variational video prediction.
\newblock In {\em 6th International Conference on Learning Representations},
  pages 1--14, 2018.

\bibitem{barshalom93estimation}
Y.~Bar-Shalom and {X. R.} Li.
\newblock {\em Estimation and Tracking: Principles, Techniques, and Software}.
\newblock Artech House, 1993.

\bibitem{barber11inferenceA}
D.~Barber, {A. T.} Cemgil, and S.~Chiappa.
\newblock Inference and estimation in probabilistic time series models.
\newblock {\em Bayesian Time Series Models}, pages 1--31, 2011.

\bibitem{blackman99design}
S.~Blackman and R.~Popoli.
\newblock {\em Design and Analysis of Modern Tracking Systems}.
\newblock Artech House, 1999.

\bibitem{chiappa06phd}
S.~Chiappa.
\newblock {\em Analysis and Classification of {EEG} Signals using Probabilistic
  Models for Brain Computer Interfaces}.
\newblock PhD thesis, EPF Lausanne, Switzerland, 2006.

\bibitem{chiappa08bayesian}
S.~Chiappa.
\newblock A {B}ayesian approach to switching linear {G}aussian state-space
  models for unsupervised time-series segmentation.
\newblock In {\em Proceedings of the Seventh International Conference on
  Machine Learning and Applications}, pages 3--9, 2008.

\bibitem{chiappa14explicit}
S.~Chiappa.
\newblock Explicit-duration {M}arkov switching models.
\newblock {\em Foundations and Trends in Machine Learning}, 7(6):803--886,
  2014.

\bibitem{chiappa17recurrent}
S.~Chiappa, S.~Racani{\`{e}}re, D.~Wierstra, and S.~Mohamed.
\newblock Recurrent environment simulators.
\newblock In {\em 5th International Conference on Learning Representations},
  2017.

\bibitem{denton17unsupervised}
{E. L.} Denton and V.~Birodkar.
\newblock Unsupervised learning of disentangled representations from video.
\newblock In {\em Advances in Neural Information Processing Systems 30}, pages
  4414--4423, 2017.

\bibitem{Finn2016}
C.~Finn, {I. J.} Goodfellow, and S.~Levine.
\newblock Unsupervised learning for physical interaction through video
  prediction.
\newblock In {\em Advances in Neural Information Processing Systems 29}, pages
  64--72, 2016.

\bibitem{fraccaro17disentangled}
M.~Fraccaro, S.~Kamronn, U.~Paquet, and O.~Winther.
\newblock A disentangled recognition and nonlinear dynamics model for
  unsupervised learning.
\newblock In {\em Advances in Neural Information Processing Systems 30}, pages
  3604--3613, 2017.

\bibitem{Fraccaro2016}
M.~Fraccaro, {S. K.} S{\o}nderby, U.~Paquet, and O.~Winther.
\newblock Sequential neural models with stochastic layers.
\newblock In {\em Advances in Neural Information Processing Systems 29}, pages
  2199--2207, 2016.

\bibitem{Gao2016}
Y.~Gao, {E. W.} Archer, L.~Paninski, and {J. P.} Cunningham.
\newblock Linear dynamical neural population models through nonlinear
  embeddings.
\newblock In {\em Advances in Neural Information Processing Systems 29}, pages
  163--171, 2016.

\bibitem{hochreiter97long}
S.~Hochreiter and J.~Schmidhuber.
\newblock Long short-term memory.
\newblock {\em Neural Computation}, 9(8):1735--1780, 1997.

\bibitem{johnson16composing}
M.~Johnson, {D. K.} Duvenaud, A.~Wiltschko, {R. P.} Adams, and {S. R.} Datta.
\newblock Composing graphical models with neural networks for structured
  representations and fast inference.
\newblock In {\em Advances in Neural Information Processing Systems 29}, pages
  2946--2954, 2016.

\bibitem{kingma14autoencoding}
{D. P.} Kingma and M.~Welling.
\newblock Auto-encoding variational {B}ayes.
\newblock In {\em 2nd International Conference on Learning Representations},
  2014.

\bibitem{Krishnan2017}
R.~Krishnan, U.~Shalit, and D.~Sontag.
\newblock Structured inference networks for nonlinear state space models.
\newblock In {\em Proceedings of the Thirty-First {AAAI} Conference on
  Artificial Intelligence}, pages 2101--2109, 2017.

\bibitem{lin18variational}
W.~Lin, N.~Hubacher, and {M. E.} Khan.
\newblock Variational message passing with structured inference networks.
\newblock In {\em 6th International Conference on Learning Representations},
  2018.

\bibitem{Oh2015}
J.~Oh, X.~Guo, H.~Lee, {R. L.} Lewis, and S.~Singh.
\newblock Action-conditional video prediction using deep networks in {A}tari
  games.
\newblock In {\em Advances in Neural Information Processing Systems 28}, pages
  2863--2871, 2015.

\bibitem{pearce2018comparing}
M.~Pearce, S.~Chiappa, and U.~Paquet.
\newblock Comparing interpretable inference models for videos of physical
  motion.
\newblock In {\em Symposium on Advances in Approximate Bayesian Inference},
  2018.

\bibitem{rezende14stochastic}
{D. J.} Rezende, S.~Mohamed, and D.~Wierstra.
\newblock Stochastic backpropagation and approximate inference in deep
  generative models.
\newblock In {\em Proceedings of the 31st International Conference on Machine
  Learning}, pages 1278--1286, 2014.

\bibitem{srivastava15unsupervised}
N.~Srivastava, E.~Mansimov, and R.~Salakhutdinov.
\newblock Unsupervised learning of video representations using {LSTMs}.
\newblock In {\em Proceedings of the 32nd International Conference on Machine
  Learning}, pages 843--852, 2015.

\bibitem{Sun2016}
W.~Sun, A.~Venkatraman, B.~Boots, and {J. A.} Bagnell.
\newblock Learning to filter with predictive state inference machines.
\newblock In {\em Proceedings of the 32nd International Conference on Machine
  Learning}, pages 1197--1205, 2016.

\bibitem{watters17visual}
N.~Watters, A.~Tacchetti, T.~Weber, R.~Pascanu, P.~Battaglia, and D.~Zoran.
\newblock Visual interaction networks.
\newblock {\em CoRR}, abs/1706.01433, 2017.

\end{thebibliography}
\bibliographystyle{plain}       


\newpage

\begin{figure}[t]
\section*{Appendix. Multi-step ahead generation of images \& inference using past and future observations}
\begin{center}
\hskip-0.3cm
\includegraphics[height=1.725cm,width=3.45cm]{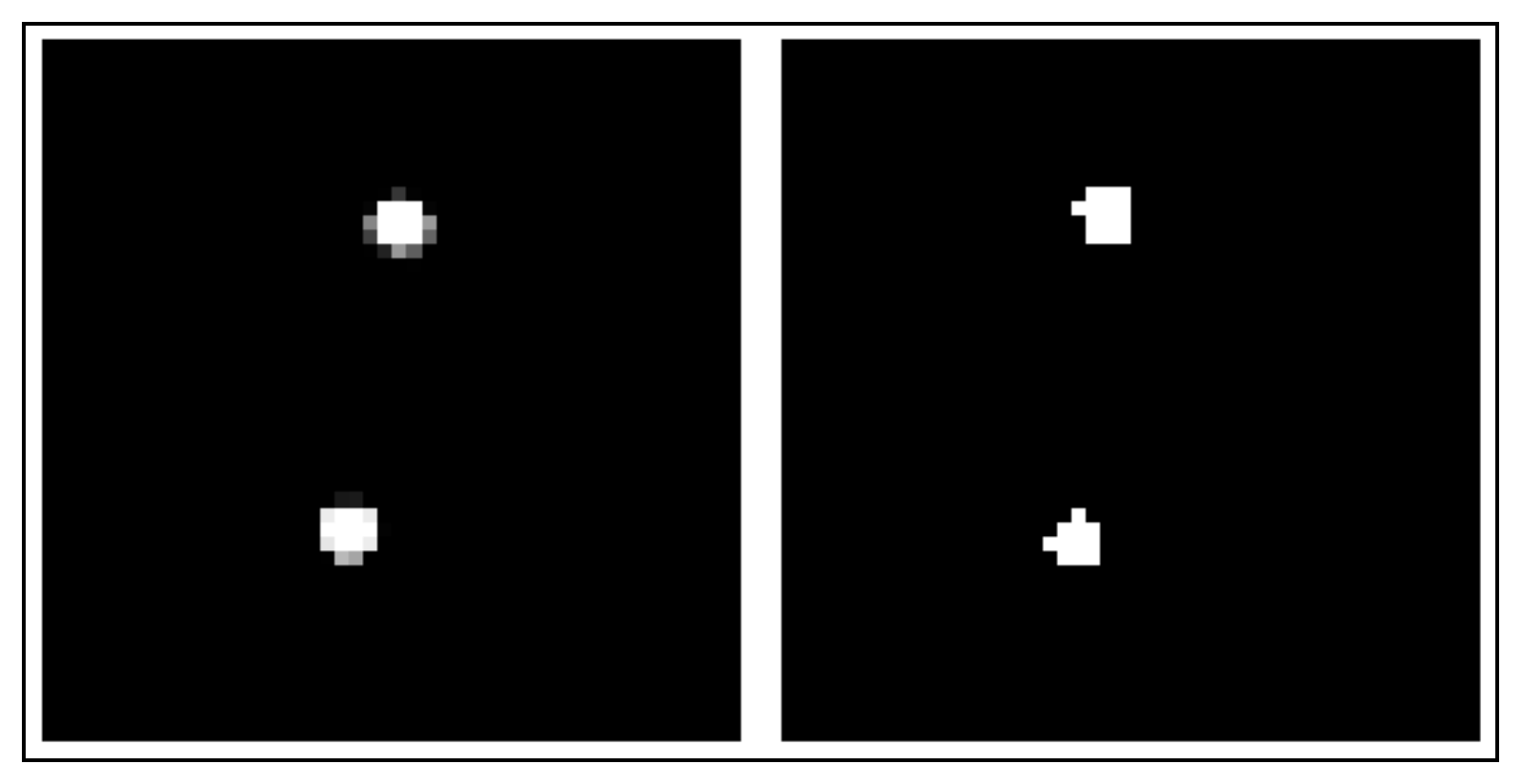}
\hskip0.04cm
\includegraphics[height=1.725cm,width=3.45 cm]{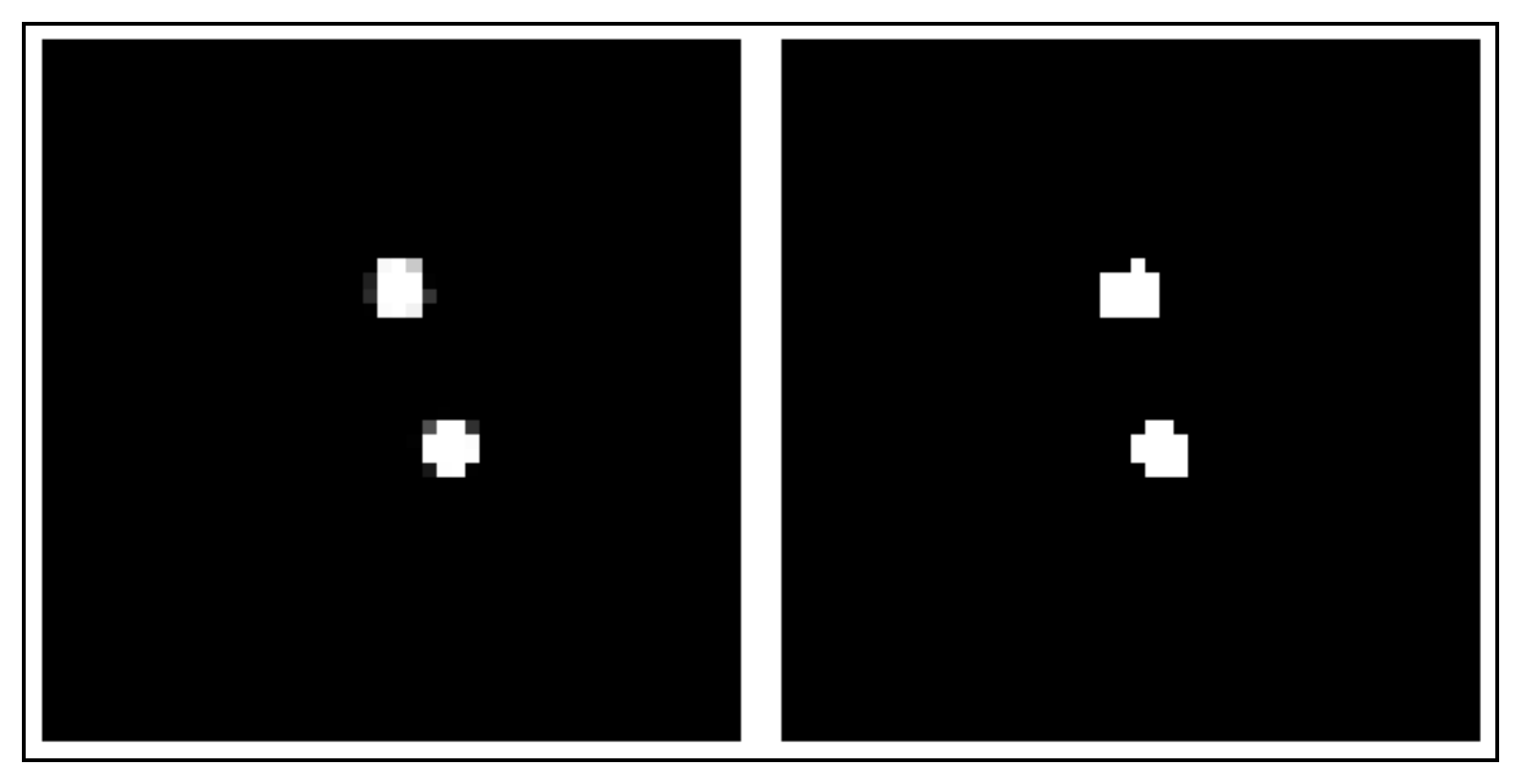}
\hskip0.04cm
\includegraphics[height=1.725cm,width=3.45 cm]{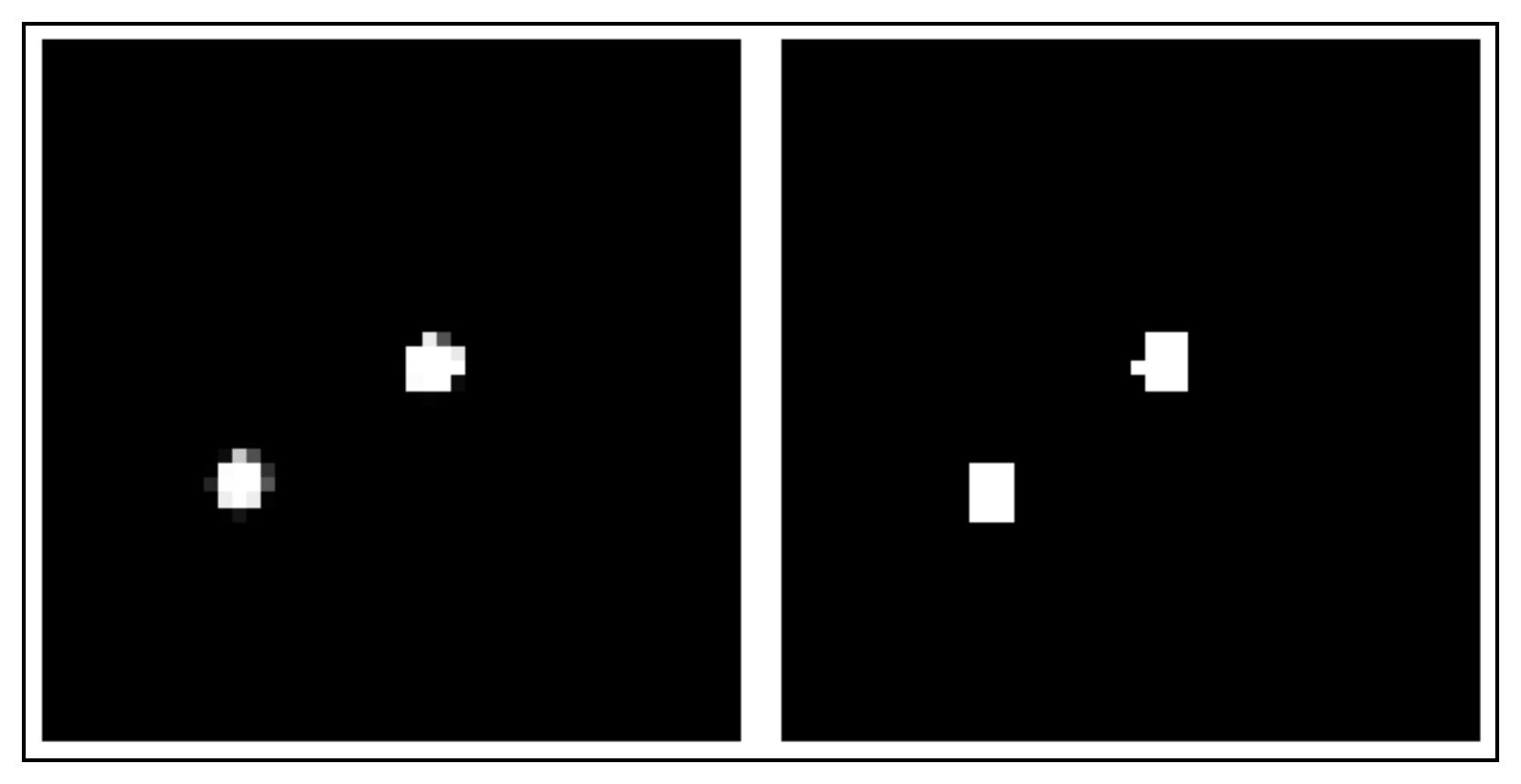}
\end{center}
\begin{center}
\hskip-0.3cm
\includegraphics[height=1.725cm,width=3.45 cm]{./fig/exp_default_no5_ccov_delta_48_tm30_h5_ppg_seed340/340_144000_gen_obs_frames_2_all.pdf}
\hskip0.04cm
\includegraphics[height=1.725cm,width=3.45 cm]{./fig/exp_default_no5_ccov_delta_48_tm30_h5_ppg_seed340/340_144000_gen_obs_frames_4_all.pdf}
\hskip0.04cm
\includegraphics[height=1.725cm,width=3.45 cm]{./fig/exp_default_no5_ccov_delta_48_tm30_h5_ppg_seed340/340_148000_gen_obs_frames_0_all.pdf}
\end{center}
\begin{center}
\hskip-0.3cm
\includegraphics[height=1.725cm,width=3.45 cm]{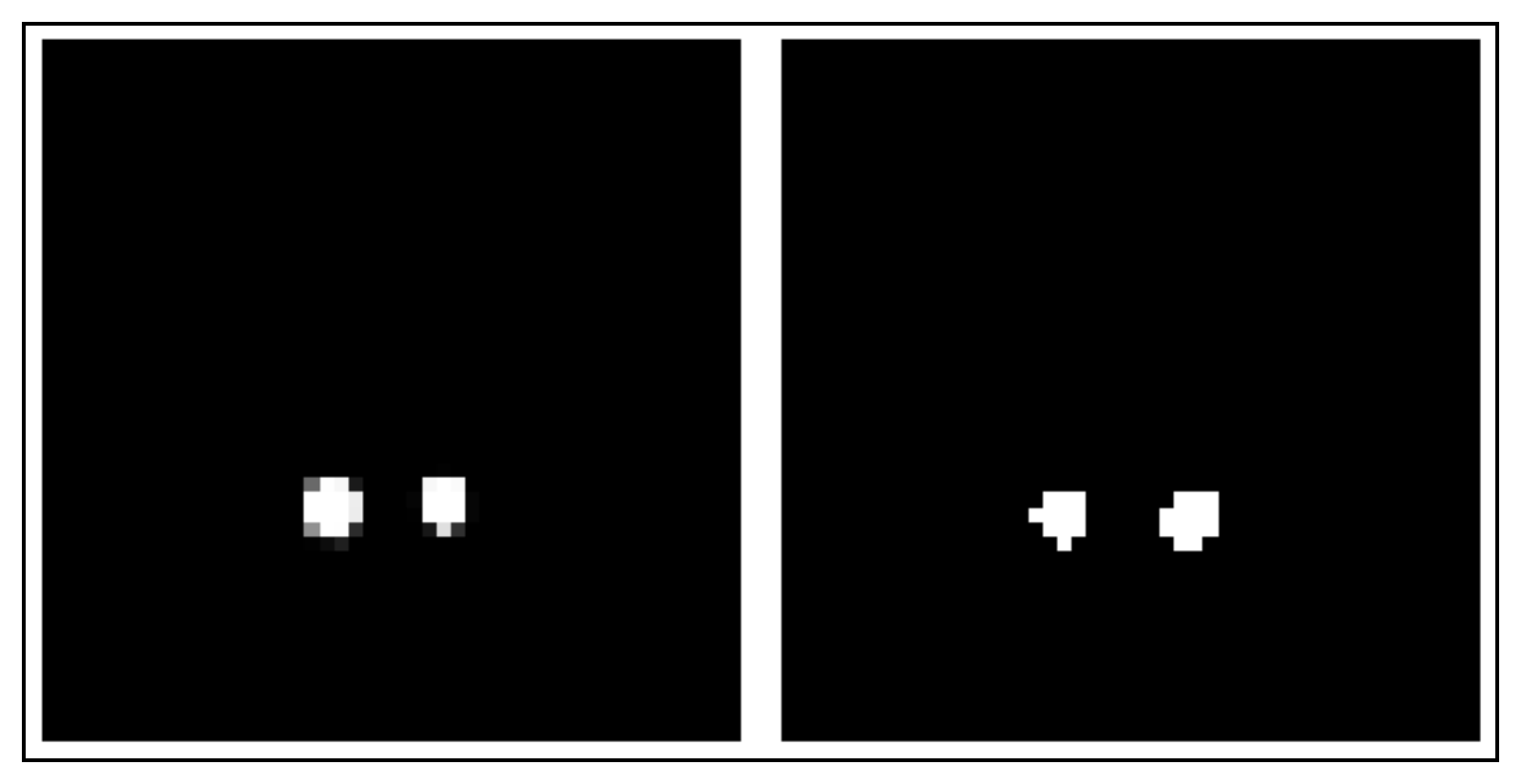}
\hskip0.04cm
\includegraphics[height=1.725cm,width=3.45 cm]{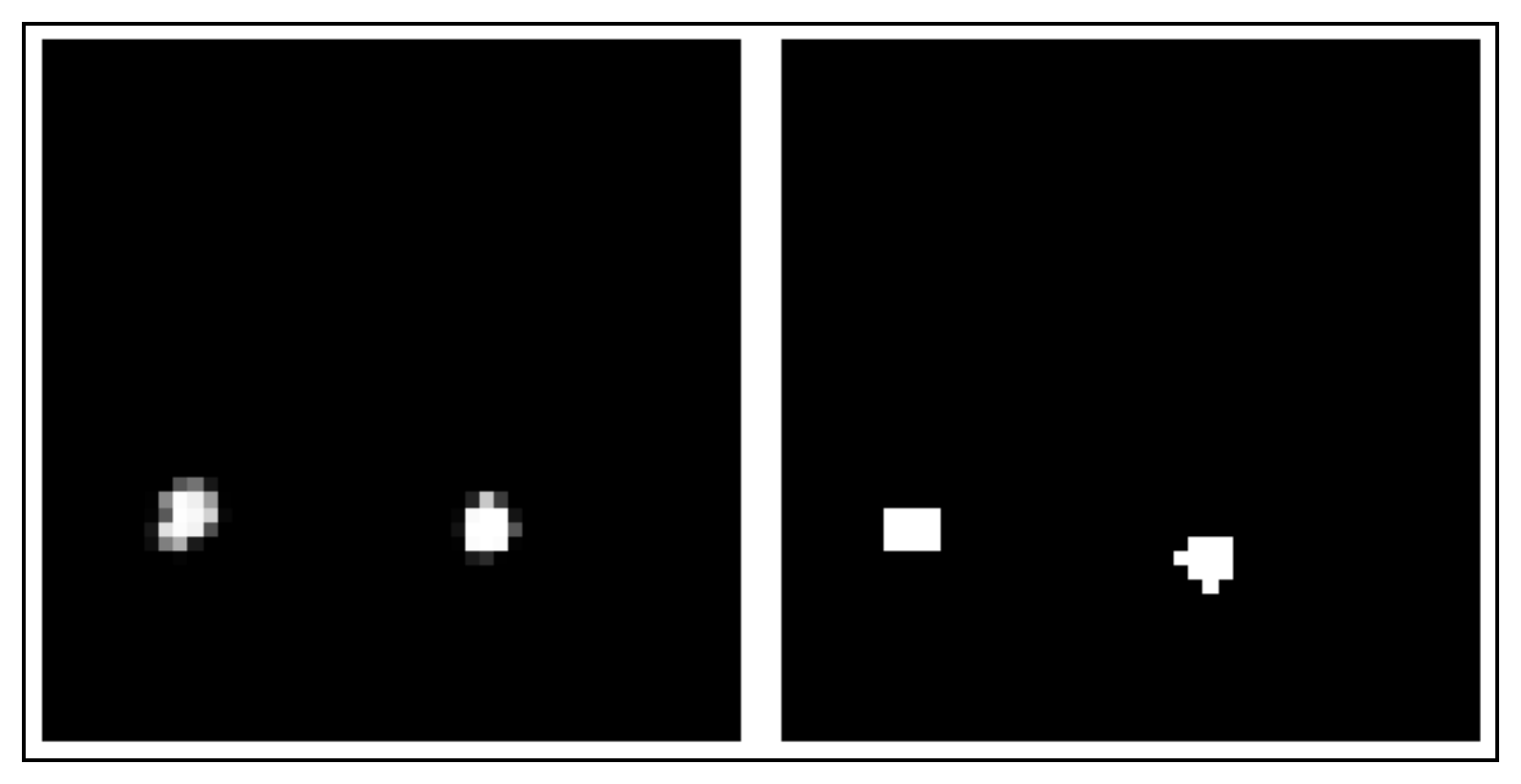}
\hskip0.04cm
\includegraphics[height=1.725cm,width=3.45 cm]{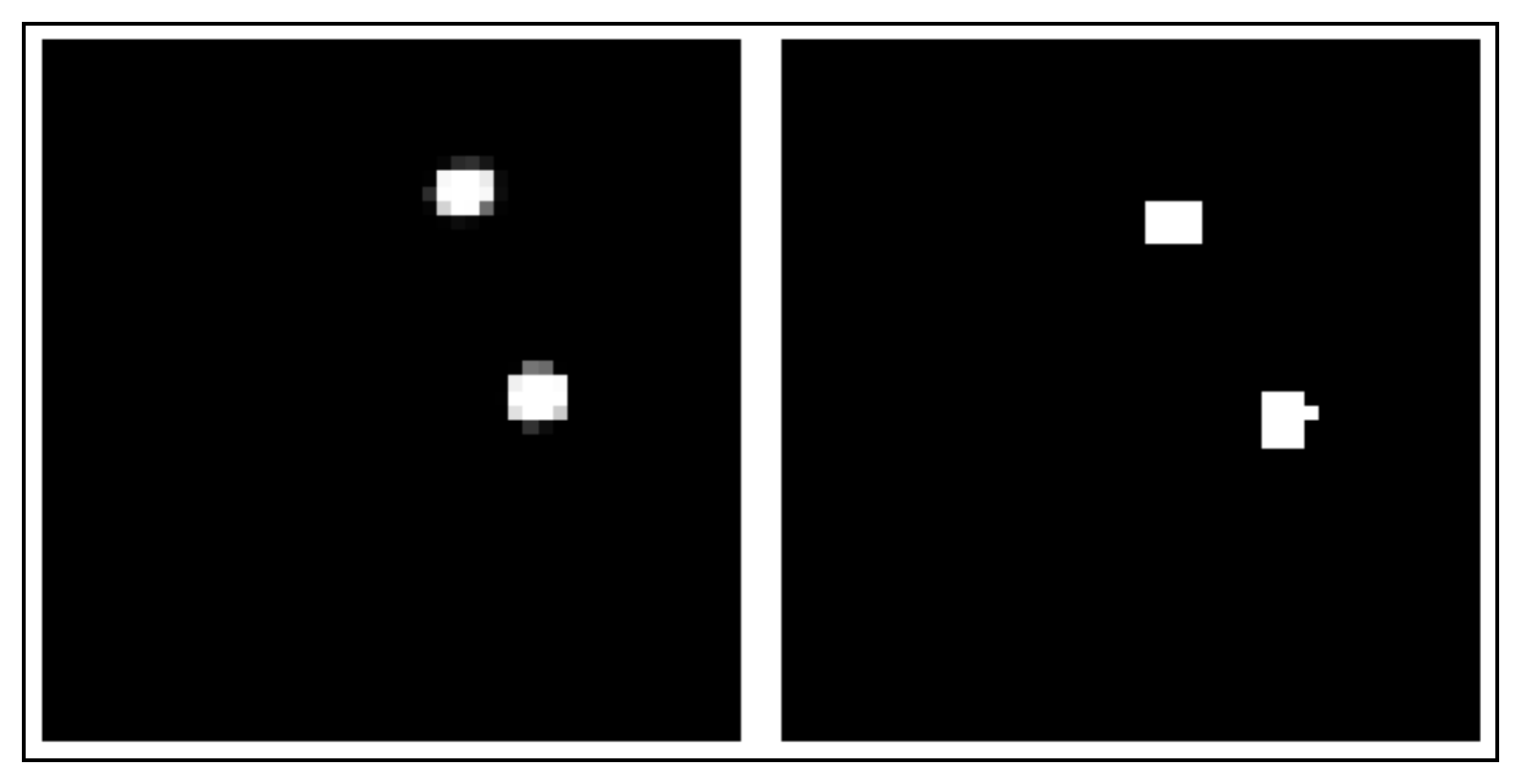}
\end{center}
\begin{center}
\hskip-0.3cm
\includegraphics[height=1.725cm,width=3.45 cm]{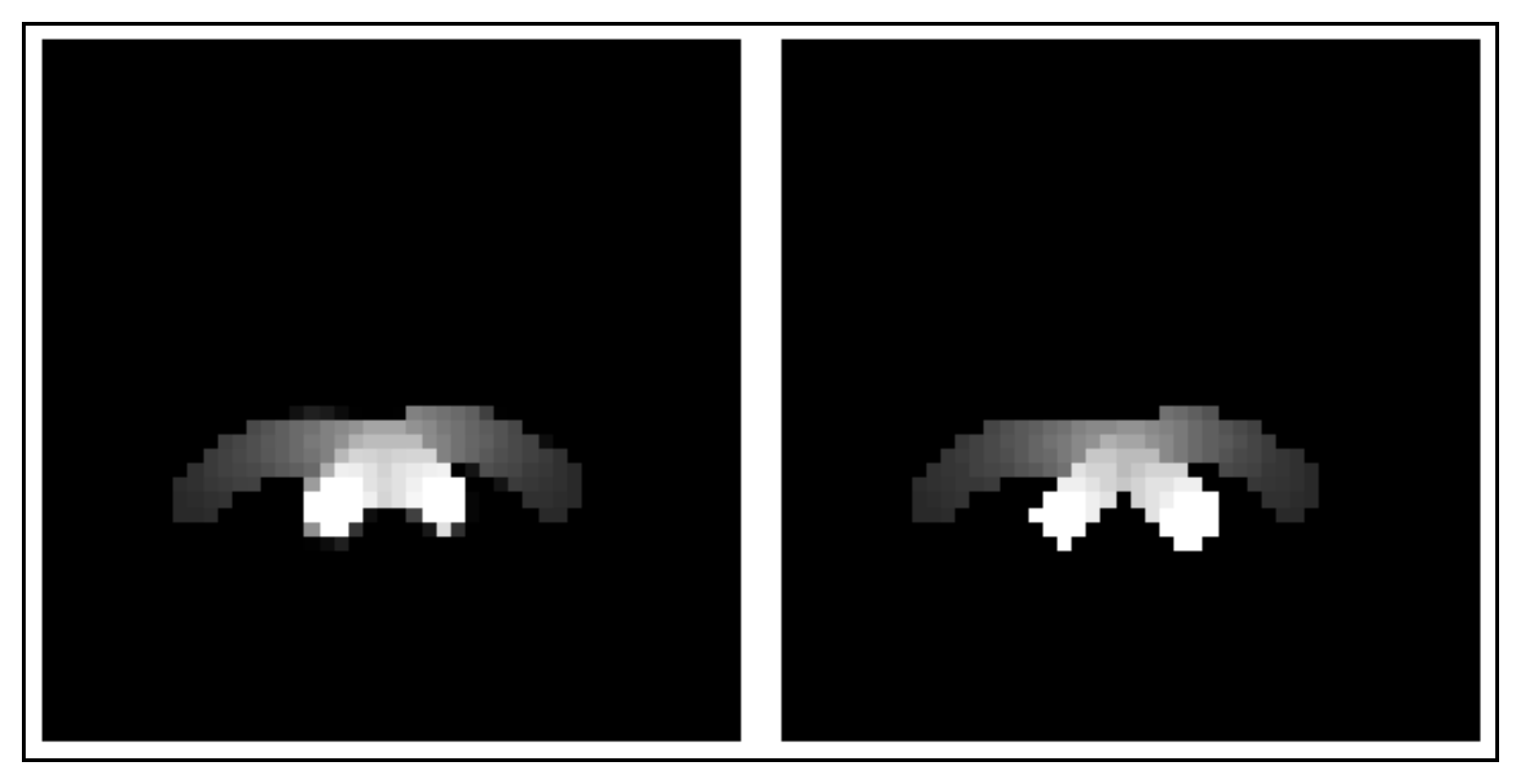}
\hskip0.04cm
\includegraphics[height=1.725cm,width=3.45 cm]{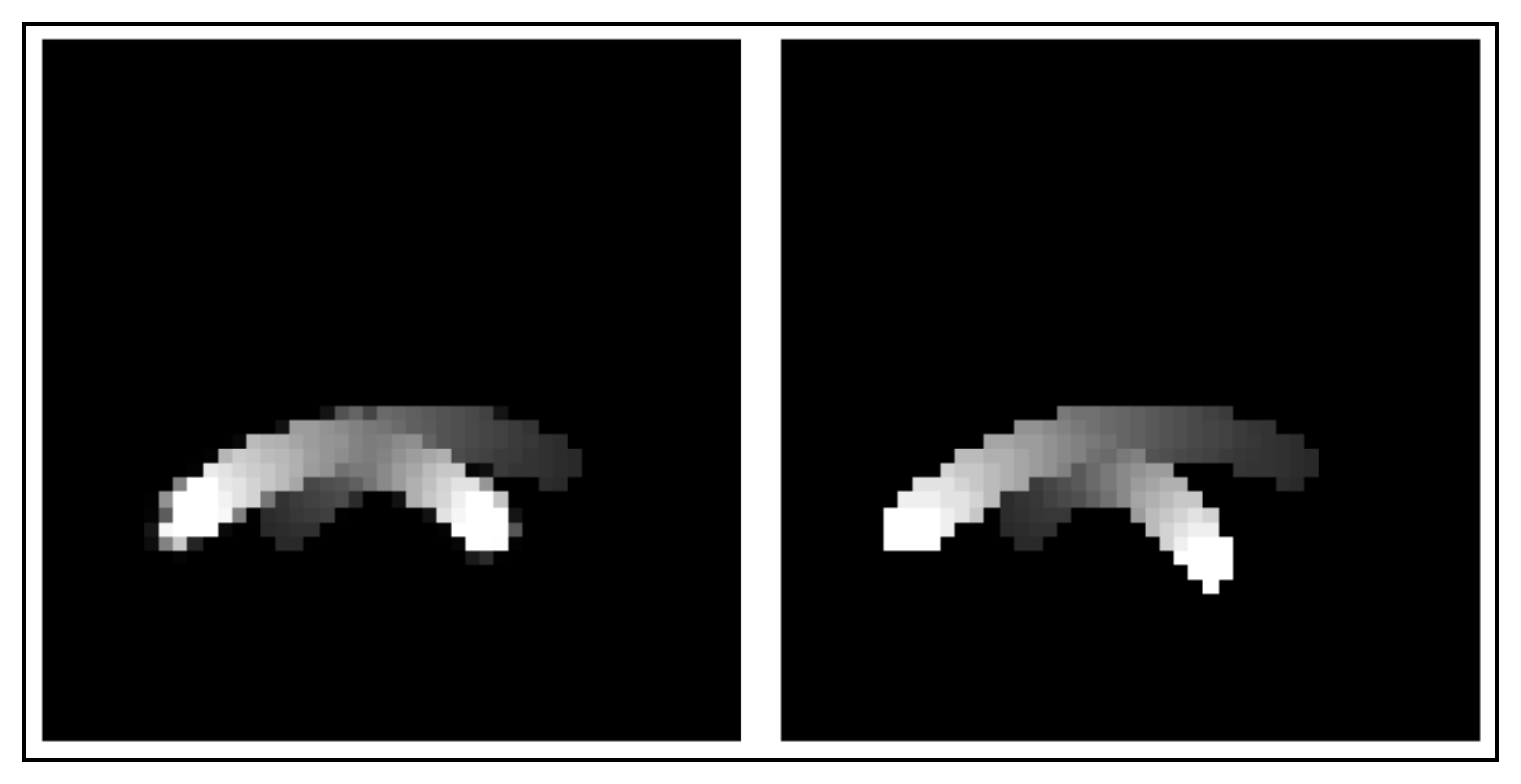}
\hskip0.04cm
\includegraphics[height=1.725cm,width=3.45 cm]{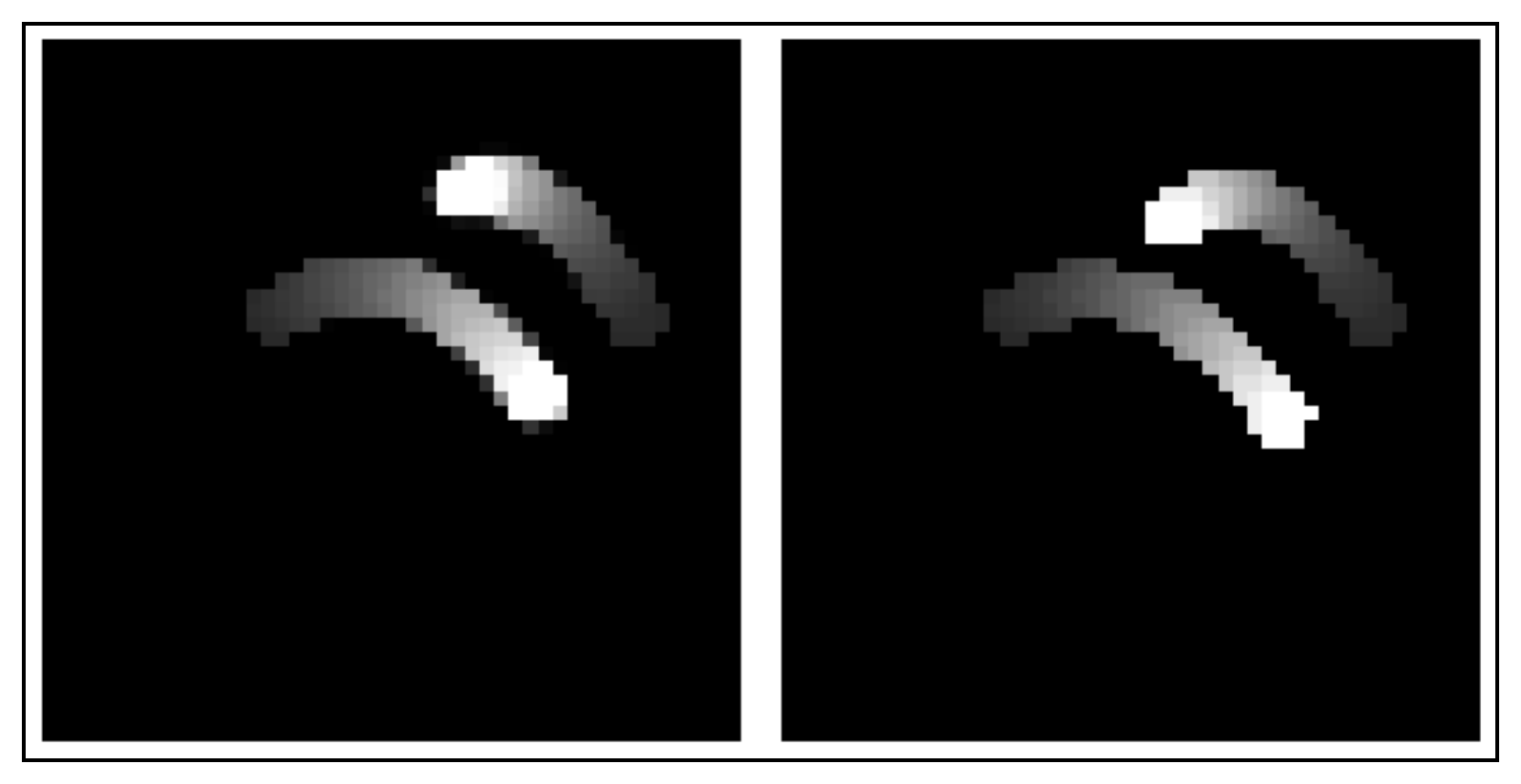}
\end{center}
\begin{center}
\hskip-0.3cm
\includegraphics[height=1.725cm,width=3.45 cm]{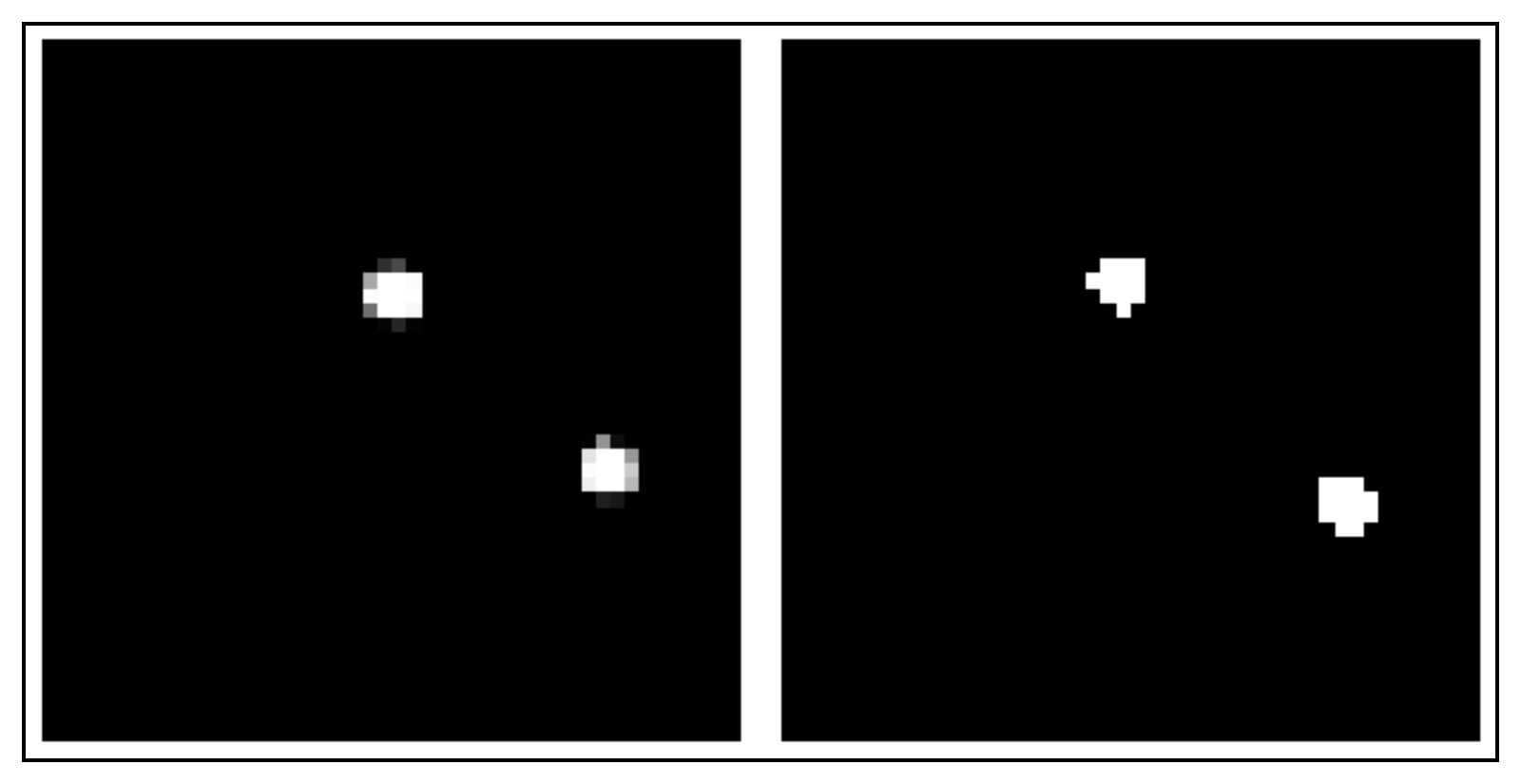}
\hskip0.04cm
\includegraphics[height=1.725cm,width=3.45 cm]{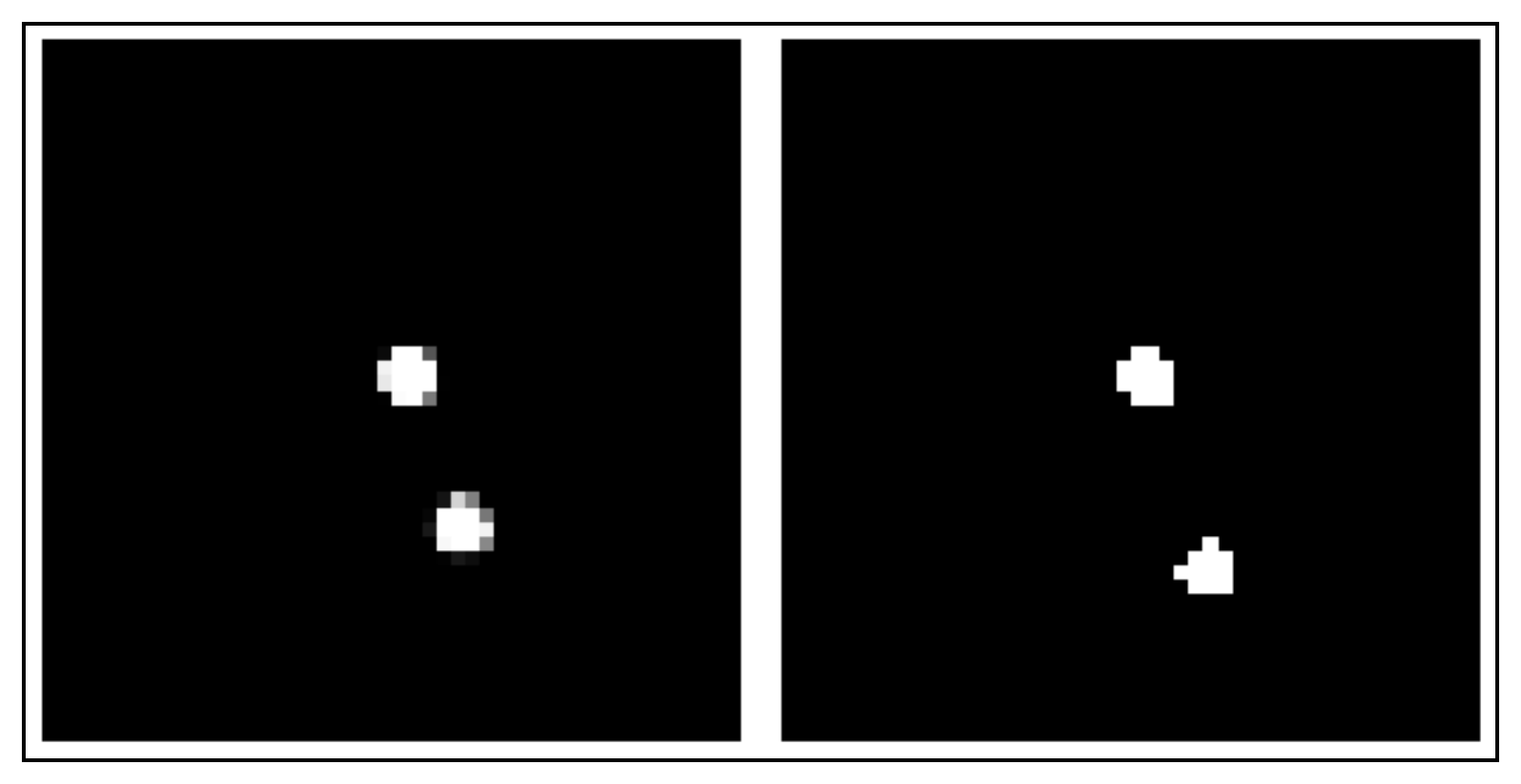}
\hskip0.04cm
\includegraphics[height=1.725cm,width=3.45 cm]{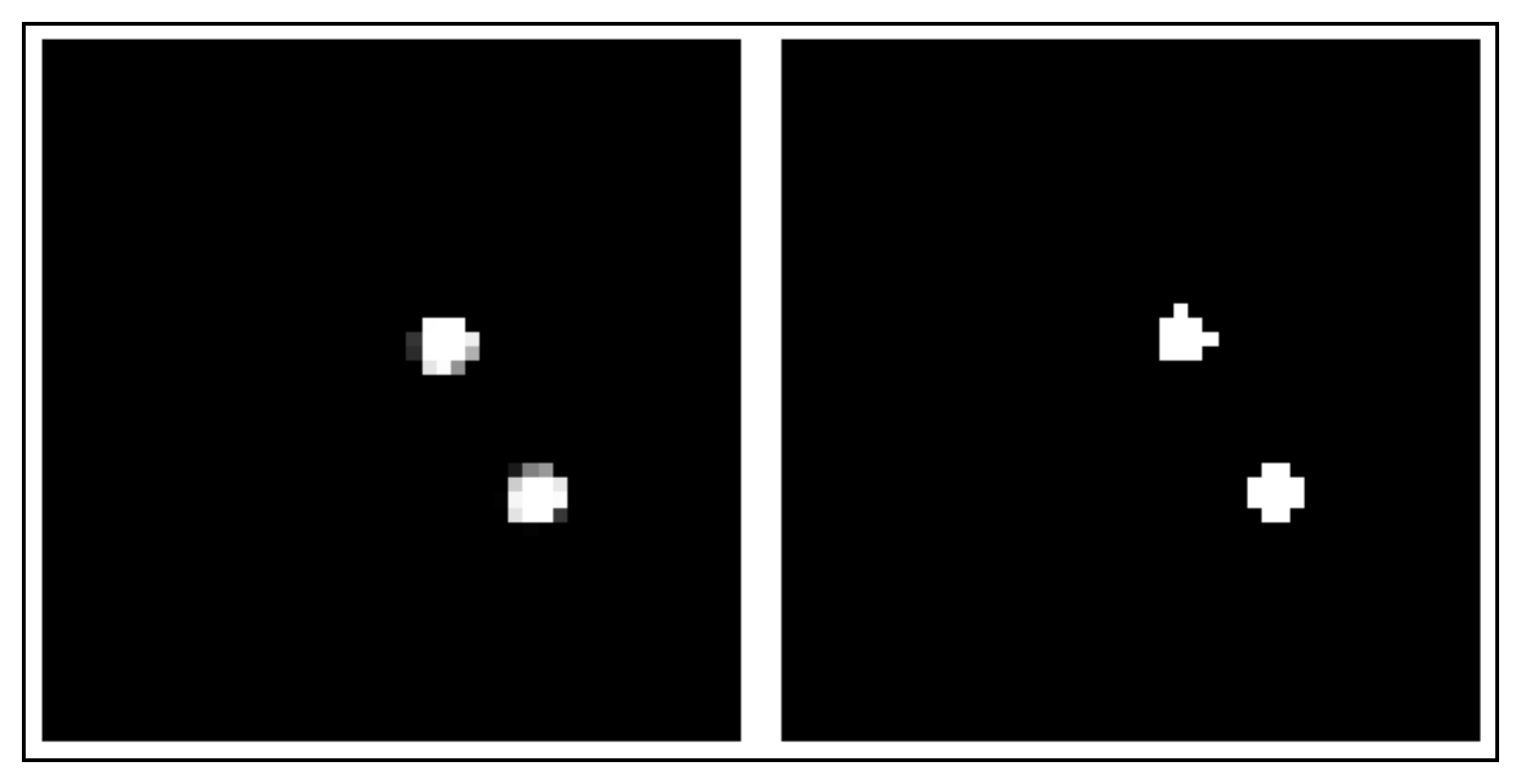}
\end{center}
\begin{center}
\hskip-0.3cm
\includegraphics[height=1.725cm,width=3.45 cm]{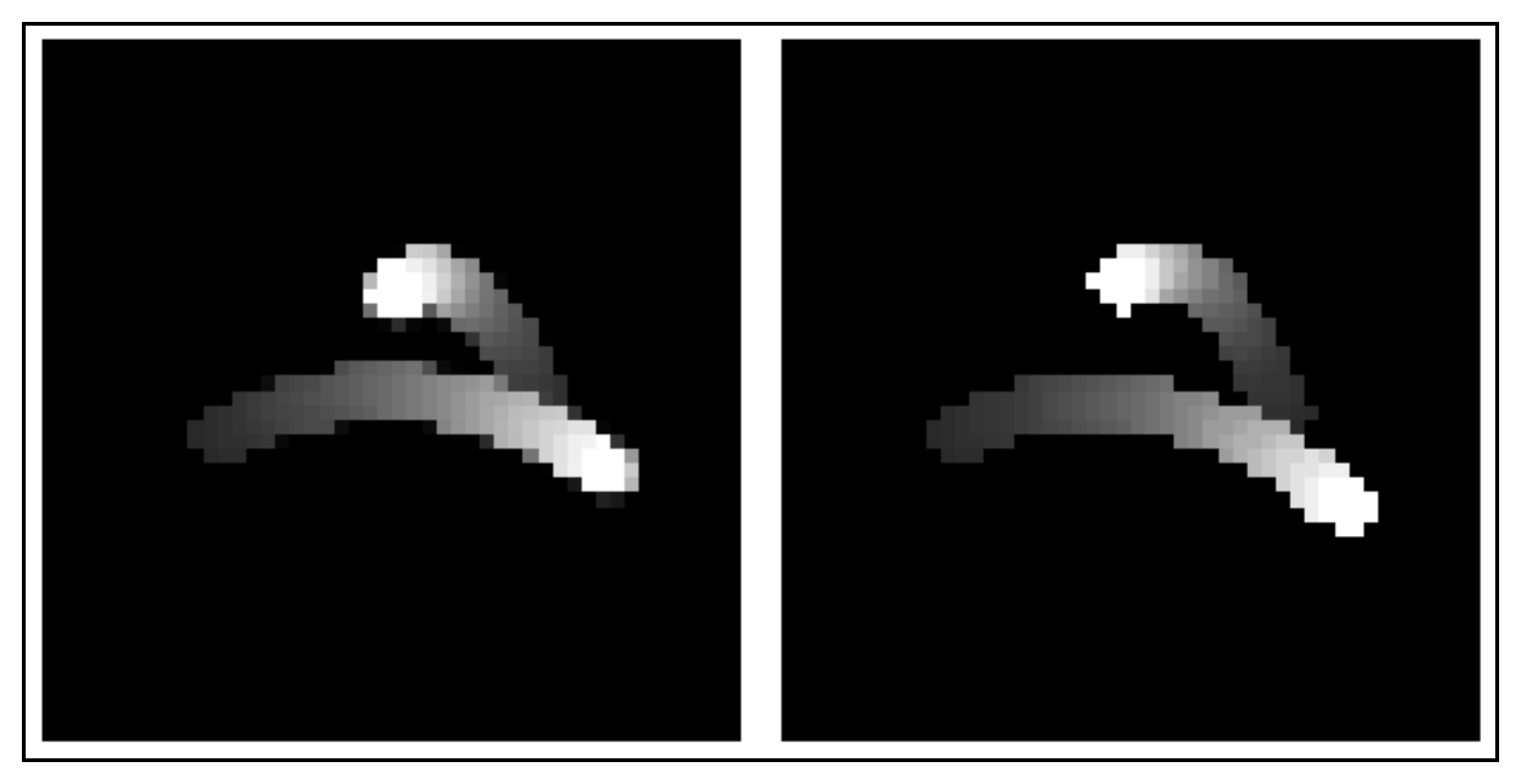}
\hskip0.04cm
\includegraphics[height=1.725cm,width=3.45 cm]{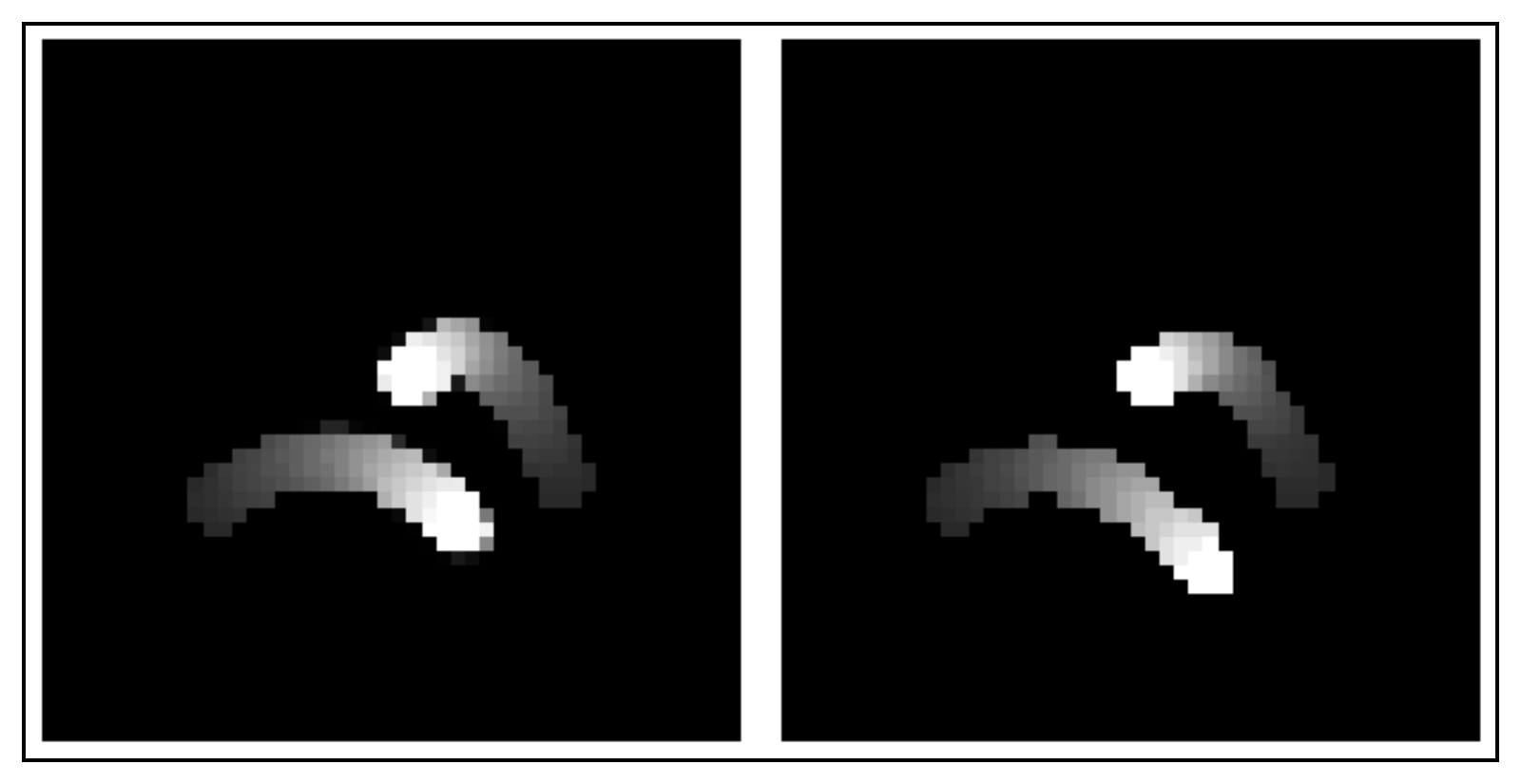}
\hskip0.04cm
\includegraphics[height=1.725cm,width=3.45 cm]{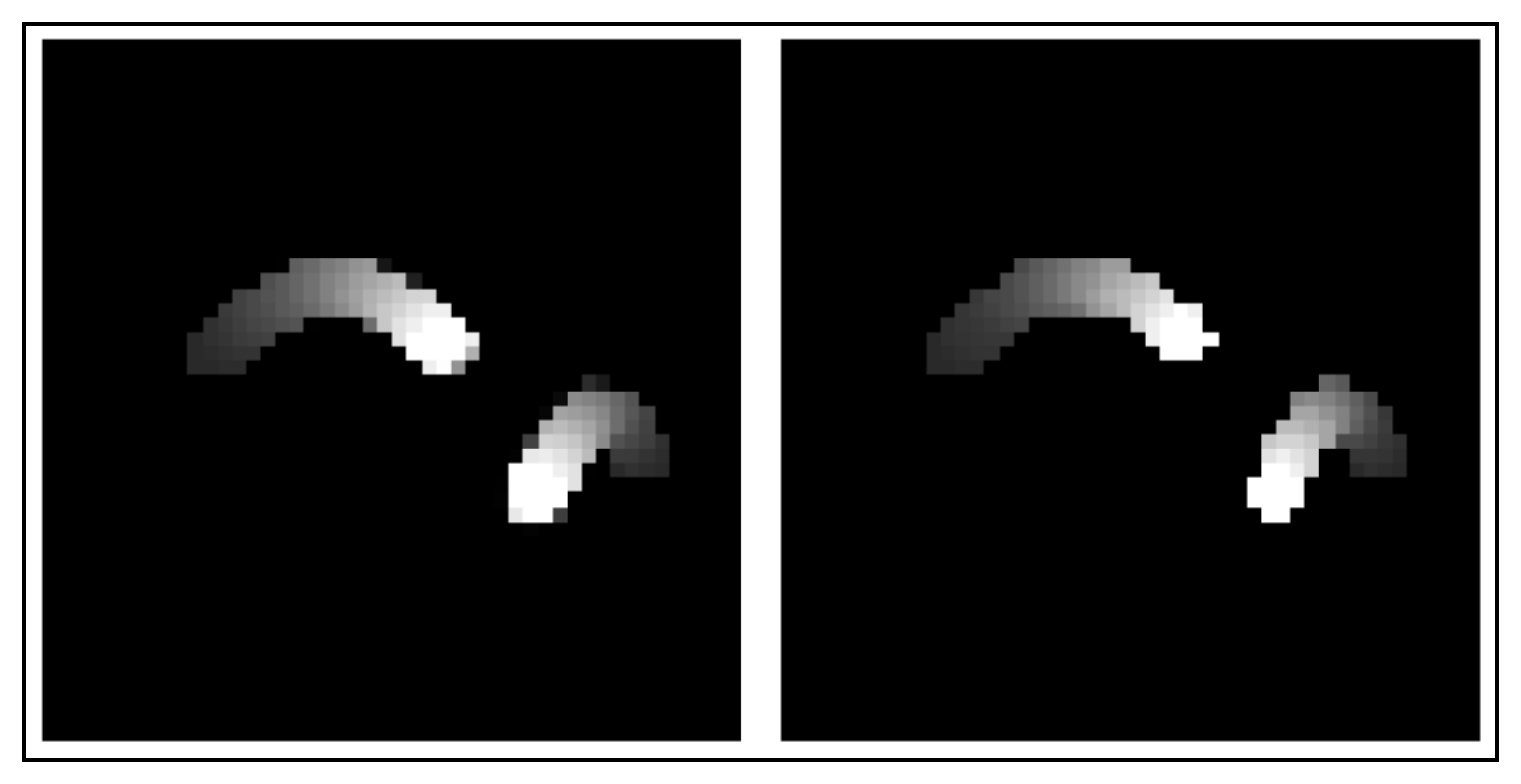}
\end{center}
\begin{center}
\hskip-0.3cm
\includegraphics[height=1.725cm,width=3.45 cm]{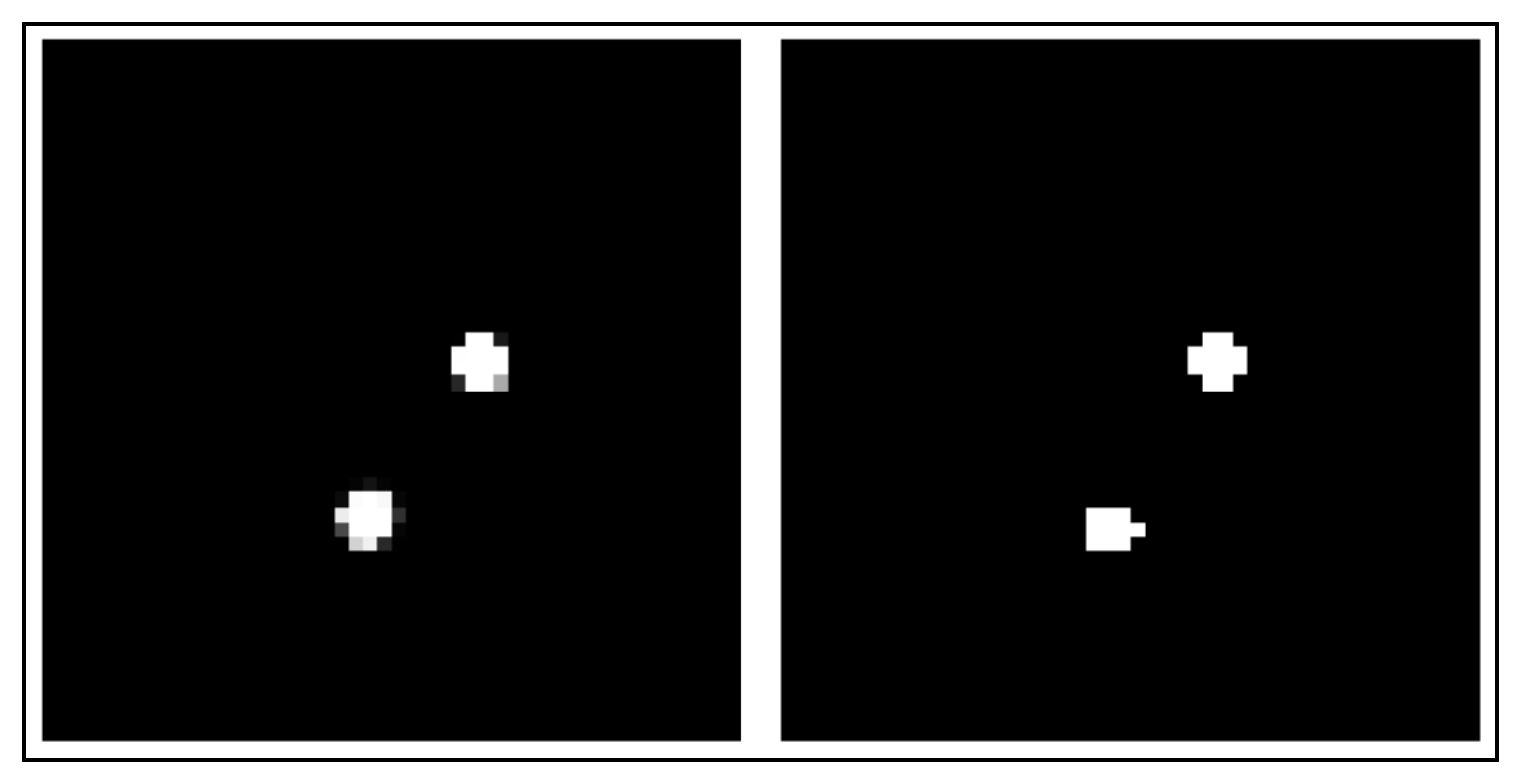}
\hskip0.04cm
\includegraphics[height=1.725cm,width=3.45 cm]{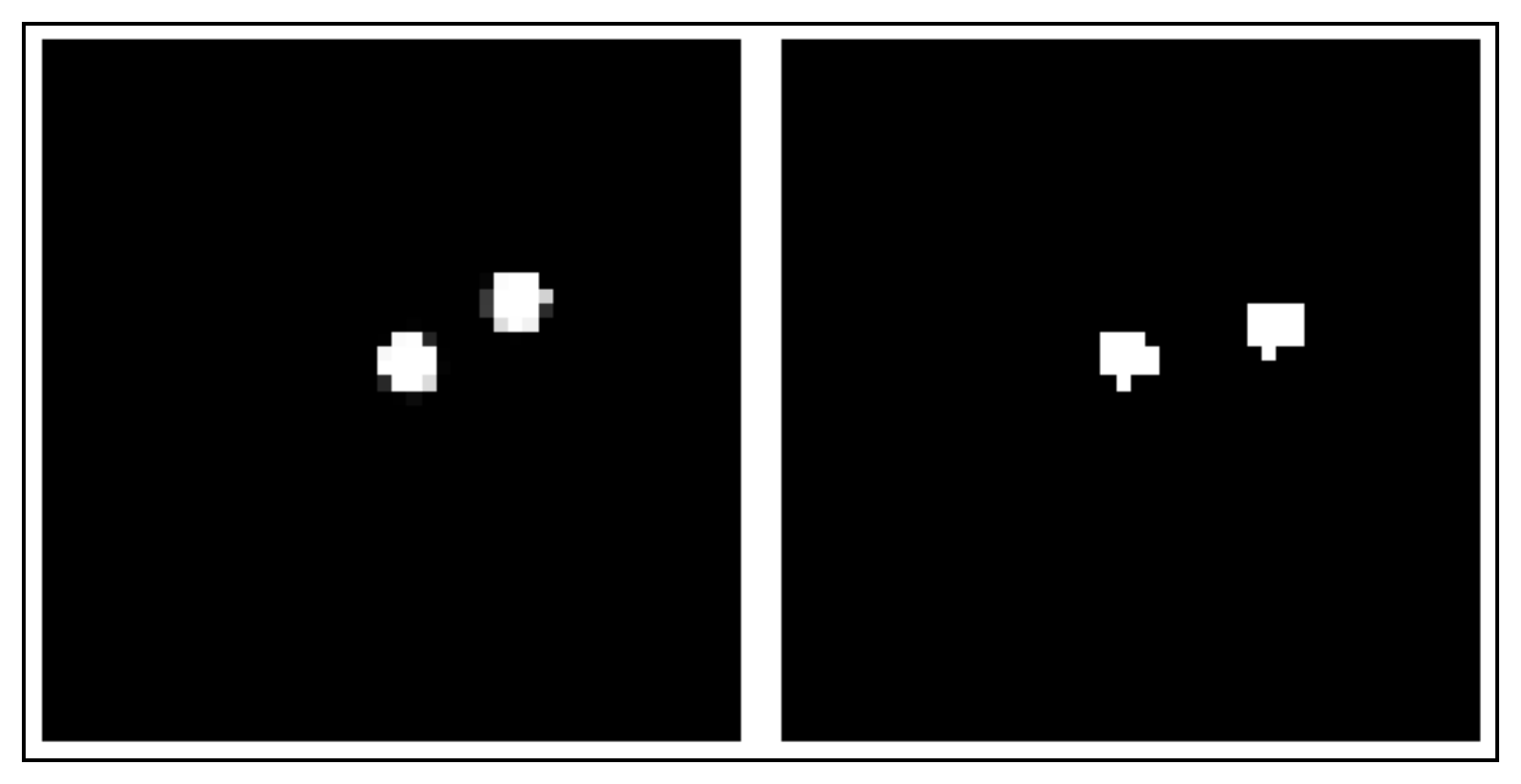}
\hskip0.04cm
\includegraphics[height=1.725cm,width=3.45 cm]{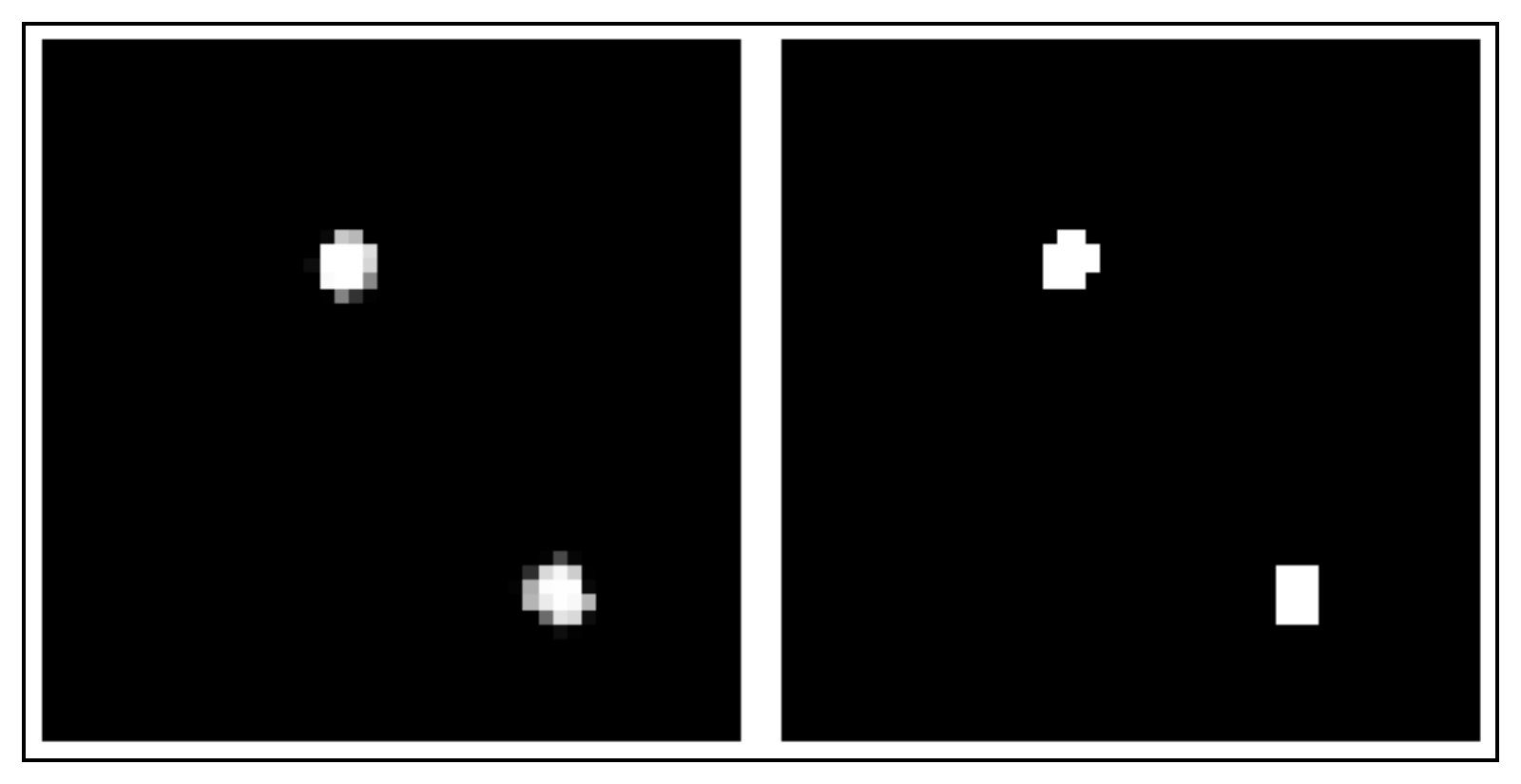}
\end{center}
\begin{center}
\hskip-0.3cm
\includegraphics[height=1.725cm,width=3.45 cm]{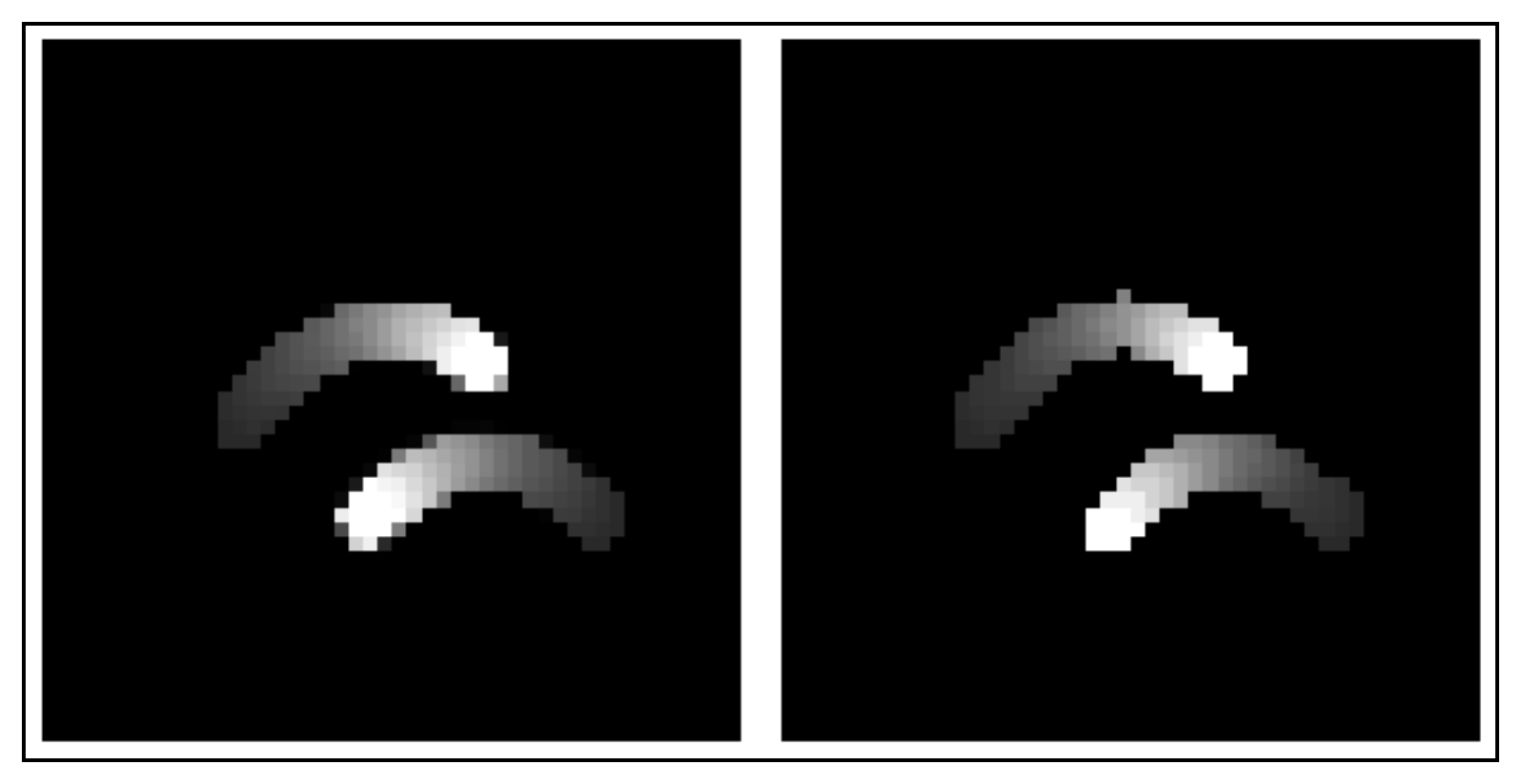}
\hskip0.04cm
\includegraphics[height=1.725cm,width=3.45 cm]{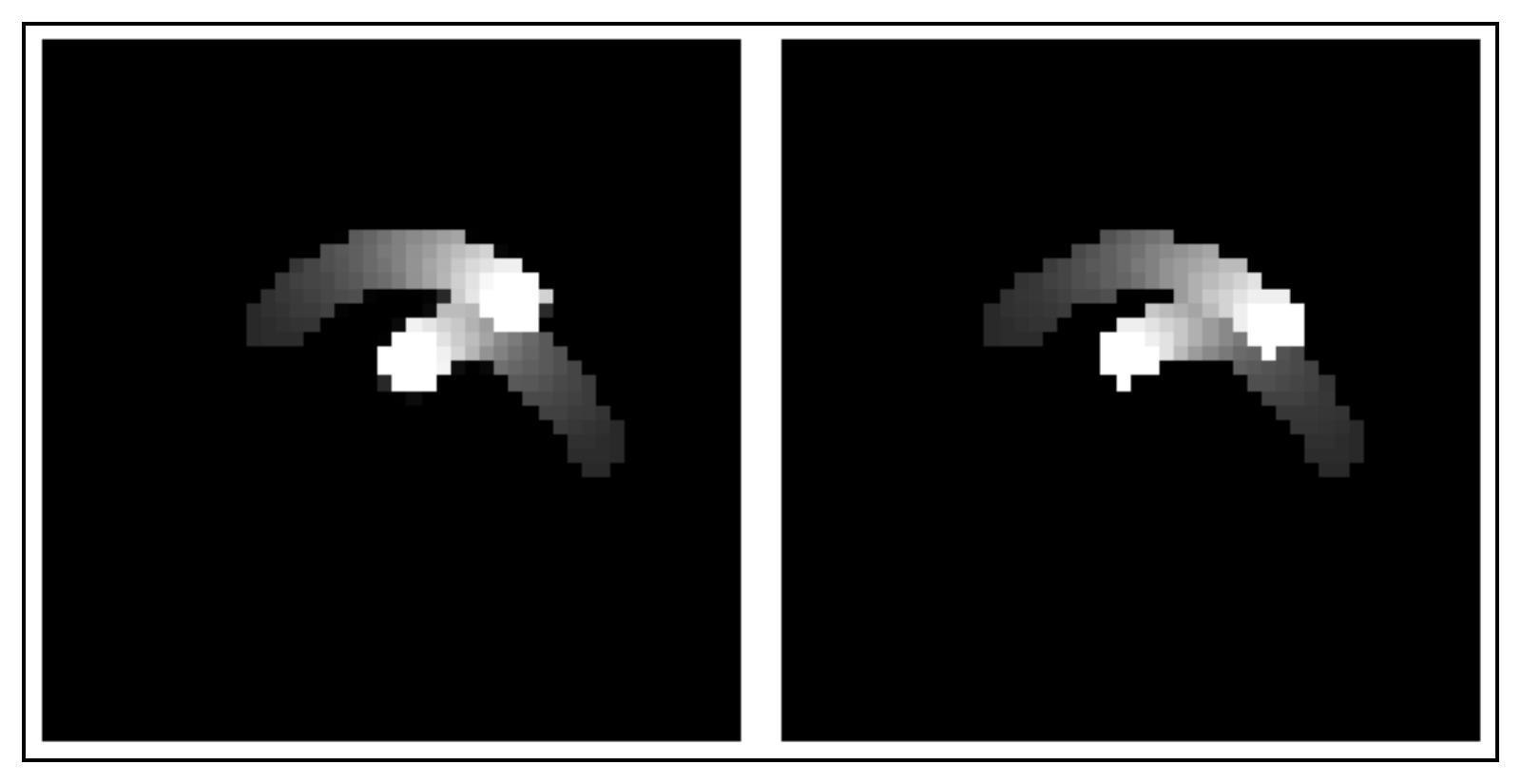}
\hskip0.04cm
\includegraphics[height=1.725cm,width=3.45 cm]{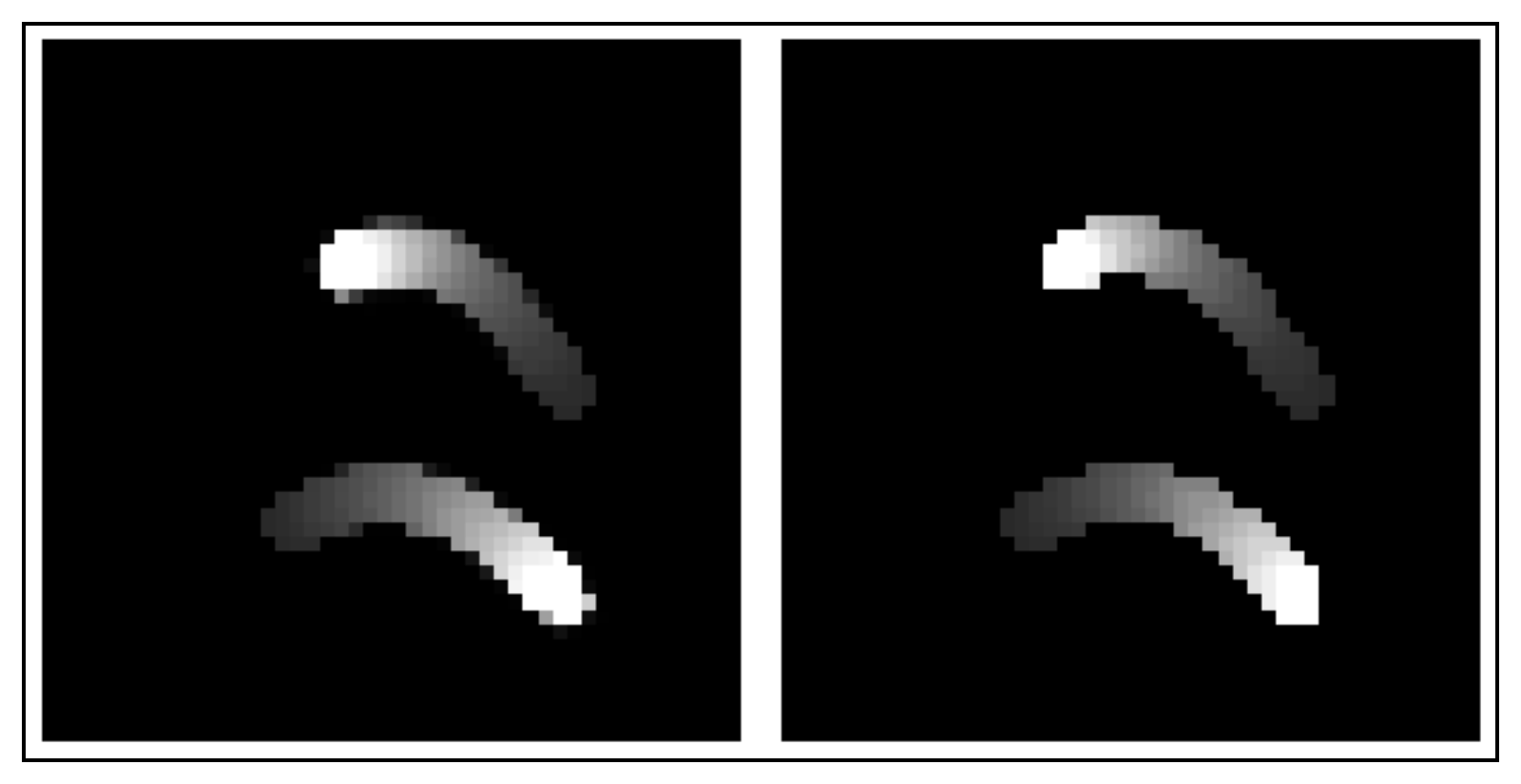}
\end{center}
\begin{center}
\hskip-0.3cm
\includegraphics[height=1.725cm,width=3.45 cm]{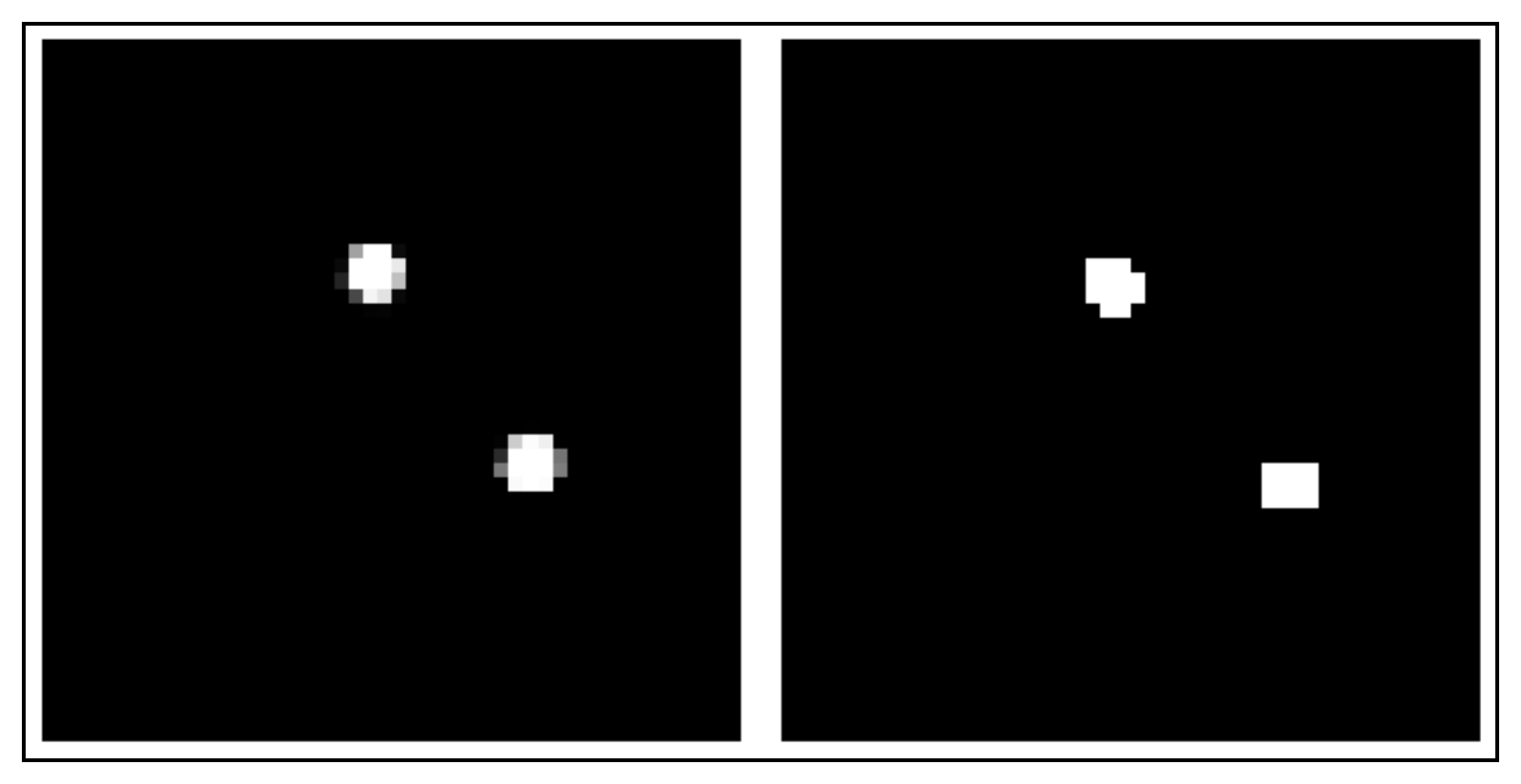}
\hskip0.04cm
\includegraphics[height=1.725cm,width=3.45 cm]{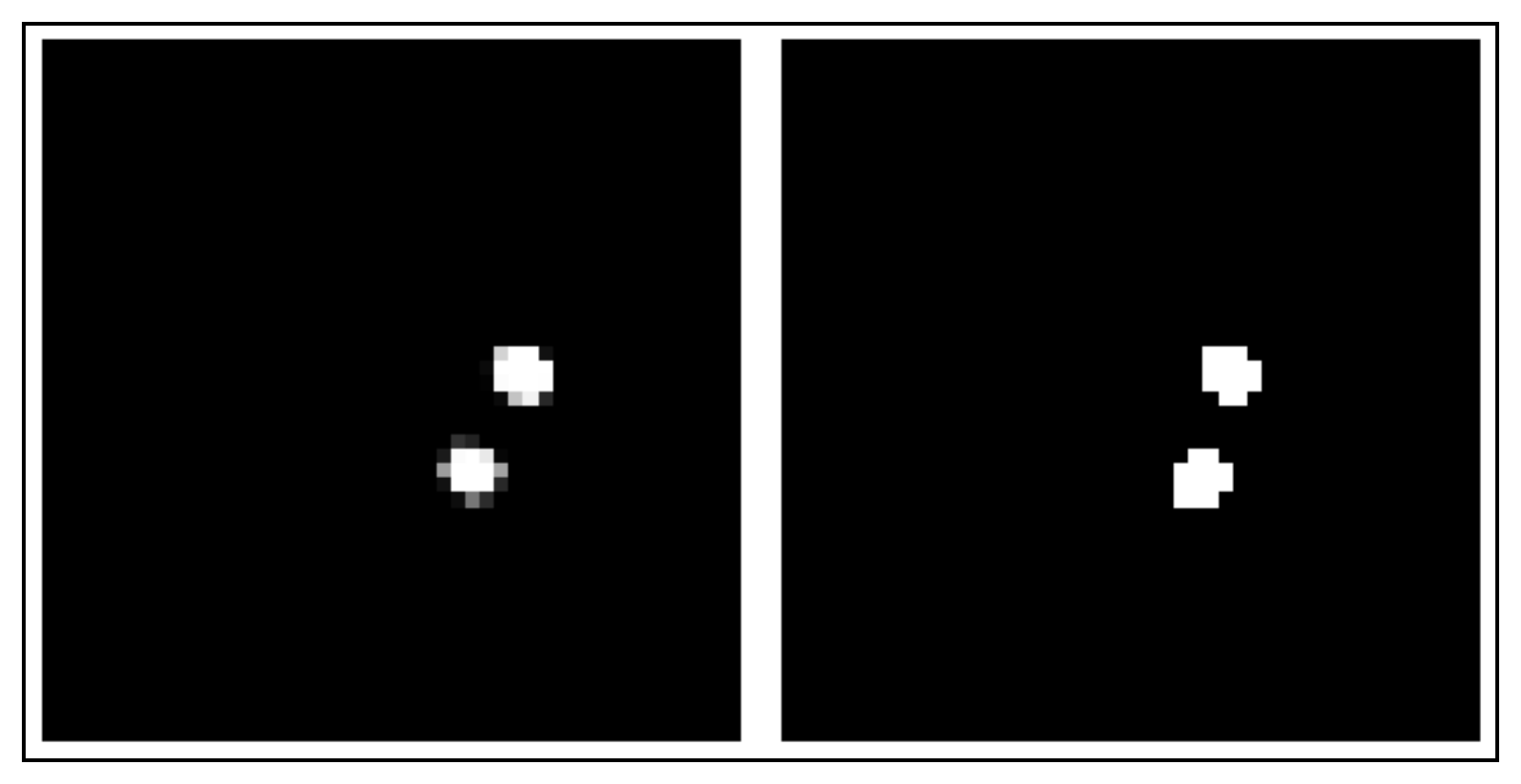}
\hskip0.04cm
\includegraphics[height=1.725cm,width=3.45 cm]{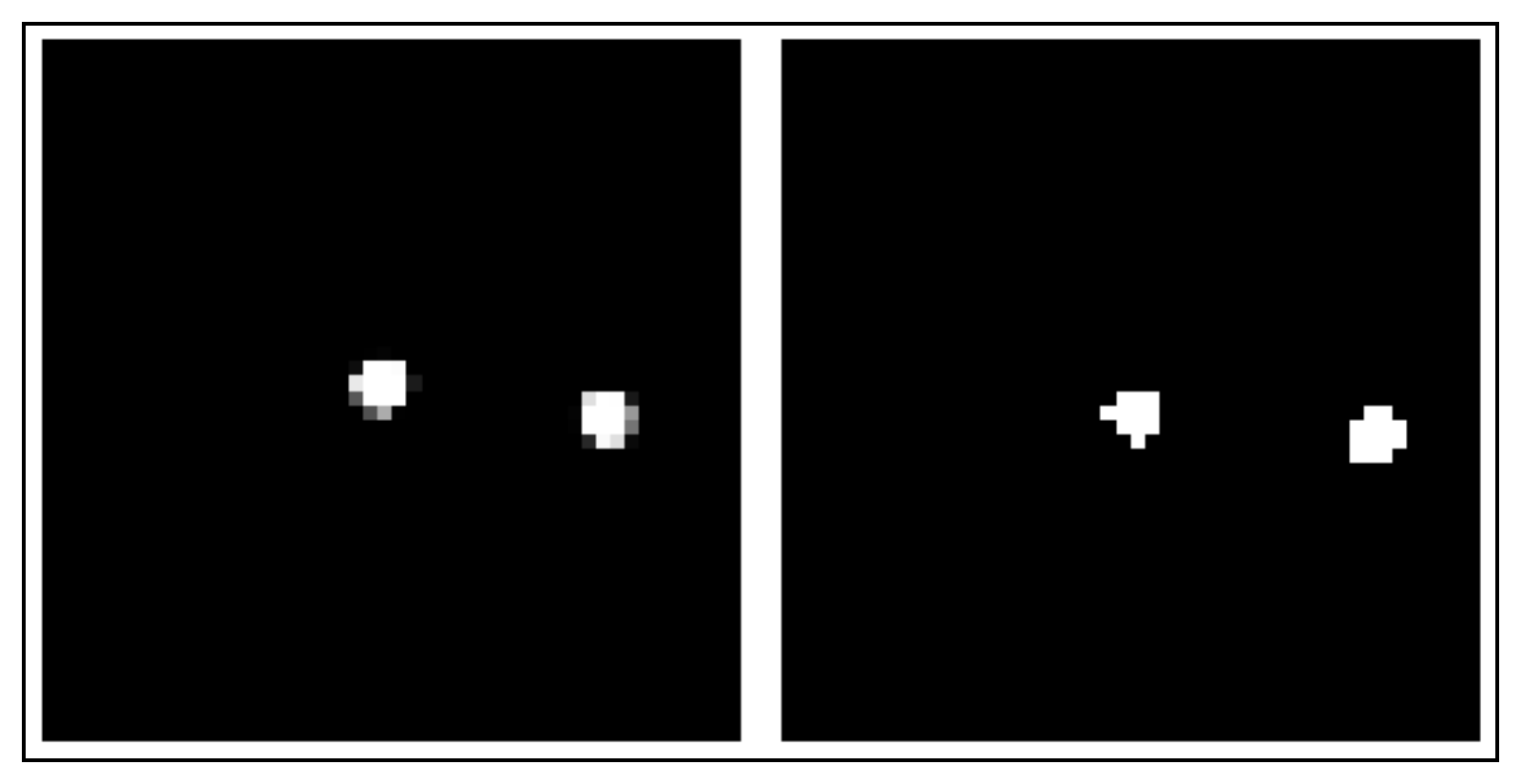}
\end{center}
\begin{center}
\hskip-0.3cm
\includegraphics[height=1.725cm,width=3.45 cm]{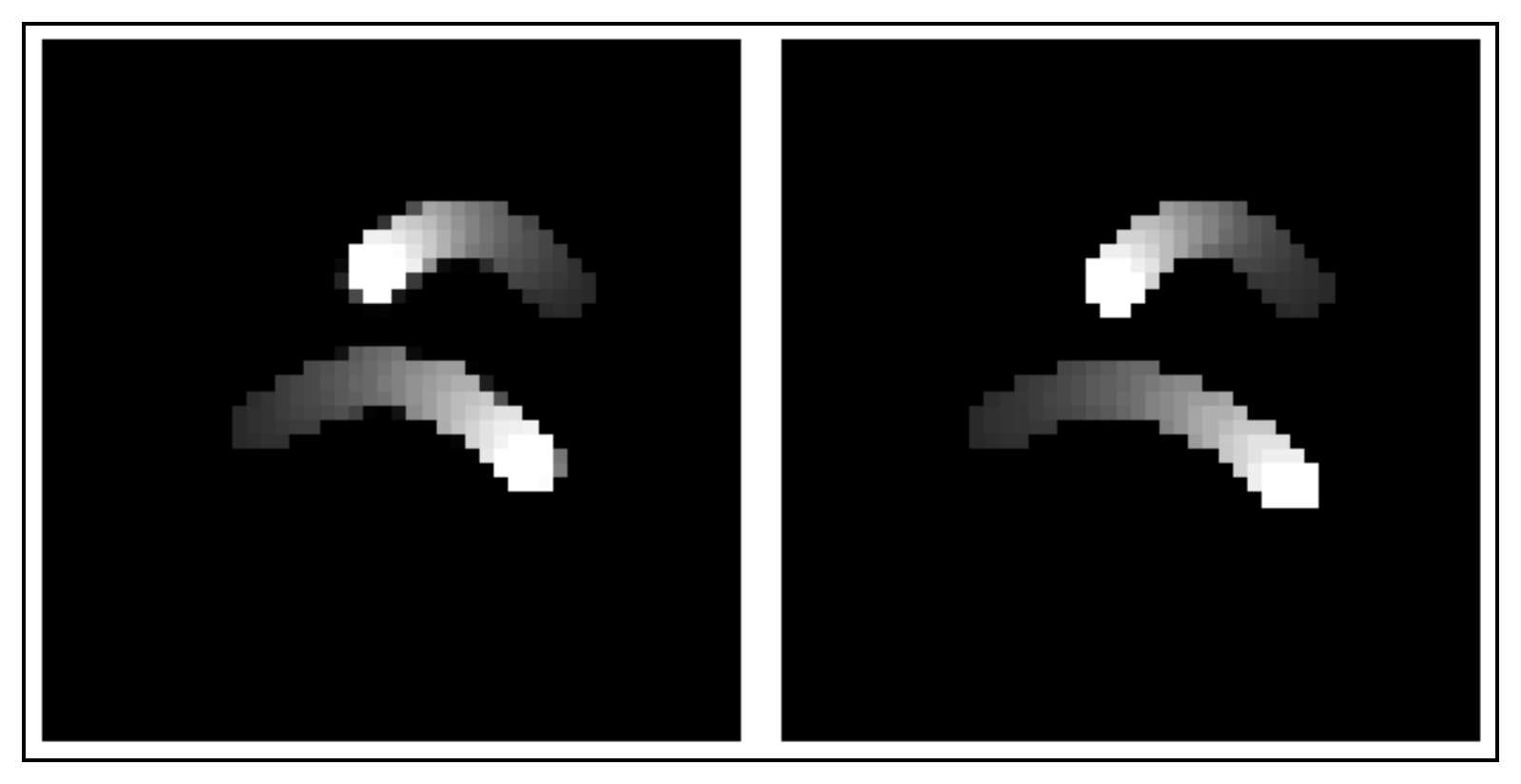}
\hskip0.04cm
\includegraphics[height=1.725cm,width=3.45 cm]{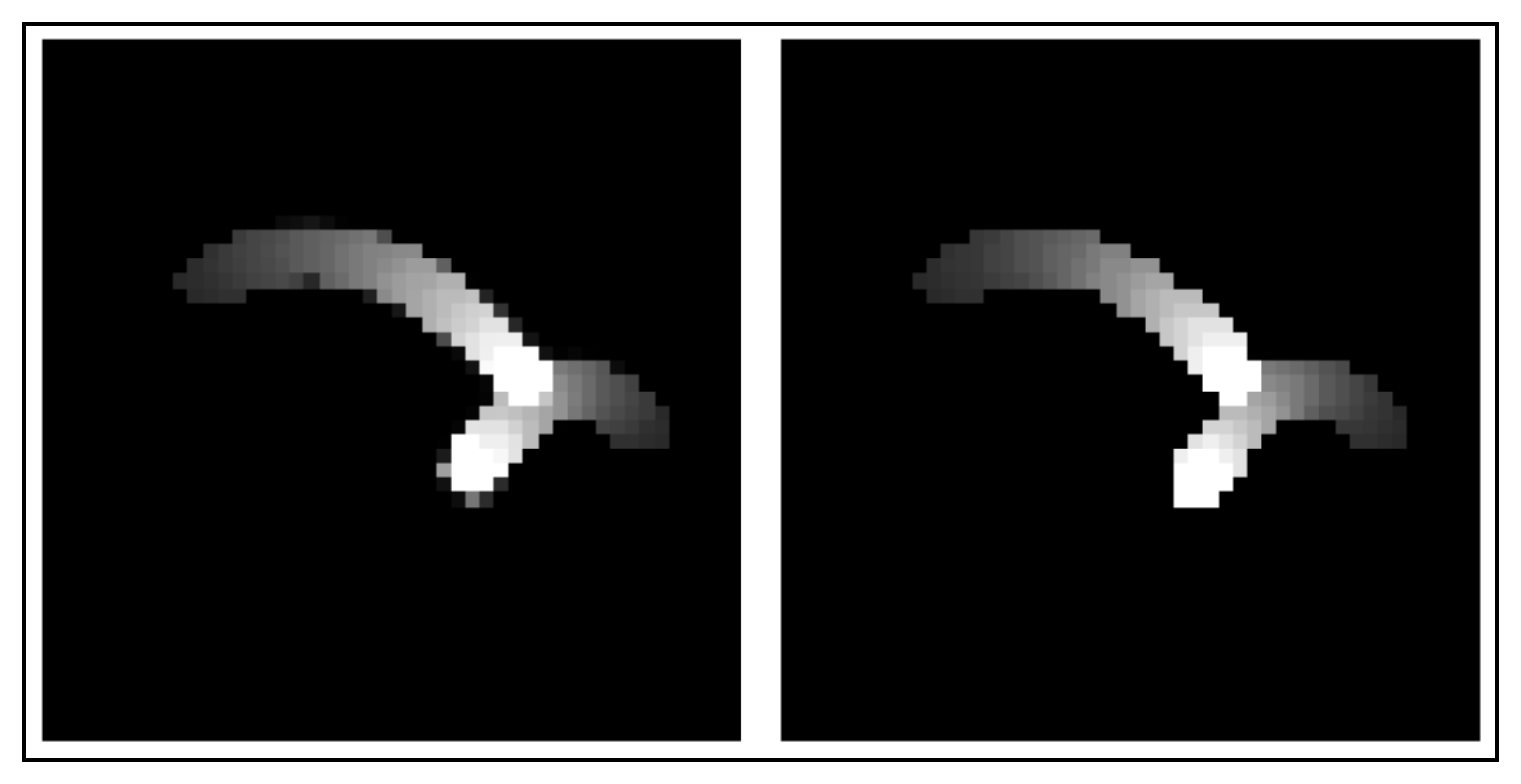}
\hskip0.04cm
\includegraphics[height=1.725cm,width=3.45 cm]{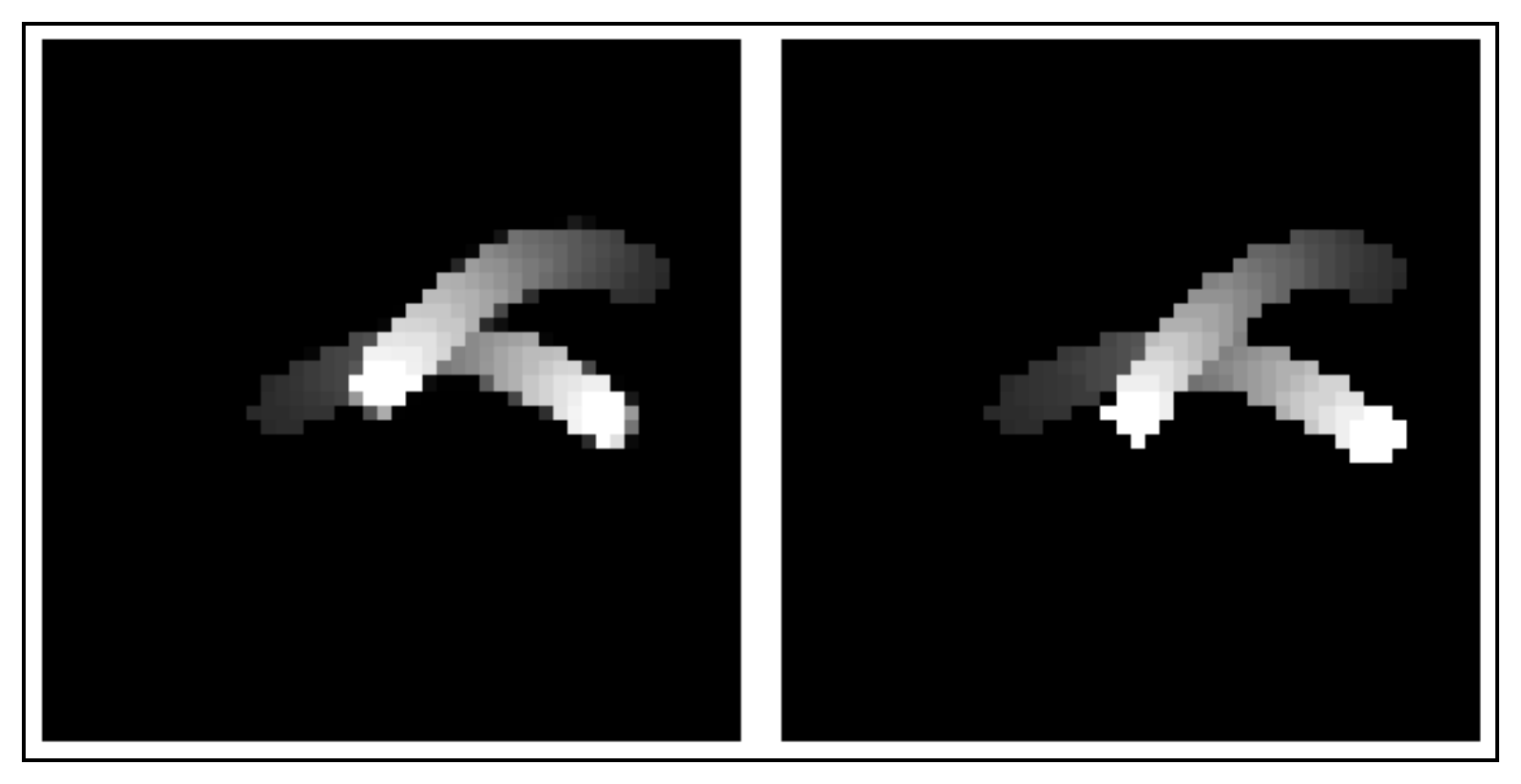}
\end{center}

\caption{Each plot shows generated (left) versus ground-truth (right) images at time-step 30 (top) and overlaid in time (bottom) for our model.}
\label{fig:GenObsFrames}
\end{figure}

\begin{figure}[t]
\begin{center}
\hskip-0.3cm
\includegraphics[height=1.725cm,width=3.45cm]{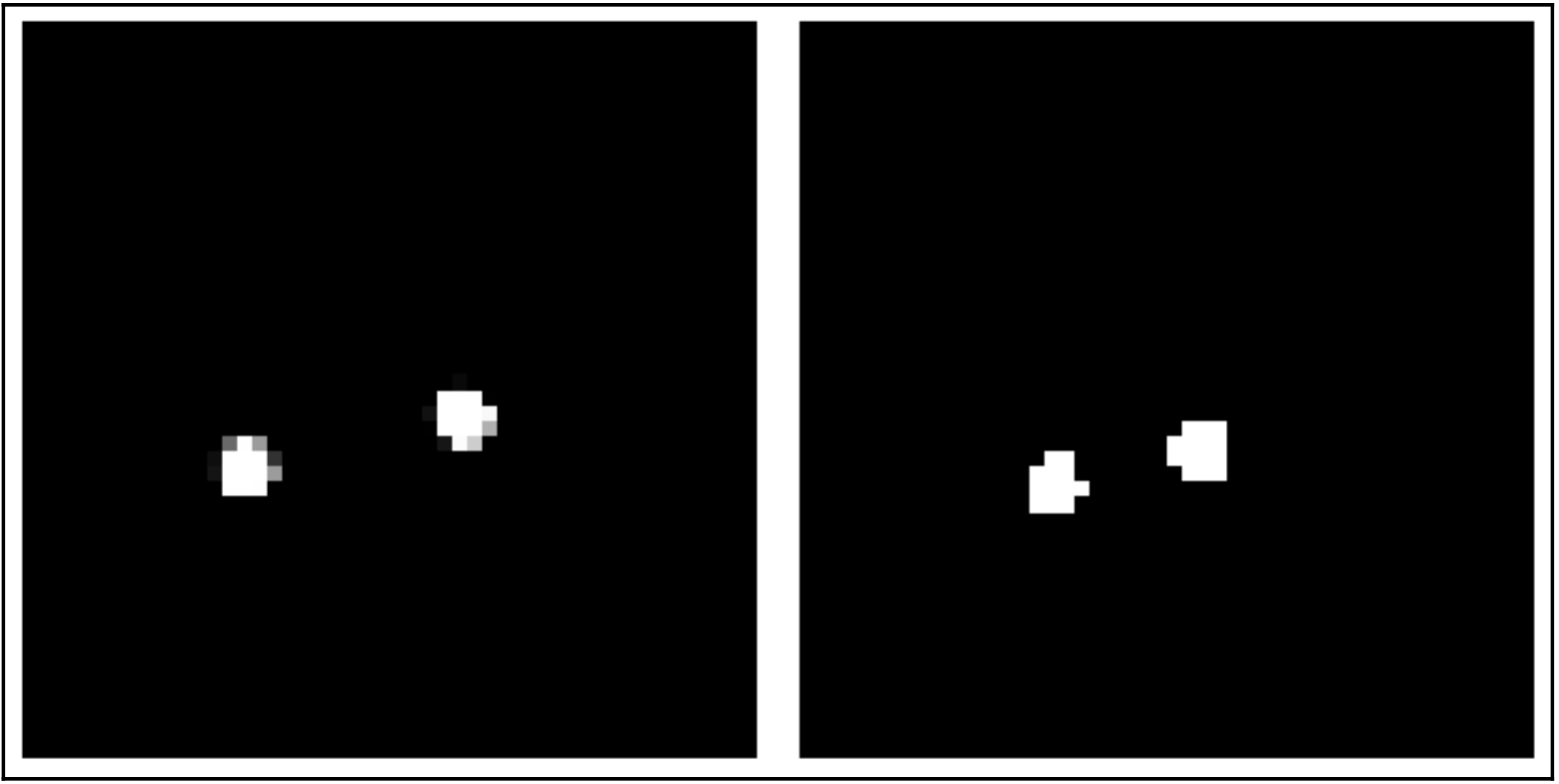}
\hskip0.04cm
\includegraphics[height=1.725cm,width=3.45cm]{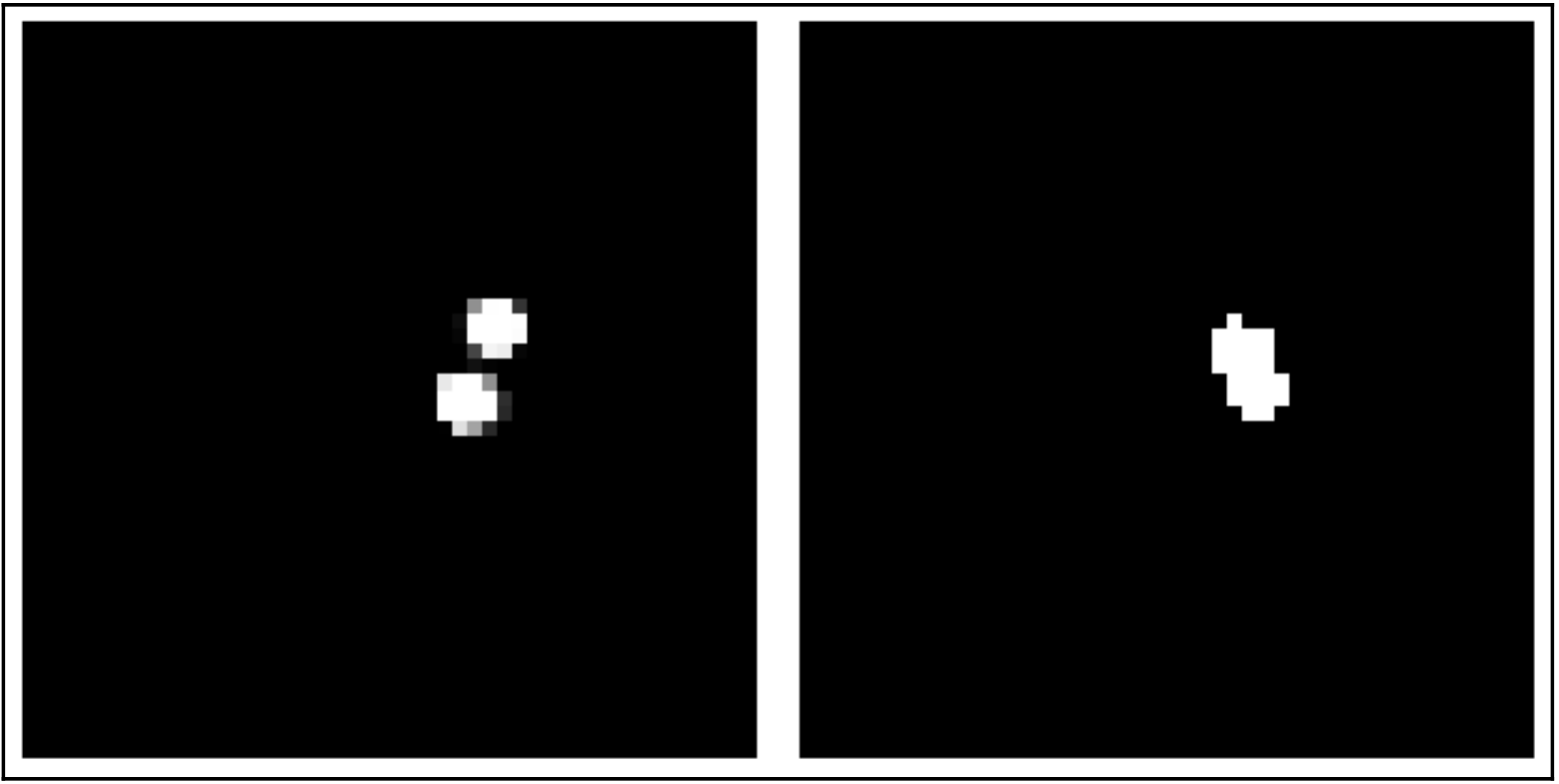}
\hskip0.04cm
\includegraphics[height=1.725cm,width=3.45cm]{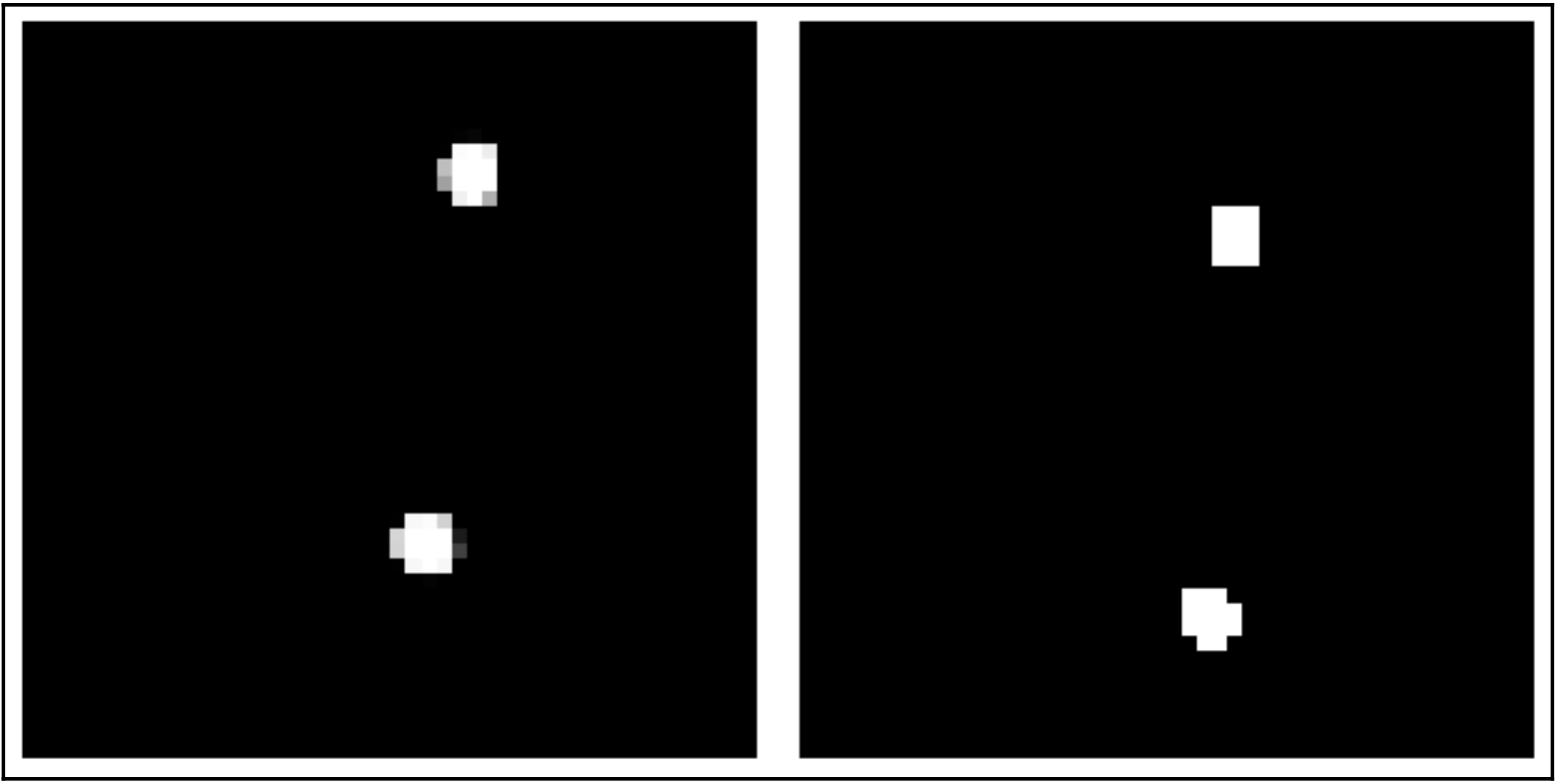}
\end{center}
\begin{center}
\hskip-0.3cm
\includegraphics[height=1.75cm,width=3.5cm]{./fig/lstm/lstm_200000_gen_obs_frames_0_all.pdf}
\includegraphics[height=1.75cm,width=3.5cm]{./fig/lstm/lstm_200000_gen_obs_frames_1_all.pdf}
\includegraphics[height=1.75cm,width=3.5cm]{./fig/lstm/lstm_200000_gen_obs_frames_2_all.pdf}
\end{center}
\begin{center}
\hskip-0.3cm
\includegraphics[height=1.725cm,width=3.45cm]{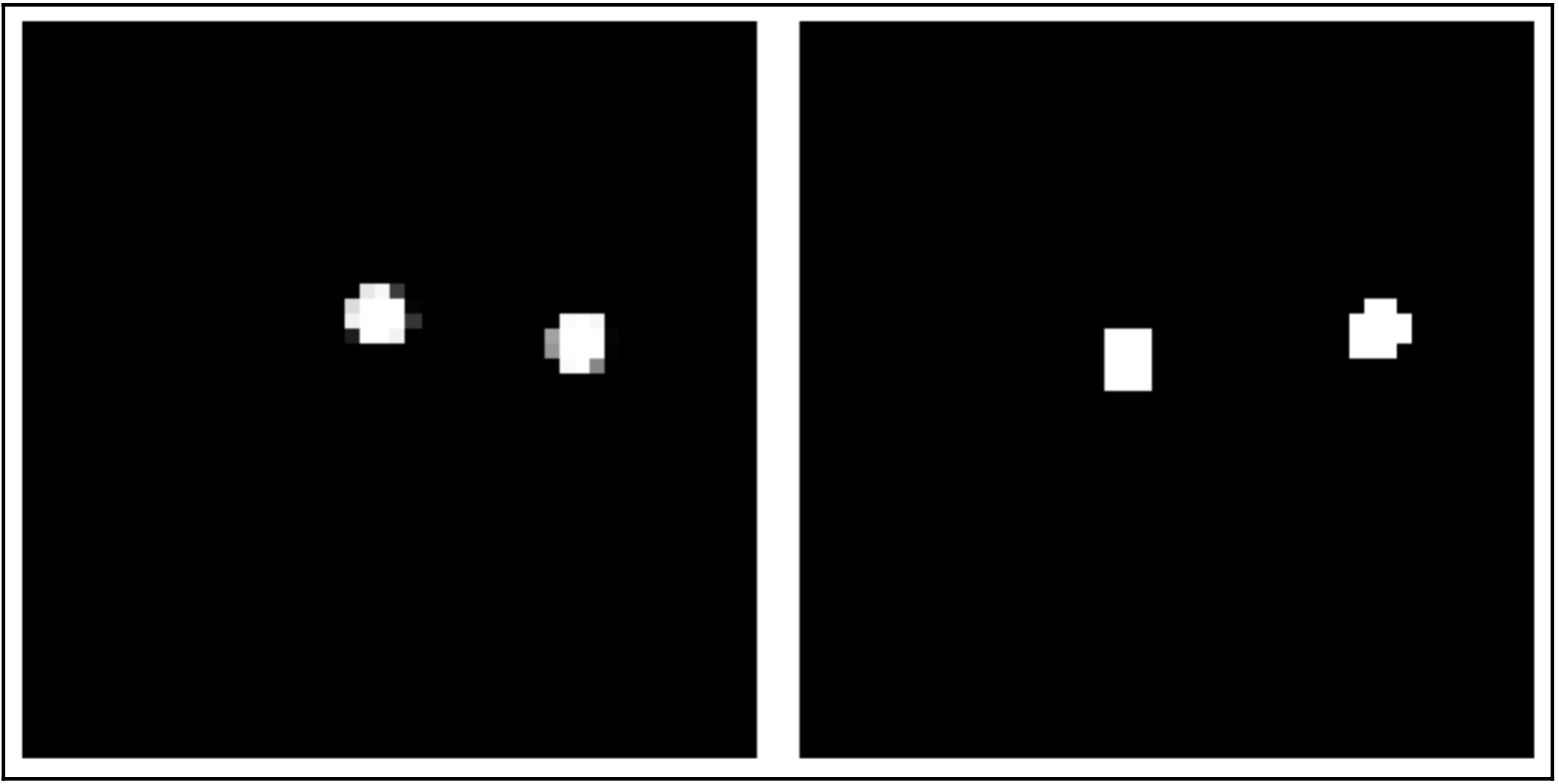}
\hskip0.04cm
\includegraphics[height=1.725cm,width=3.45cm]{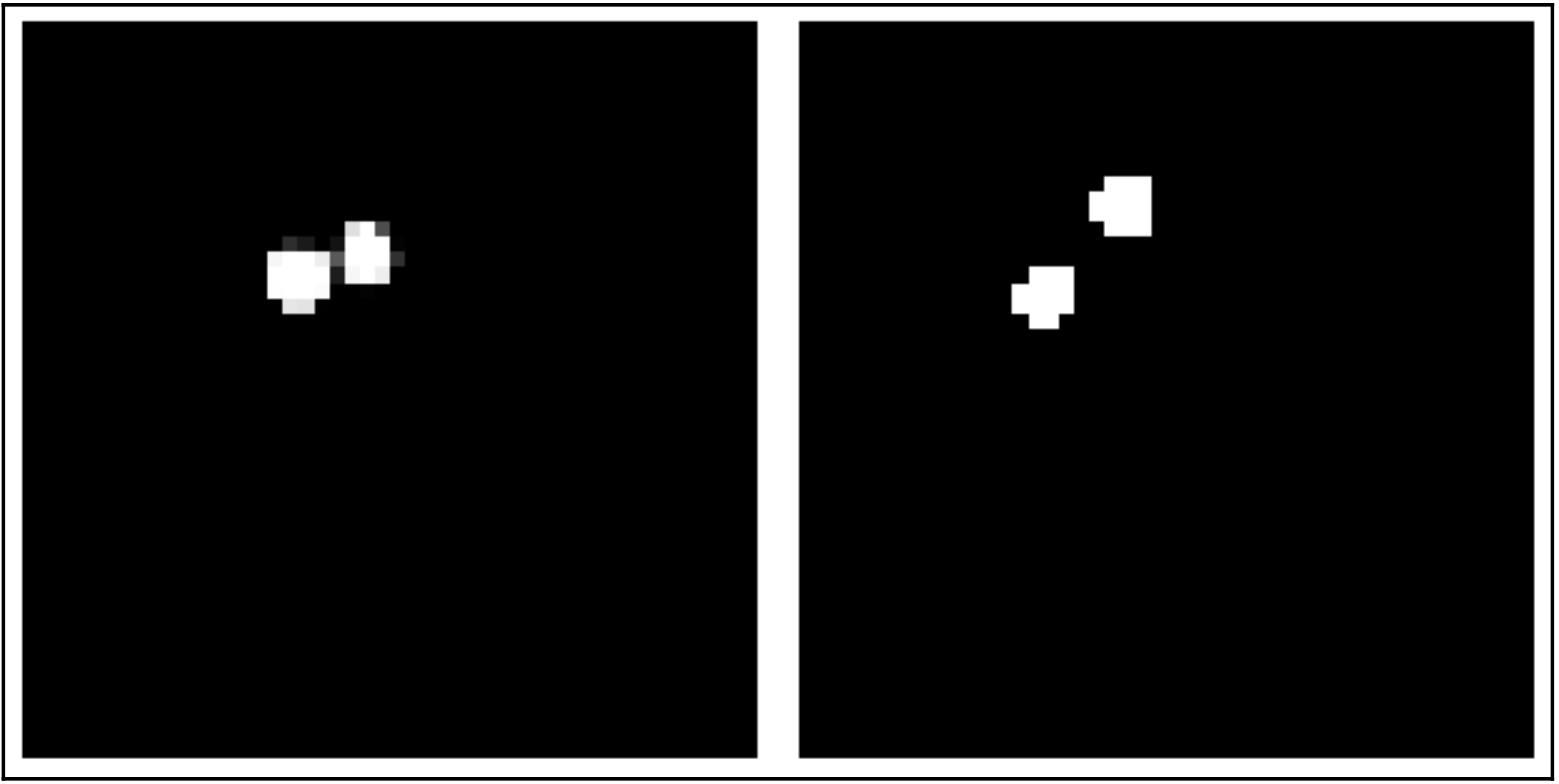}
\hskip0.04cm
\includegraphics[height=1.725cm,width=3.45cm]{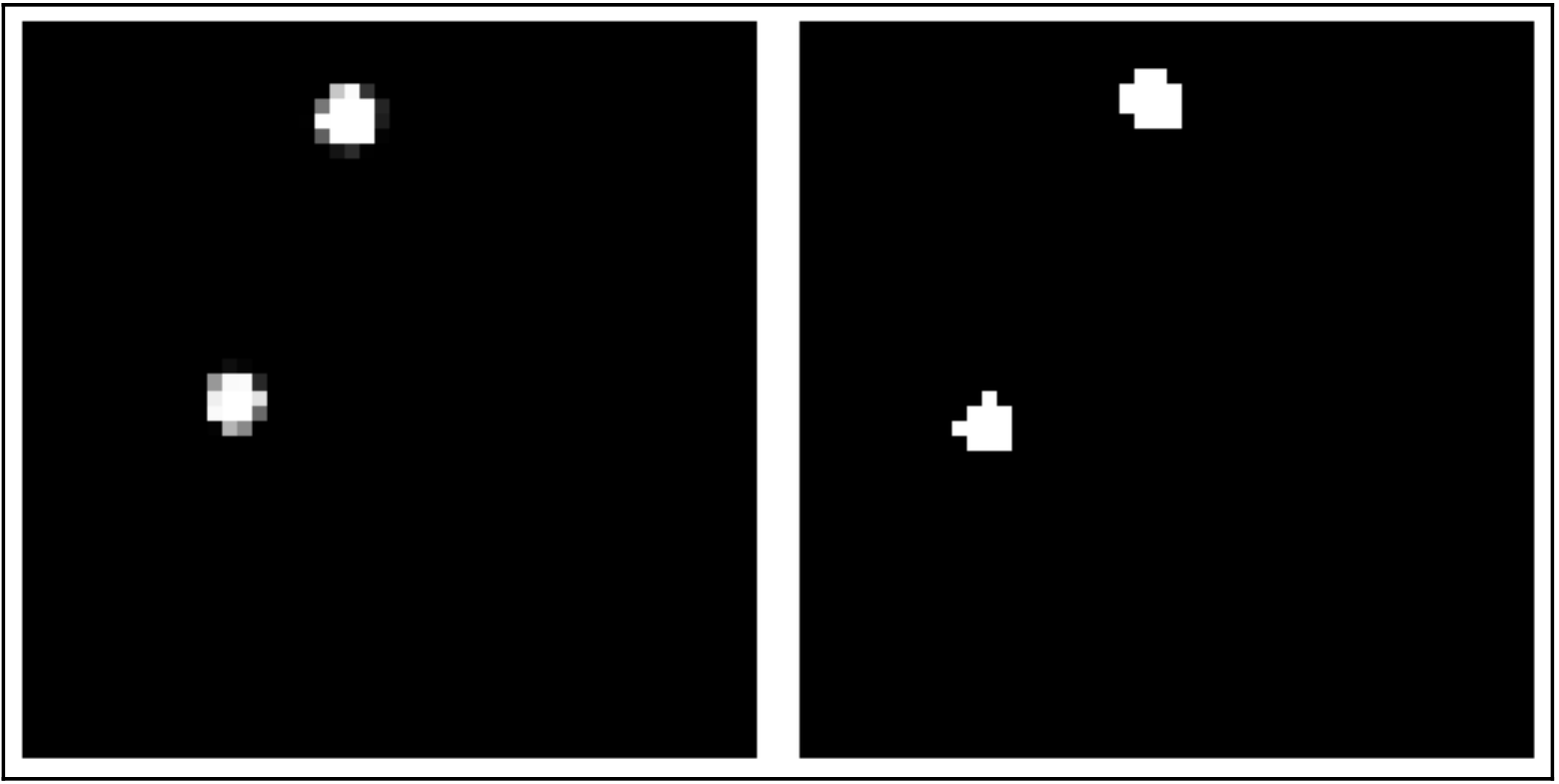}
\end{center}
\begin{center}
\hskip-0.3cm
\includegraphics[height=1.75cm,width=3.5cm]{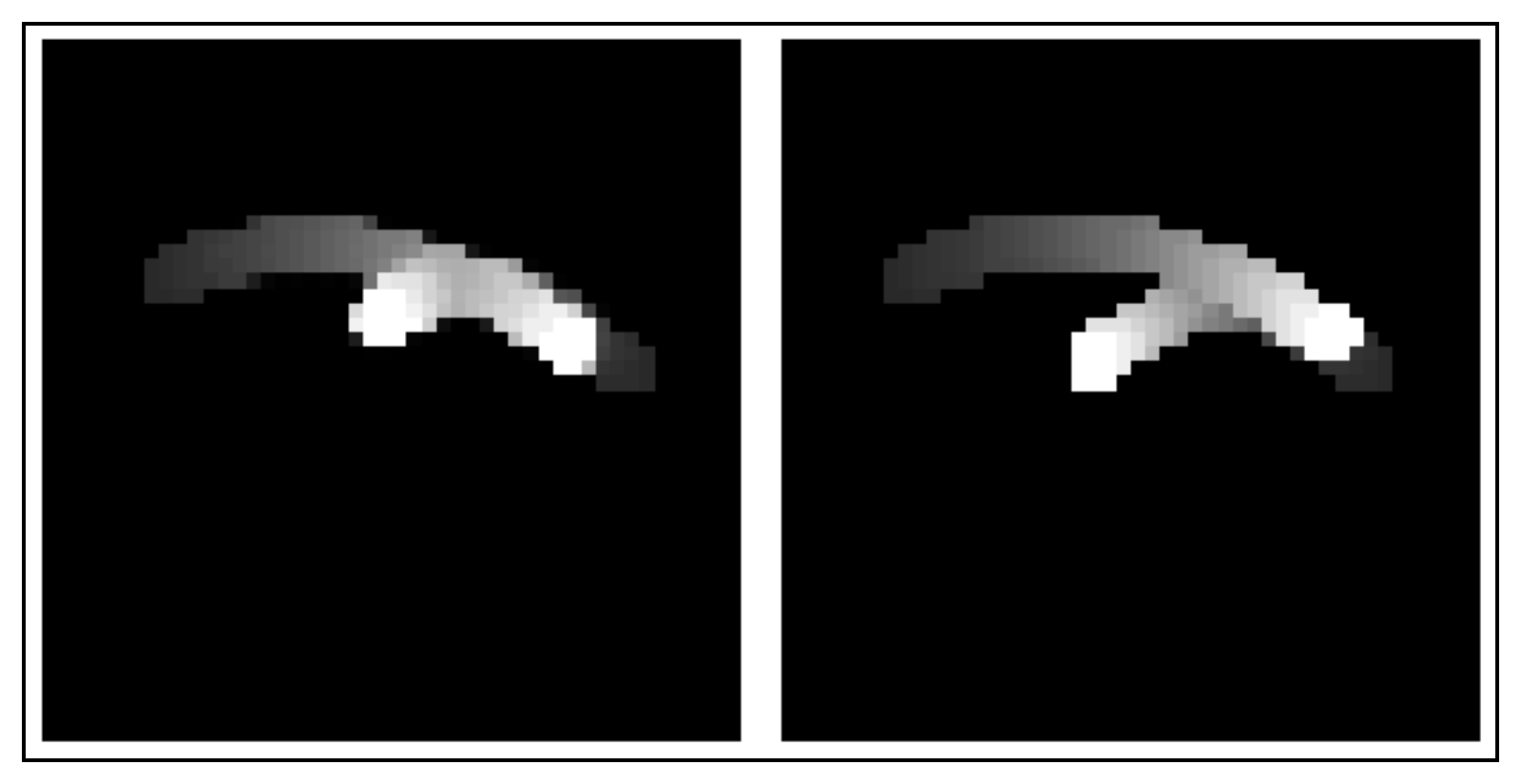}
\includegraphics[height=1.75cm,width=3.5cm]{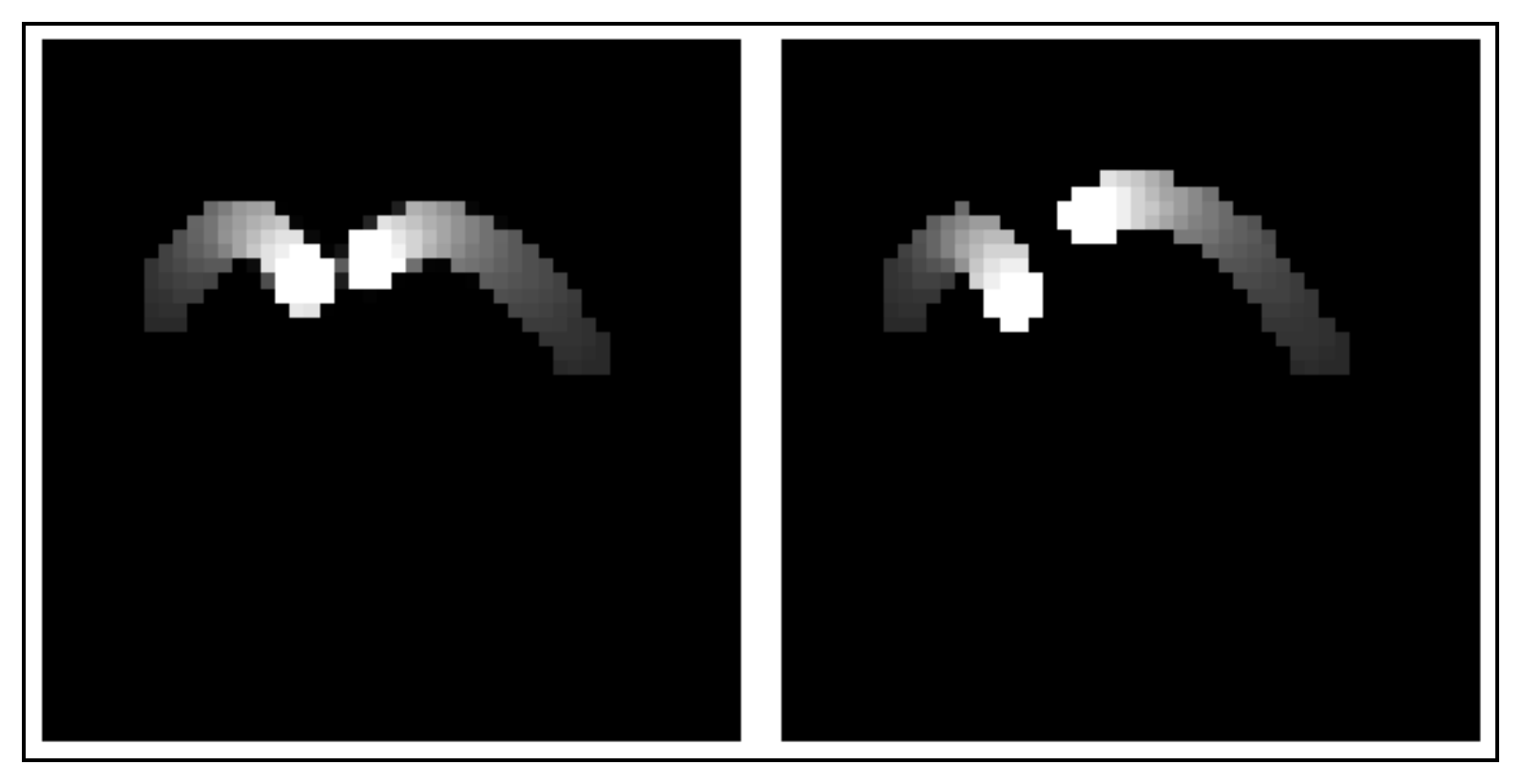}
\includegraphics[height=1.75cm,width=3.5cm]{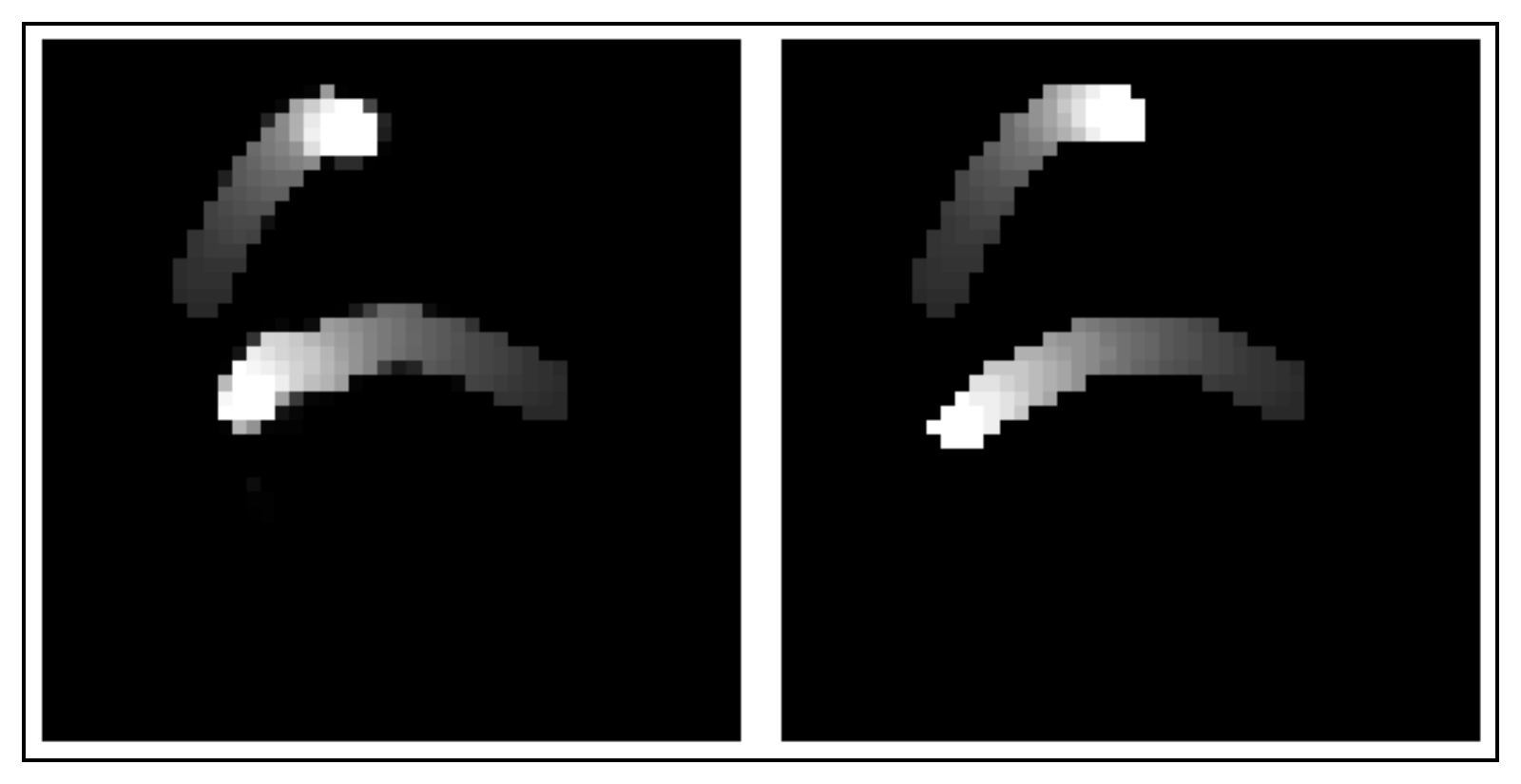}
\end{center}
\begin{center}
\hskip-0.3cm
\includegraphics[height=1.725cm,width=3.45cm]{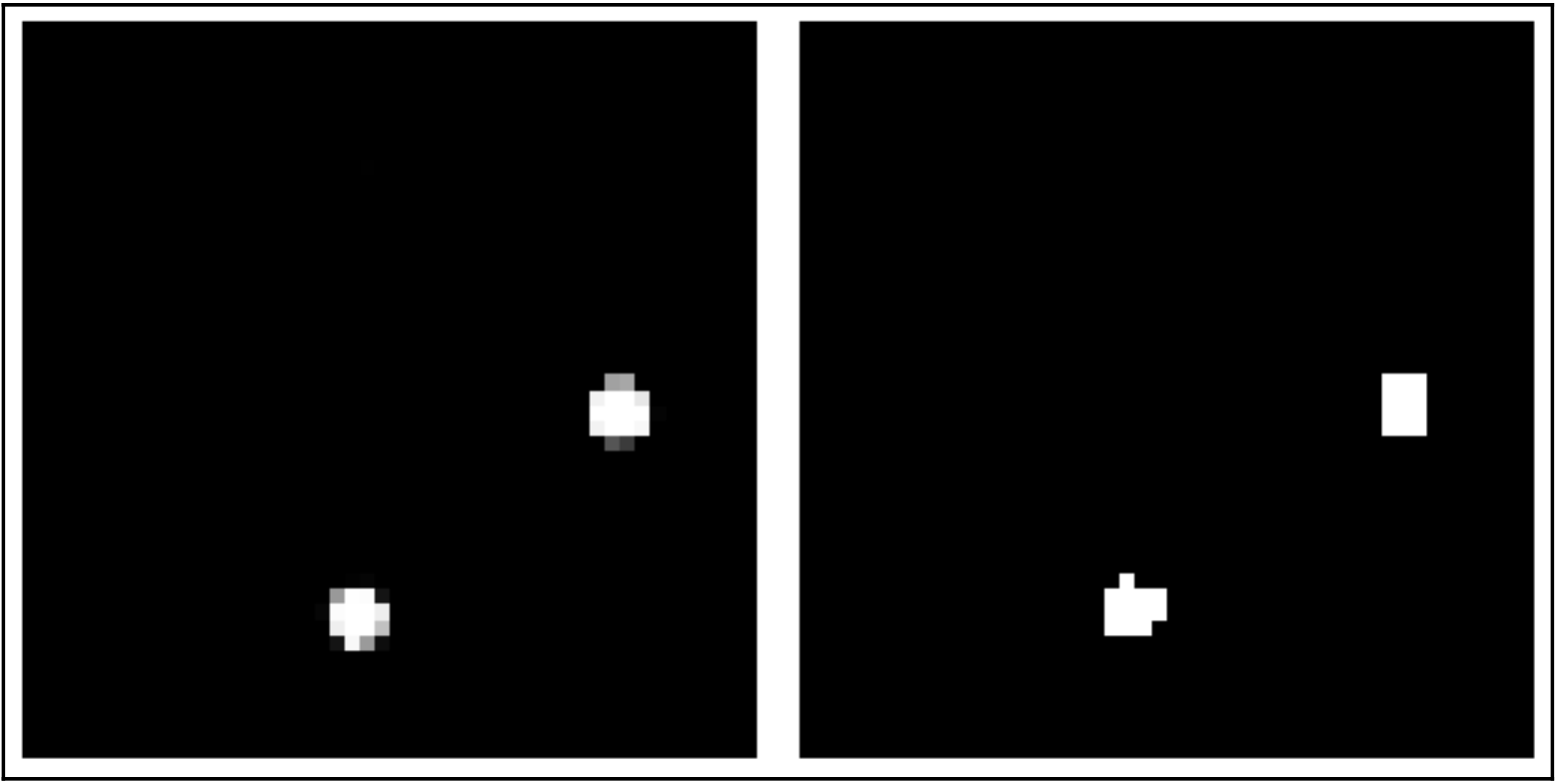}
\hskip0.04cm
\includegraphics[height=1.725cm,width=3.45cm]{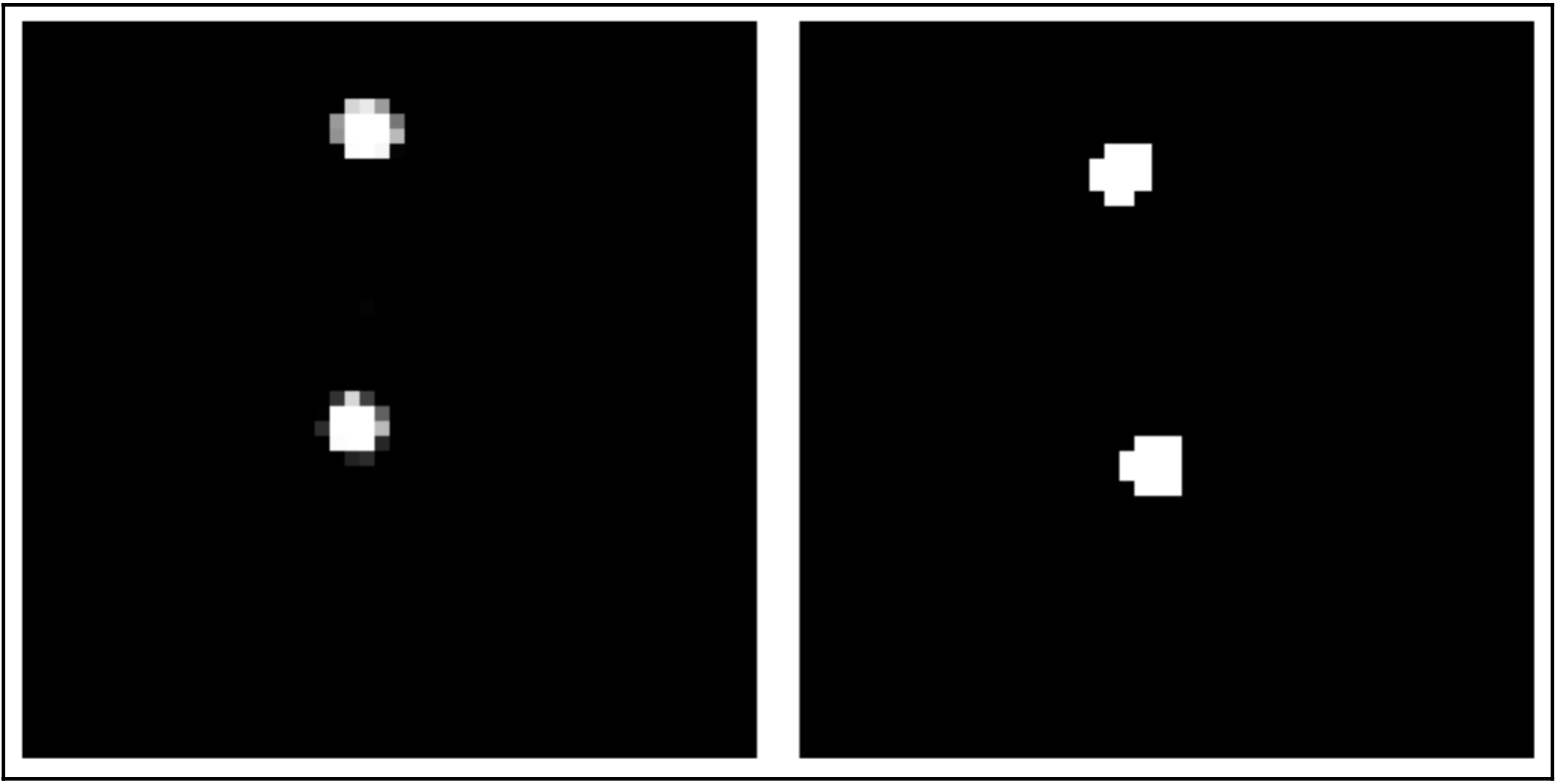}
\hskip0.04cm
\includegraphics[height=1.725cm,width=3.45cm]{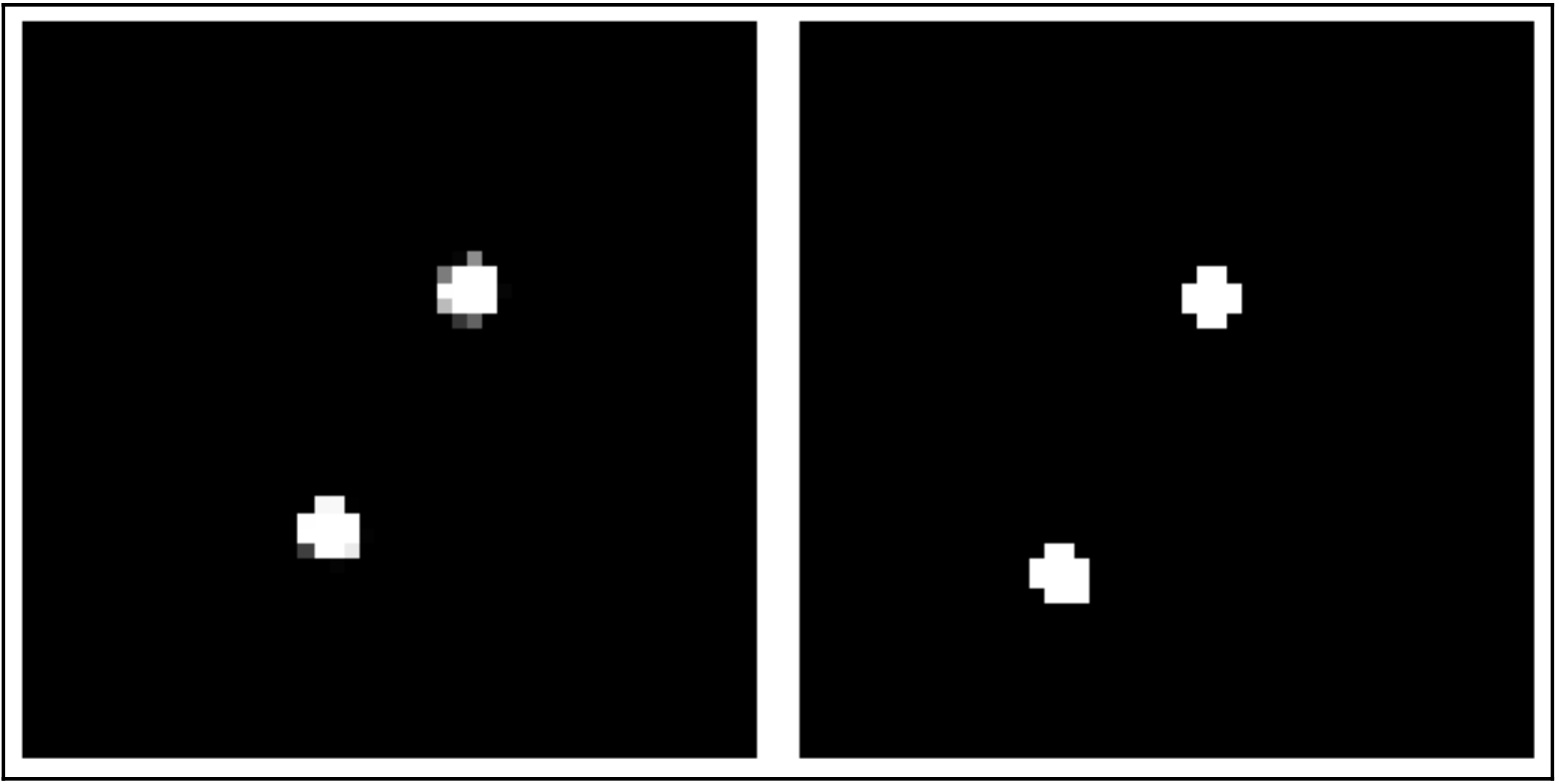}
\end{center}
\begin{center}
\hskip-0.3cm
\includegraphics[height=1.75cm,width=3.5cm]{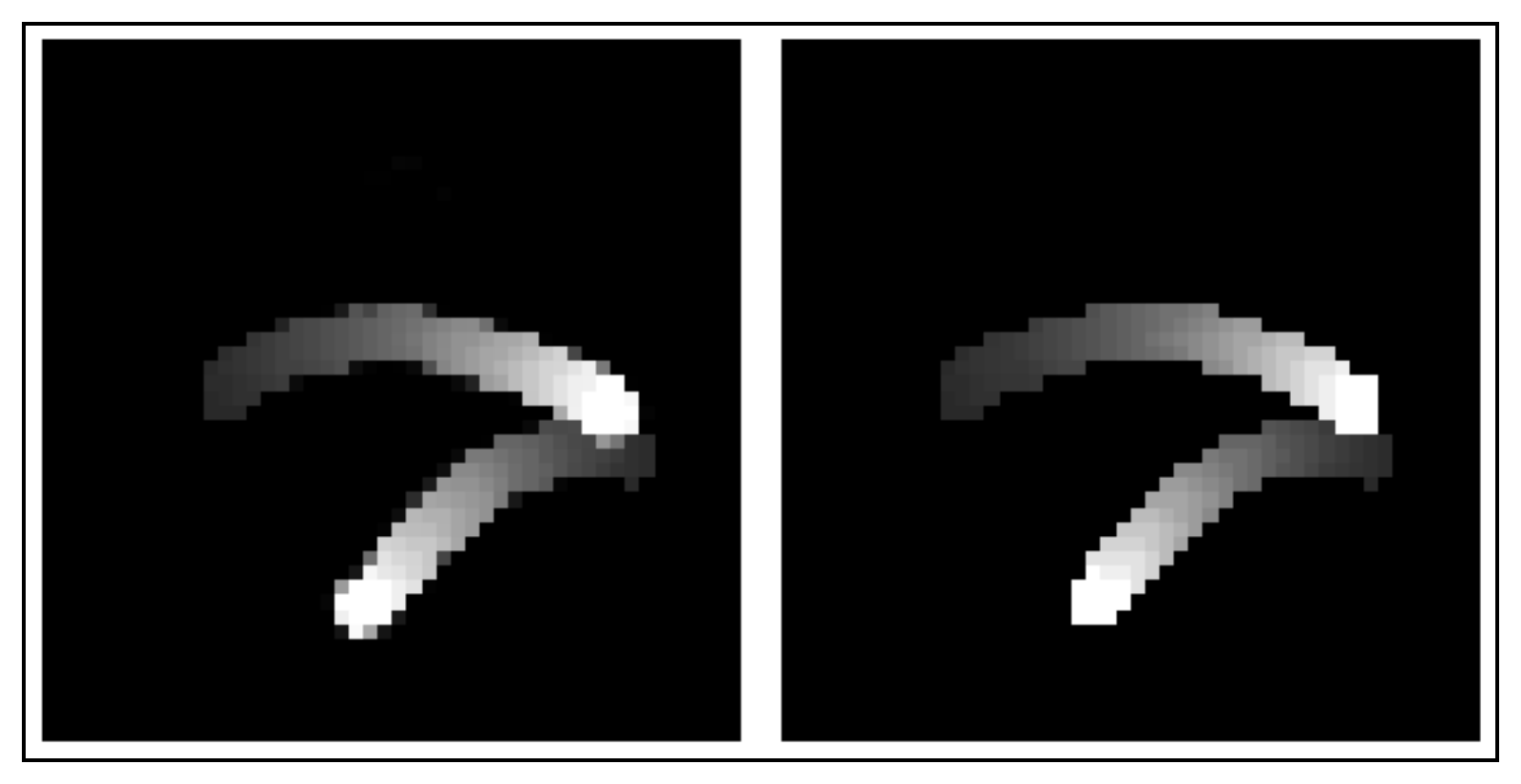}
\includegraphics[height=1.75cm,width=3.5cm]{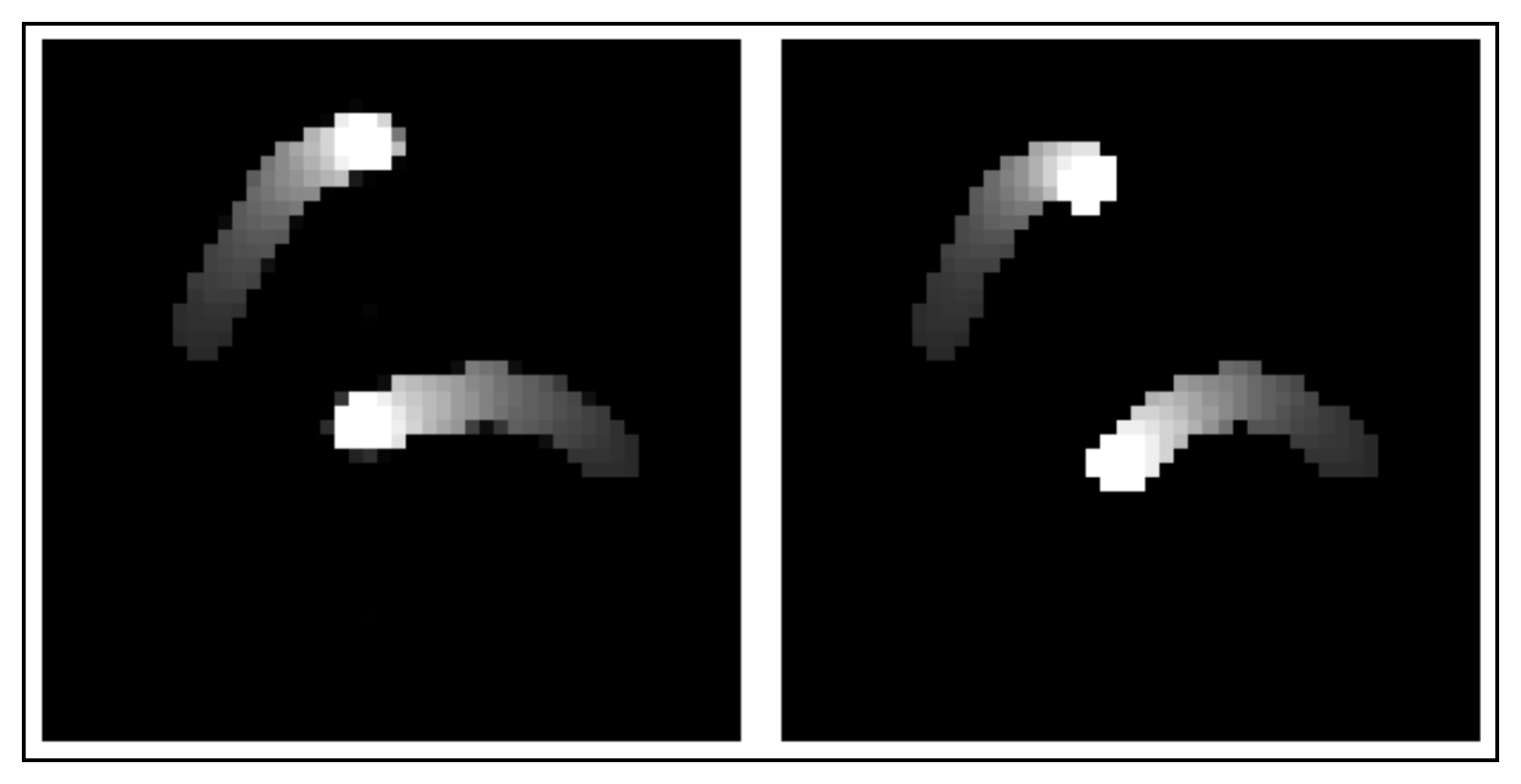}
\includegraphics[height=1.75cm,width=3.5cm]{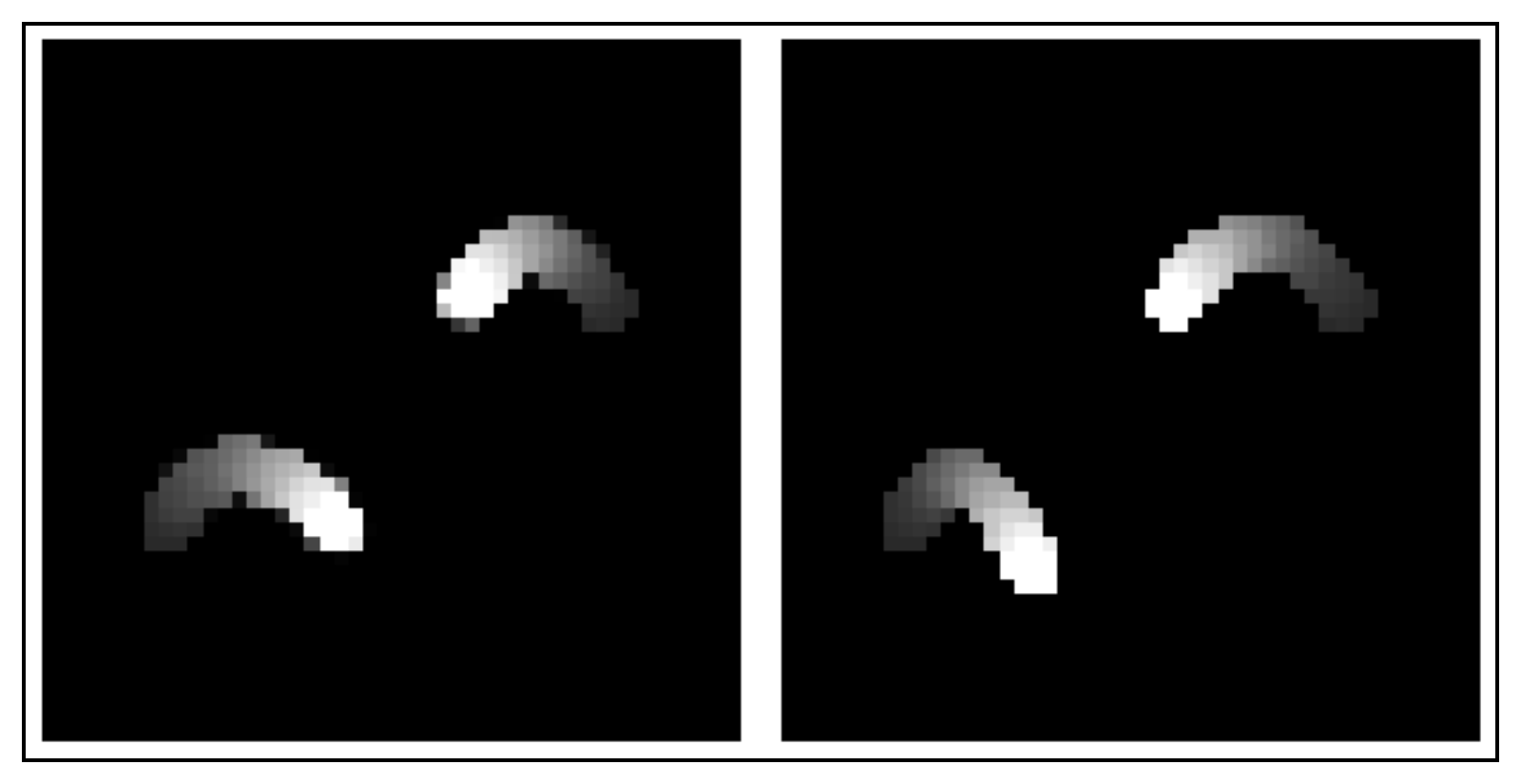}
\end{center}
\begin{center}
\hskip-0.3cm
\includegraphics[height=1.725cm,width=3.45cm]{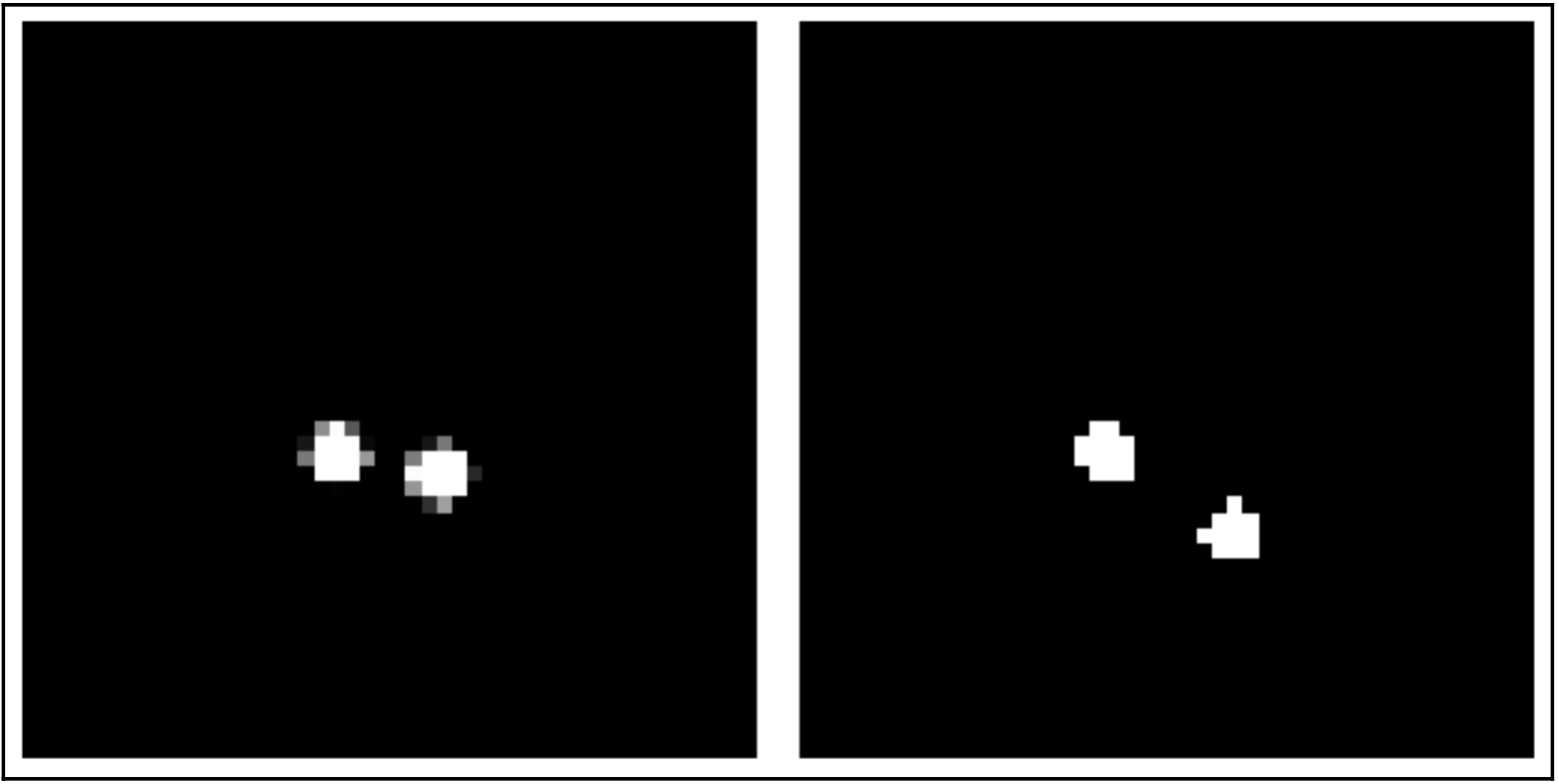}
\hskip0.04cm
\includegraphics[height=1.725cm,width=3.45cm]{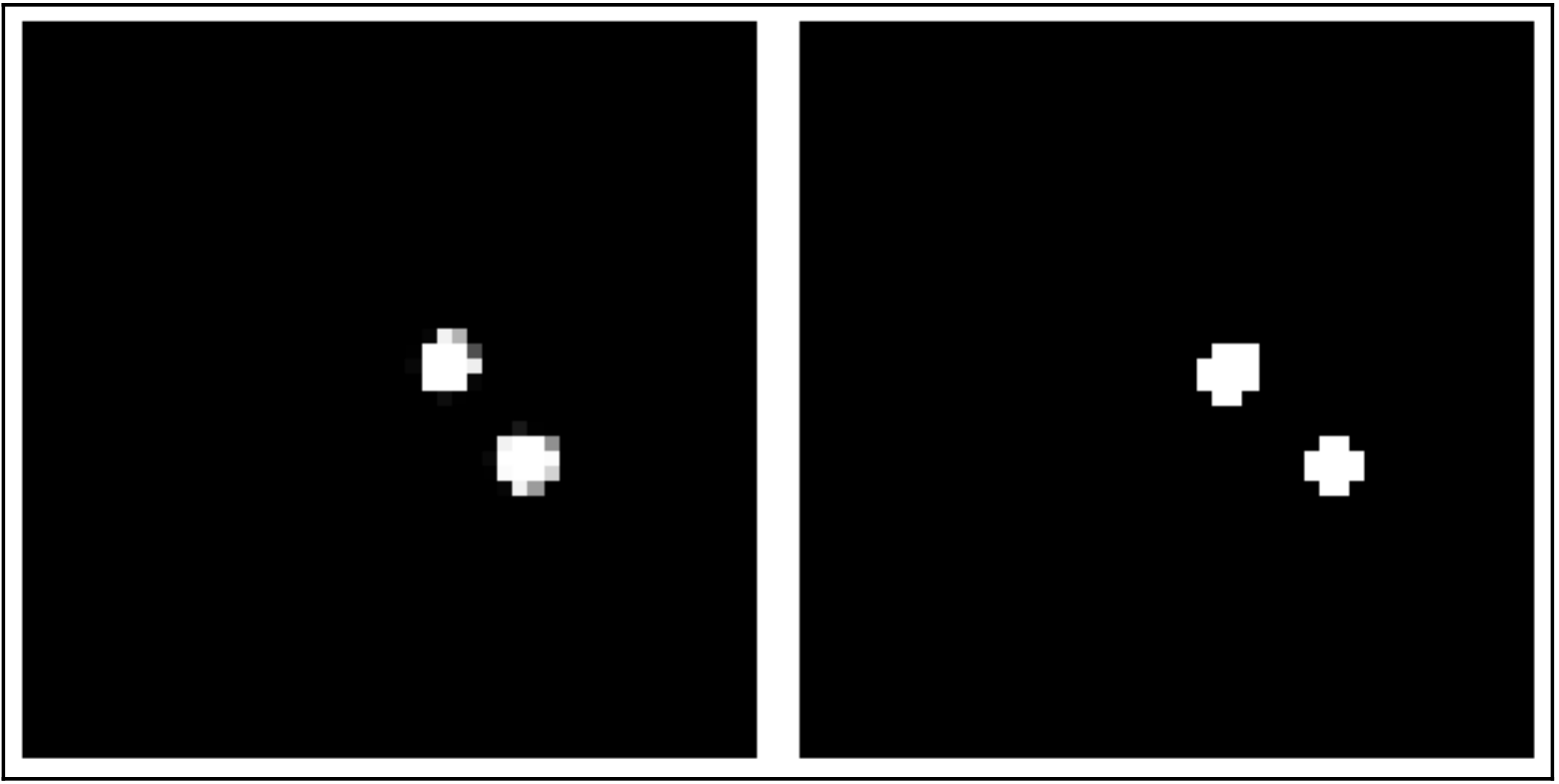}
\hskip0.04cm
\includegraphics[height=1.725cm,width=3.45cm]{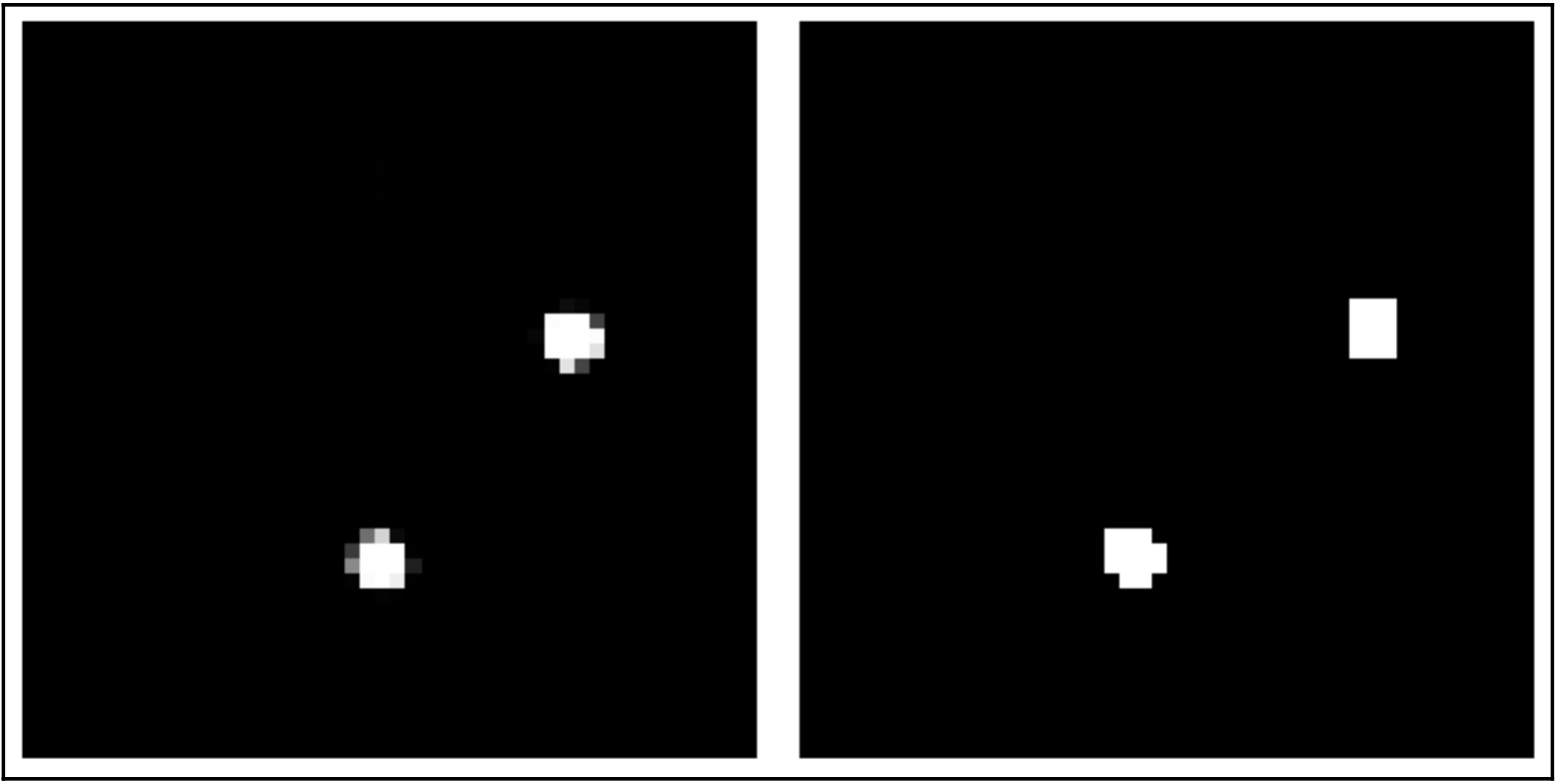}
\end{center}
\begin{center}
\hskip-0.3cm
\includegraphics[height=1.75cm,width=3.5cm]{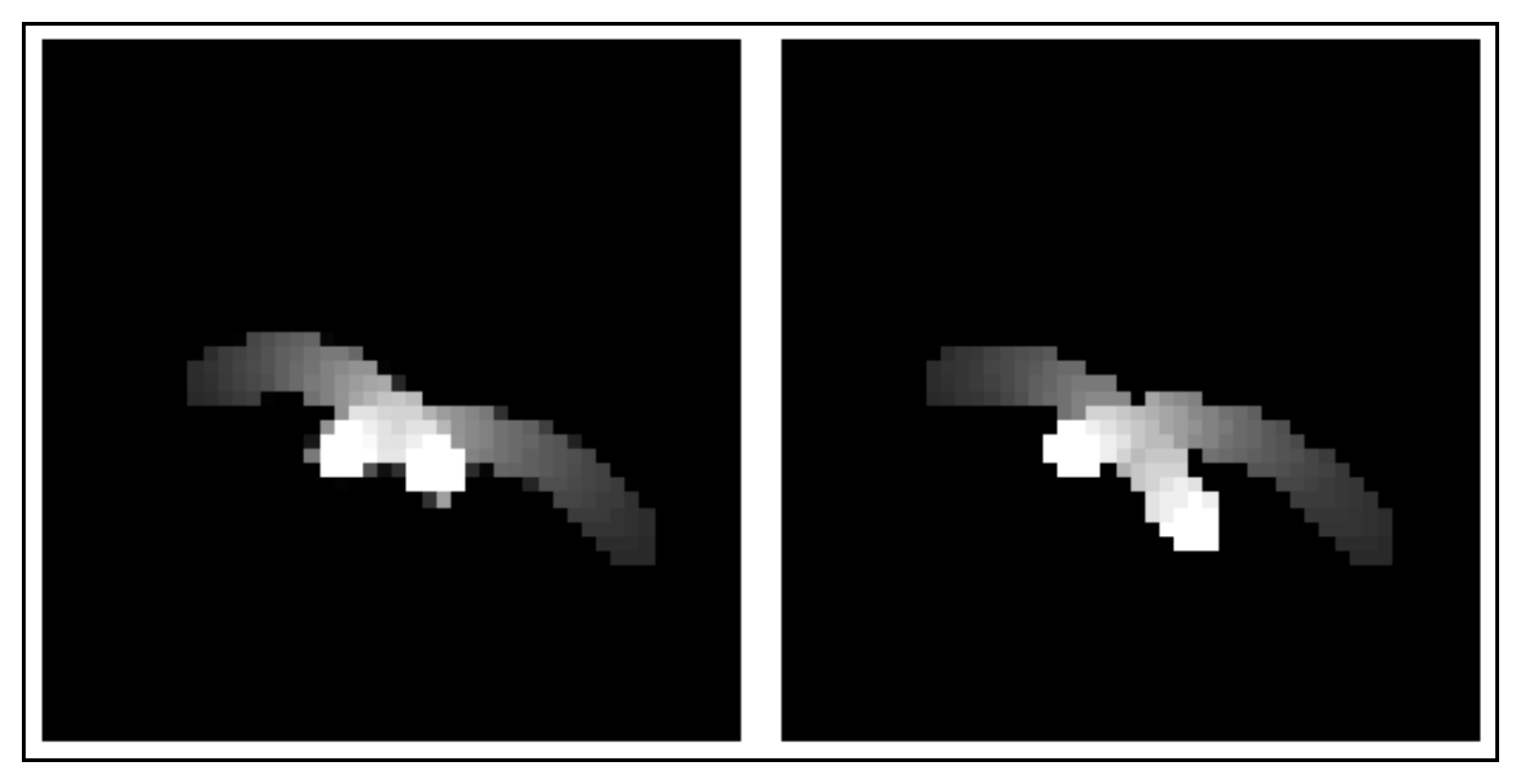}
\includegraphics[height=1.75cm,width=3.5cm]{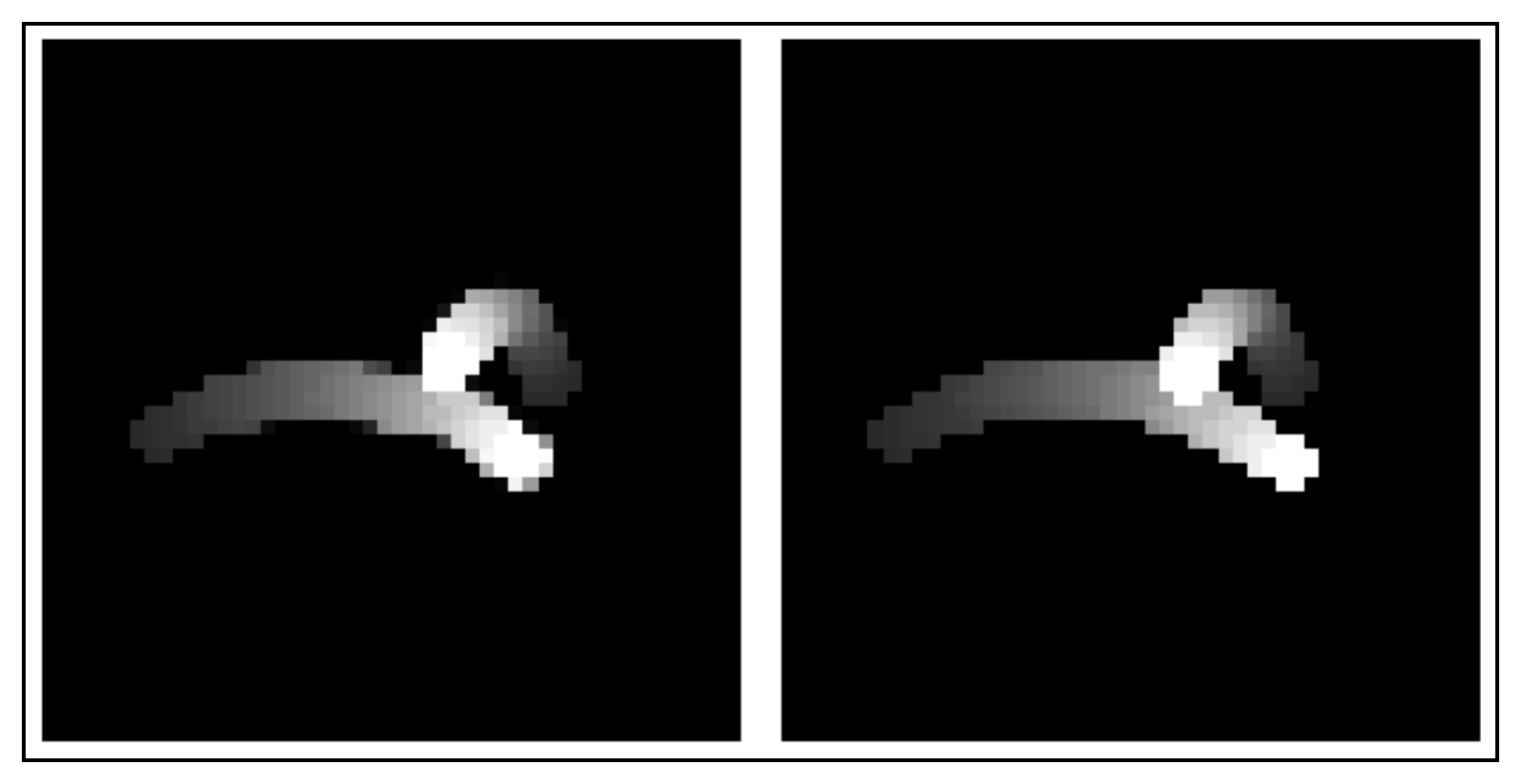}
\includegraphics[height=1.75cm,width=3.5cm]{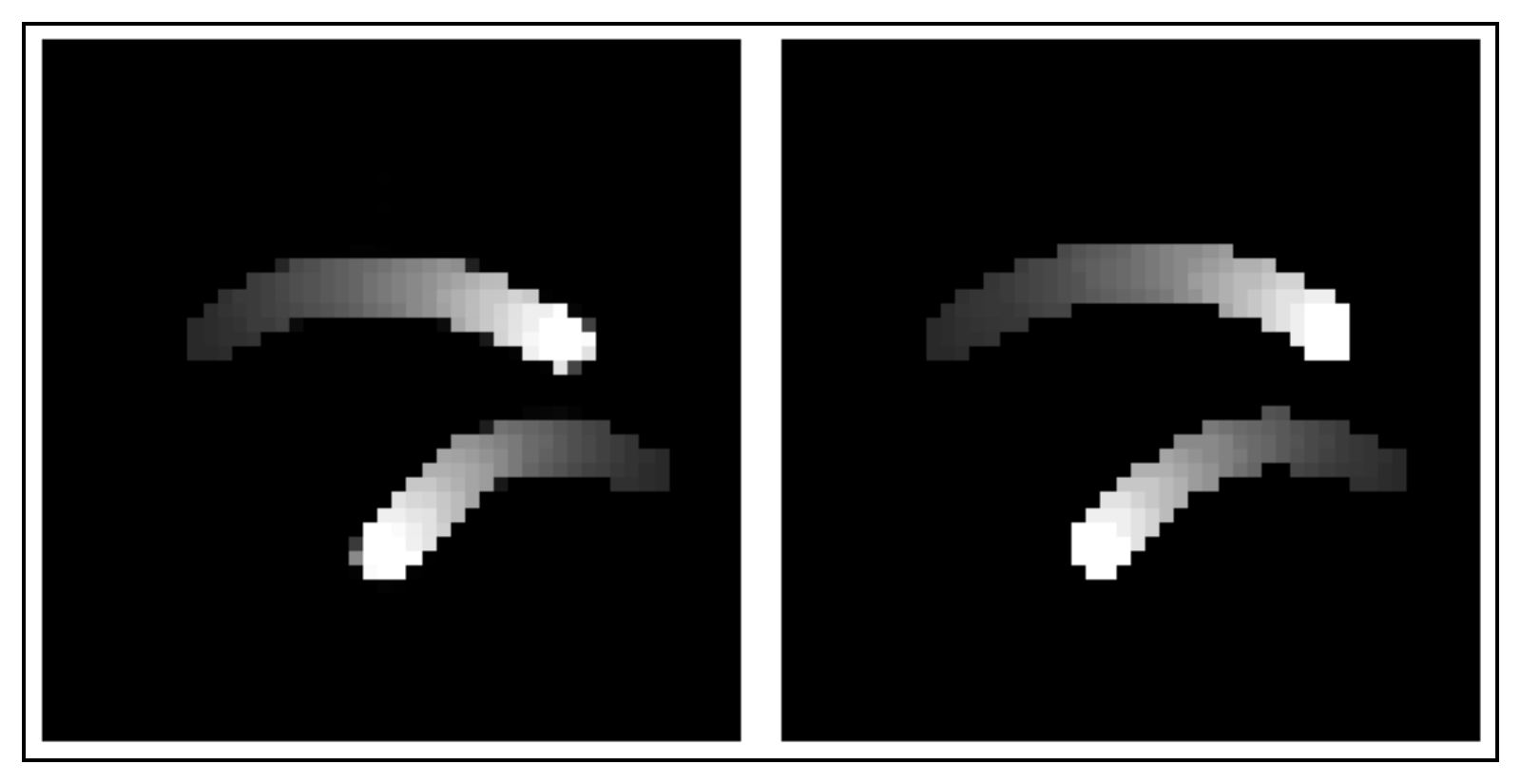}
\end{center}
\begin{center}
\hskip-0.3cm
\includegraphics[height=1.725cm,width=3.45cm]{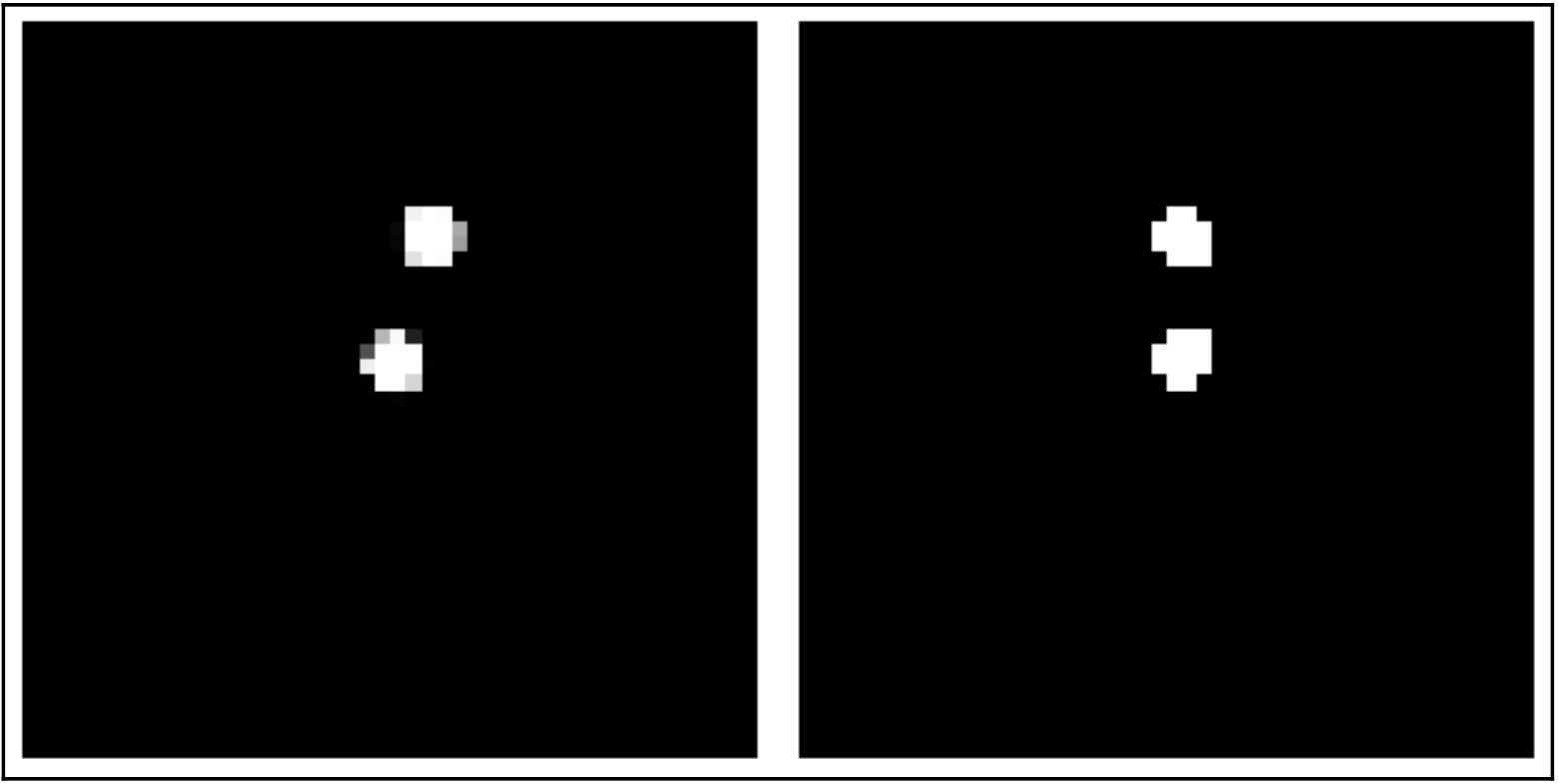}
\hskip0.04cm
\includegraphics[height=1.725cm,width=3.45cm]{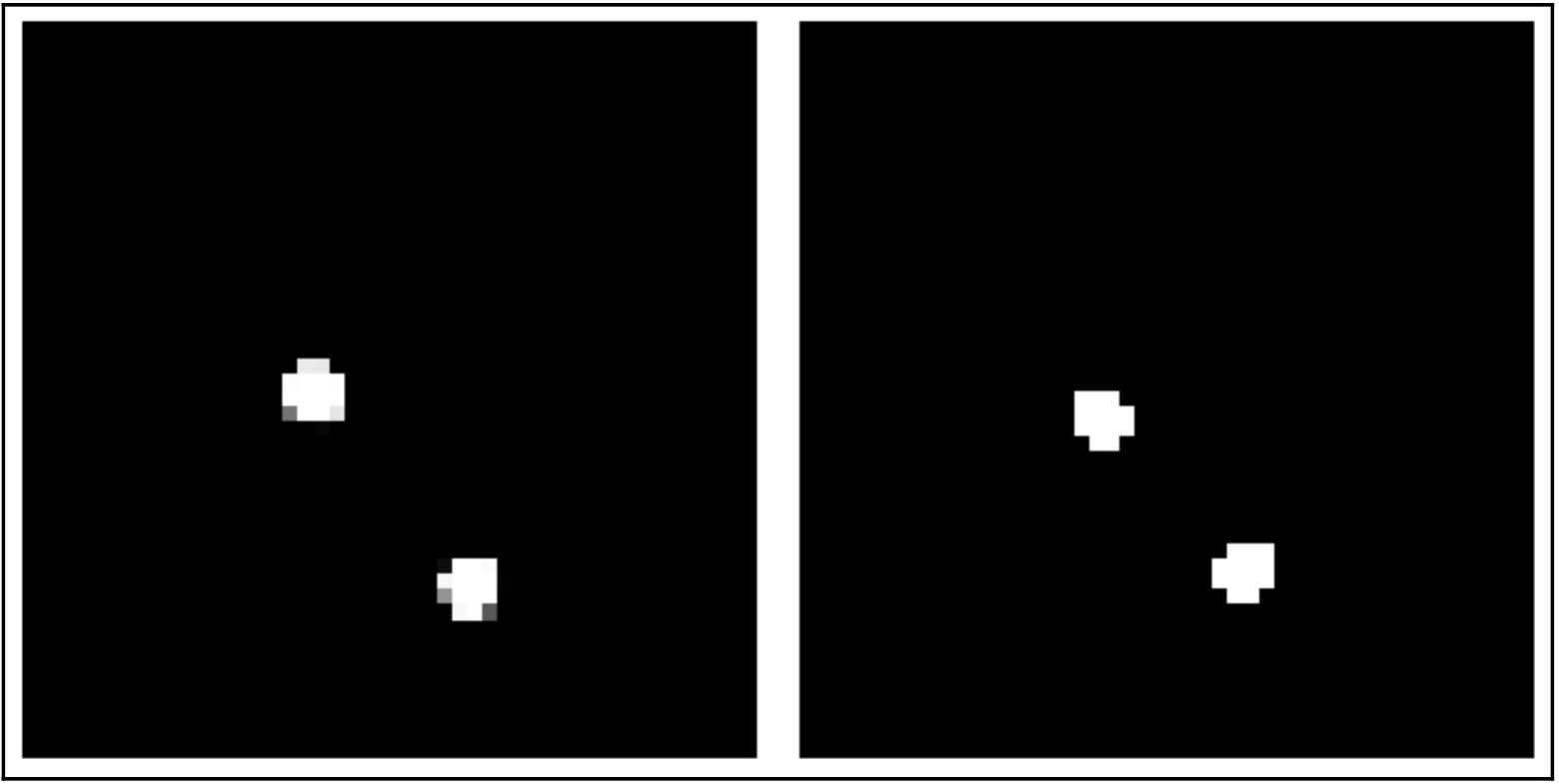}
\hskip0.04cm
\includegraphics[height=1.725cm,width=3.45cm]{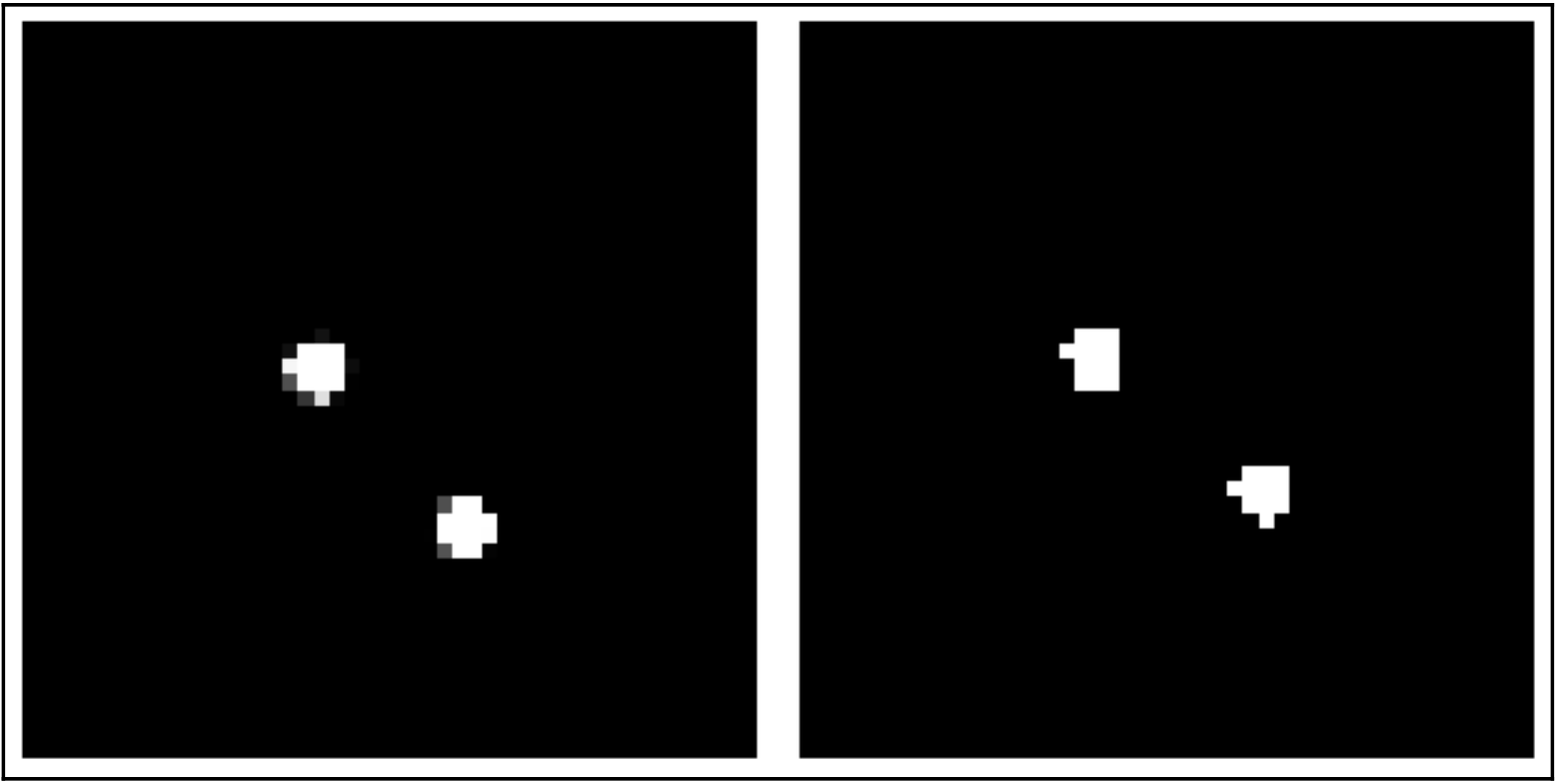}
\end{center}
\begin{center}
\hskip-0.3cm
\includegraphics[height=1.75cm,width=3.5cm]{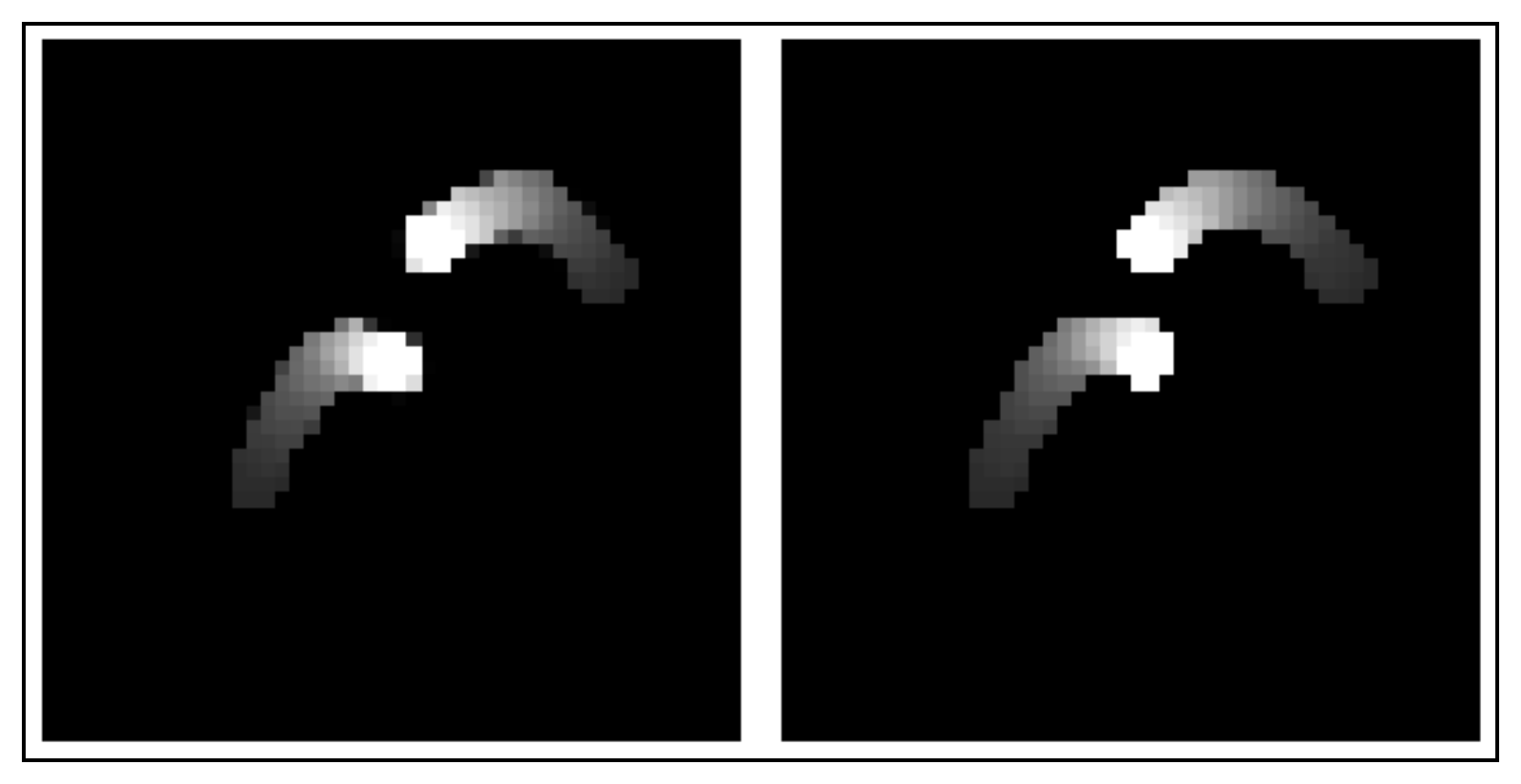}
\includegraphics[height=1.75cm,width=3.5cm]{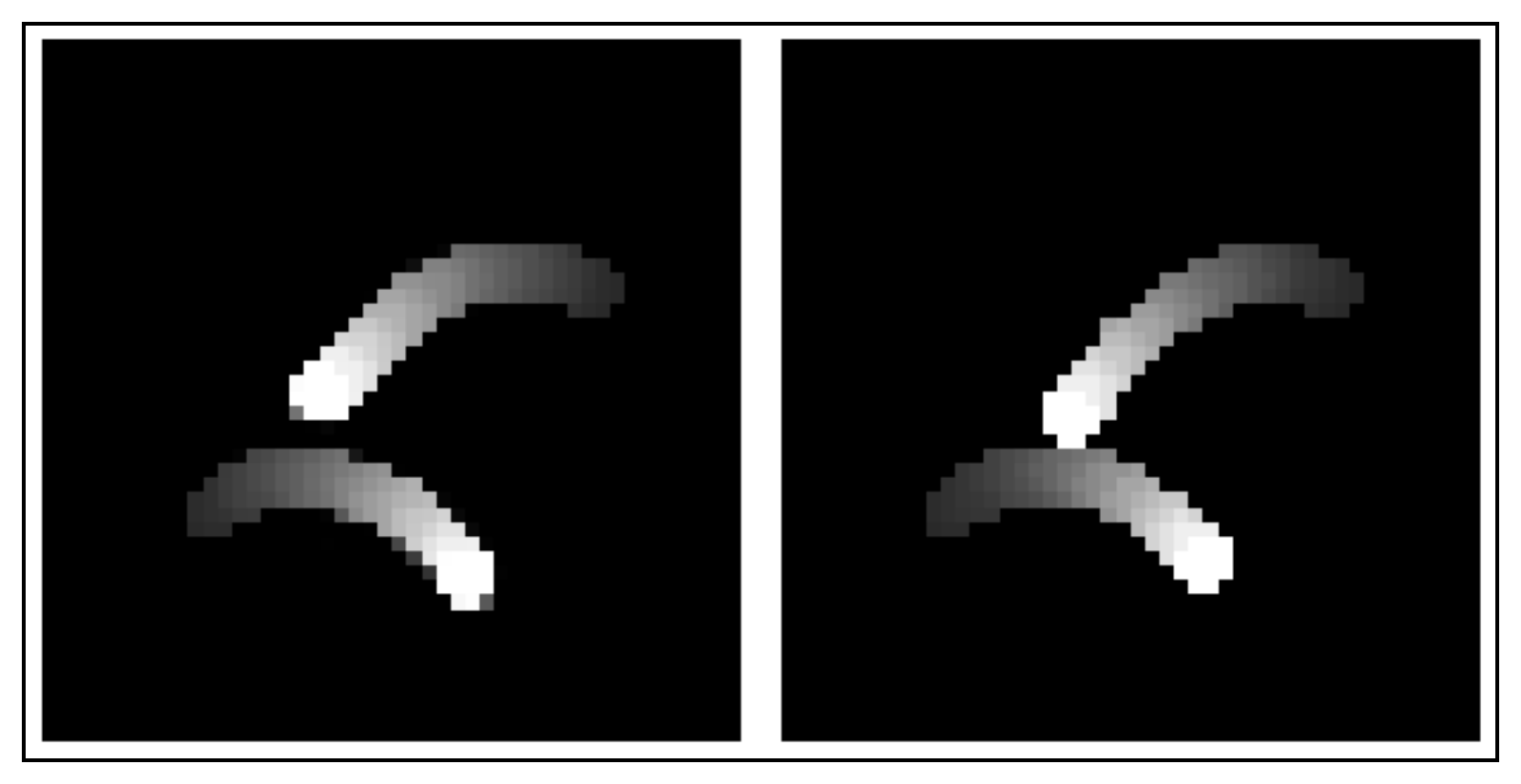}
\includegraphics[height=1.75cm,width=3.5cm]{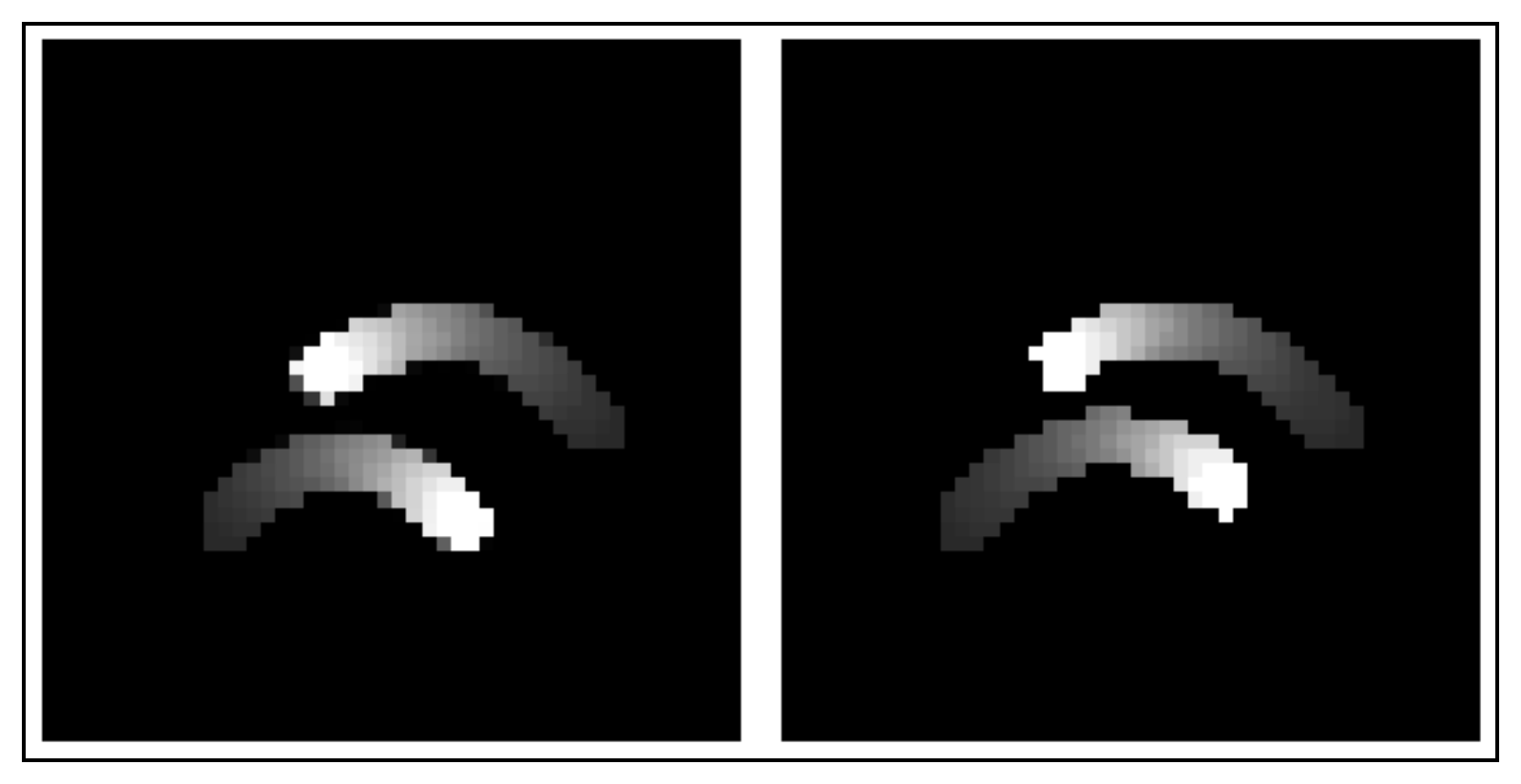}
\end{center}
\caption{Each plot shows generated (left) versus ground-truth (right) images at time-step 30 (top) and overlaid in time (bottom) for the ED-LSTM.}
\label{fig:GenObsFramesEDLSTM}
\end{figure}

\begin{figure}[t]
\begin{center}
\includegraphics[height=2.5cm,width=2.6cm]{./fig/exp_default_no5_ccov_delta_48_tm30_h5_ppg_seed340/340_152000_r_trj_2-crop.pdf}
\hskip0.1cm
\includegraphics[height=2.5cm,width=2.6cm]{./fig/exp_default_no5_ccov_delta_48_tm30_h5_ppg_seed340/340_152000_g_i_inter_r_trj_2-crop.pdf}\\
\includegraphics[height=2.2cm,width=4.4cm]{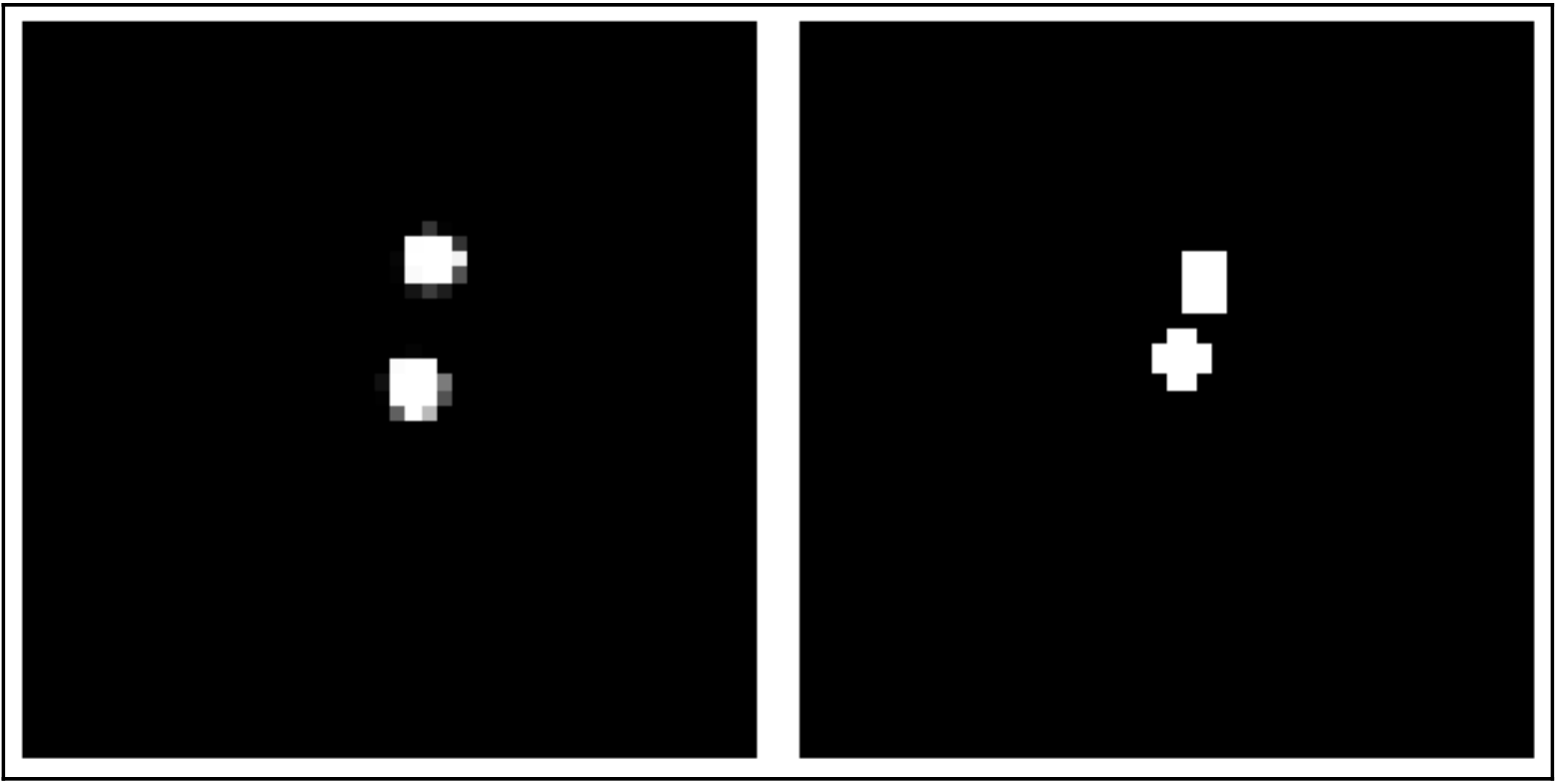}
\hskip0.1cm
\includegraphics[height=2.2cm,width=4.4cm]{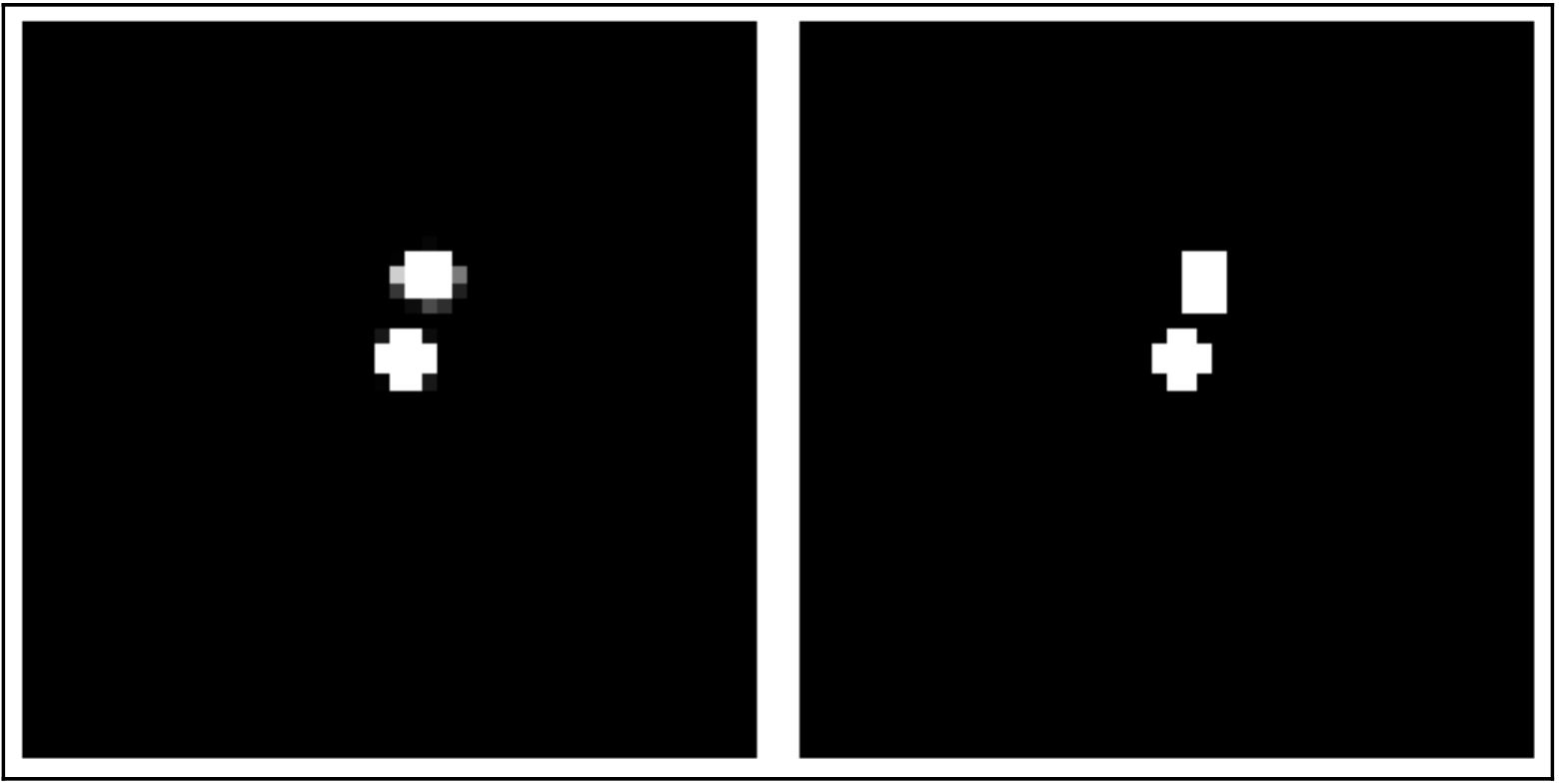}\\
\includegraphics[height=2.25cm,width=4.5cm]{./fig/exp_default_no5_ccov_delta_48_tm30_h5_ppg_seed340/340_152000_gen_obs_frames_2_all.pdf}
\hskip0.0cm
\includegraphics[height=2.25cm,width=4.5cm]{./fig/exp_default_no5_ccov_delta_48_tm30_h5_ppg_seed340/340_152000_inter_obs_frames_2_all.pdf}
\end{center} 
\begin{center}
\hskip0.1cm
\includegraphics[height=2.5cm,width=2.6cm]{./fig/exp_default_no5_ccov_delta_48_tm30_h5_ppg_seed340/340_152000_r_trj_4-crop.pdf}
\hskip0.15cm
\includegraphics[height=2.5cm,width=2.6cm]{./fig/exp_default_no5_ccov_delta_48_tm30_h5_ppg_seed340/340_152000_g_i_inter_r_trj_4-crop.pdf}\\
\hskip0.1cm
\includegraphics[height=2.2cm,width=4.4cm]{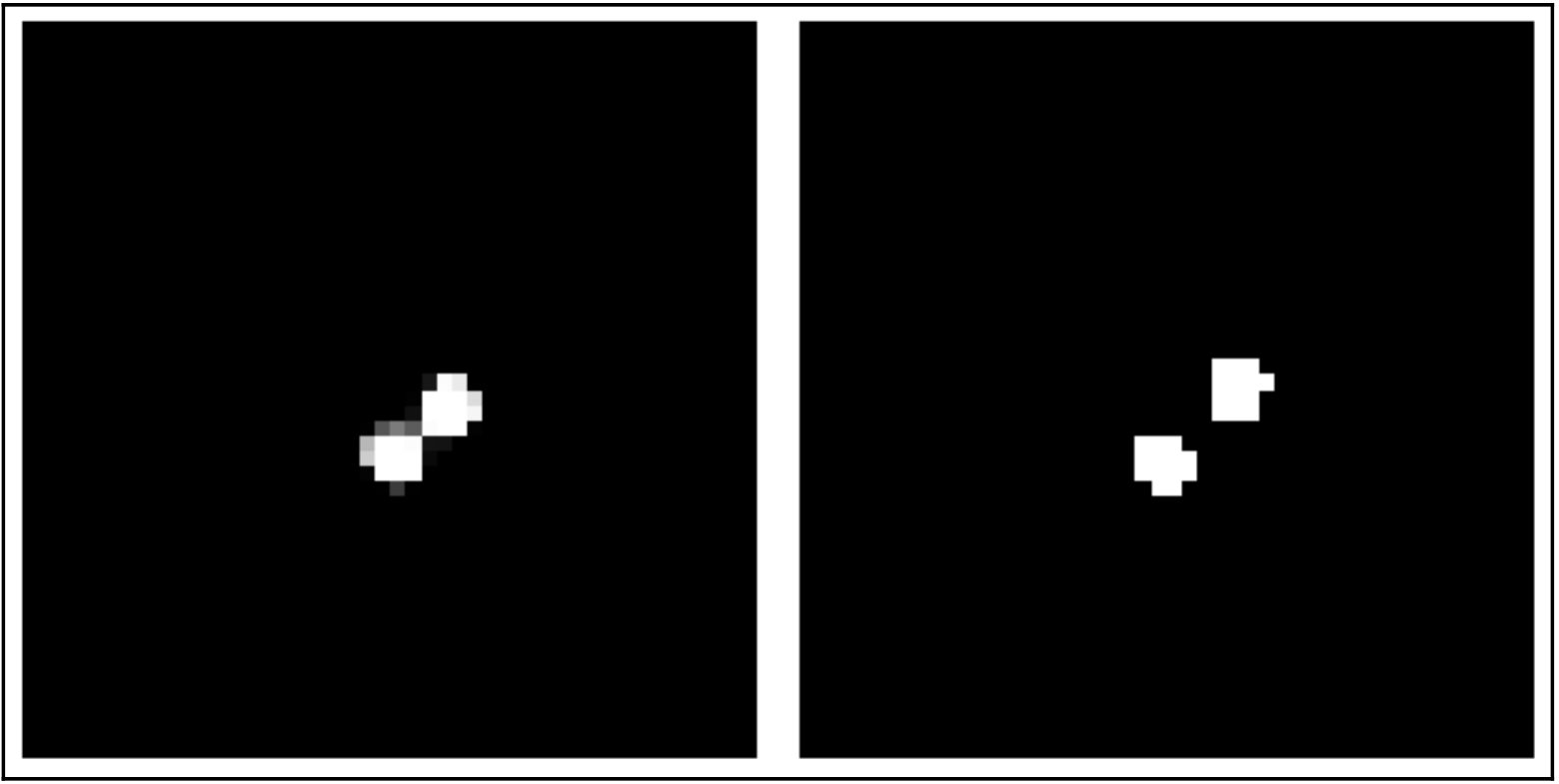}
\hskip0.1cm
\includegraphics[height=2.2cm,width=4.4cm]{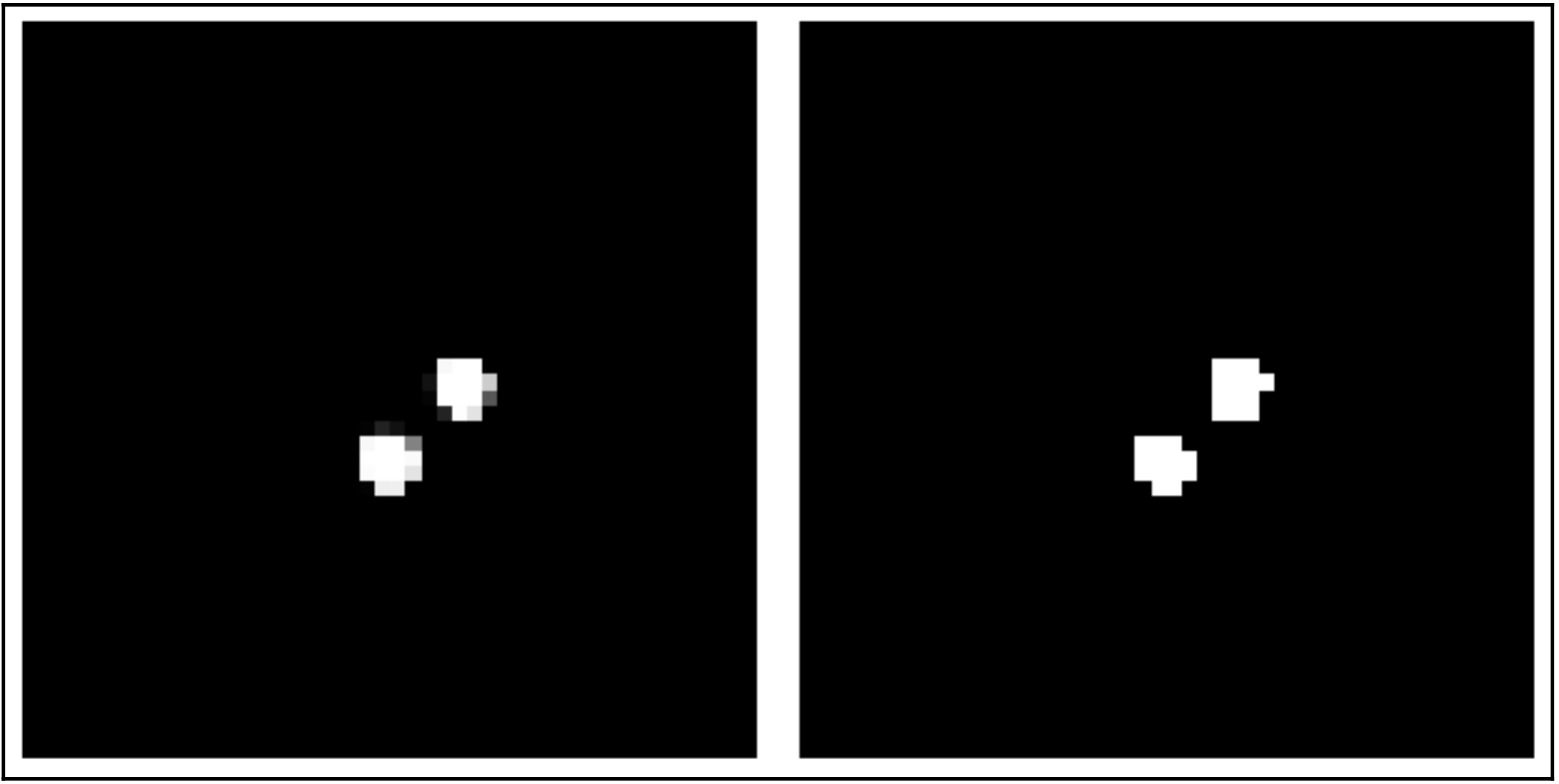}\\
\hskip0.1cm
\includegraphics[height=2.25cm,width=4.5cm]{./fig/exp_default_no5_ccov_delta_48_tm30_h5_ppg_seed340/340_152000_gen_obs_frames_4_all.pdf}
\hskip0.0cm
\includegraphics[height=2.25cm,width=4.5cm]{./fig/exp_default_no5_ccov_delta_48_tm30_h5_ppg_seed340/340_152000_inter_obs_frames_4_all.pdf}
\end{center}
\begin{center}
\hskip0.1cm
\includegraphics[height=2.5cm,width=2.6cm]{./fig/exp_default_no5_ccov_delta_48_tm30_h5_ppg_seed340/340_154000_r_trj_0-crop.pdf}
\hskip0.1cm
\includegraphics[height=2.5cm,width=2.6cm]{./fig/exp_default_no5_ccov_delta_48_tm30_h5_ppg_seed340/340_154000_g_i_inter_r_trj_0-crop.pdf}\\
\hskip0.1cm
\includegraphics[height=2.2cm,width=4.4cm]{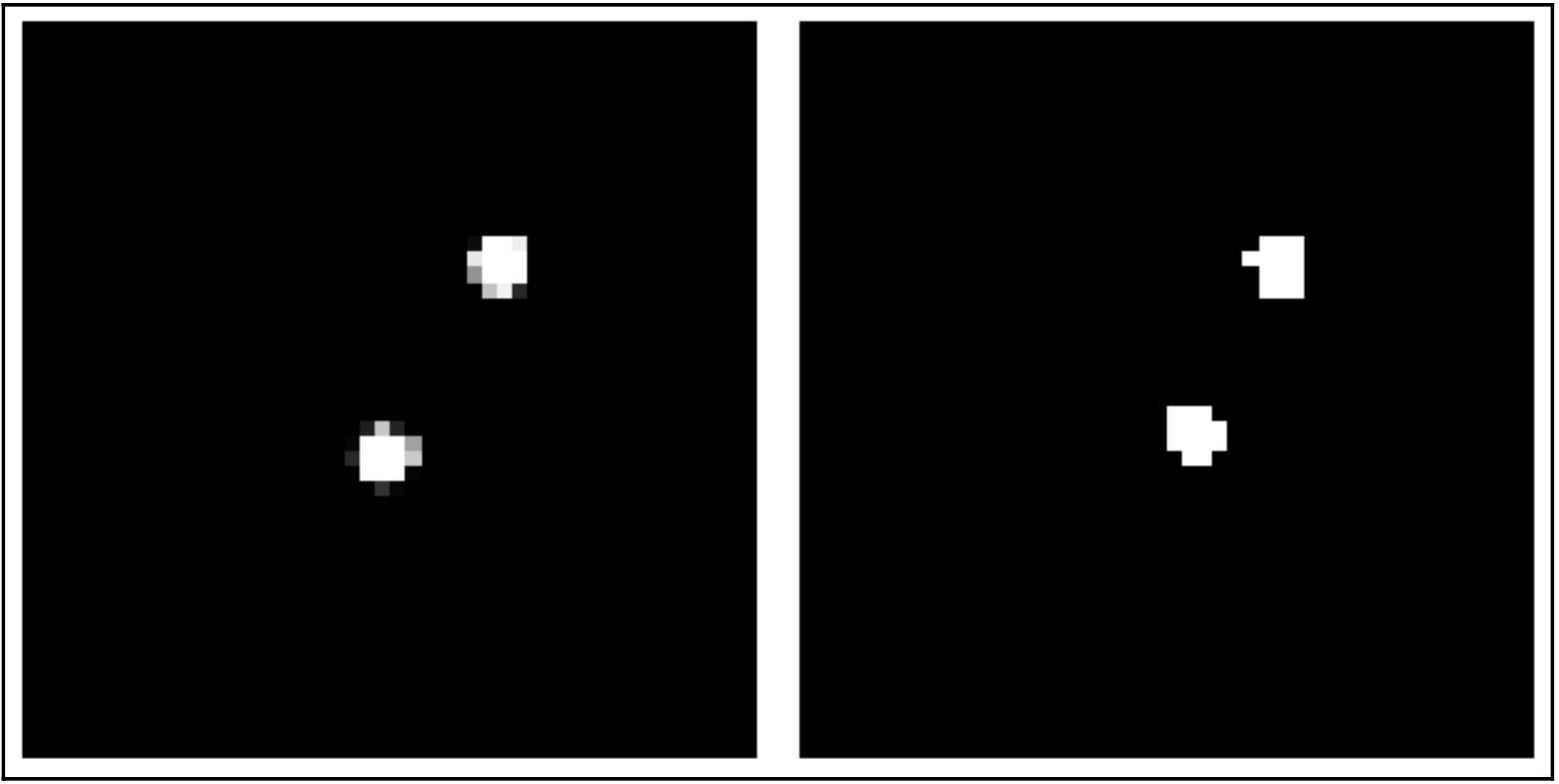}
\hskip0.1cm
\includegraphics[height=2.2cm,width=4.4cm]{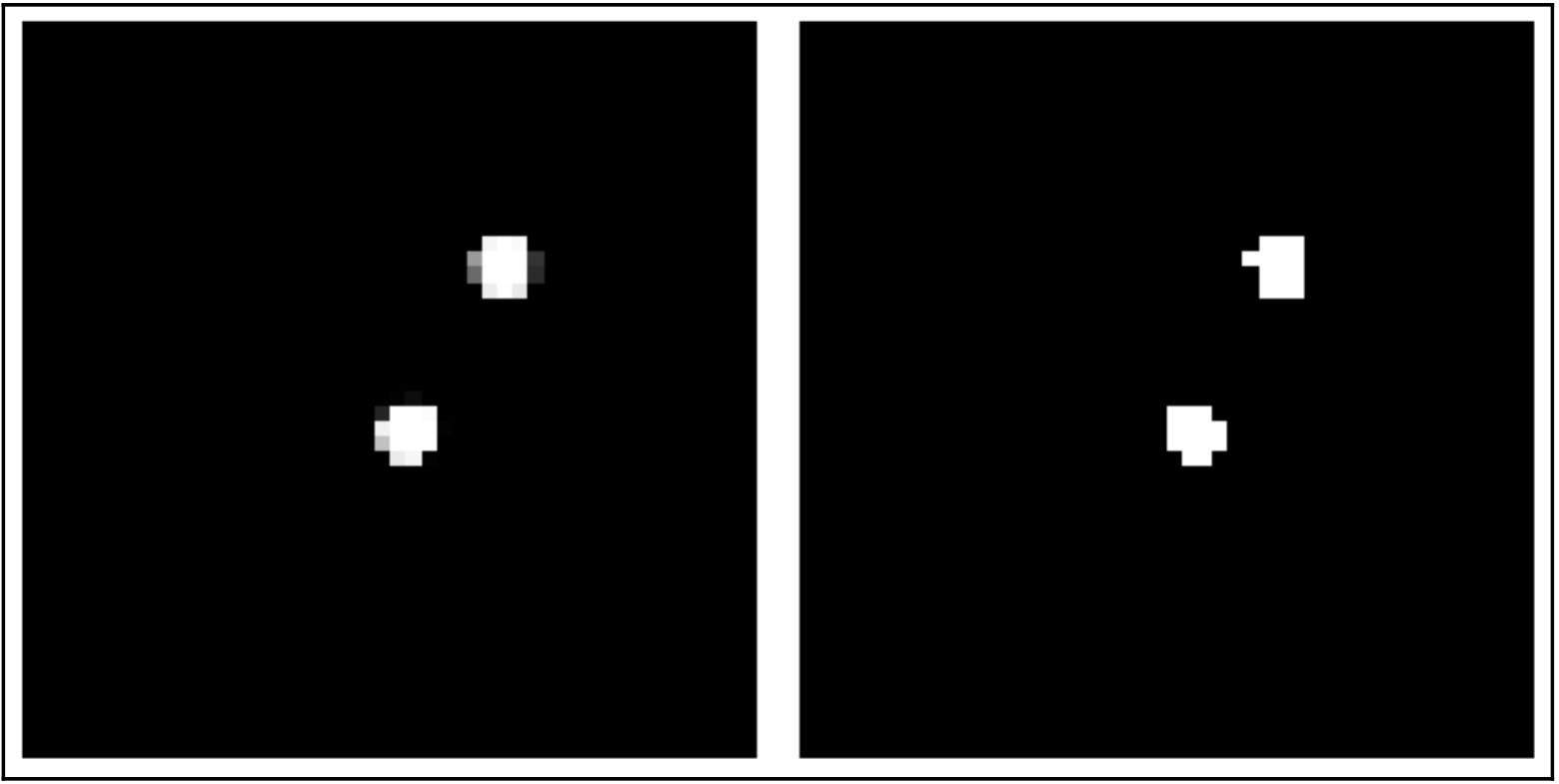} \\
\hskip0.1cm
\includegraphics[height=2.25cm,width=4.5cm]{./fig/exp_default_no5_ccov_delta_48_tm30_h5_ppg_seed340/340_154000_gen_obs_frames_0_all.pdf}
\hskip0.0cm
\includegraphics[height=2.25cm,width=4.5cm]{./fig/exp_default_no5_ccov_delta_48_tm30_h5_ppg_seed340/340_154000_inter_obs_frames_0_all.pdf}
\end{center}
\caption{Top: Ground-truth (black), inferred (blue), generated (red), and interpolated (cyan) trajectories. 
Middle: Generated versus ground-truth images and interpolated versus ground-truth images at time-step 30.
Bottom: Generated versus ground-truth images and interpolated versus ground-truth images overlaid in time.}
\label{fig:Inter1}
\end{figure}

\begin{figure}[t]
\begin{center}
\includegraphics[height=2.5cm,width=2.6cm]{./fig/exp_default_no5_ccov_delta_48_tm30_h5_ppg_seed340/340_154000_r_trj_4-crop.pdf}
\hskip0.1cm
\includegraphics[height=2.5cm,width=2.6cm]{./fig/exp_default_no5_ccov_delta_48_tm30_h5_ppg_seed340/340_154000_g_i_inter_r_trj_4-crop.pdf}\\
\includegraphics[height=2.2cm,width=4.4cm]{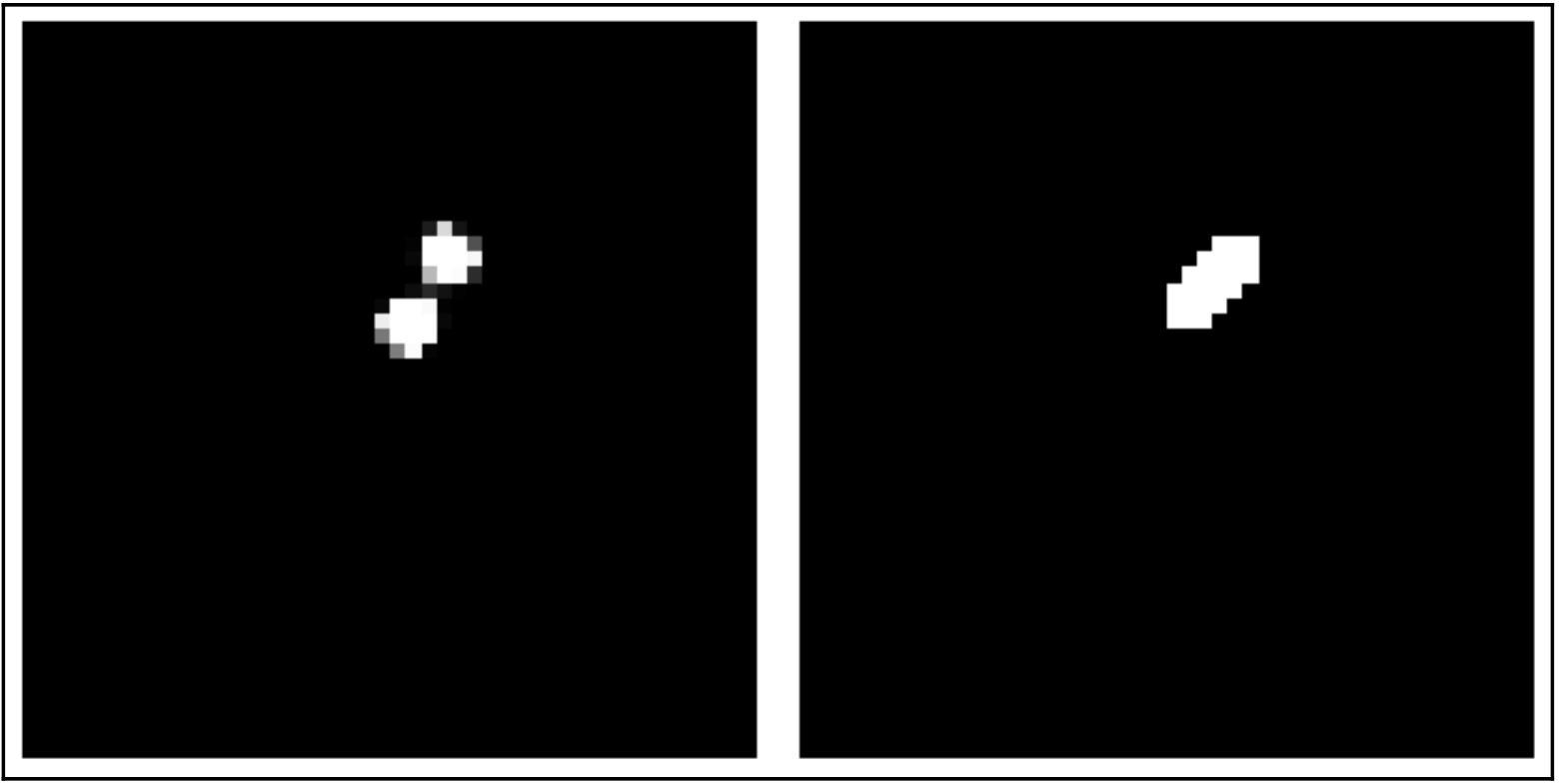}
\hskip0.1cm
\includegraphics[height=2.2cm,width=4.4cm]{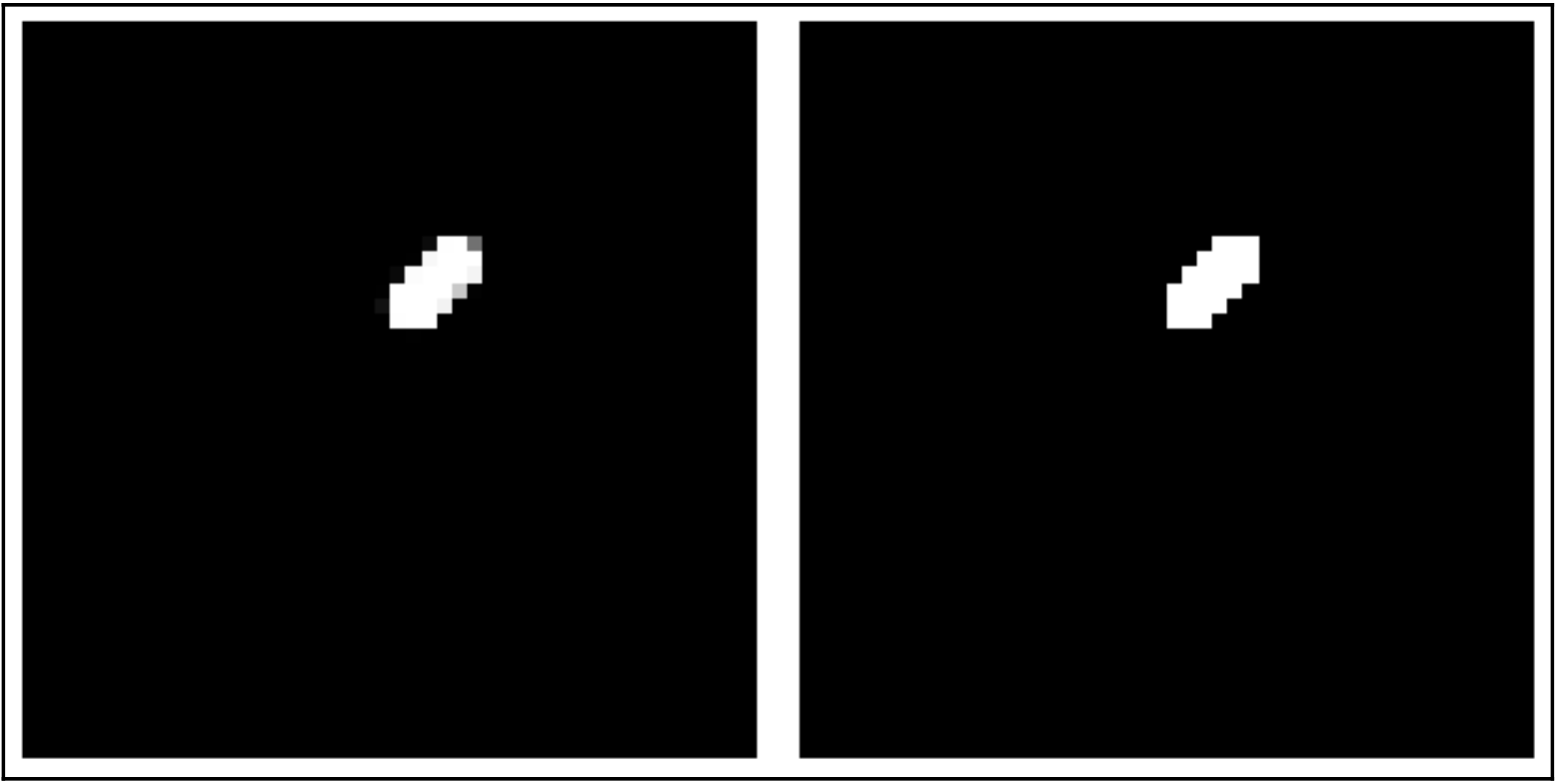}\\
\hskip0.1cm
\includegraphics[height=2.25cm,width=4.5cm]{./fig/exp_default_no5_ccov_delta_48_tm30_h5_ppg_seed340/340_154000_gen_obs_frames_4_all.pdf}
\hskip0.0cm
\includegraphics[height=2.25cm,width=4.5cm]{./fig/exp_default_no5_ccov_delta_48_tm30_h5_ppg_seed340/340_154000_inter_obs_frames_4_all.pdf}
\end{center}
\begin{center}
\includegraphics[height=2.5cm,width=2.6cm]{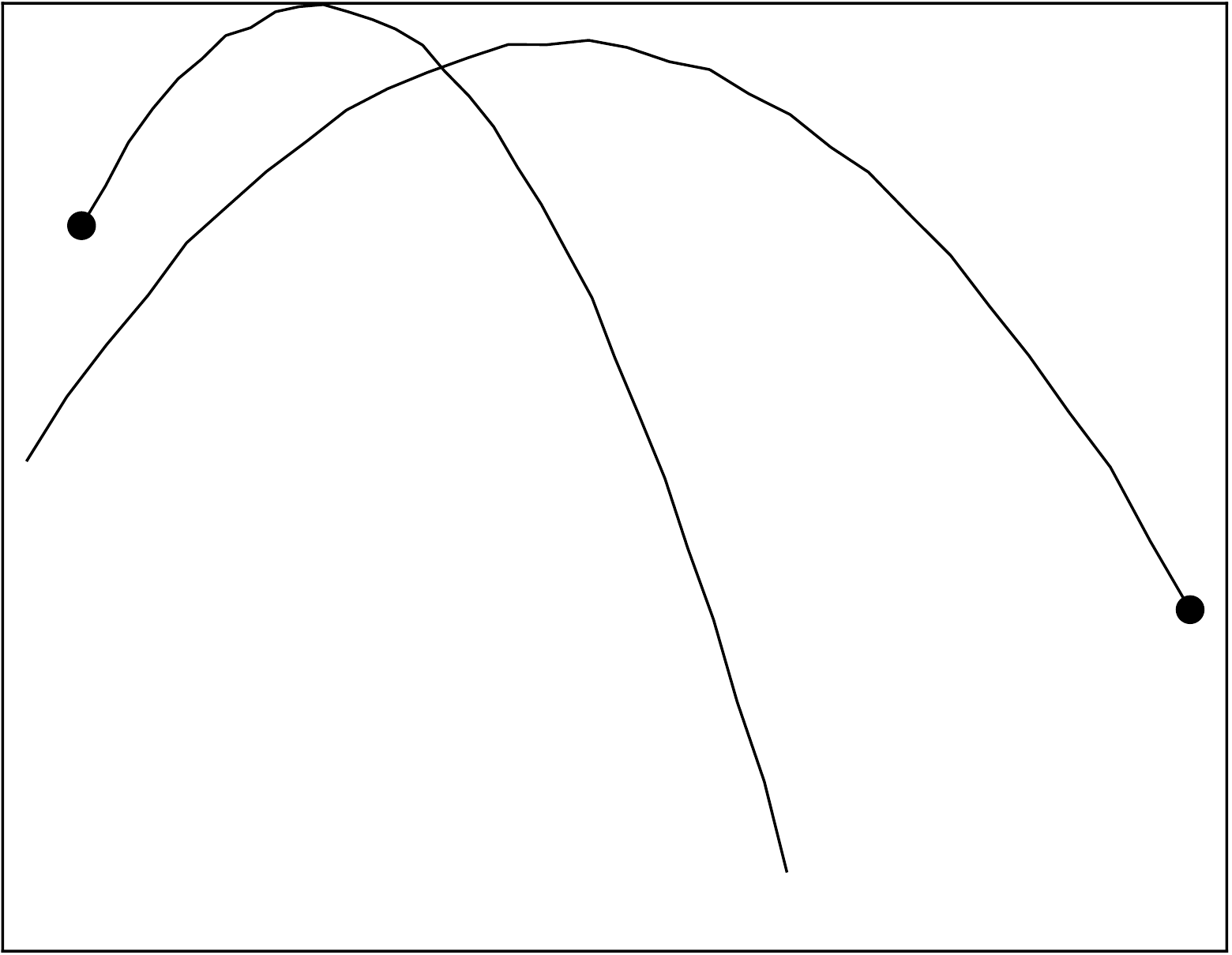}
\hskip0.1cm
\includegraphics[height=2.5cm,width=2.6cm]{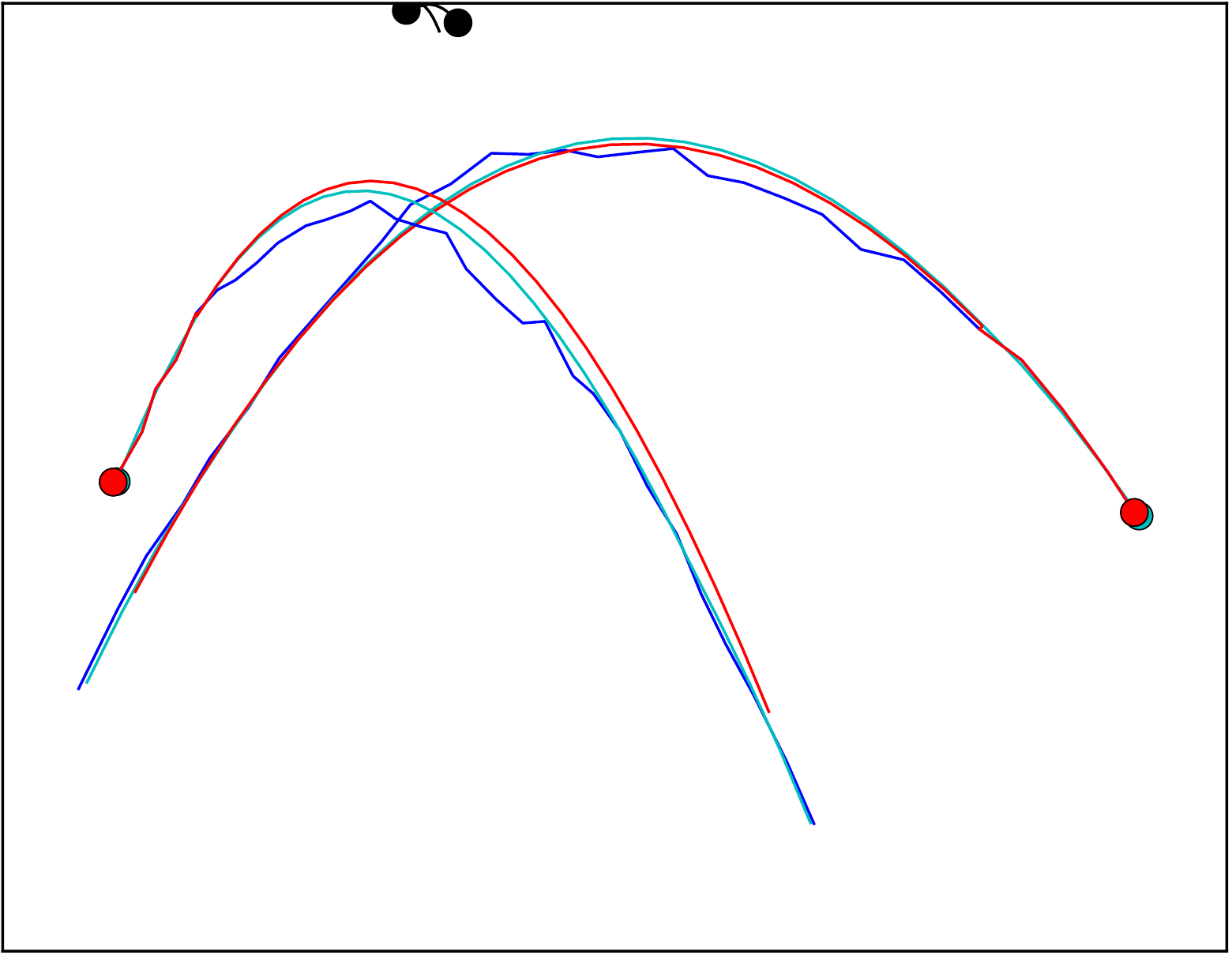}\\
\includegraphics[height=2.2cm,width=4.4cm]{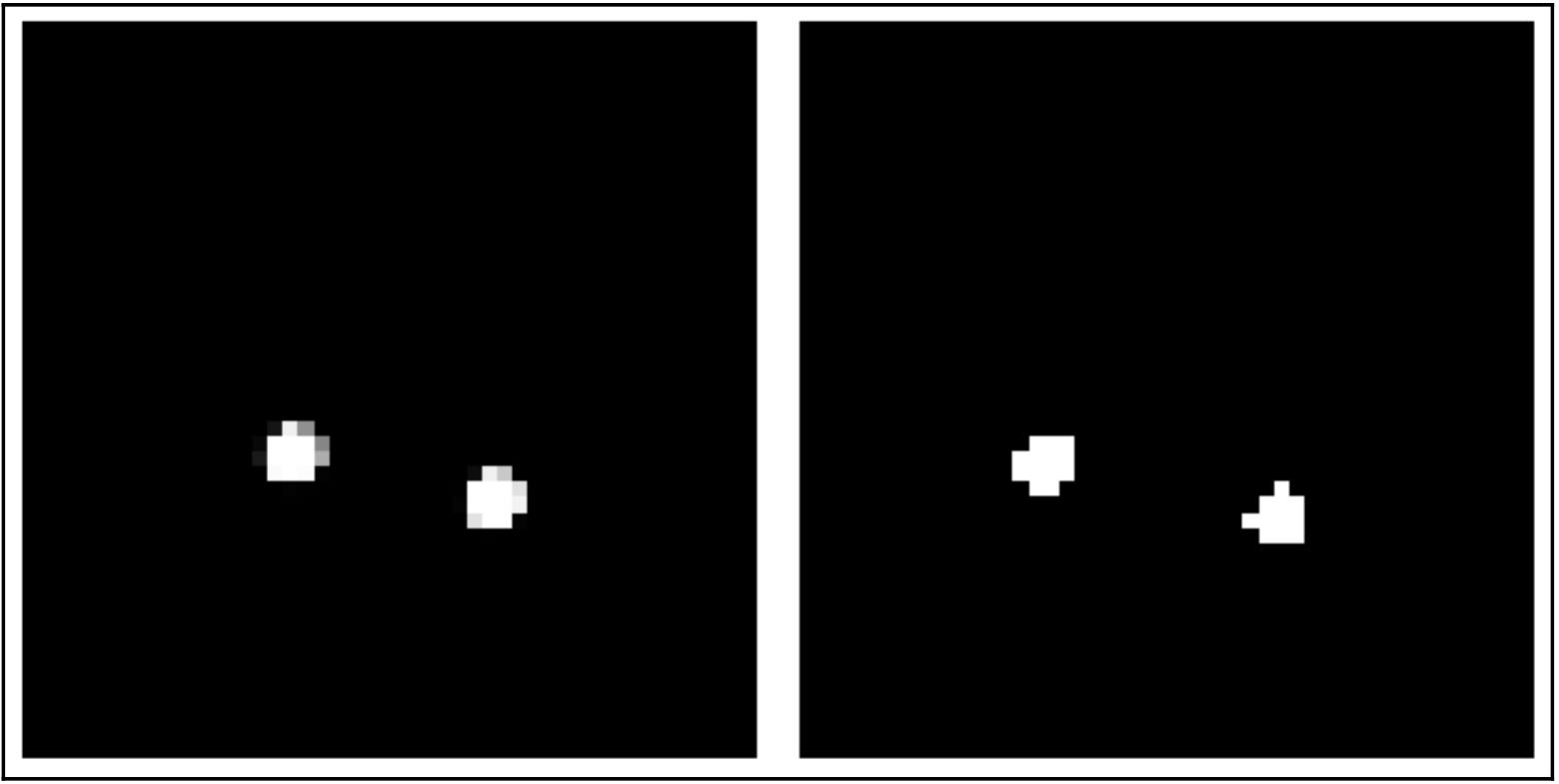}
\hskip0.1cm
\includegraphics[height=2.2cm,width=4.4cm]{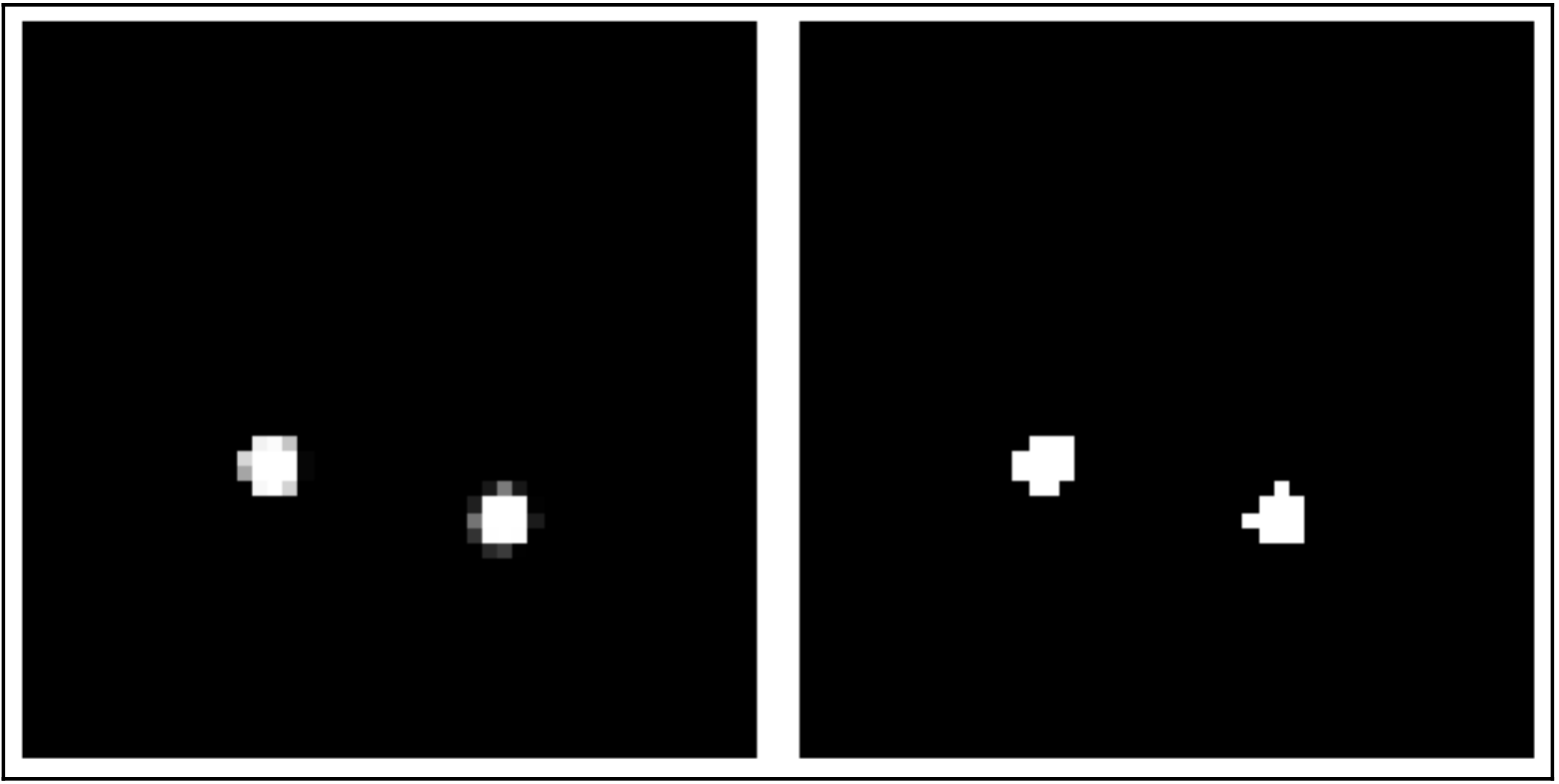}\\
\hskip0.0cm
\includegraphics[height=2.25cm,width=4.5cm]{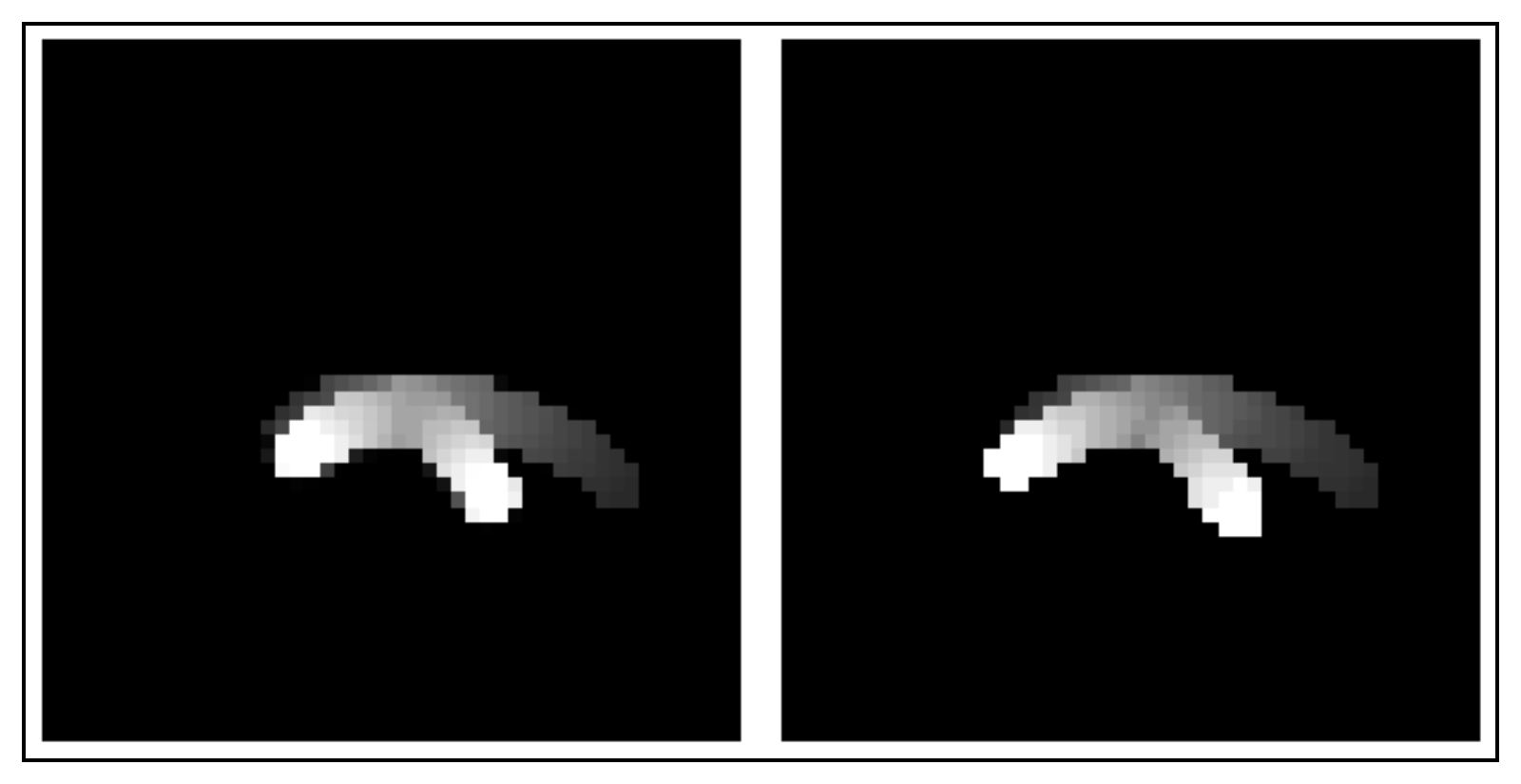}
\hskip0.0cm
\includegraphics[height=2.25cm,width=4.5cm]{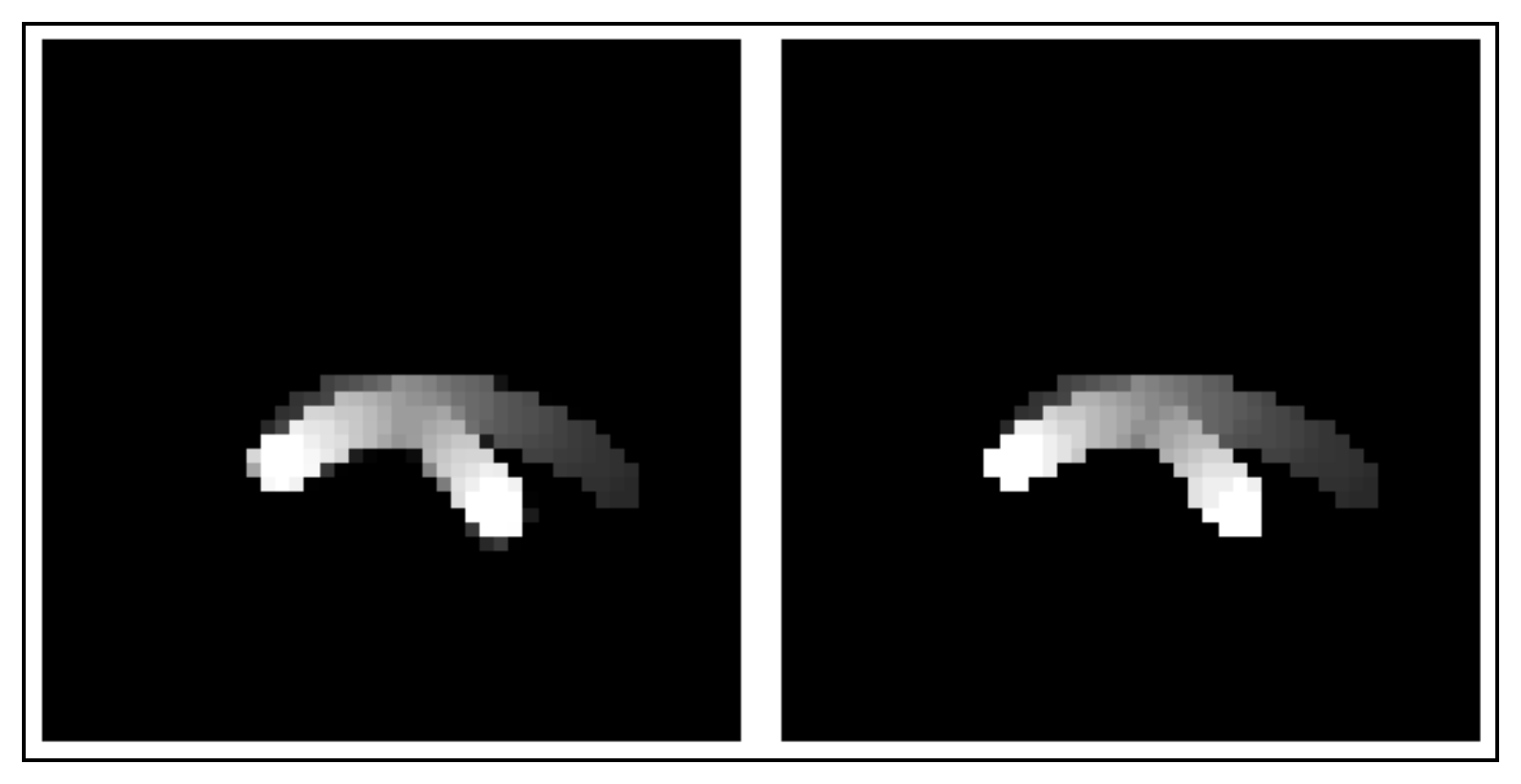}
\end{center}
\begin{center}
\includegraphics[height=2.5cm,width=2.6cm]{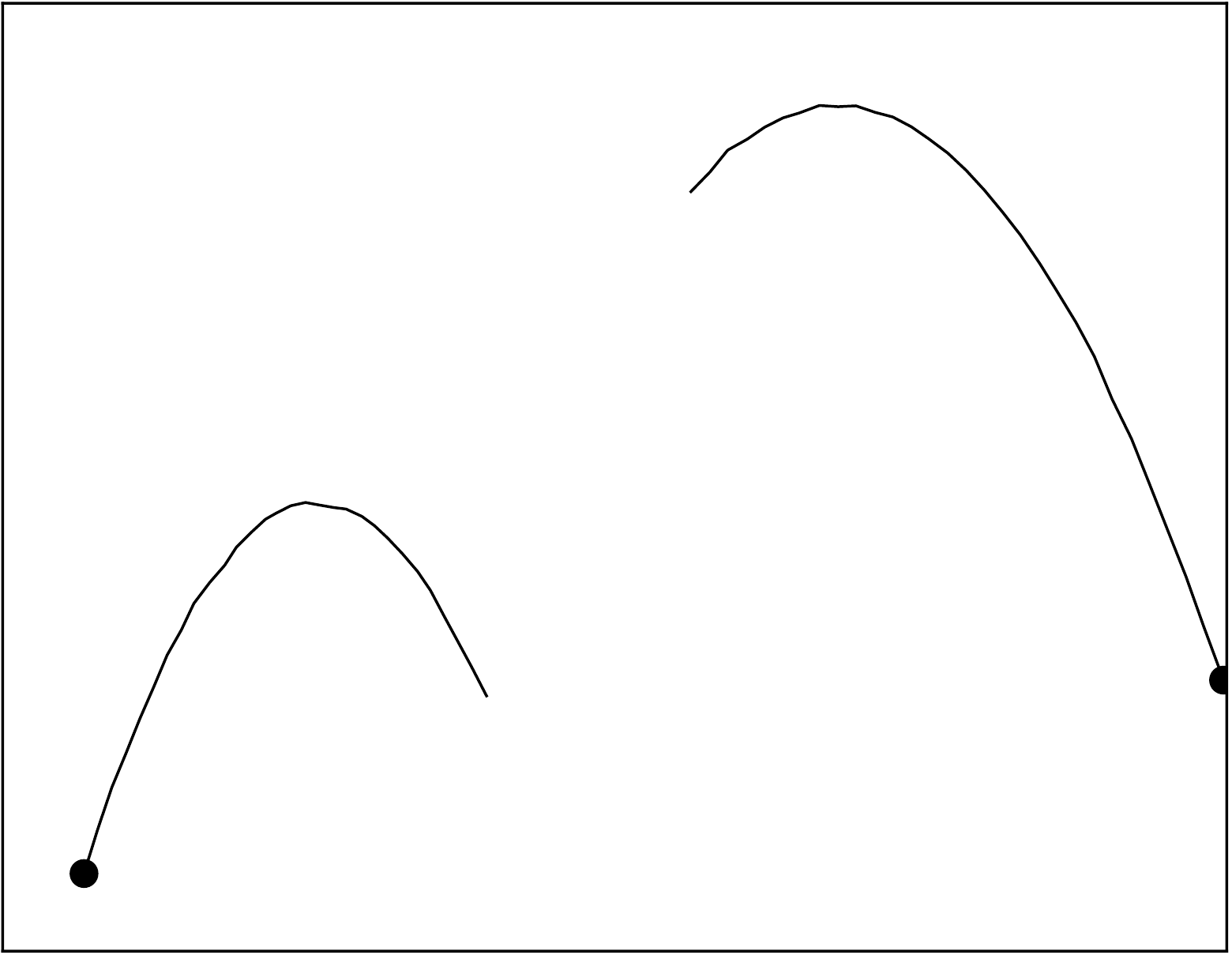}
\hskip0.1cm
\includegraphics[height=2.5cm,width=2.6cm]{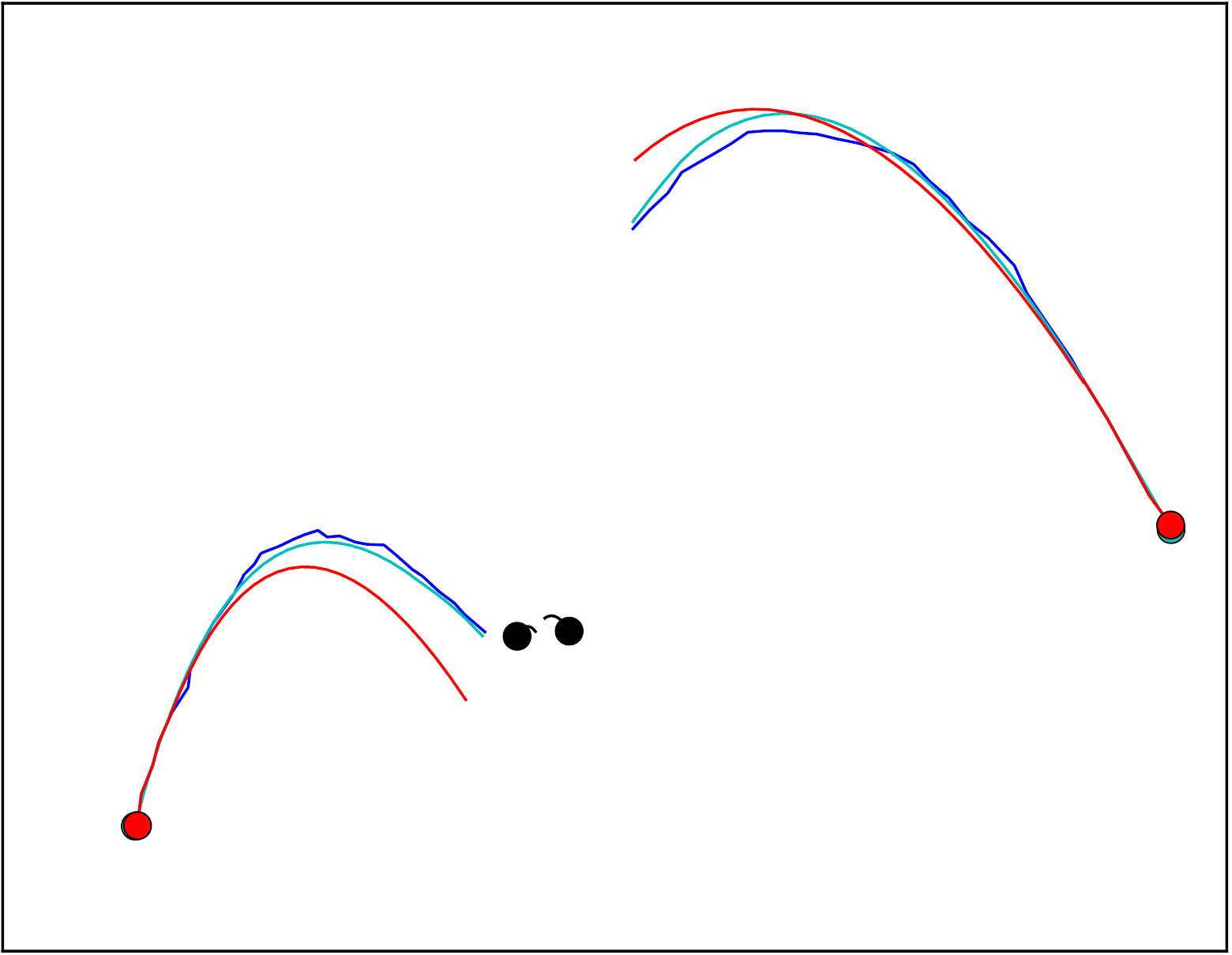}\\
\includegraphics[height=2.2cm,width=4.4cm]{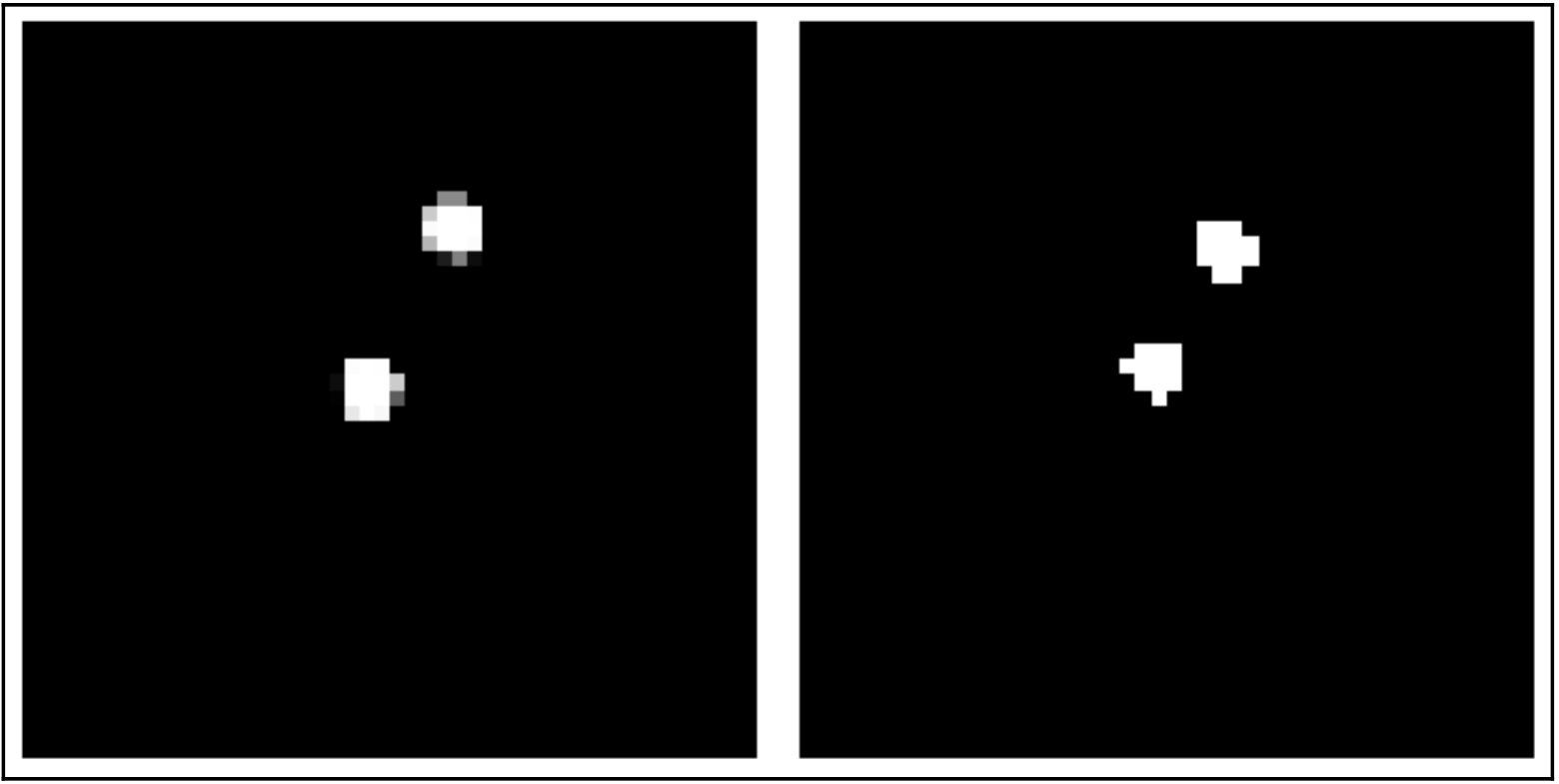}
\hskip0.1cm
\includegraphics[height=2.2cm,width=4.4cm]{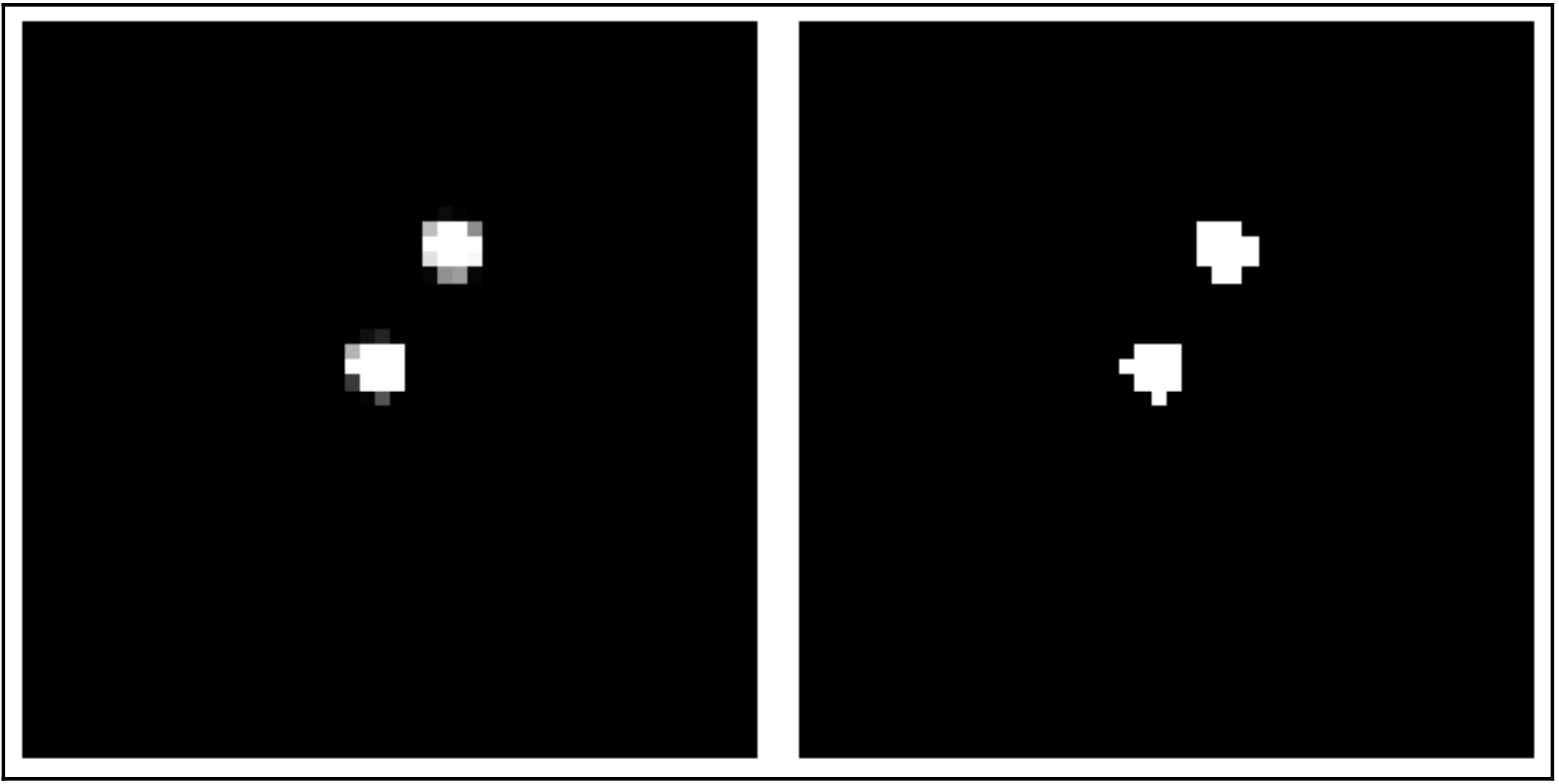} \\
\hskip0.0cm
\includegraphics[height=2.25cm,width=4.5cm]{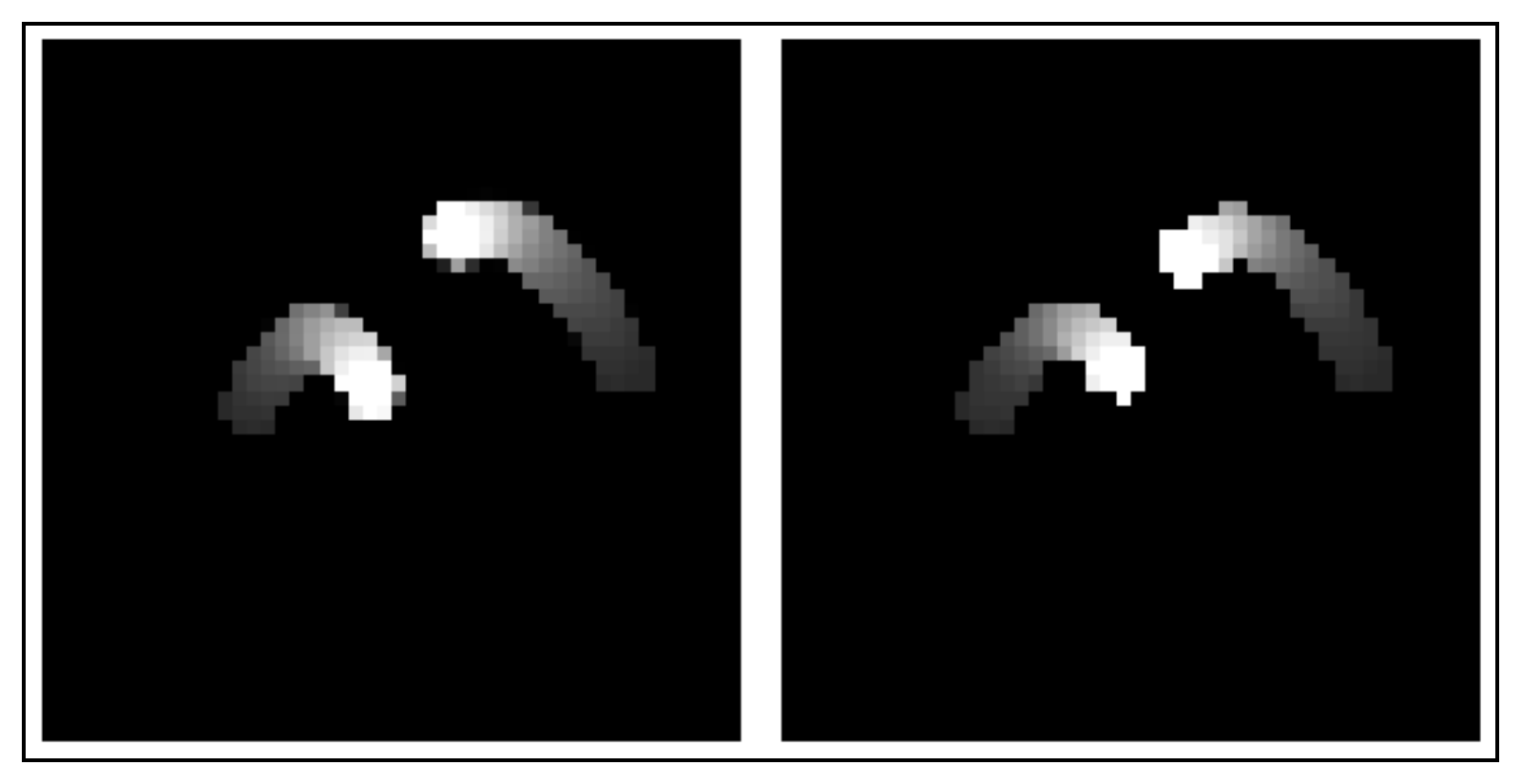}
\hskip0.0cm
\includegraphics[height=2.25cm,width=4.5cm]{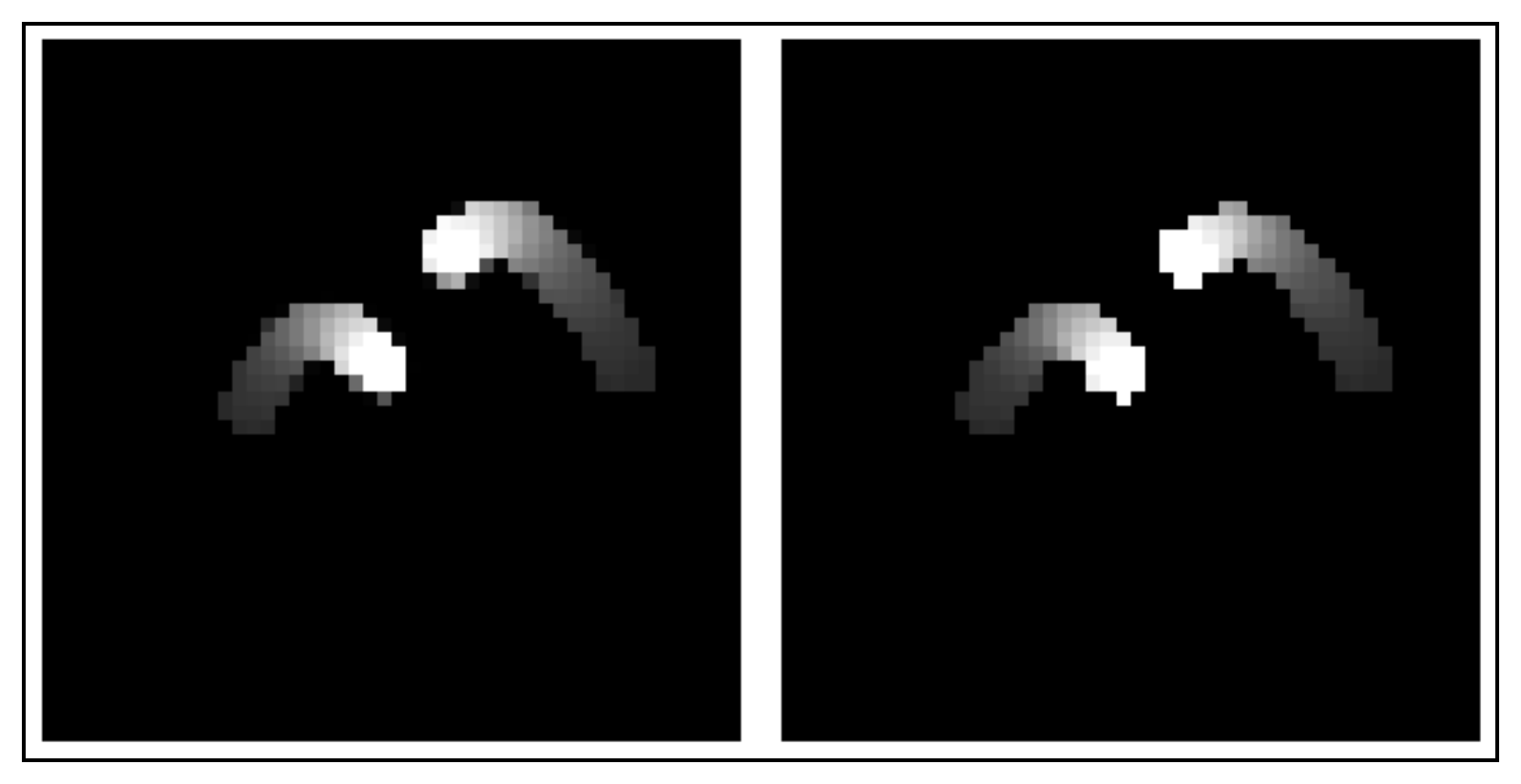}
\end{center}
\caption{Top: Ground-truth (black), inferred (blue), generated (red), and interpolated (cyan) trajectories. 
Middle: Generated versus ground-truth images and interpolated versus ground-truth images at time-step 30.
Bottom: Generated versus ground-truth images and interpolated versus ground-truth images overlaid in time.}
\label{fig:Inter2}
\end{figure}

\end{document}